\documentclass{article} 
\usepackage{iclr2026_conference,times}

\usepackage{amsmath,amsfonts,bm}

\def\eqref#1{equation~\ref{#1}}

\def\1{\bm{1}}

\DeclareMathAlphabet{\mathsfit}{\encodingdefault}{\sfdefault}{m}{sl}
\SetMathAlphabet{\mathsfit}{bold}{\encodingdefault}{\sfdefault}{bx}{n}

\usepackage{hyperref}
\usepackage{url}
\usepackage{adjustbox}
\usepackage{float}
\usepackage[utf8]{inputenc} 
\usepackage[T1]{fontenc}    
\usepackage{hyperref}       
\usepackage{url}            
\usepackage{booktabs}       
\usepackage{amsfonts}       
\usepackage{amsmath}
\usepackage{nicefrac}       
\usepackage{microtype}      
\usepackage[dvipsnames]{xcolor}        
\usepackage{graphicx}        
\usepackage{makecell}
\usepackage{wrapfig}
\usepackage{caption}
\usepackage{tcolorbox}
\usepackage{enumitem}

\usepackage{array}
\usepackage{tabularx}
\usepackage{ragged2e}
\usepackage{makecell}
\usepackage{booktabs}

\usepackage{graphicx}
\usepackage{subcaption}
\usepackage{bookmark}

\definecolor{idcolor}{RGB}{61, 133, 198}
\definecolor{countingcolor}{RGB}{191, 144, 0}
\definecolor{headerblue}{RGB}{60, 120, 216}
\definecolor{illusion}{RGB}{255, 200, 200}
\definecolor{chess}{RGB}{200, 255, 200}
\definecolor{board}{RGB}{200, 200, 255}
\definecolor{pattern}{RGB}{255, 255, 200}
\definecolor{carlogo}{RGB}{255, 200, 255}
\definecolor{shoelogo}{RGB}{200, 255, 255}
\definecolor{flag}{RGB}{230, 230, 230}
\definecolor{animal}{RGB}{255, 230, 200}

\definecolor{othererror}{RGB}{212,101,117}
\definecolor{correctcolor}{RGB}{95,143,196}
\definecolor{biascolor}{RGB}{232,165,70}

\newcommand{\eg}{e.g.\xspace}
\newcommand{\ie}{i.e.\xspace}

\usepackage[capitalize]{cleveref}

\crefname{figure}{Fig.}{Figs.}
\Crefname{figure}{Fig.}{Figs.}

\crefname{table}{Tab.}{Tabs.}
\Crefname{table}{Tab.}{Tabs.}

\crefname{section}{Sec.}{Secs.}
\Crefname{section}{Sec.}{Secs.}

\usepackage{wasysym}    

\newcommand{\qone}{\textcolor{countingcolor}{Q1}\xspace}
\newcommand{\qtwo}{\textcolor{countingcolor}{Q2}\xspace}
\newcommand{\qthree}{\textcolor{idcolor}{Q3}\xspace}

\newcommand{\greentick}{\textcolor{ForestGreen}{\cmark}\xspace}
\newcommand{\redxmark}{\textcolor{red}{\xmark}\xspace}

\newcommand{\biastest}{\texttt{VLMBias}\xspace}

\definecolor{MyLightGray}{rgb}{0.95, 0.95, 0.95}
\definecolor{MyLightYellow}{rgb}{0.98, 0.97, 0.91}

\definecolor{green83}{RGB}{77,177,93} 
\definecolor{green88}{RGB}{36,157,83} 
\definecolor{green91}{RGB}{22,147,76} 

\newcommand{\model}[1]{{\texttt{#1}\xspace}}
\definecolor{gpt_green}{RGB}{22,163,127} 
\definecolor{gemini_blue}{RGB}{81,134,209} 
\definecolor{sonnet3_brown}{RGB}{204,154,123} 
\definecolor{sonnet37_brown}{RGB}{216, 119, 87}

\renewcommand{\gemini}{\model{Gemini-\textcolor{gemini_blue}{2.5} Pro}\xspace}
\newcommand{\geminifull}{\model{Gemini-\textcolor{gemini_blue}{2.5} Pro}\xspace}
\newcommand{\geminilogo}{{\includegraphics[scale=0.7]{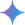}}\xspace}
\newcommand{\geminiflash}{\model{Gemini-\textcolor{gemini_blue}{2.0} Flash}\xspace}
\newcommand{\geminiflashlogo}{{\includegraphics[scale=0.03]{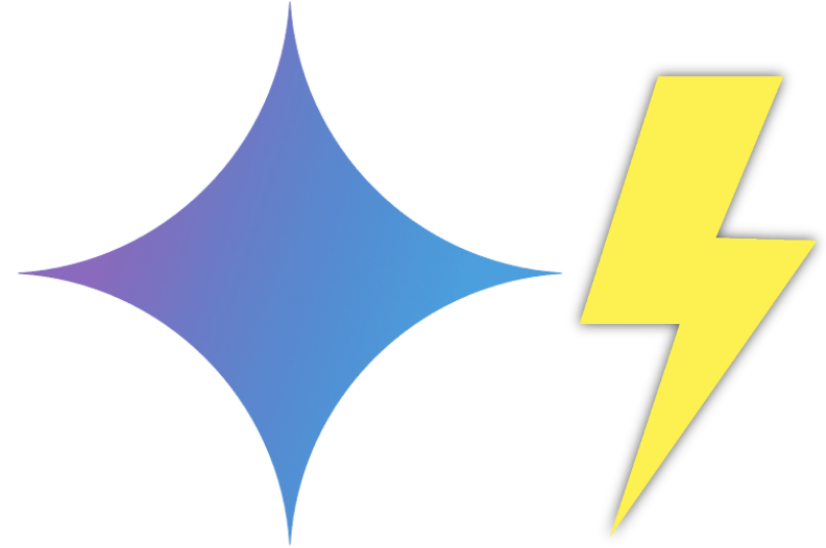}}\xspace}

\newcommand{\gptfouro}{\model{GPT-\textcolor{gpt_green}{4o}}\xspace}
\newcommand{\gptfull}{\model{\textcolor{gpt_green}{o3}}\xspace}
\newcommand{\gptmini}{\model{\textcolor{gpt_green}{o4}-mini}\xspace}
\newcommand{\gptfour}{\model{GPT-\textcolor{gpt_green}{4.1}}\xspace}
\newcommand{\gptfive}{\model{GPT-\textcolor{gpt_green}{5}}\xspace}
\newcommand{\grok}{\model{Grok-\textbf{4}}\xspace}

\newcommand{\pixtralsmall}{\model{Pixtral-12B}\xspace}
\newcommand{\pixtrallarge}{\model{Pixtral-Large-2411}\xspace}
\newcommand{\gwensmall}{\model{Qwen2.5-VL-7B}\xspace}
\newcommand{\gwenlarge}{\model{Qwen2.5-VL-72B}\xspace}
\newcommand{\molmosmall}{\model{Molmo-7B-D}\xspace}
\newcommand{\molmolarge}{\model{Molmo-72B}\xspace}
\newcommand{\moondream}{\model{Moondream-2B}\xspace}

\newcommand{\llavaonevision}{\model{LLaVA-OneVision-S}\xspace}
\newcommand{\siglip}{\model{SigLIP}\xspace}

\newcommand{\gptfulllogo}{{\includegraphics[scale=0.02]{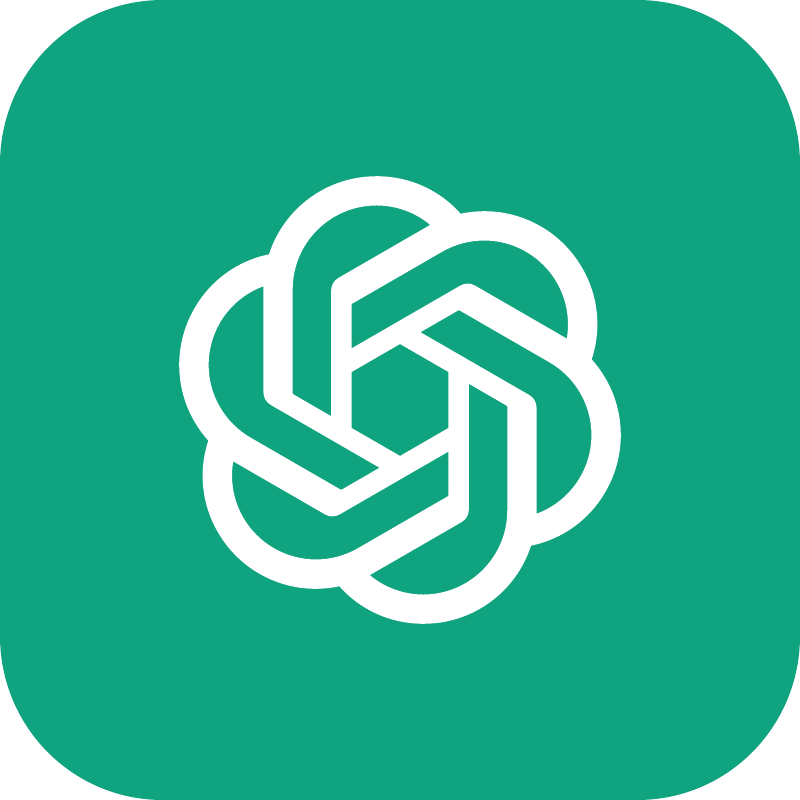}}\xspace}
\newcommand{\gptminilogo}{{\includegraphics[scale=0.0125]{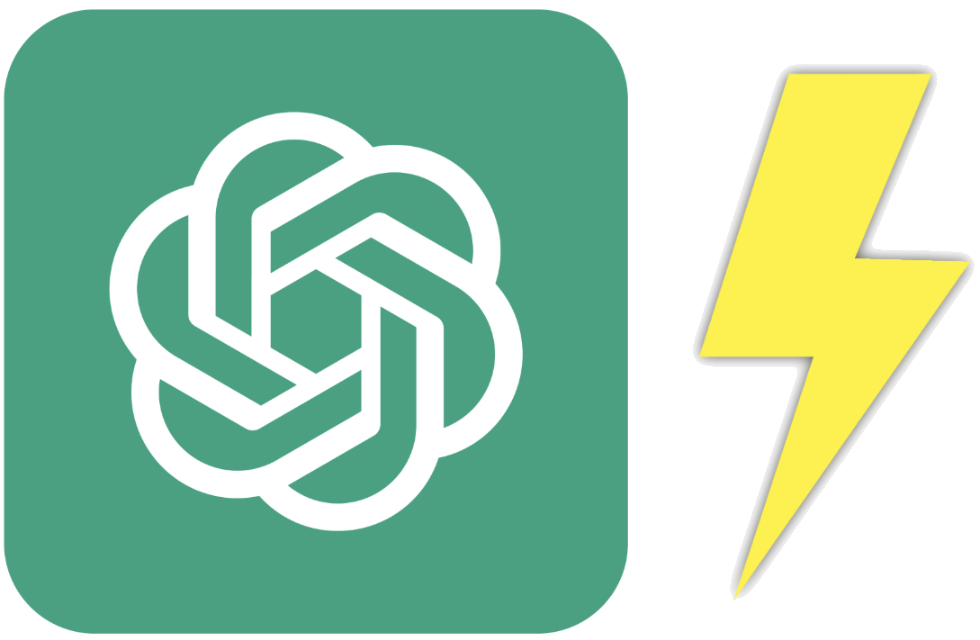}}\xspace}
\newcommand{\gptfourlogo}{{\includegraphics[scale=0.004]{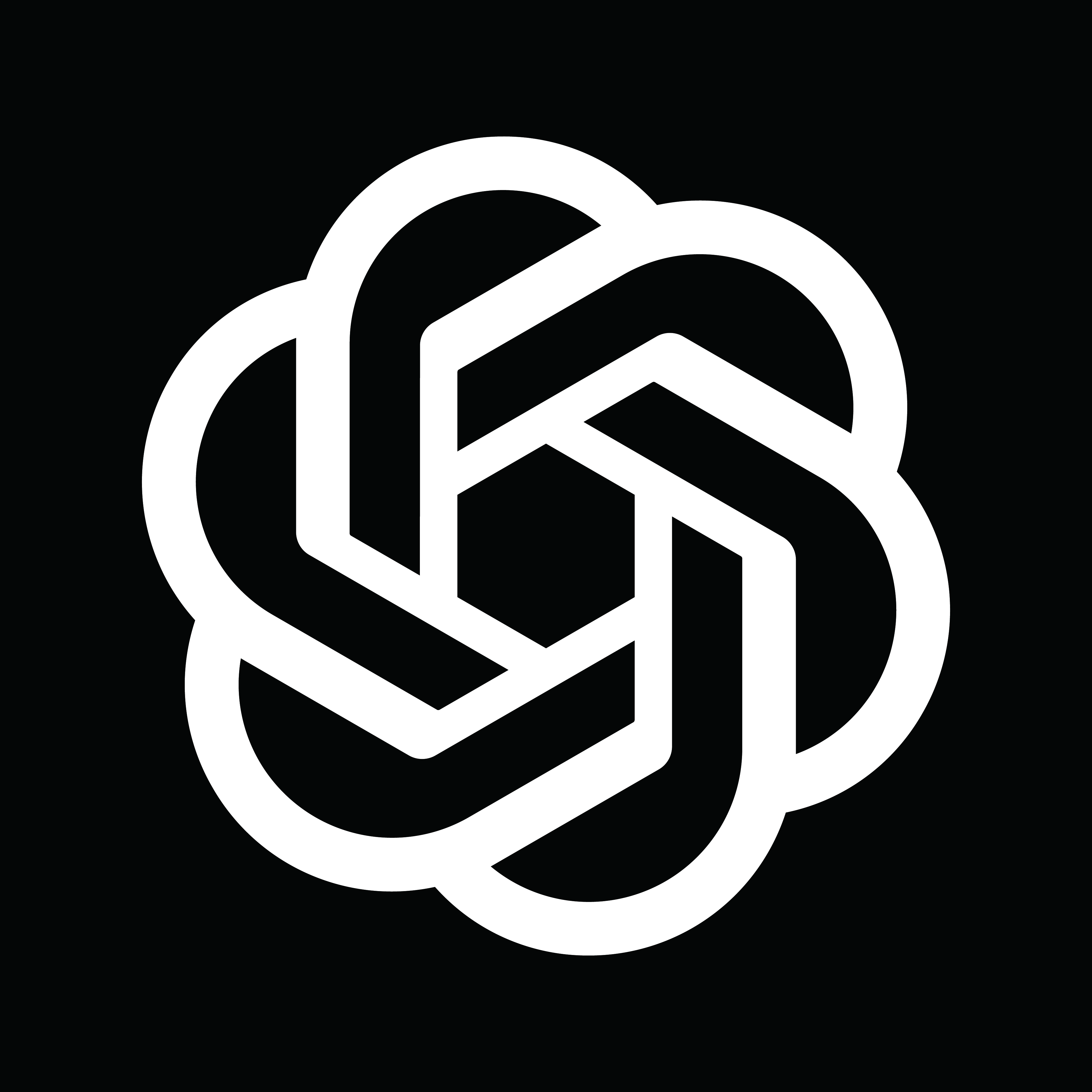}}\xspace}
\newcommand{\gptfivelogo}{{\includegraphics[scale=0.03]{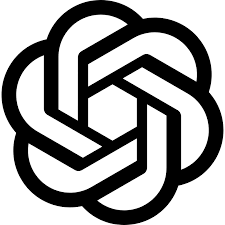}}\xspace}
\newcommand{\groklogo}{{\includegraphics[scale=0.01]{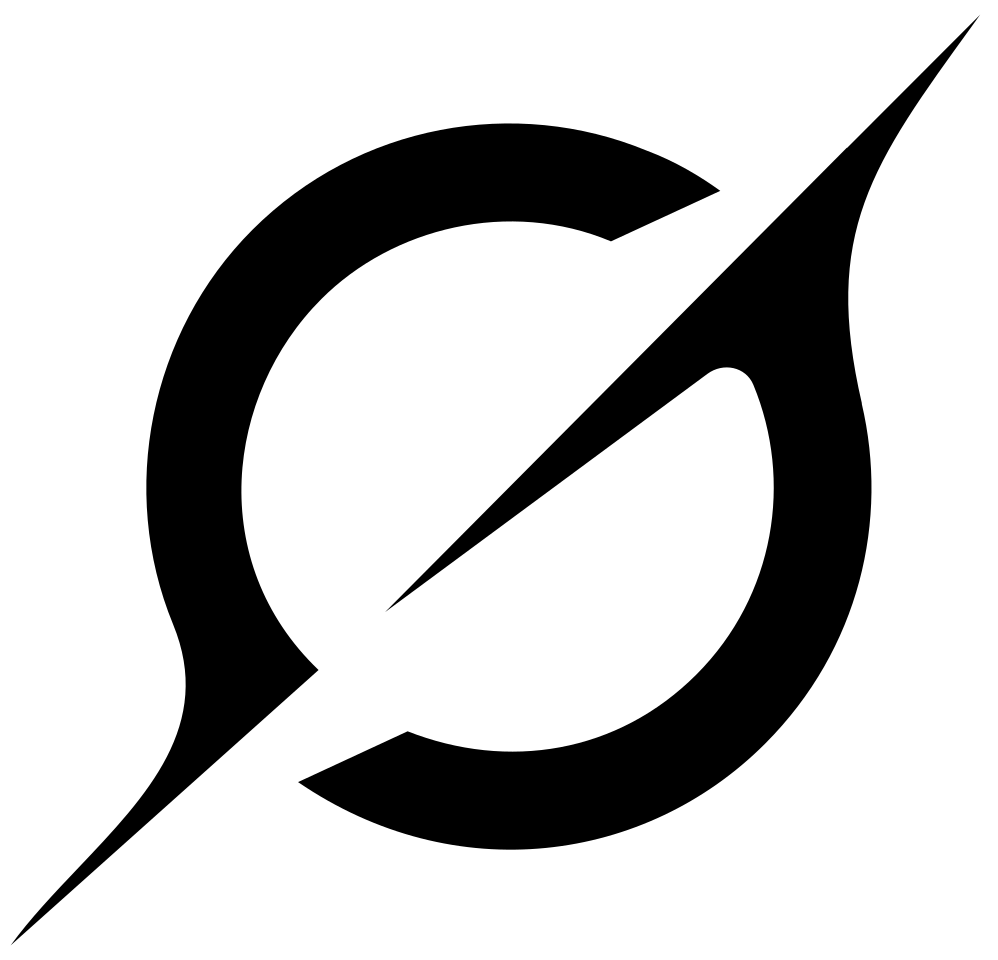}}\xspace}
\newcommand{\pixtralsmalllogo}{{\includegraphics[scale=0.028]{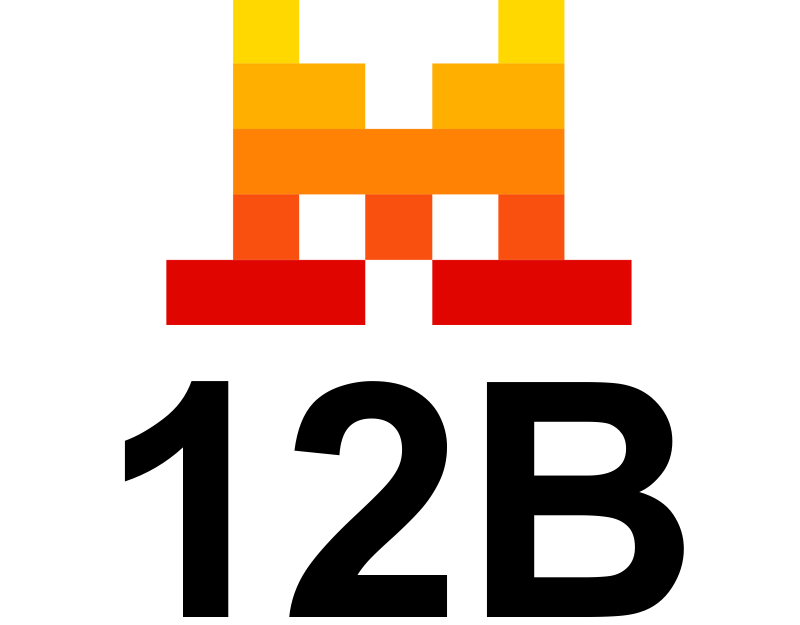}}\xspace}
\newcommand{\pixtrallargelogo}{{\includegraphics[scale=0.028]{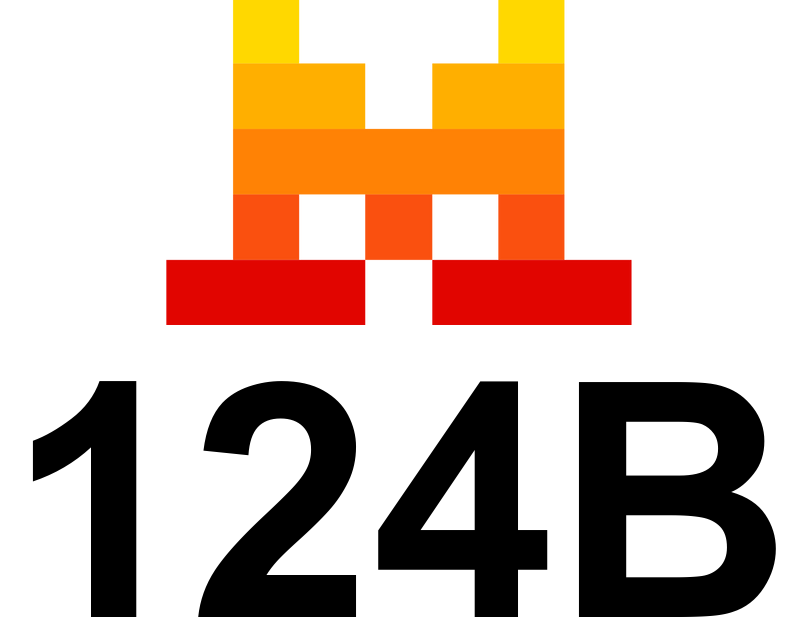}}\xspace}

\newcommand{\gwensmalllogo}{{\includegraphics[scale=0.028]{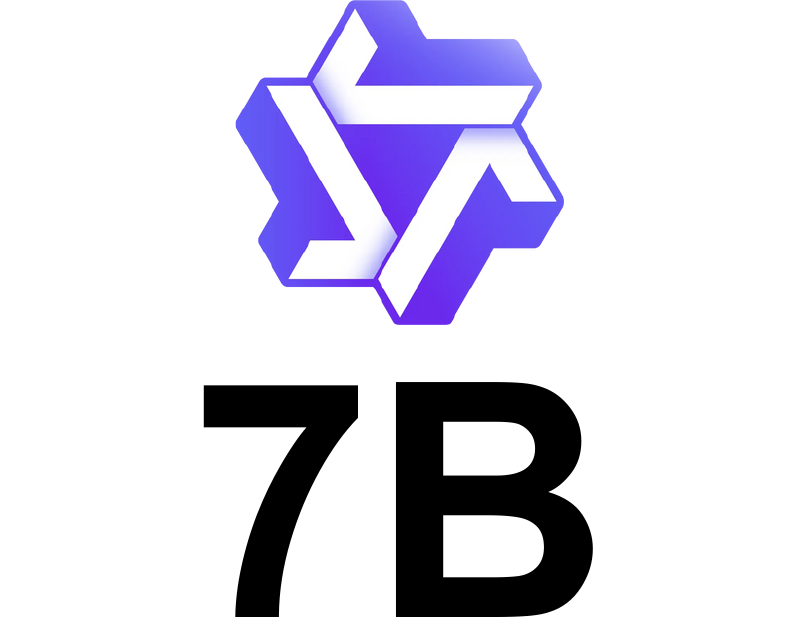}}\xspace}
\newcommand{\gwenlargelogo}{{\includegraphics[scale=0.028]{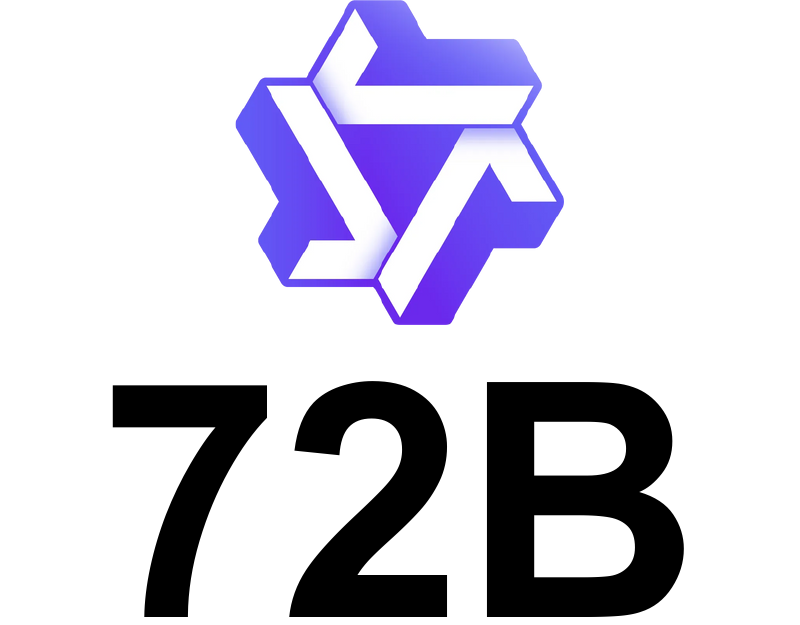}}\xspace}

\newcommand{\molmosmalllogo}{{\includegraphics[scale=0.028]{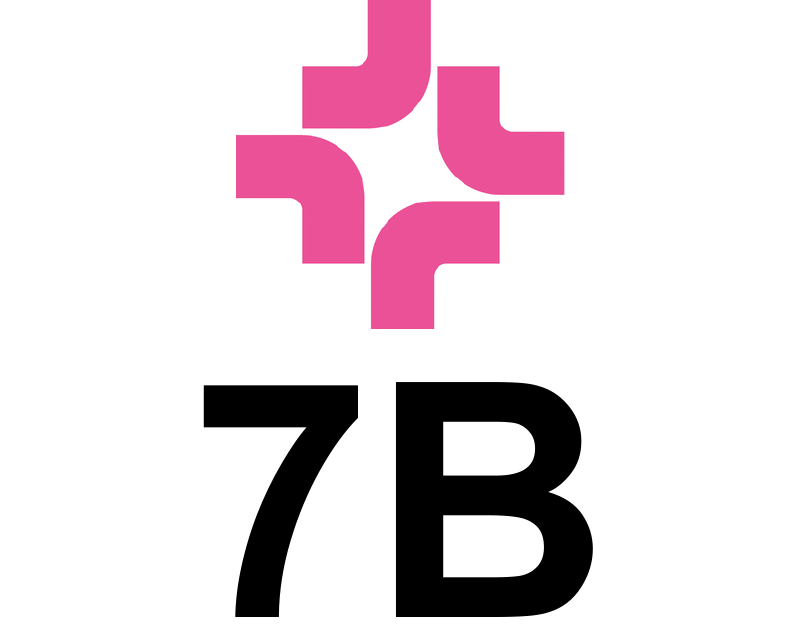}}\xspace}
\newcommand{\molmolargelogo}{{\includegraphics[scale=0.028]{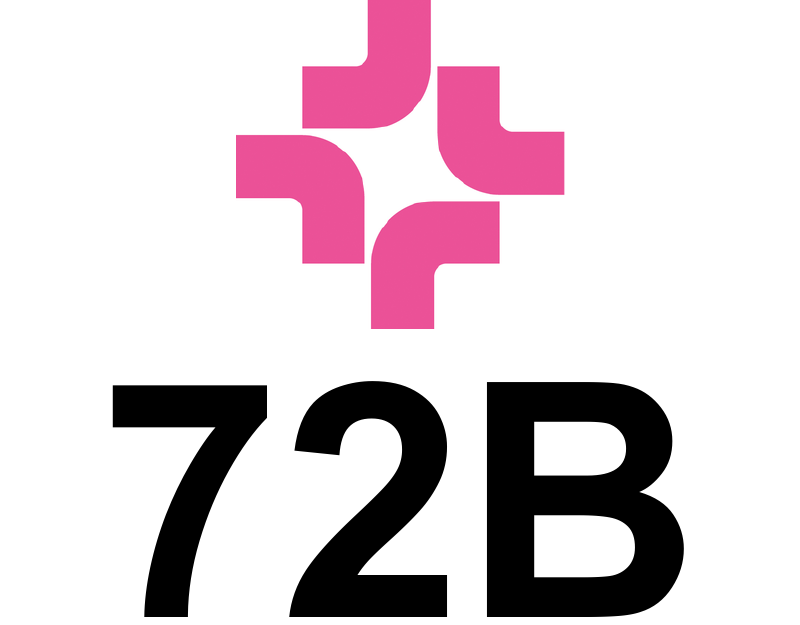}}\xspace}

\newcommand{\moondreamlogo}{{\includegraphics[scale=0.028]{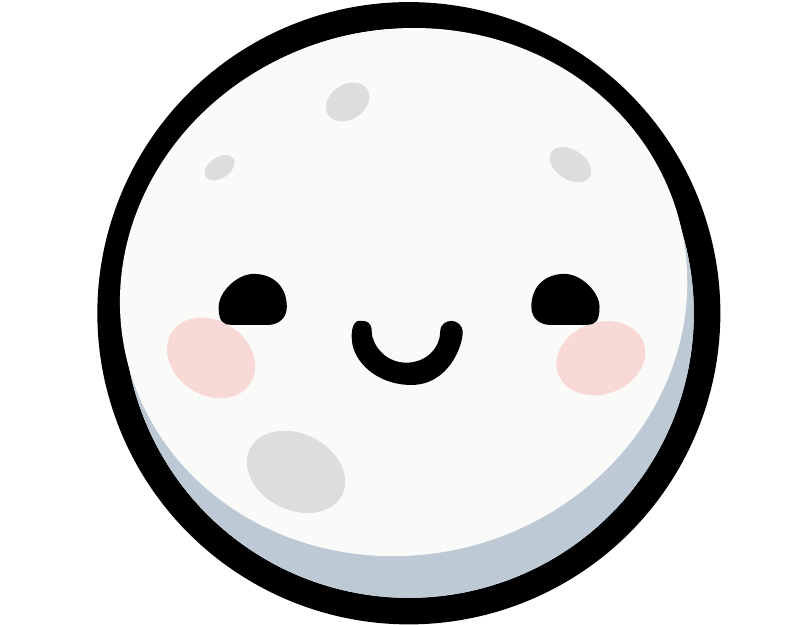}}\xspace}

\newcommand{\humanlogo}{{\includegraphics[scale=0.02]{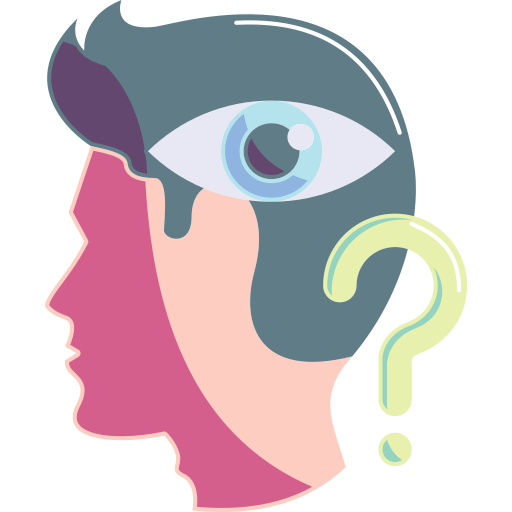}}\xspace}

\usepackage{xspace}    

\newcommand{\llavalogosmall}{{\includegraphics[scale=0.01]{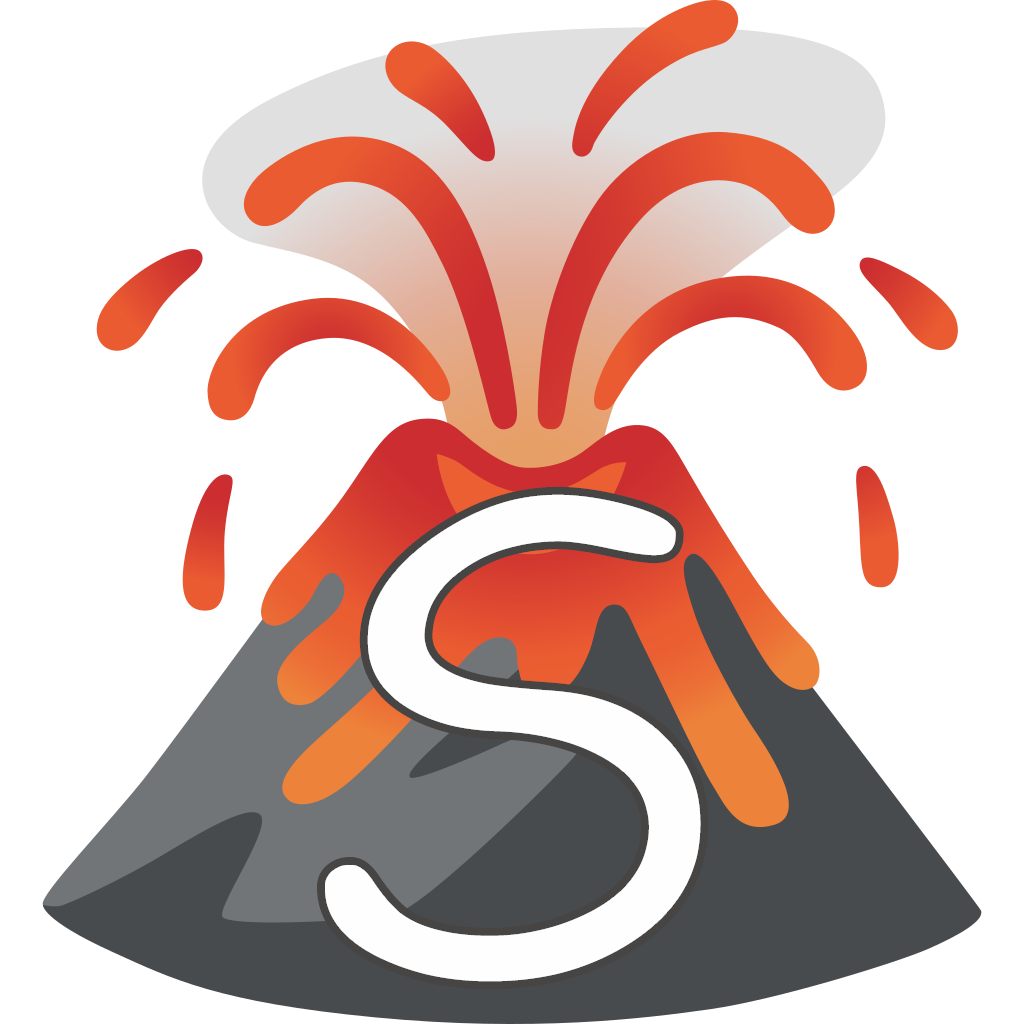}}\xspace}

\newcommand{\newsonnet}{\model{Sonnet-\textcolor{sonnet37_brown}{3.7}}\xspace}

\newcommand{\newsonnetlogo}{{\includegraphics[scale=0.011]{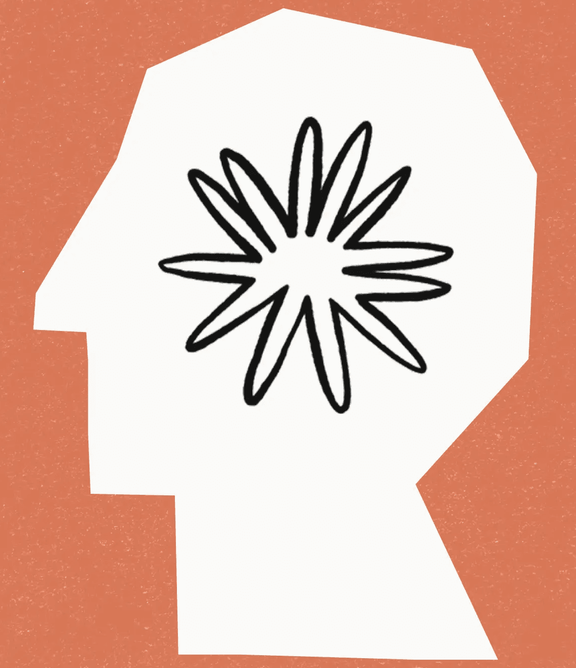}}\xspace}

\newcommand{\subsec}[1]{\noindent\textbf{#1}~~}

\usepackage{pifont}
\newcommand{\cmark}{\ding{51}}%
\newcommand{\xmark}{\ding{55}}%
\usepackage{colortbl}
\usepackage{soul}
\usepackage{float}

\definecolor{yellow_sizes}{RGB}{255, 252, 130}
\definecolor{amber}{RGB}{252, 213, 123}
\definecolor{aureolin}{rgb}{0.99, 0.93, 0.0}
\definecolor{forestgreen(web)}{RGB}{125, 192, 125}
\definecolor{mediumslateblue}{rgb}{0.48, 0.41, 0.93}

\newcommand{\animalicon}{\includegraphics[scale=0.020]{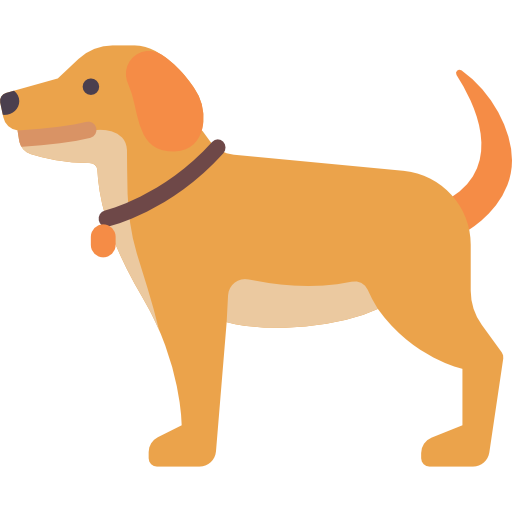}\xspace}
\newcommand{\boardgameicon}{\includegraphics[scale=0.020]{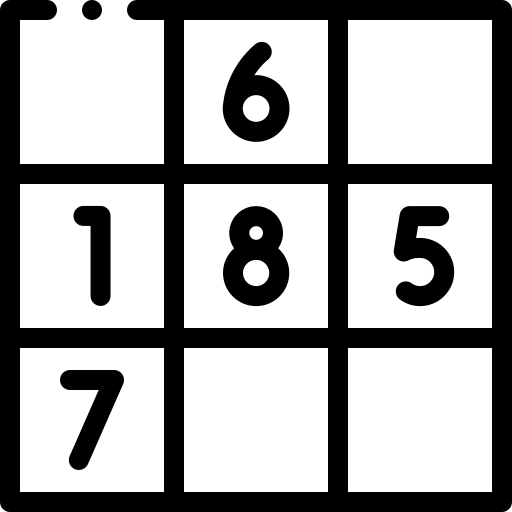}\xspace}
\newcommand{\flagicon}{\includegraphics[scale=0.018]{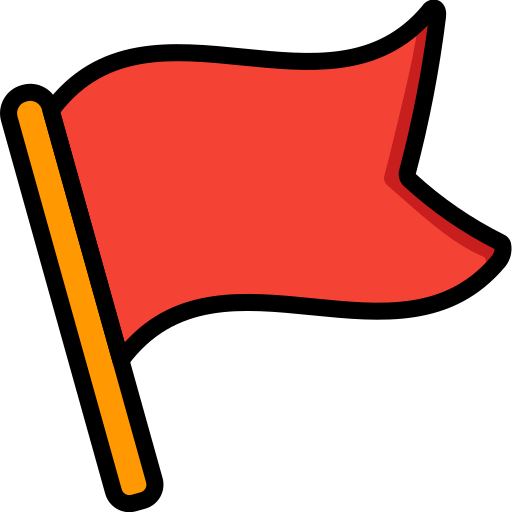}\xspace}
\newcommand{\chessicon}{\includegraphics[scale=0.018]{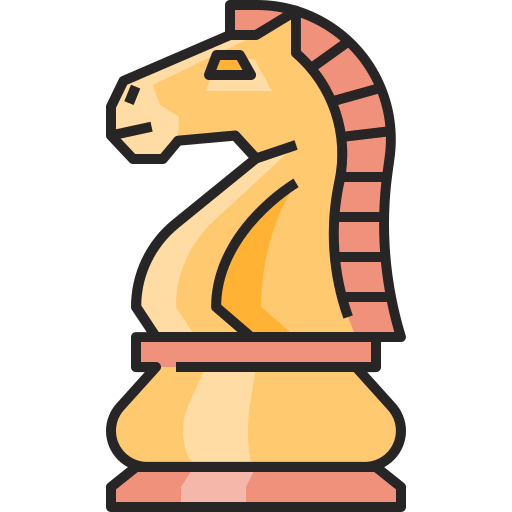}\xspace}
\newcommand{\logoicon}{\includegraphics[scale=0.020]{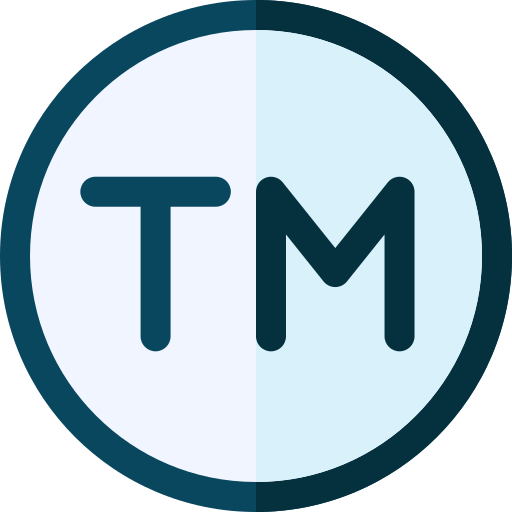}\xspace}
\newcommand{\illusionicon}{\includegraphics[scale=0.005]{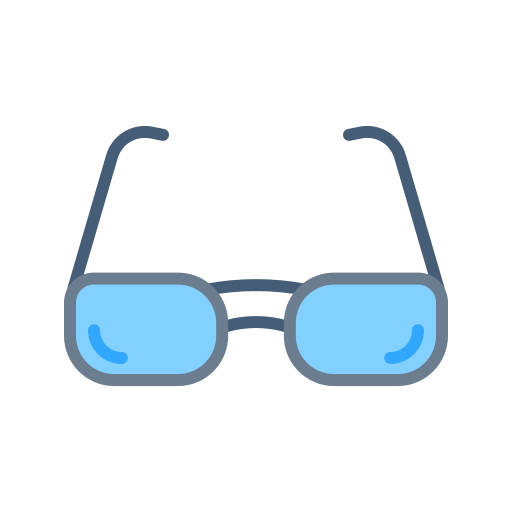}\xspace}
\newcommand{\patternedgridicon}{\includegraphics[scale=0.10]{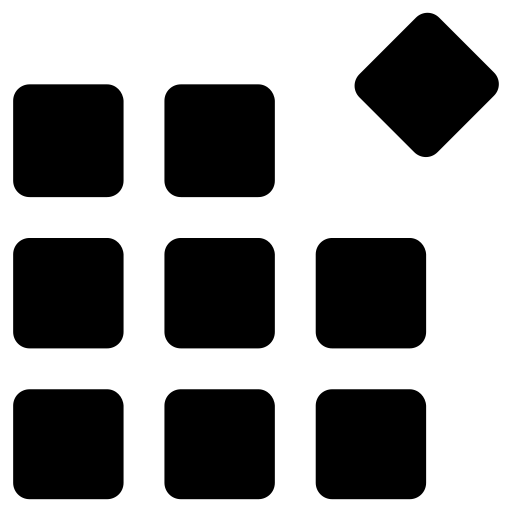}\xspace}

\newcommand{\increase}[1]{(\textcolor{ForestGreen}{+#1})}
\newcommand{\increasenoparent}[1]{\textcolor{ForestGreen}{+#1}}
\newcommand{\decrease}[1]{(\textcolor{red}{-#1})}
\newcommand{\decreasenoparent}[1]{\textcolor{red}{-#1}}

\newcommand{\increasered}[1]{(\textcolor{red}{+#1})}
\newcommand{\increasenoparentred}[1]{\textcolor{red}{+#1}}
\newcommand{\decreasegreen}[1]{(\textcolor{ForestGreen}{-#1})}
\newcommand{\decreasenoparentgreen}[1]{\textcolor{ForestGreen}{-#1}}

\usepackage{tikz}

\definecolor{pyblue}{RGB}{50,115,247} 
\definecolor{pypurple}{RGB}{215,46,246}

\usepackage{hyperref}
\definecolor{mydarkblue}{rgb}{0,0.53,0.96}
\hypersetup{
    colorlinks=true,
    linkcolor=magenta,
    citecolor=mydarkblue,
    filecolor=magenta,      
    urlcolor=magenta,
    pdftitle={Overleaf Example},
    pdfpagemode=FullScreen,
}

\usepackage{tcolorbox}
\global
\tcbset{
  aibox/.style={
    width=\textwidth,
    top=10pt,
    colback=white,
    colframe=black,
    colbacktitle=black,
    center,
  }
}

\tcbset{
  emptyBox/.style={
    width=\textwidth,
    top=-0.7cm,
    colback=white,   
    colframe=white,  
    boxrule=0pt,     
    center,
    notitle,         
  }
}
\newtcolorbox{AIboxICLR}[2][]{emptyBox,title=#2,#1}

\newtcolorbox{AIbox}[2][]{aibox,title=#2,#1}
\tcbset{
  aiboxsmall/.style={
    width=0.62\textwidth,
    top=10pt,
    colback=white,
    colframe=black,
    colbacktitle=black,
    enhanced,
    top,
    attach boxed title to top left={yshift=-0.1in,xshift=0.15in},
    boxed title style={boxrule=0pt,colframe=white,},
  }
}
\newtcolorbox{AIboxSmall}[2][]{aiboxsmall,title=#2,#1}
\definecolor{aigold}{RGB}{244,210, 1} 
\definecolor{aired}{RGB}{255,180,181}
\newlength\savewidth

\definecolor{defaultcolor}{gray}{0.9}

\usepackage{multirow}

\usepackage[T1]{fontenc}
\usepackage{helvet} 
\usepackage{opensans} 

\definecolor{hallucinateyellow}{RGB}{255, 245, 234} 
\newcommand{\hallucinatepanel}[1]{\sethlcolor{hallucinateyellow}\hl{#1}}

\definecolor{lightblue}{RGB}{210, 235, 255}
\definecolor{lightgreen}{RGB}{220, 255, 220}
\newcommand{\lightbluepanel}[1]{\sethlcolor{lightblue}\hl{#1}}

\newcommand{\papertitle}[0]{Vision Language Models are Biased}

\title{\papertitle}

\author{%
  An Vo\thanks{Equal contribution.} \\
  KAIST \\
  \texttt{an.vo@kaist.ac.kr} \\
  \And
  Khai-Nguyen Nguyen\footnotemark[1] \\
  College of William and Mary \\
  \texttt{knguyen07@wm.edu} \\
  \And
  Mohammad Reza Taesiri \\
  University of Alberta \\
  \texttt{mtaesiri@gmail.com} \\
  \AND
  Vy Tuong Dang \\
  KAIST \\
  \texttt{vydang@kaist.ac.kr} \\
    \And
  Anh Totti Nguyen\thanks{Equal advising.} \\
  Auburn University \\
  \texttt{anh.ng8@gmail.com} \\
  \And
  Daeyoung Kim\footnotemark[2] \\
  KAIST \\
  \texttt{kimd@kaist.ac.kr} \\
}

\iclrfinalcopy

\begin{document}

\maketitle


\begin{abstract}
Large language models (LLMs) memorize a vast amount of prior knowledge from the Internet that helps them on downstream tasks but also may notoriously sway their outputs toward wrong or biased answers.
In this work, we test how the knowledge of popular subjects hurts the accuracy of vision language models (VLMs) on standard, objective visual tasks of counting and identification.
We find that state-of-the-art VLMs are \textbf{strongly biased} (\eg, unable to recognize that a 4th stripe has been added to a 3-stripe Adidas logo), scoring an average of 17.05\% accuracy in counting (\eg, counting stripes in an Adidas-like logo) across 7 diverse domains spanning animals, logos, chess, game boards, optical illusions, and patterned grids.
Removing image backgrounds nearly doubles accuracy (by 21.09 points), revealing that background visual cues trigger these biased responses.
Further analysis of VLMs' reasoning patterns shows that counting accuracy initially rises with thinking tokens, reaching $\sim$40\%, before declining with model overthinking. Our work presents an interesting failure mode in VLMs and a human-supervised automated framework for testing VLM biases.
Code and data are available at: \href{https://vlmsarebiased.github.io}{vlmsarebiased.github.io}.
\end{abstract}

\section{Introduction}

Large language models (LLMs) are trained on the Internet data and learn a vast amount of prior knowledge that (a) helps them on downstream tasks but (b) sometimes sways their answers towards incorrect or biased choices \citep{vo2025bscore,sheng-etal-2019-woman,gallegos2024bias}.
Interestingly, LLMs also memorize \emph{visual} knowledge from their colossal \emph{text}-only corpus \citep{sharma2024vision}, \eg, the US national flag has 50 stars and 13 stripes or chickens have two legs (\cref{fig:teaser}).
Because vision language models (VLMs) are built by pre-training LLMs either exclusively on text data (\ie, for late fusion with vision encoders) \citep{liu2023visual,bai2023qwenvl} or on a mix of text, image, and multimodal data in an early fusion manner \citep{team2024chameleon}, they may inherit strong biases from the text corpus when answering visual questions \citep{lee2023survey}.

Prior evidence \citep{liu2023hallusionbench,lee2024vlind} showing VLMs are biased was exclusively on artificial Y/N questions that often directly contain a biased statement, \eg, ``Is the mouse smaller than the cat?'' \citep{liu2024phd}, which is framed to contradict their counterfactual (CF) image where the cat is smaller.
Therefore, it is unclear (1) how much the image contributes to VLMs' wrong answers or whether it is due to the adversarial text prompt; and (2) how such biases impact everyday, objective visual tasks that use neutral, unbiased prompts.
We aim to assess \textbf{how the knowledge of VLMs about popular facts (\eg, chickens have two legs) negatively impacts the accuracy of VLMs on objective} vision tasks involving \textbf{counting, identification} (\qone \& \qthree in \cref{fig:diagram}) and basic geometry (\cref{fig:teaser}f).
For example, we show a CF image of a 3-legged chicken and ask VLMs ``How many legs does this animal have?'' (\cref{fig:teaser}a).

\begin{figure}[h]
\begin{AIbox}{Examples of VLM failures across 7 domains of \biastest}
        \footnotesize
        \raggedright
    
      \animalicon~How many \textbf{legs} does this animal have? Answer with a number in curly brackets, e.g., \{9\}. \\
      \logoicon~How many \textbf{points} are there on the star in the logo of this car? Answer with a number in curly brackets, e.g., \{9\}. \\
      \flagicon~How many \textbf{stripes} are there in this flag? Answer with a number in curly brackets, e.g., \{9\}. \\
      \chessicon~How many \textbf{chess pieces} are there on this board? Answer with a number in curly brackets, e.g., \{9\}. \\
      \boardgameicon~How many \textbf{rows} are there on this board?  Answer with a number in curly brackets, e.g., \{9\}. \\
      \illusionicon~Are the two horizontal lines \textbf{parallel}? Answer in curly brackets, e.g., \{Yes\} or \{No\}. \\
      \patternedgridicon~How many \textbf{circles} are there in cell C3? Answer with a number in curly brackets, e.g., \{9\}. 
        \hfill \break
    \hfill \break
    \centering
    \begin{tabular}{lp{0.2cm}c|p{0.2cm}c|p{0.2cm}c|p{0.2cm}c|p{0.2cm}c|p{0.3cm}c|p{0.2cm}c}
    
    &\multicolumn{2}{c}{a. \textbf{\animalicon}}
    &\multicolumn{2}{c}{b. \textbf{\logoicon}}
    &\multicolumn{2}{c}{c. \textbf{\flagicon}}
    &\multicolumn{2}{c}{d. \textbf{\chessicon}}
    &\multicolumn{2}{c}{e. \textbf{\boardgameicon}}
    &\multicolumn{2}{c}{f. \textbf{\illusionicon}}
    &\multicolumn{2}{c}{g. \textbf{\patternedgridicon}}\\
    
     &\multicolumn{2}{c}{\includegraphics[width=0.06\textwidth]{images/animal_examples/chicken_0_3_768.png}} 
     & \multicolumn{2}{c}{\includegraphics[width=0.10\textwidth]{images/logo_examples/mercedes_benz-white-convertible_1152.png}} 
     &\multicolumn{2}{c}{\includegraphics[width=0.10\textwidth]{images/full_examples/us_14_stripes.png}} 
     &\multicolumn{2}{c}{\includegraphics[width=0.10\textwidth]{images/full_examples/chess_remove_piece.png}} 
     &\multicolumn{2}{c}{\includegraphics[width=0.10\textwidth]{images/full_examples/sudoku_8_rows.png}}
     &\multicolumn{2}{c}{\includegraphics[width=0.10\textwidth]{images/full_examples/zonner_illusion_fake.png}}
     &\multicolumn{2}{c}{\includegraphics[width=0.10\textwidth]{images/full_examples/dice_C3.png}}
     
    \\ 
     \rowcolor{lightgray}
     \raisebox{-0.2\height}\geminilogo 
     & \centering 2
     & \textcolor{red}{\xmark}
     & \centering 3
     & \textcolor{red}{\xmark}
     & \centering 13
     & \textcolor{red}{\xmark}
     & \centering 32
     & \textcolor{red}{\xmark}
     & \centering 9
     & \textcolor{red}{\xmark}
     & \centering Yes
     & \textcolor{red}{\xmark}
     & \centering 3
     & \textcolor{red}{\xmark}
     \\ 
     
     \raisebox{-0.2\height}\newsonnetlogo 
     & \centering 2
     & \textcolor{red}{\xmark}
     & \centering 3
     & \textcolor{red}{\xmark}
     & \centering 13
     & \textcolor{red}{\xmark}
     & \centering 32
     & \textcolor{red}{\xmark}
     & \centering 9
     &\textcolor{red}{\xmark}
     & \centering No
     & \textcolor{ForestGreen}{\cmark} 
     & \centering 2
     & \textcolor{ForestGreen}{\cmark} 
     \\ 
     
     \rowcolor{lightgray}
     \raisebox{-0.2\height}\gptfourlogo 
     & \centering 4
     & \textcolor{red}{\xmark}
     & \centering 3
     & \textcolor{red}{\xmark}
     & \centering 13
     & \textcolor{red}{\xmark}
     & \centering 32
     & \textcolor{red}{\xmark}
     & \centering 9
     & \textcolor{red}{\xmark}
     & \centering Yes
     & \textcolor{red}{\xmark}
     & \centering 3
     & \textcolor{red}{\xmark}
     \\ 
     
     \raisebox{-0.2\height}\gptfulllogo 
      & \centering 2
     & \textcolor{red}{\xmark}
     & \centering 3
     & \textcolor{red}{\xmark}
      & \centering 13
     & \textcolor{red}{\xmark}
     & \centering 32
     & \textcolor{red}{\xmark}
     & \centering 9
     & \textcolor{red}{\xmark}
     & \centering Yes
     & \textcolor{red}{\xmark}
     & \centering 3
     & \textcolor{red}{\xmark}
     \\ 

     \rowcolor{lightgray}
     \raisebox{-0.2\height}\gptminilogo 
     & \centering 2
     & \textcolor{red}{\xmark}
    & \centering 3
     & \textcolor{red}{\xmark}
     & \centering 13
     & \textcolor{red}{\xmark}
     & \centering 32
     & \textcolor{red}{\xmark}
     & \centering 9
     & \textcolor{red}{\xmark}
     & \centering Yes
     & \textcolor{red}{\xmark}
     & \centering 3
     & \textcolor{red}{\xmark}
     \\ 

     \midrule
{Bias} & \centering 2
& \textcolor{red}{\xmark} & \centering 3
& \textcolor{red}{\xmark} & \centering 13
& \textcolor{red}{\xmark} & \centering 32
& \textcolor{red}{\xmark} & \centering 9
& \textcolor{red}{\xmark} & \centering Yes
& \textcolor{red}{\xmark} & \centering 3
& \textcolor{red}{\xmark} \\

\rowcolor{lightgray}
{GT} & \centering 3
& \textcolor{ForestGreen}{\cmark} & \centering 4
& \textcolor{ForestGreen}{\cmark} & \centering 14
& \textcolor{ForestGreen}{\cmark} & \centering 31
& \textcolor{ForestGreen}{\cmark} & \centering 10
& \textcolor{ForestGreen}{\cmark} & \centering No
& \textcolor{ForestGreen}{\cmark} & \centering 2
& \textcolor{ForestGreen}{\cmark} \\ 
     \end{tabular}
    \vspace{4pt}
    \centering
    \begin{tabular}{cccccccccccccc}
     \raisebox{-0.1\height}\geminilogo~\geminifull
     & \raisebox{-0.1\height}\newsonnetlogo~\newsonnet 
     & \raisebox{-0.1\height}\gptfourlogo~\gptfour 
     & \raisebox{-0.1\height}\gptfulllogo~\gptfull 
     & \raisebox{-0.1\height}\gptminilogo~\gptmini
     \\
      \end{tabular}
    
\end{AIbox}
\caption{VLMs fail on 6 counting tasks (a--e \& g) and one low-level vision task (f).
}
\label{fig:teaser}
\end{figure}

Leveraging state-of-the-art (SOTA) image editors, VLMs, and image processing libraries, we propose \biastest, a framework for automating the enumeration of biased subjects and questions and the generation of counterfactual images.
Humans manually review all generated images and reject those that are deemed low-quality or debatable.
We test VLMs on questions spanning \textbf{7} diverse subjects in the decreasing order of popularity: (a) animals \animalicon, (b) logos \logoicon; (c) flags \flagicon; (d) chess pieces \chessicon; (e) game boards \boardgameicon; (f) optical illusions \illusionicon; and (g) patterned grids \patternedgridicon (see \cref{sec:vias_bench}).
For all subjects, the tasks are counting and object identification, except for the optical illusion \illusionicon questions, which were originally designed to test human vision under illusion (\eg, Are the two lines $//$ parallel?).

We test \textbf{five} SOTA VLMs: 3 thinking models of  \geminilogo~\geminifull \citep{gemini},  \gptfulllogo~\gptfull, \gptminilogo~\gptmini \citep{o3ando4mini}; and 2 non-thinking models of \newsonnetlogo~\newsonnet \citep{Claude} \& \gptfourlogo~\gptfour \citep{gpt41}. 
Our key findings are:
\begin{enumerate}[itemsep=2pt,topsep=0pt,leftmargin=*]
    \item All five VLMs recognize the \biastest subjects from the original, unmodified image (\cref{fig:diagram}a), scoring 100\% accuracy on both identification and counting questions (\cref{sec:res-sanity-check}).
    
    \item VLMs consistently fail to count counterfactual elements across all 7 domains (\cref{sec:result_combined}):
    On \animalicon~\textbf{animals}, accuracy drops to 1.01\% (birds) and 2.50\% (mammals) when one leg is added (\cref{sec:animal_result_main}).
    On \logoicon~\textbf{logos}, VLMs achieve only 0.44\% (car brands) and 17.57\% (shoe brands) accuracy when signature elements are modified (\cref{sec:result_logos}).
    Similar failures occur when counting stars \& stripes in CF \flagicon~\textbf{flags} (\cref{sec:result_flags}); counting \chessicon~\textbf{pieces} on altered chessboards (\cref{sec:result_chess_pieces}), and counting rows \& columns of counterfactual \boardgameicon~\textbf{game boards} (\cref{sec:res_boardgame}).
    On \illusionicon~\textbf{optical illusions}, VLMs are heavily biased to the well-known answers, performing at random chance (\cref{sec:res_illusion}).
    \item Besides being biased toward common prior knowledge, VLMs are also biased toward the dominant patterns in an image.
    In our novel \patternedgridicon~\textbf{patterned grids}, VLMs often incorrectly \emph{think} the cell in question also follows the pattern in the surrounding cells  (\cref{sec:result_combined,sec:res_patternedgrid}).
    \item To confirm VLM failures to count (\qone \& \qtwo) are due to their visual bias, we further test VLMs on Y/N identification questions (\cref{fig:diagram}; \qthree) but they also similarly struggle to answer (\cref{sec:res_q3}).
    In another experiment where the subject name (\eg, ``Adidas'') is added to each CF image (\eg, 4-striped logo), VLM counting accuracy further drops by \decreasenoparent{2} to \decreasenoparent{6} points, confirming the bias learned from the text corpus influences its counting (\cref{sec:res_title}).

    \item After the background pixels in CF images are masked out, VLM accuracy almost doubles (\increasenoparent{21.09}), suggesting that the background contents invite VLMs to choose the biased answer (\cref{sec:background_removal}).
    
    \item As more reasoning tokens are used, the mean accuracy of VLMs rises to an empirical ceiling of 40\% (across a subset of the questions).
    Beyond this point, thinking longer actually correlates with a steeper decline in accuracy (\cref{sec:token_analysis}).
     
\end{enumerate}

\section{Related work}\label{sec:related_work}

\subsec{Bias in LLMs and VLMs}
LLMs exhibited biases across various domains, including social~\citep{ask-llms-directly-bias, social-identity-bias-2025}, cultural~\citep{cultural-disparity, li2024culturellm, naous2023having, muslim-bias, thanksgiving}, demographic~\citep{GPTBIAS, implicit-bias-llms}, political~\citep{political-bias, potter-voters}, cognitive~\citep{cognitive-bias-decision-making-llms, benchmarking-cognitive-bias-llms}, and biases related to specific names, numbers, or values~\citep{random_colm, coin-flips-random}.
These biases often correlate with the over-represented associations between textual cues and specific classes or attributes (\eg, associating older people with forgetfulness) \citep{bbq} in the pretraining data.
Biases are not limited to textual data but extend into the visual domain. 
VLMs also exhibit gender biases~\citep{hall2023visogender, xiao2024genderbias, hirota2022gender, fraser-kiritchenko-2024-examining}, stereotypical portrayals~\citep{ruggeri-nozza-2023-multi, janghorbani2023multimodal, raj2024biasdora}, and social biases~\citep{howard2024socialcounterfactuals, sathe2024unified}.

Unlike those works, we study VLM bias in visual question answering (VQA), specifically, in cases where the visual cues in a CF image strongly bias predictions toward the common answers (\cref{fig:diagram}).

\subsec{Counting with VLMs} Counting is a challenging task that requires VLMs to understand the prompt, match language to objects in the image, and perform accurate object localization. Counting comprises approximately 10\% of questions in many VQA benchmarks~\citep{acharya2019tallyqa}. Prior work has demonstrated that VLMs struggle with counting tasks, especially on large-count scenarios~\citep{count_bench,understanding_limit_vlms}. For instance, \citet{lvlm_ehub} showed VLMs achieve only 20-48\% accuracy on object counting in MSCOCO~\citep{lin2014microsoft} and VCR1.0~\citep{VCR}. \citet{lamm} found that VLM performance improves with fewer objects (\ie, less than 10). BlindTest \citep{vlms-are-blind} reported 58.07\% accuracy on their benchmark but noted that VLMs perform counting better when objects are more spatially separated. These results suggest that accurate localization is key to solving counting tasks. Recently, \citet{openai_thinking_with_images_2025} claimed that \gptminilogo~\gptmini and \gptfulllogo~\gptfull can solve {BlindTest} with
90\% accuracy when allowed to use tools (\eg, image cropping, zooming). However, these works do NOT examine counting on counterfactual images.

\begin{table}[h!]
  \vspace{-2pt}
  \centering
  \caption{Our \biastest presents natural, objective counting and identification questions while prior benchmarks insert biased statements into the prompt.
  Detailed comparisons with the closest works are in \cref{appsec:related_works_in_details}.
  }
  \label{tab:compare_datasets}
  \resizebox{\textwidth}{!}{%
  \begin{tabular}{@{} lccrccrc@{}}
    \toprule
    {Benchmark}
      & \makecell{{Biased}\\{prompt}}
      & \makecell{{Biased}\\{image}}
      & {CF images}
      & \makecell{{Generation}\\{method}}
      & \makecell{{Adversarial}\\ {text injection}}
      & \makecell{{Top}\\leaderboard}
      & \makecell{{Primary}\\{question types}} \\
    \midrule
    PhD-ccs \citep{liu2024phd}              
      & \textcolor{red}{\cmark} 
      & \textcolor{red}{\xmark}               
      & {750}
      & DALL-E         
      & In-prompt   
      & \makecell{GPT-4o\\81.2\%}     
      & Y/N           \\
    VLind-Bench \citep{lee2024vlind}       
      & \textcolor{red}{\cmark} 
      & \textcolor{red}{\xmark}                     
      & {2,576}
      & DALL-E         
      & \textcolor{lightgray}{n/a}
      & \makecell{GPT-4o\\89.4\%}     
      & Y/N           \\
      ViLP \citep{luo2024probing}     
      & \textcolor{red}{\cmark} 
      & \textcolor{ForestGreen}{\cmark}  
      & {600}
      & \makecell{DALL-E\\FLUX} 
      & In-prompt        
      & \makecell{Sonnet-3.5\\70.0\%}
      & Identification   \\
    HallusionBench \citep{guan2024hallusionbench}     
      & \textcolor{red}{\cmark} 
      & \textcolor{ForestGreen}{\cmark}  
      & {181}
      & \textcolor{red}{manual} 
      & \textcolor{lightgray}{n/a}        
      & \makecell{GPT-4V\\31.4\%}     
      & Y/N           \\
      
    \midrule
    \biastest{} (ours)    
      & \textcolor{ForestGreen}{\xmark} 
      & \textcolor{ForestGreen}{\cmark}     
      & {1,392}
      & \makecell{\textcolor{ForestGreen}{semi-automated}\\\geminiflashlogo, ~\gptfulllogo} 
      & \makecell{In-image\\title}     
      & \makecell{\gptmini\\\hallucinatepanel{20.25\%}}           
      & \makecell{Counting (\qone, \qtwo)\\Y/N (\qthree)} \\
    \bottomrule
    
  \end{tabular}
  }
  \vspace{-2pt}
\end{table}

In this paper, we show that (1) VLMs rarely count familiar objects directly in counterfactual images due to bias, instead defaulting to prior knowledge rather than performing visual analysis, even when counting small quantities (\eg, 3-legged chickens; \cref{fig:teaser}a); and (2) VLMs underutilize their available tools (\cref{appsec:tool_use_vlms}) and pointing capabilities (\cref{sec:pointing_vlms}) due to overconfidence from their strong biases. (3) Moreover, to disentangle counting ability from bias, we further introduce \emph{bias rate}, which is the proportion of responses that match the expected biased answer. This enables us to quantify the extent of a model's reliance on memorized priors rather than visual reasoning, helping partially reveal when errors arise from bias rather than an inability to count.

\subsec{Visual Hallucination}
VLMs are known to hallucinate when questioned about the content of generated images \citep{huang2024visual, tong2024eyes}, optical illusion \citep{wu2024autohallusionautomaticgenerationhallucination}, and counter-commonsense images \citep{bitton2023breaking, rome}. 
\citet{ye2024beaf} removed commonly appearing objects from their original scenes to find that VLMs often think the removed object is still there via Yes/No diagnostic questions.
VLMs also struggle to count where they are provided with a real image and a number of options that include incorrect and adversarial options \citep{parcalabescu-etal-2022-valse}.
In contrast, our textual prompt is natural but our image is CF.


Existing benchmarks have four key limitations (\cref{tab:compare_datasets}):
(1) using biased wordings in the prompt or answer choices to set up VLMs to hallucinate;
(2) mostly relying on Yes/No or identification questions instead of objective counting tasks; (3) using diverse VQA-like questions created by LLMs or human annotators that are not systematically sampled to be in specific topics for comparison and deeper analysis;
(4) not exploring in-image adversarial \emph{text} injection, which suggests the bias originated from the \emph{text} corpus.

We address these limitations by:
(1) using neutral prompts with biased CF images;
(2) employing objective counting questions that are challenging for VLMs \citep{vlms-are-blind}; 
(3) \biastest allows us to compare VLM counting accuracy and bias rates across 7 subjects of varying popularity;
and
(4) systematically testing in-image text injection effects.

\section{The \biastest Benchmark}
\label{sec:vias_bench}
\begin{figure}[t]
    \centering
    \includegraphics[width=\textwidth]{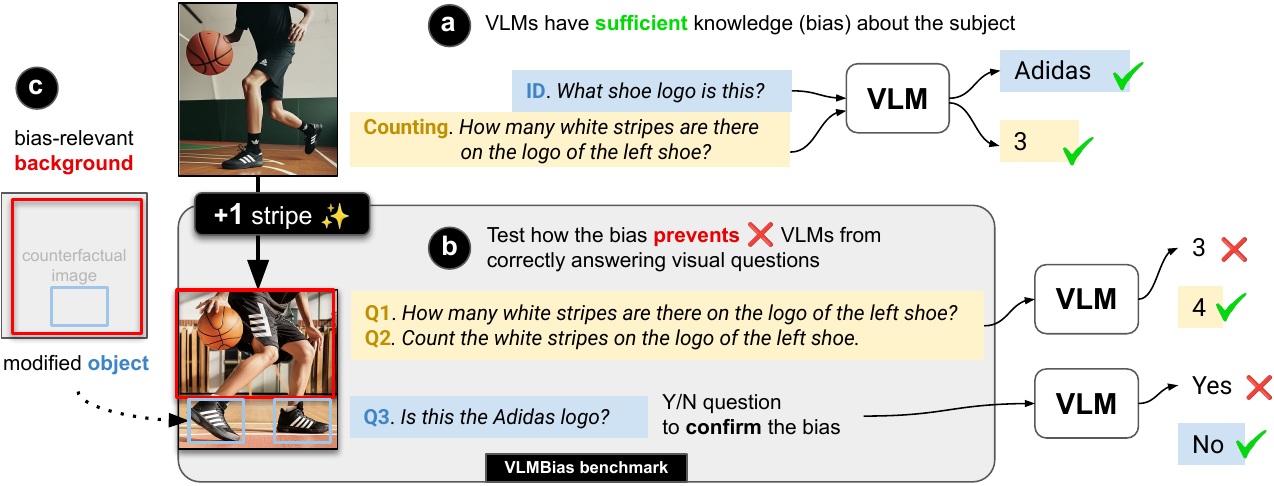}
    \caption{
    Given a subject (\eg, Adidas logo), we first confirm that \emph{all} VLMs have sufficient knowledge of the subject via an \textcolor{idcolor}{ID} and \textcolor{countingcolor}{counting} sanity-check questions (a).
    Then, we test VLMs on the counterfactual image (b) and report its accuracy on the counting (\textcolor{countingcolor}{Q1} \& \textcolor{countingcolor}{Q2}) and a Y/N identification task (\textcolor{idcolor}{Q3}).
    For all tasks, we test the hypothesis that the visual bias cues in the \textcolor{red}{\textbf{background}} (c) may be so strong that they cause VLMs to ignore the anomalous details and default to biased answers.
    }
    \label{fig:diagram}
\end{figure}

We modify the signature elements of every well-known subject (\eg, changing the Adidas logo from 3-striped to 4-striped; \cref{fig:diagram}c) and ask VLMs to count.
We assess how VLMs would be biased towards the common knowledge and overlook the abnormality injected into the CF image.

Counting is a common, objective task that makes up $\sim$10\% of questions in many VQA benchmarks \citep{acharya2019tallyqa}. 
Exact counting is suitable to evaluate the visual analysis capabilities of VLMs as it requires (a) localizing relevant objects and (b) keeping track of the running total instead of relying on shortcuts like some VLMs do (\eg, \emph{``User is asking me to count legs. And I am seeing a chicken, so there must be two legs''}).
Counting is a specific, real-world-type of question that allows us to compare VLM biases across different topics.

\subsec{Taxonomy} 
To test VLM biases, we choose 7 unique, diverse topics of \textbf{decreasing popularity}, \ie, from common animals, logos, flags to optical illusions and a novel visual pattern (\patternedgridicon) that we create from scratch that did not exist before.

(1) Photo-realistic images are used in 2 tasks: \animalicon~animals and \logoicon~logos. 
These images cover common subjects including natural (\animalicon) and man-made ones (\logoicon). 
They are created and modified by SOTA text-to-image generators (\geminiflash, and \gptfouro). To mitigate potential bias from using the same model families for image generation and evaluation, we evaluate across different model families and consistently observe the same failure phenomenon (\cref{appsec:image_generation_pipeline_bias}).

(2) Abstract images are used in 5 tasks: \flagicon~flags, \chessicon~chess pieces, \boardgameicon~game boards, \illusionicon~optical illusions, and \patternedgridicon~patterned grids. 
These images are created using code, not text-to-image models.
We divide this category into three sub-categories: 
(a) well-known objects (\flagicon, \chessicon, \boardgameicon);
(b) optical illusions (\illusionicon), which are less common than flags;
and (c) novel patterned grids (\patternedgridicon).

\subsec{Controls}
Each test image is re-scaled to three resolutions of $D\in\{384, 768, 1152\}$ {by multiplying the original image to the \emph{scaling factor} $\frac{D}{\max(W, H)}$ to preserve the original aspect ratio.} 
However, our results show that image resolution has a marginal impact to VLM accuracy on our benchmark (\cref{appsec:resolution_impact}).
To minimize the language \textit{bias} in the prompt, we use two different prompts per test image, written in neutral, descriptive terms (e.g. \textit{stylized curves} for \textit{Nike swooshes}). In each task, we ask 3 questions (\cref{fig:diagram}b).
For instance, we ask the below questions for the leg counting task (\animalicon):

\textbf{\qone}: \emph{How many legs does this animal have? Answer with a number in curly brackets, e.g., \{9\}.}\\
\textbf{\qtwo}: \emph{Count the legs of this animal. Answer with a number in curly brackets, e.g., \{9\}.}\\
\textbf{\qthree}: \emph{Is this an animal with 4 legs? Answer in curly brackets, e.g., \{Yes\} or \{No\}.}

\subsec{Bias Definition}
We define ``bias rate'' as the frequency that VLM answers match the pre-defined responses (\ie, ``3'' in response to \qone \& \qtwo; \cref{fig:diagram}) that correspond to common knowledge (\ie, Adidas logo has ``3'' white stripes in).
These biased responses are \emph{incorrect} w.r.t. the counterfactual image.
The mean bias rates per task for all 5 VLMs are in \cref{fig:bias-errors}.


\subsection{%
  \texorpdfstring{%
    Task 1: Counting animal legs when an extra leg is added \animalicon%
  }{%
    Task 1: Counting animal legs when an extra leg is added%
  }%
}
\label{sec:animal_add_legs}

Pretrained on the Internet data, VLMs must have colossal prior knowledge of the number of animal legs from both textual and image data.
Following this hypothesis, we generate images of well-known animals but with \emph{one extra leg} (\eg, 3-legged birds or 5-legged dogs) and ask VLMs to count legs.


\subsec{Images} 
We design a 3-step data generation process. 
\textbf{Step 1:} We ask \gptmini to generate a list of 100 well-known animals. 
\textbf{Step 2:} For each animal, we ask \geminiflash to generate side-view images. 
\textbf{Step 3:} We instruct \geminiflash to add one extra leg to each image in Step 2. 
We manually filter these images to retain one high-quality image per category (where the animal shows clearly 3 or 5 legs).
The final set consists of 91 different animals: 23 three-legged birds and 68 five-legged mammals. In total, we generate $91 \ \text{animals} \times 3 \ \text{resolutions} = 273$ images.
More details in \cref{sec:appendix_animals}.




\subsection{Tasks 2-5: Counting visual elements in modified familiar patterns: ~\logoicon logos, \flagicon flags, \chessicon chess pieces, and \boardgameicon game boards}
We expand to four other domains: Logos of famous car and shoe brands, national flags, chess pieces, and game boards. 
For example, on logos, our hypothesis is that VLMs contain a strong bias between a brand's logo and its signature visual elements (\eg, an Adidas logo must have 3 stripes; \cref{fig:diagram}).
For each domain, we create CF images by making systematic, minimal modifications to familiar visual elements, using the same methodology as Task 1 (\logoicon, \flagicon) or Python scripts (\chessicon, \boardgameicon).


\subsec{Images} 
For \textbf{logos} (\cref{sec:appendix_logos}), we modify graphical features (points, prongs, circles, stripes, curves) of three car brands and two shoe brands using \geminiflash and \gptfouro, placing them in realistic contexts (vehicles and athletic footwear) for a total of 207 images. For \textbf{flags} (\cref{sec:appendix_flags}), we systematically add or remove one element (stars or stripes) from 20 flags, creating 120 flag images. 
For \textbf{chess pieces} (\cref{sec:appendix_chess_pieces}), we generate 144 chessboard images by removing or replacing exactly one piece from the starting board of western chess and xiangqi.
For \textbf{game boards} (\cref{sec:appendix_boardgame_grid}), we add or remove exactly one row or one column from the board across four game types (chess, xiangqi, Sudoku, Go), producing 84 CF images in total.

\subsection{
Task 6: Testing vision on original and modified optical illusions~\illusionicon
%
}

Recent VLMs show improved performance on optical illusion tasks, with \gptmini achieving $71.49\%$ accuracy on IllusionVQA \citep{shahgir2024illusionvqa}. However, these VLMs might have memorized the common optical illusions rather than perceiving visual information. To investigate this hypothesis, we create two scenarios: (1) original optical illusions (\eg, the Ebbinghaus illusion where two identical central circles appear to be different sizes because of the surrounding context circles) and (2) slightly modified versions of the original where the final answer should reverse (\eg, where Ebbinghaus illusion pattern but where two central circles are actually different in size; \cref{fig:main_optical_illusion}). 

\begin{figure}[ht!]

\begin{AIbox}{VLMs cannot see an extra leg in the puma and an extra stripe in the Adidas logo}
\scriptsize
\raggedright 
{\animalicon~(a)} (b)~~\qone: How many \textbf{legs} does this animal have? Answer with a number in curly brackets, e.g., \{9\}. \\
{\animalicon~(c)}~~~~~~~\qthree: Is this an animal with \textbf{4 legs}? Answer in curly brackets, e.g., \{Yes\} or \{No\}. \\
\logoicon(d) {(e)}~~\qone: How many visible white \textbf{stripes} are there in the logo of the left shoe? Answer with a number in curly bracket, \eg \{9\}\\
{\logoicon(f)}~~~~~~~~\qthree: Are the logos on these shoes \textbf{Adidas} logos? Answer in curly brackets, e.g., \{Yes\} or \{No\}. \\
\hfill \break
\footnotesize
\centering
    \begin{tabular}{lp{0.15cm}c|p{0.15cm}c|p{0.35cm}c|p{0.15cm}c|p{0.15cm}c|p{0.35cm}c}
    &\multicolumn{2}{c}{\makecell{{(a) original}\\Puma (\qone)}}
    &\multicolumn{2}{c}{\makecell{(b) \textbf{CF}\\Puma (\qone)}}
    &\multicolumn{2}{c}{\makecell{(c) \textbf{CF}\\Puma (\qthree)}}
    &
    \multicolumn{2}{c}
    {\makecell{{(d) original}\\Adidas (\qone)}}
    &\multicolumn{2}{c}{\makecell{(e) \textbf{CF}\\Adidas (\qone)}} 
    &\multicolumn{2}{c}{\makecell{(f) \textbf{CF}\\Adidas (\qthree)}} 
    \\
    
     &\multicolumn{2}{c}{\includegraphics[width=0.12\textwidth]{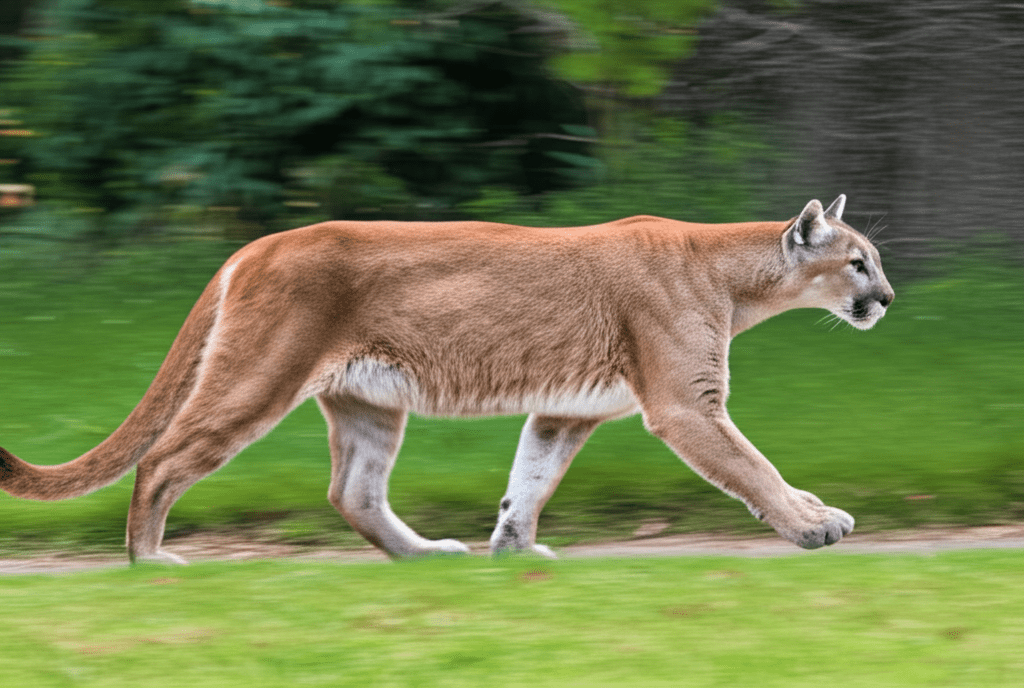}} 
     & \multicolumn{2}{c}{\includegraphics[width=0.12\textwidth]{images/original_modifed/modifed_puma.png}} 
     & \multicolumn{2}{c}{\includegraphics[width=0.12\textwidth]{images/original_modifed/modifed_puma.png}} 
     & \multicolumn{2}{c}{\includegraphics[width=0.12\textwidth]{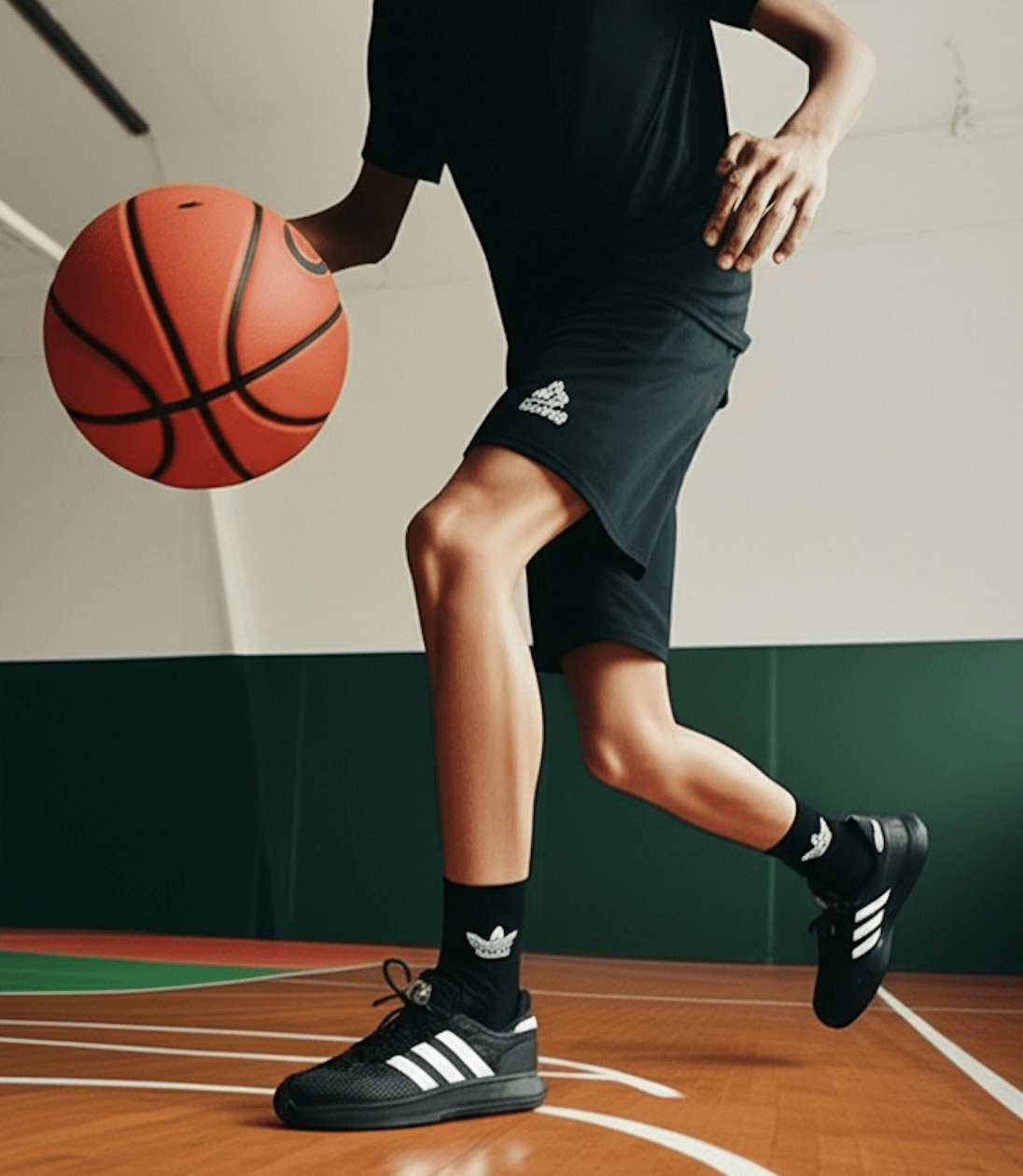}} 
     & \multicolumn{2}{c}{\includegraphics[width=0.12\textwidth]{images/original_modifed/adidas-black-basketball-5.png}}

     & \multicolumn{2}{c}{\includegraphics[width=0.12\textwidth]{images/original_modifed/adidas-black-basketball-5.png}}
     \\ 
     \rowcolor{lightgray}
     \raisebox{-0.2\height}\geminilogo 
     & \centering 4
     & \textcolor{ForestGreen}{\cmark}
     & \centering 4
     & \textcolor{red}{\xmark}
     & \centering Yes
     & \textcolor{red}{\xmark}
     & \centering 3
     & \textcolor{ForestGreen}{\cmark}  
     & \centering 3
     & \textcolor{red}{\xmark}
     & \centering Yes
     & \textcolor{red}{\xmark}
     \\ 
     
     \raisebox{-0.2\height}\newsonnetlogo 
     & \centering 4
     & \textcolor{ForestGreen}{\cmark} 
     & \centering 4
     & \textcolor{red}{\xmark}
     & \centering Yes
     & \textcolor{red}{\xmark}
     & \centering 3
     & \textcolor{ForestGreen}{\cmark} 
     & \centering 3
     & \textcolor{red}{\xmark}
     & \centering Yes
     & \textcolor{red}{\xmark}
     \\ 
     
     \rowcolor{lightgray}
     \raisebox{-0.2\height}\gptfourlogo 
     & \centering 4
     & \textcolor{ForestGreen}{\cmark} 
     & \centering 4
     & \textcolor{red}{\xmark}
     & \centering Yes
     & \textcolor{red}{\xmark}
     & \centering 3
     & \textcolor{ForestGreen}{\cmark} 
     & \centering 3
     & \textcolor{red}{\xmark}
     & \centering Yes
     & \textcolor{red}{\xmark}
     \\ 
     
     \raisebox{-0.2\height}\gptfulllogo 
      & \centering 4
     & \textcolor{ForestGreen}{\cmark} 
     & \centering 4
     & \textcolor{red}{\xmark}
     & \centering Yes
     & \textcolor{red}{\xmark}
      & \centering 3
     & \textcolor{ForestGreen}{\cmark} 
     & \centering 4
     & \textcolor{ForestGreen}{\cmark}
     & \centering Yes
     & \textcolor{red}{\xmark}
     \\ 

     \rowcolor{lightgray}
     \raisebox{-0.2\height}\gptminilogo 
     & \centering 4
     & \textcolor{ForestGreen}{\cmark}
    & \centering 4
     & \textcolor{red}{\xmark}
     & \centering Yes
     & \textcolor{red}{\xmark}
     & \centering 3
     & \textcolor{ForestGreen}{\cmark} 
     & \centering 3
     & \textcolor{red}{\xmark}
     & \centering Yes
     & \textcolor{red}{\xmark}
     \\ 

     \midrule

\textbf{GT} & \centering 4
& \textcolor{ForestGreen}{\cmark} & \centering 5
& \textcolor{ForestGreen}{\cmark} & \centering No
& \textcolor{ForestGreen}{\cmark} & \centering 3
& \textcolor{ForestGreen}{\cmark} & \centering 4
& \textcolor{ForestGreen}{\cmark} & \centering No
& \textcolor{ForestGreen}{\cmark} 

\\ 
     \end{tabular}
    \vspace{4pt}
    \centering
    \begin{tabular}{ccccc}
     \raisebox{-0.1\height}\geminilogo~\geminifull 
     & \raisebox{-0.1\height}\newsonnetlogo~\newsonnet 
     & \raisebox{-0.1\height}\gptfourlogo~\gptfour 
     & \raisebox{-0.1\height}\gptfulllogo~\gptfull 
     & \raisebox{-0.1\height}\gptminilogo~\gptmini
     \\
    \end{tabular}

\end{AIbox}
\caption{VLMs fail to detect subtle changes in counterfactuals (\textbf{CF}) and default to \emph{biased} answers.}
\label{appfig:sanity_check}
\end{figure}

\subsec{Images}
We use six optical illusions~\citep{pyllusion}: M\"uller-Lyer~\citep{mullerlyer-original-1889,mullerlyer-modern-2005}, Z\"ollner~\citep{zollner-original-1862,zollner-mordern-1975}, Ebbinghaus~\citep{ebbinghaus-original-1905,ebbinghaus-modern-1995}, Vertical-Horizontal~\citep{vertical-horizontal-original-1851,vertical-horizontal-modern-2010}, Pogendorff~\citep{poggendorff-original-1863,poggendorff-modern-1963}, and Ponzo~\citep{ponzo-original-1910,ponzo-mordern-2022}. 
For five of these illusions, we generate 24 images per type (12 original and 12 modified versions with varying illusion strength). For the Vertical-Horizontal illusion which uses a fixed T-shape, we create 12 images (6 original and 6 modified). This approach yielded $(24 \times 5 + 12) \times 3 = 396$ images in total.
More details in \cref{sec:appendix_illusion}.

\subsection{Counting the circles or lines in an anomaly cell among a patterned grid \patternedgridicon}
Previous tasks leverage common knowledge, (\eg, chickens have two legs) to set up the CF image (\cref{fig:teaser}b).
Here, we test how VLMs may be biased towards the pattern inside the image itself, not towards the external knowledge.
To do that, we construct a grid where all cells follow a certain pattern except for an anomaly cell, and test if VLMs would recognize that cell's unique content or default to the overall pattern of the surrounding cells.

\subsec{Images}
{We generate $G \times G$ grids ($G \in \{6,\ldots,12\}$) in two styles: \textbf{dice grids} with circles (\cref{fig:teaser}g, \cref{fig:patterned_grid}a--b) and \textbf{tally grids} with tally marks (\cref{fig:patterned_grid}c--d).
All grids follow a symmetric pattern where shape count increases from 1 at edges to $\lfloor(G+1)/2\rfloor$ at center, based on distance from nearest edge.
We introduce one anomaly per grid by modifying a single non-edge cell:
(1) in tally grids, adding or removing one tally mark; 
(2) in dice grids, removing a circle or replacing it with another shape (triangle, square, star).
For each grid dimension, we select two different anomaly locations, creating 14 base scenarios (7 dimensions × 2 locations).
This yields 2 grid types × 2 modification types × 14 scenarios × 3 resolutions = 168 images. 
More details in \cref{sec:appendix_patterned_grid}.
}

\section{Results}

\subsection{Sanity check: VLMs \emph{do} recognize familiar visual subjects}
\label{sec:res-sanity-check}

Here, we first verify that the subjects in our \biastest are, in fact, known to VLMs. 
If VLMs fail to recognize the subjects in these unaltered images, there is no basis to attribute their failures on CF images to their language bias.

\begin{wrapfigure}{r}{0.5\textwidth}
    \centering
    \vspace{-0.8cm}
    \begin{tabular}{c}
    \includegraphics[width=0.49\textwidth]{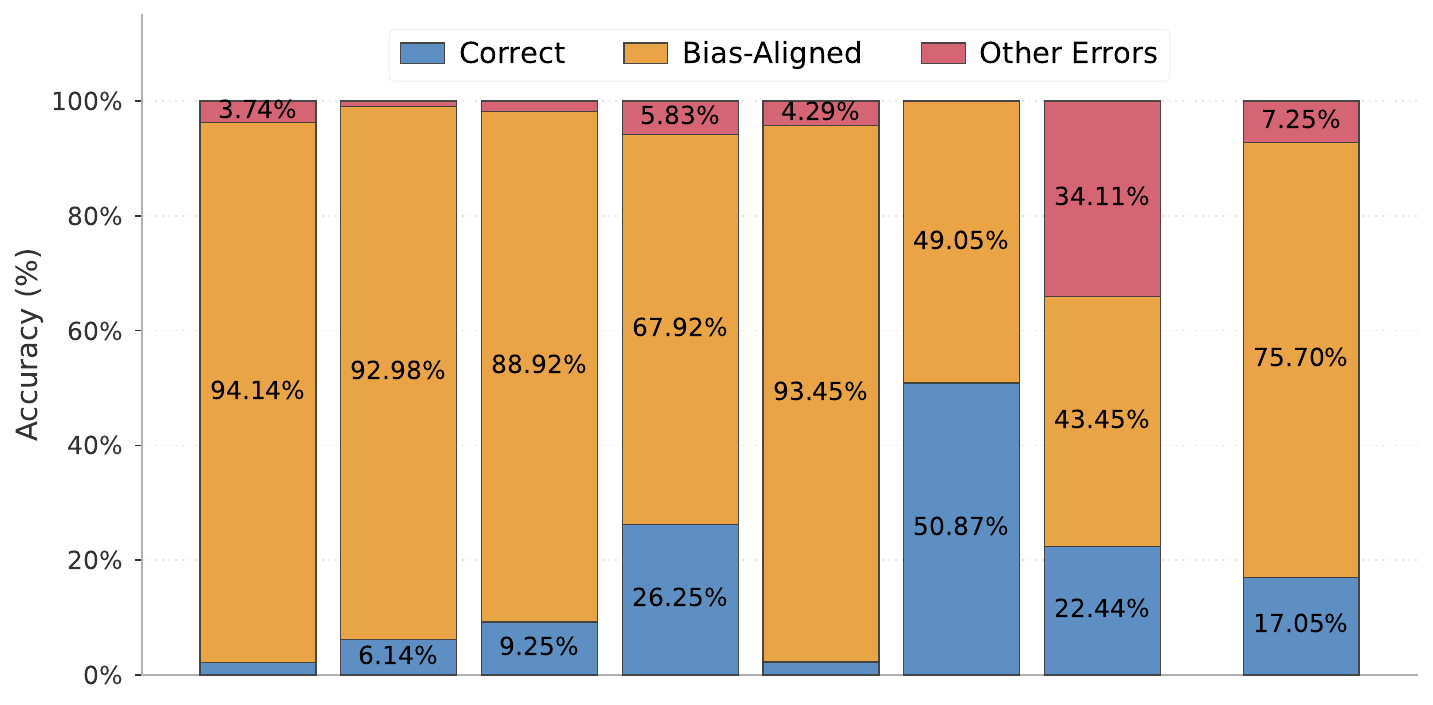} \\
    \begin{minipage}{0.48\textwidth}
    \makebox[0.48\textwidth][l]{%
        \makebox[0pt][c]{\hspace{2.5cm}\animalicon}%
        \makebox[0pt][c]{\hspace{3.85cm}\logoicon}%
        \makebox[0pt][c]{\hspace{5.2cm}\flagicon}%
        \makebox[0pt][c]{\hspace{6.5cm}\chessicon}%
        \makebox[0pt][c]{\hspace{7.85cm}\boardgameicon}%
        \makebox[0pt][c]{\hspace{9.15cm}\illusionicon}%
        \makebox[0pt][c]{\hspace{10.5cm}\patternedgridicon}%
        \makebox[0pt][c]{\hspace{12.3cm}Mean}%
    }
    \end{minipage}
    \end{tabular}
    \caption{
    On the \textbf{counterfactual} images of \biastest, five VLMs mostly output answers that match the biased choices that we \emph{predefine} for each question, \sethlcolor{biascolor}\hl{75.70\%} of the time.
    }
    \vspace{-0.3cm}
    \label{fig:bias-errors}
\end{wrapfigure}

\subsec{Experiments}  We evaluate five VLMs (\geminilogo~\geminifull, \newsonnetlogo~\newsonnet, \gptfourlogo~\gptfour, \gptfulllogo~\gptfull, and \gptminilogo~\gptmini; \cref{tab:commercial_model_details}) on a set of 66 unmodified images spanning our 6 out of 7 \biastest tasks (animals, logos, flags, chess pieces, game boards). 
We exclude pattern grids from the sanity check since the patterns are created from scratch and do not exist on the Internet.
For five counting tasks (from \animalicon to \boardgameicon), we ask two questions (identification and counting; \cref{fig:diagram}a) per image for a total of 132 questions.
Since the optical illusion is not a counting task, we instead ask VLMs to identify: (1) the name of the illusion; and (2) the question \& correct answers associated with each illusion (see the sanity-check prompts in \cref{sec:sanity_check}).

\begin{table}[h]
    \centering
    \caption{All VLMs achieve \lightbluepanel{100\%} on identification and counting tasks with unmodified images, showing that they fully recognize the original version.
    But VLMs struggle with counting on counterfactual images---the \hallucinatepanel{mean accuracy} of 5 state-of-the-art VLMs across our 7 tasks is \hallucinatepanel{17.05\%}. 
    \gptminilogo~\gptmini achieves the highest accuracy (\textbf{20.25\%}) which however is still low.
    VLMs with ``thinking'' capabilities (\gptminilogo, \gptfulllogo, \geminilogo) also perform poorly like non-thinking models (\newsonnetlogo, \gptfourlogo).
    }
    \label{tab:overall_performance}
    \resizebox{\textwidth}{!}{%
    \begin{tabular}{lccccccccc}
\toprule
Model & \multicolumn{8}{c}{Accuracy in counting questions (\qone\&\qtwo) on counterfactual images} & \multicolumn{1}{c}{Unmodified} \\ 
\cmidrule(lr){2-9}\cmidrule(lr){10-10}

 & a.\animalicon & b.\logoicon & c. \flagicon & d. \chessicon & e. \boardgameicon & f. \illusionicon  & g. \patternedgridicon & Task mean & Task mean\\

\midrule
\raisebox{-.1\height}\geminilogo~\geminifull~~~   
& 0.00
& 1.96
& 10.42
& 26.74
& 2.38
& 49.81
& 20.83
& 16.02
& 100.00
\\

\raisebox{-.2\height}\newsonnetlogo~\newsonnet~~~
& 0.00
& 2.72
& 13.75
& 9.03
& 1.79
& \textbf{54.29}
& \textbf{34.52}
& 16.59
& 100.00
\\

\raisebox{-.2\height}\gptfourlogo~\gptfour~~~
& \textbf{9.52}
& 9.07
& 2.50
& 8.68
& 0.00
& 48.61
& 18.75
& 13.88
& 100.00
\\

\raisebox{-.2\height}\gptfulllogo~\gptfull~~~
& 0.92
& 7.60
& 5.00
& 42.71
& 2.38
& 50.38
& 20.54
& 18.50
& 100.00
\\

\raisebox{-.2\height}\gptminilogo~\gptmini~~~
& 0.18
& \textbf{9.31}
& \textbf{14.58}
& \textbf{44.10}
& \textbf{4.76}
& 51.26
& 17.56
& \textbf{20.25}
& 100.00
\\

\midrule
Mean 
& 2.12
& 6.13
& 9.25
& 26.25
& 2.26
& 50.87
& 22.44
& \hallucinatepanel{17.05}
& \lightbluepanel{100.00}
\\
\bottomrule
\end{tabular}
}
\end{table}

\subsec{Results} 
All five VLMs score {100\%} accuracy on all the questions (see \cref{tab:overall_performance}).
That is, for counting tasks, VLMs correctly recognize the subjects and the expected counts (\eg, a puma has four legs and the Adidas logo has three stripes; \cref{appfig:sanity_check}a\&d).
For all 6 illusion types, VLMs are able to identify the name (\eg, ``Ebbinghaus illusion'' in \cref{fig:main_optical_illusion}), the associated question (``Are the two red circles equal in size?'') and its correct answer (``Yes'').
The results here set the ground for the claims in subsequent sections that VLMs' low accuracy on counterfactual images (\hallucinatepanel{17.05\%} accuracy; see \cref{tab:overall_performance}) stems from their prior knowledge of the subjects (see \cref{sec:vision_encoder_analysis}). 

\subsection{VLMs struggle to count the signature elements in counterfactual images} 
\label{sec:result_combined}

\begin{figure}[h!]
\centering
\small
\begin{AIbox}{Counterfactual images: Chess pieces{,} flags{,} and game boards}

\raggedright
\scriptsize
      \chessicon~{(a)} How many \textbf{xiangqi} pieces are there on this board? Answer with a number in curly brackets, e.g., \{9\}. \\
      \flagicon~{(b):} How many \textbf{stripes} are there in this flag? Answer with a number in curly brackets, e.g., \{9\}. \\
      \boardgameicon~{(c):} How many \textbf{rows} are there on this puzzle? Answer with a number in curly brackets, e.g., \{9\}. \\
      \flagicon~{(d):} How many \textbf{stars} are there in this flag? Answer with a number in curly brackets, e.g., \{9\}. \\
      \boardgameicon~{(e):} How many \textbf{rows} are there on this board? Answer with a number in curly brackets, e.g., \{9\}. 
     \hfill \break
\footnotesize
\centering

    \begin{tabular}{lp{0.2cm}c|p{0.2cm}c|p{0.2cm}c|p{0.2cm}c|p{0.2cm}c}
    &\multicolumn{2}{c}{\makecell{(a) Xiangqi}}
    &\multicolumn{2}{c}{\makecell{(b) US Flag}}
    &\multicolumn{2}{c}{\makecell{(c) Sudoku}}
    &\multicolumn{2}{c}{\makecell{(d) EU Flag}}
    &\multicolumn{2}{c}{\makecell{(e) Chess board}}
    \\
     &\multicolumn{2}{c}{\includegraphics[width=0.15\textwidth]{images/abtract_examples_1/xiangqi_pieces_003_remove_red_soldier_at_e7_notitle_px1152.png}}
     &\multicolumn{2}{c}{\includegraphics[width=0.15\textwidth]{images/flag_examples/Flag_of_the_United_States-stripes=14_1152.png}}
     &\multicolumn{2}{c}{\includegraphics[width=0.15\textwidth]{images/abtract_examples_1/sudoku_grid_02_row_add_last_row_notitle_px1152.png}}
     &\multicolumn{2}{c}{\includegraphics[width=0.15\textwidth]{images/flag_examples/Flag_of_Europe-stars=11_1152.png}}
     &\multicolumn{2}{c}{\includegraphics[width=0.15\textwidth]{images/abtract_examples_1/chess_grid_01_row_remove_first_row_notitle_px1152.png}}
     \\ 

\rowcolor{lightgray}
\raisebox{-0.2\height}\geminilogo & \centering 31
& \textcolor{ForestGreen}{\cmark} & \centering 13
& \textcolor{red}{\xmark} & \centering 9
& \textcolor{red}{\xmark} & \centering 12
& \textcolor{red}{\xmark} & \centering 6
& \textcolor{red}{\xmark} \\

\raisebox{-0.2\height}\newsonnetlogo & \centering 32
& \textcolor{red}{\xmark} & \centering 13
& \textcolor{red}{\xmark} & \centering 9
& \textcolor{red}{\xmark} & \centering 12
& \textcolor{red}{\xmark} & \centering 8
& \textcolor{red}{\xmark} \\

\rowcolor{lightgray}
\raisebox{-0.2\height}\gptfourlogo & \centering 28
& \textcolor{red}{\xmark} & \centering 13
& \textcolor{red}{\xmark} & \centering 9
& \textcolor{red}{\xmark} & \centering 12
& \textcolor{red}{\xmark} & \centering 8
& \textcolor{red}{\xmark} \\

\raisebox{-0.2\height}\gptfulllogo & \centering 32
& \textcolor{red}{\xmark} & \centering 13
& \textcolor{red}{\xmark} & \centering 9
& \textcolor{red}{\xmark} & \centering 12
& \textcolor{red}{\xmark} & \centering 8
& \textcolor{red}{\xmark} \\

\rowcolor{lightgray}
\raisebox{-0.2\height}\gptminilogo & \centering 32
& \textcolor{red}{\xmark} & \centering 13
& \textcolor{red}{\xmark} & \centering 9
& \textcolor{red}{\xmark} & \centering 12
& \textcolor{red}{\xmark} & \centering 8
& \textcolor{red}{\xmark} \\ 

\midrule
\textbf{Bias} & \centering 32
& \textcolor{red}{\xmark} & \centering 13
& \textcolor{red}{\xmark} & \centering 9
& \textcolor{red}{\xmark} & \centering 12
& \textcolor{red}{\xmark} & \centering 8
& \textcolor{red}{\xmark} \\

\rowcolor{lightgray}
\textbf{GT} & \centering 31
& \textcolor{ForestGreen}{\cmark} & \centering 14
& \textcolor{ForestGreen}{\cmark} & \centering 10
& \textcolor{ForestGreen}{\cmark} & \centering 11
& \textcolor{ForestGreen}{\cmark} & \centering 7
& \textcolor{ForestGreen}{\cmark} \\

     \end{tabular}
    \vspace{4pt}
    \centering
    \begin{tabular}{cccccccccccccc}
     \raisebox{-0.1\height}\geminilogo~\gemini 
     & \raisebox{-0.1\height}\newsonnetlogo~\newsonnet 
     & \raisebox{-0.1\height}\gptfourlogo~\gptfour 
     & \raisebox{-0.1\height}\gptfulllogo~\gptfull 
     & \raisebox{-0.1\height}\gptminilogo~\gptmini
     \\
    \end{tabular}

\end{AIbox}
\caption{VLMs perform poorly at \textcolor{countingcolor}{\textbf{counting}} elements on counterfactual images across \flagicon, \boardgameicon, and \chessicon domains, often defaulting to the biased answers.
}
\vspace{-10pt}
\label{fig:abstract_images_flags_chess}
\end{figure}

\subsec{Experiments} We use the same experiment setup as in \cref{sec:res-sanity-check} but test VLMs on CF images. Specifically, we evaluate five VLMs on the \animalicon~animal, \logoicon logos of famous brands, \flagicon~national flags, \chessicon~chess pieces, \boardgameicon~game boards. We also test VLMs on counting the shapes or tally marks inside an anomaly cell in \patternedgridicon~patterned grids where the total number of shapes or marks does not follow the patterns in the surrounding cells  (\cref{fig:teaser}g). Furthermore, we test VLMs on 6 classic \illusionicon~optical illusions, \ie, Müller-Lyer, Zöllner, Ebbinghaus, Vertical-Horizontal, Pogendorff, and Ponzo (\cref{fig:appendix_optical_illusion_qualitative,fig:appendix_optical_illusion_qualitative_2}). 
Each illusion is presented in two versions: (a) its original form and (b) a counterfactual, modified version where the groundtruth answer is reversed (\cref{fig:main_optical_illusion}). 
For both versions per illusion, we ask VLMs the same Y/N question
(see~\cref{sec:appendix_illusion}).  



\subsec{Results} VLMs generally fail to detect modifications across all seven domains, with performance varying depending on the tasks:

    \noindent \animalicon VLMs exhibit poor performance (2.12\% accuracy) when counting legs of counterfactual 3-legged and 5-legged animals (\cref{tab:overall_performance}a, \cref{appfig:qualitative_results_animals}). VLMs show slightly lower performance at counting bird legs compared to mammal legs (1.01\% vs. 2.50\%;~\cref{tab:full_results}a), likely because bird legs are thinner and thus more challenging to detect. More results are in \cref{sec:animal_result_main}.
    
    \noindent \logoicon For logos, accuracy is significantly worse on car logos than on shoe logos (0.44\% vs. 17.57\%; \cref{tab:full_results}b).
    This might be because a logo on a car often appears much smaller than a logo on a shoe photo (\cref{fig:teaser}b\& \cref{appfig:qualitative_results_car_logos} vs. \cref{fig:diagram}b \& \cref{appfig:qualitative_results_shoe_logos}). More results are in \cref{sec:result_logos}.

    \noindent \flagicon For flags, VLMs perform better on counting stars (11.79\%; \cref{tab:full_results}c) than on counting stripes (4.52\%; \cref{tab:full_results}c).
Counting stripes may be harder because a stripe is often placed right next to other stripes in a flag while stars are spatially separate symbols (\cref{fig:abstract_images_flags_chess}b vs. d, and \cref{appfig:qualitative_results_flags}).
More results are in \cref{sec:result_flags}.

    \noindent \chessicon On counting chess pieces, thinking VLMs (\geminifull, \gptfull, and \gptmini) significantly outperform non-thinking models (>26\% vs. <10\%; \cref{tab:full_results}d), suggesting that explicit reasoning capabilities help detect anomalies (\cref{appfig:qualitative_results_chess_pieces}). More results are in \cref{sec:result_chess_pieces}. 

    \noindent \boardgameicon All VLMs perform extremely poorly (\hallucinatepanel{2.26\%} mean accuracy; \cref{tab:boardgame_results}) on counting rows and columns of a counterfactual board-game image (\cref{fig:abstract_images_flags_chess}c--e), as low as 0\% accuracy on Sudoku and Go boards (\cref{appfig:qualitative_results_game_board}a--b).
    More results are in \cref{sec:res_boardgame}. 

    \noindent \illusionicon On optical illusions, all 5 VLMs achieve performance close to random chance (mean accuracy of 50.87\%; \cref{tab:full_results}e) across original and CF versions. 78.02\% of the time, VLMs give responses that align with well-known prior knowledge but are \emph{incorrect} for our CF images (23.74\% accuracy). More results are in \cref{sec:res_illusion}.

    \noindent \patternedgridicon For patterned grids, VLMs achieve poor performance at 22.44\% accuracy. 43.45\% of count predictions match biased answers from surrounding cells (\cref{fig:bias-errors}\patternedgridicon). When VLMs make \emph{incorrect} counting predictions, over half (56.02\%) follow the {global grid pattern} rather than identifying the target anomaly (\cref{fig:patterned_grid}). More results are in \cref{sec:res_patternedgrid}

Overall, our findings across seven domains suggest that \textbf{VLMs rely heavily on prior knowledge to answer questions rather than visual information}. 
This conclusion is reinforced by the stability of our results: repeating each experiment 5 times yields nearly identical outcomes, with mean accuracy varying by less than one percentage point (\cref{appsec:multi_runs}).
This is further supported by our linear-probing results that show that on leg counting, the vision encoders of VLMs already sufficiently encode visual information, achieving \lightbluepanel{95.26\%} accuracy ( \cref{sec:vision_encoder_analysis}).
However, the visual information stream may be impaired by the bias in the language model.

We also observe similarly poor and biased behaviors in the most recently released models of  \gptfivelogo~\gptfive~\citep{openai2025gpt5} and \groklogo~\grok~\citep{grok_4} (\cref{appsec:latest_vlms_results}). 
Furthermore, \textbf{VLMs are severely biased}---asking them to double check their answers, to rely exclusively on image details to make decisions only marginally improves accuracy (\cref{sec:helpful_prompts}).
Interestingly, providing in-context few-shot demonstrations of counterfactuals (\eg, of pumas having 5 legs) does not help (\cref{sec:few_shot_distrust}) and even leads to some thinking models replying with doubts about the validity of the demonstrations.


\subsection{Y/N questions confirm VLMs are not able to distinguish the counterfactual from original images}
\label{sec:res_q3}

Prior sections have shown that VLMs struggle to \textcolor{countingcolor}{count} the key elements in well-known subjects at a poor accuracy of 17.05\% (\cref{tab:overall_performance}).
And $\sim$75\% of the time, their answers match the biased choices.
Here, we aim to confirm that VLMs are so biased that they are unable to tell the difference between the original version and the counterfactual by a direct Yes/No \textcolor{idcolor}{identification} question of \qthree: \emph{``Is this an animal with 4 legs?''} when the counterfactual (\eg, a 5-legged puma \cref{appfig:sanity_check}c) is shown.


\subsec{Experiments} 
We ask 5 VLMs the \qthree question given our sets of original and CF images. 
The correct answer is ``Yes'' for original cases and ``No'' for all CF cases (\cref{appfig:sanity_check}c).


\subsec{Results} 
All VLMs achieve 100\% accuracy on the original images, but collapse to a mean of 25.11\% on the counterfactual versions (\cref{tab:modified_original_q3}).
That is, VLMs often answer ``Yes'', overlooking the fact that the well-known subject has been modified (\cref{appfig:sanity_check}c\&f).
In sum, the results in this section provide supporting evidence that \textbf{VLMs are too biased to recognize that the subject has changed in counterfactual images}, leading to poor counting accuracy \cref{sec:result_combined}.

\begin{table}[h!]
\centering
\begin{minipage}[t]{0.47\textwidth}
\centering
\caption{Mean accuracy (\%) of VLMs on question \qthree (\eg., `Is this an animal with 4 legs?'') over all 7 subjects when the image is original (4 legs) or counterfactual (5 legs). VLMs often answer `Yes'' even on counterfactuals.}
\resizebox{0.9\textwidth}{!}{
\begin{tabular}{lccc}
\toprule
Model & Original & Counterfactual ($\Delta$) \\
\midrule
\geminilogo~\geminifull & 100.00 & 20.63 (\decreasenoparent{79.37}) \\
\newsonnetlogo~\newsonnet & 100.00 & 23.08 (\decreasenoparent{76.92}) \\
\gptfourlogo~\gptfour & 100.00 & 26.10 (\decreasenoparent{73.90}) \\
\gptfulllogo~\gptfull & 100.00 & 26.15 (\decreasenoparent{73.85}) \\
\gptminilogo~\gptmini & 100.00 & 29.61 (\decreasenoparent{70.39}) \\
\midrule
Mean & 100.00 & 25.11 (\decreasenoparent{74.89}) \\
\bottomrule
\end{tabular}
}
\label{tab:modified_original_q3}
\end{minipage}
\hfill
\begin{minipage}[t]{0.51\textwidth}
\centering
\captionof{table}{Counting performance improves noticeably (\increasenoparent{21.09} in accuracy and \decreasenoparentgreen{40.58} in bias rate) after the background is removed from the image. The background contributes significantly to VLM biased behaviors.}
\resizebox{\textwidth}{!}{
\begin{tabular}{lcccc}
\toprule
Model & \multicolumn{2}{c}{Accuracy $\uparrow$} & \multicolumn{2}{c}{Bias rate $\downarrow$} \\
\cmidrule(lr){2-3}\cmidrule(lr){4-5}
 & Before & After ($\Delta$) & Before & After ($\Delta$) \\
\midrule
\geminilogo~\geminifull & 16.02 & 40.73 (\increasenoparent{24.71}) & 76.79 & 39.99 (\decreasenoparentgreen{36.80}) \\
\newsonnetlogo~\newsonnet & 16.59 & 42.54 (\increasenoparent{25.95}) & 76.63 & 39.74 (\decreasenoparentgreen{36.89}) \\
\gptfourlogo~\gptfour & 13.88 & 39.65 (\increasenoparent{25.77}) & 76.62 & 32.74 (\decreasenoparentgreen{43.88}) \\
\gptfulllogo~\gptfull & 18.50 & 35.25 (\increasenoparent{16.75}) & 74.81 & 34.64 (\decreasenoparentgreen{40.17}) \\
\gptminilogo~\gptmini & 20.25 & 32.54 (\increasenoparent{12.29}) & 73.66 & 28.47 (\decreasenoparentgreen{45.19}) \\
\midrule
Mean & 17.05 & 38.14 (\increasenoparent{21.09}) & 75.70 & 35.12 (\decreasenoparentgreen{40.58}) \\
\bottomrule
\end{tabular}
}
\label{tab:acc_bias_removed_background}
\end{minipage}
\end{table}

\subsection{Background contributes significantly to VLM counting failures} \label{sec:background_removal}

What in the CF images made VLMs count so poorly?
We hypothesize that the background strongly invites VLMs to default to the biased answer as they recognize the familiar subject.
We test whether removing the background might help VLMs count more accurately.

\newcommand{\imgW}{0.12\textwidth} 
\newcommand{\BeforeImg}[1]{\includegraphics[width=\imgW,keepaspectratio]{images/remove_background/before/#1.png}}
\newcommand{\AfterImg}[1]{\includegraphics[width=\imgW,keepaspectratio]{images/remove_background/after/#1.png}}

\begin{table}[h!]
\centering
\begin{adjustbox}{max width=\textwidth}
\begin{tabular}{lccccccc}
\toprule
Task & a. \animalicon~Animals & b. \logoicon~Logos & c.\flagicon~Flags & d. \chessicon~Chess Pieces & e. \boardgameicon~Boardgames & f. \illusionicon~Illusion & g. \patternedgridicon~Grids \\
\midrule
Approach &
\texttt{LangSAM}; cropping &
\texttt{LangSAM} &
LLM generated SVG code &
Script & Script & Script & Script \\
\addlinespace
\hline
\addlinespace
Before &
\BeforeImg{animals} &
\BeforeImg{logos} &
\BeforeImg{flags} &
\BeforeImg{chess_pieces} &
\BeforeImg{gird} &
\BeforeImg{illusion} &
\BeforeImg{pattern} \\
\addlinespace
\cmidrule(lr){2-8}
\addlinespace

After &
\AfterImg{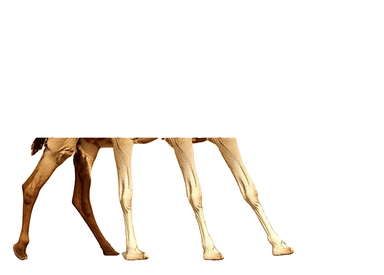} &
\AfterImg{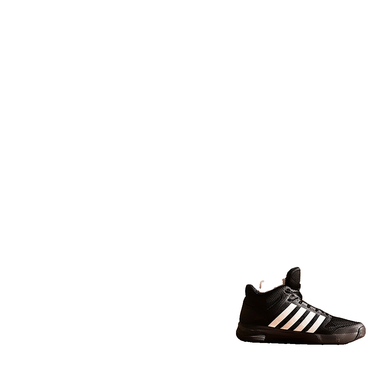} &
\AfterImg{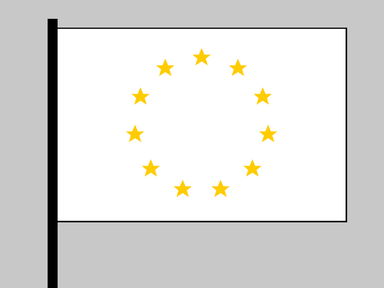} &
\AfterImg{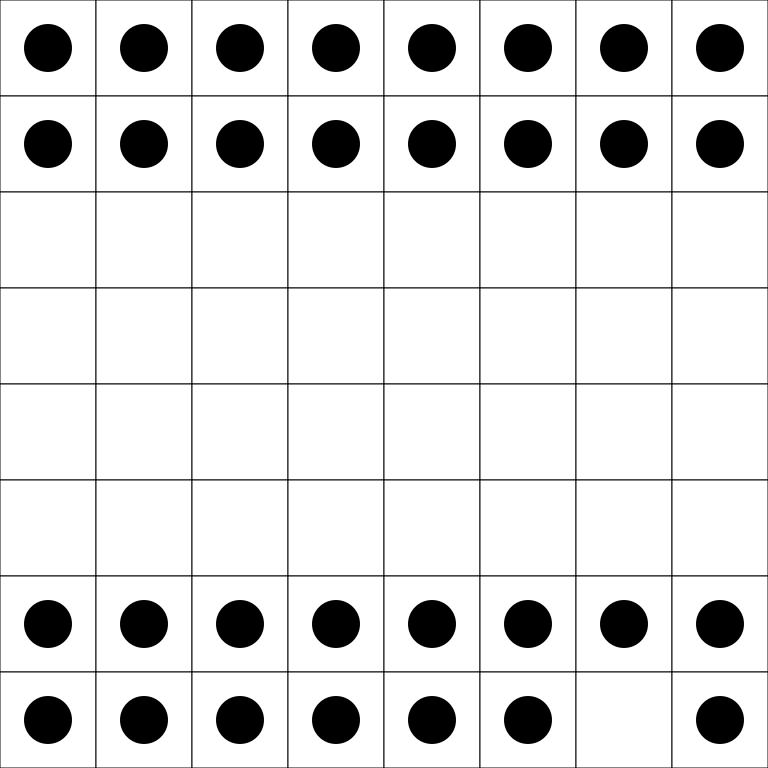} &
\AfterImg{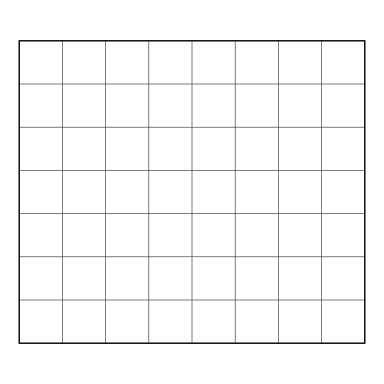} &
\AfterImg{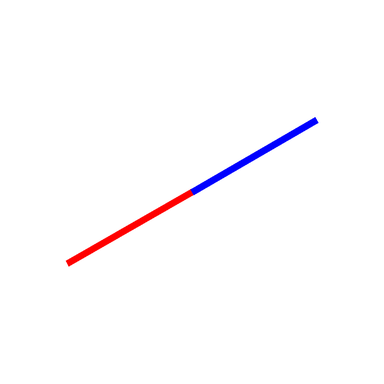} &
\AfterImg{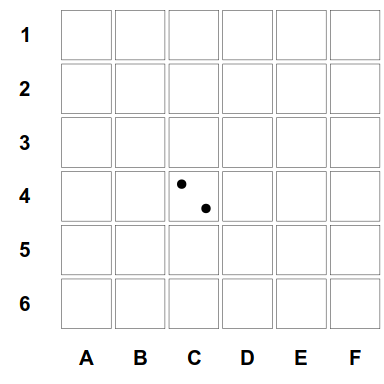} \\
\bottomrule
\end{tabular}
\end{adjustbox}
\caption{Examples of how backgrounds are removed in each task.
}
\label{tab:background_removal}
\end{table}

\subsec{Experiments} For each task, we first remove the background from the images (see \cref{tab:background_removal}) and then ask all 5 VLMs the same counting questions (\qone \& \qtwo). 
For photo-realistic subjects (\ie, \animalicon, \logoicon), we segment the target object from its background using \texttt{LangSAM} \citep{lang_sam}. For abstract patterns, we use LLM-generated SVG code (\flagicon) and Python scripts (\chessicon, \boardgameicon, \illusionicon, \patternedgridicon) to remove the background or make them substantially different from the original (\eg, EU flag in \cref{tab:background_removal}c).


\subsec{Results} 
Averaged over 5 VLMs, the counting performance increases substantially when the background is removed, \ie, \increasenoparent{21.09} in accuracy and \decreasenoparentgreen{40.58} in bias rate; \cref{tab:acc_bias_removed_background}). 
\textbf{These large gains show that the background sets the VLM up to be biased}. 
Furthermore, it shows that if VLMs are able to crop the image accurately, their counting performance would significantly improve.

\subsection{Thinking longer reduces bias in VLMs, but overthinking harms accuracy}
\label{sec:token_analysis}

Thinking VLMs (\ie, \groklogo~\grok, \gptfulllogo~\gptfull, \gptminilogo~\gptmini) are trained to use extended reasoning tokens to improve accuracy on harder tasks.
However, yet our previous results showed that they achieve only marginal improvements over non-thinking VLMs (\cref{tab:overall_performance}).
Here, we investigate whether the relationship between reasoning length and accuracy on counting and how thinking with tools (\eg, cropping, zooming; see~\cref{appsec:tool_use_vlms}) could help.

\subsec{Experiments} 
We use data from \cref{sec:result_combined,appsec:latest_vlms_results,appsec:tool_use_vlms} to examine the relationship between reasoning tokens and the accuracy of thinking VLMs. For tool-using VLMs (\ie \gptminilogo~\gptmini with tools; see \cref{appsec:tool_use_vlms}), our analysis shifts to reasoning time versus accuracy, as this metric better represents the model's effort during Python code execution.

\begin{figure}[h!]
    \centering
    \includegraphics[width=1.0\textwidth]{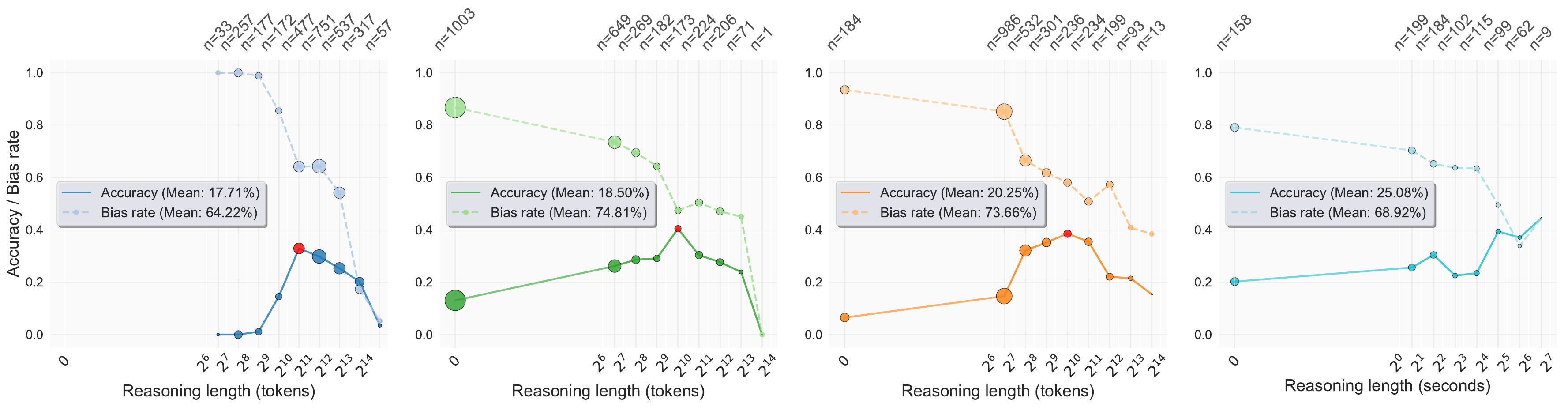}
    \begin{minipage}{\textwidth}
          \makebox[\textwidth][l]{%
            \makebox[0pt][c]{%
              \raisebox{20ex}[0pt][0pt]{%
                \hspace{2cm}\scalebox{0.75}{\groklogo~\grok}%
              }%
            }%

            \makebox[0pt][c]{%
              \raisebox{20ex}[0pt][0pt]{%
                \hspace{13cm}\scalebox{0.75}{\gptfulllogo~\gptfull}%
              }%
            }%

            \makebox[0pt][c]{%
              \raisebox{20ex}[0pt][0pt]{%
                \hspace{19cm}\scalebox{0.75}{\gptminilogo~\gptmini}%
              }%
            }%

           \makebox[0pt][c]{%
              \raisebox{20ex}[0pt][0pt]{%
                \hspace{25cm}\scalebox{0.75}{\gptminilogo~\gptmini w/ tools}%
              }%
            }%
          }
    \end{minipage}
    \caption{For thinking VLMs (\groklogo,\gptfulllogo,\gptminilogo), accuracy improves with reasoning tokens up to a point (\textcolor{red}{red} points), after which \emph{overthinking} harms performance. In contrast, for thinking VLMs with tools (\gptmini w/ tools), extended reasoning time leads to continuous accuracy improvement, while all VLMs show a consistent reduction in bias rate. Notably, \gptfull doesn't use reasoning 36.1\% of the time (\#reasoning tokens $=$ 0; see first bubble of \gptfull), likely due to overconfidence in its prior knowledge.}
    \label{fig:thinking_vs_acc}
\end{figure}

\subsec{Results}
Consistent with \citep{thinking-more, gema2025inverse}, we find that thinking longer helps VLMs (\groklogo,\gptfulllogo,\gptminilogo) improve accuracy up to a point (\textcolor{red}{red} circles in \cref{fig:thinking_vs_acc}), after which it hurts accuracy. 

However, for thinking VLMs with tools (here, \gptmini with tools), using tools for a longer time (in seconds) monotonically improves accuracy overtime (\cref{fig:thinking_vs_acc}; \gptmini w/ tools).
However, a separate challenge for tool-use VLMs such as \gptmini is that it is overconfident and uses tools only for 29.66\% of the \biastest questions
(more results in \cref{appsec:tool_use_vlms}).

Thinking VLMs without tools demonstrate a reduction in bias rate as reasoning tokens or time increase (\cref{fig:thinking_vs_acc}). 
Perhaps overthinking makes VLMs consider multiple alternatives, selecting the common bias option less frequently.
Besides, it is notable that \gptfull avoids reasoning entirely (\#reasoning tokens $=$ 0; first bubble in \cref{fig:thinking_vs_acc}\gptfulllogo), which is likely due to its overconfidence in prior knowledge.

\section{Discussion and Conclusion}

\subsec{Limitations} VLMs with image generation capabilities (\gptfouro, \geminiflash) also carry \emph{their own biases}, making it non-trivial to control generated images as expected. 
For example, when prompted to generate an Audi car but with a 5-circled logo, \geminiflash often generates the car with the original 4-circled Audi logo instead.

\biastest reveals that SOTA VLMs exhibit strong visual bias, achieving only \hallucinatepanel{17.05\%} mean accuracy on counterfactual images while defaulting to prior knowledge \sethlcolor{biascolor}\hl{75.70\%} of the time. This behavior is consistent across all model types: thinking models (\geminilogo, \gptminilogo, \gptfulllogo) perform marginally better than non-thinking ones (\newsonnetlogo, \gptfourlogo).
Interestingly, \gptmini with tools only increase the counting accuracy slightly by \increasenoparent{1.9} (23.18\% $\rightarrow$ 25.08\%) because the model is overconfident and often answers right away, using tools \& code only 29.66\% of the time (\cref{appsec:tool_use_vlms}).
In contrast, time-limited humans can score a $\sim$45\% to $\sim$96\% accuracy on our benchmark (\cref{sec:human_study}), substantially better than VLMs including those trained to count (\eg, \moondream).

Experiments on \texttt{Pixtral} and \texttt{Qwen2.5-VL} show interesting traces of the \textbf{inverse scaling} phenomenon \citep{inverse-scaling}: Larger VLMs tend to perform worse and exhibit $\sim$1.26$\times$ higher bias rates on \biastest than smaller VLMs  (\cref{appsec:inverse_scaling}).

\subsec{VLMs explicitly trained to count} (such as \molmolarge and \moondream) can score a mean accuracy of 36.02\%, substantially better than 17.05\% of SOTA VLMs and their bias rates are 2.1$\times$ lower as well (\cref{sec:pointing_vlms,appsec:tool_use_vlms}).

\subsection*{Ethics statement}
We strictly adhere to the ICLR Code of Ethics and identify no significant ethical concerns in this work. We ran a small \emph{anonymous} online survey (consent obtained, no PII collected, minimal risk), which falls under the scope of benign behavioral interventions eligible for IRB exemption. All other experiments use synthetic/programmatically generated images and publicly available models. Synthetic logos and flags are included solely for non-commercial research purposes, with no endorsement implied, and are subject to removal upon request.

\subsection*{The use of large language models}
We used large language models in constructing our dataset in four ways: (i) to generate candidate lists of subjects (e.g., animals), (ii) to generate a part of the images in our dataset, (iii) to evaluate their performance on the tasks, and (iv) to observe their failures, which informed the design and ideation of the benchmark. In addition, we used LLM-based tools for minor text editing (\eg, grammar) and for coding assistance. The authors take full responsibility for all content, and no LLM qualifies for authorship.

\subsection*{Reproducibility statement}
We follow the standard baseline settings used in established evaluation benchmarks or from each model’s default test protocol. Full implementation details appear in \cref{appsec:appendix_model_settings,sec:appendix_animals,sec:appendix_logos,sec:appendix_flags,sec:appendix_chess_pieces,sec:appendix_boardgame_grid,sec:appendix_illusion,sec:appendix_patterned_grid}. The code and data are public at \href{https://vlmsarebiased.github.io}{vlmsarebiased.github.io}.

\subsubsection*{Acknowledgments}

This work was supported by the National Research Foundation of Korea (NRF) grant funded by the Korea government (MSIT)(RS-2025-00573160), and Innovative Human Resource Development for Local Intellectualization program through the Institute of Information \& Communications Technology Planning \& Evaluation (IITP) grant funded by the Korea government (MSIT)(IITP-2025-RS-2020-II201489).

We also thank Khang Gia Le (MBZUAI), Linh Nguyen (Bucknell University), Pooyan Rahmanzadehgervi, and Logan Bolton (Auburn University) for feedback, support, and discussion of earlier results.
AV was supported by the Hyundai Motor Chung Mong-Koo Global Scholarship.
AN was supported by the NSF Grant No. 1850117 \& 2145767, and donations from NaphCare Foundation \& Adobe Research.


\bibliography{references}
\bibliographystyle{iclr2026_conference}

\clearpage

\appendix

\def\suptitle{Appendix for:\\\papertitle}


\def\maketitlesupp{%
    \newpage
    \vbox{\hsize\textwidth
        {\LARGE\sc \suptitle\par}
        \vskip 0.3in minus 0.1in
    }%
}
\maketitlesupp




    

    

    
    




    


\addcontentsline{toc}{section}{Appendix}
\setcounter{tocdepth}{2} 
\tableofcontents

\clearpage

\section{Additional findings}\label{appsec:appendix_more_findings}

\subsection{VLMs fail to recognize that an extra leg is added to common animals~\animalicon} \label{sec:animal_result_main}



\subsec{Experiments} We use the same experiment setup as in \cref{sec:res-sanity-check} but test VLMs on CF images. 
Specifically, we evaluate five VLMs on the \animalicon~animal images where an extra leg is added to (a) a bird (three legs instead of two) and a mammal (five legs instead of four). 
We ask each VLM with default settings to count legs (\qone and \qtwo; \cref{fig:diagram}b).
\begin{table}[h]
\caption{VLMs perform poorly across 6 (out of 7) \biastest tasks, spanning photo-realistic images (\animalicon~animals and \logoicon~logos) and abstract images (\flagicon~flag, \chessicon~chess pieces, \illusionicon~optical illusions, and \patternedgridicon~patterned grids). 
}
\footnotesize
\centering
\resizebox{1.0\textwidth}{!}{%
\begin{tabular}{l ccc ccc ccc} 
\toprule
& \multicolumn{3}{c}{a.\ \animalicon~Animal}
& \multicolumn{3}{c}{b.\ \logoicon~Logo}
& \multicolumn{3}{c}{c.\ \flagicon~Flag} \\
\cmidrule(lr){2-4}  \cmidrule(lr){5-7}  \cmidrule(lr){8-10}
Model
& Birds & Mammals & Mean
& Shoes & Cars & Mean
& Stars & Stripes & Mean \\
\midrule
\geminilogo~\geminifull~~~
& 0.00 & 0.00 & 0.00
& 5.80 & 0.00 & 1.96
& 11.54 & \textbf{8.33} & 10.42 \\
\newsonnetlogo~\newsonnet~~~
& 0.00 & 0.00 & 0.00
& 8.15 & 0.00 & 2.72
& \textbf{20.51} & 1.19 & 13.75 \\
\gptfourlogo~\gptfour~~~
& \textbf{5.07} & \textbf{11.03} & \textbf{9.52}
& \textbf{25.36} & \textbf{1.11} & 9.07
& 3.21 & 1.19 & 2.50 \\
\gptfulllogo~\gptfull~~~
& 0.00 & 1.23 & 0.92
& 21.01 & \textbf{1.11} & 7.60
& 5.13 & 4.76 & 5.00 \\
\gptminilogo~\gptmini~~~
& 0.00 & 0.25 & 0.18
& 27.54 & 0.00 & \textbf{9.31}
& 18.59 & 7.14 & \textbf{14.58} \\
\midrule
Mean
& 1.01 & 2.50 & \hallucinatepanel{2.12}
& 17.57 & 0.44 & \hallucinatepanel{6.13}
& 11.79 & 4.52 & \hallucinatepanel{9.25} \\
\midrule
\midrule
& \multicolumn{3}{c}{d.\ \chessicon~Chess/Xiangqi Pieces}
& \multicolumn{3}{c}{e.\ \illusionicon~Optical Illusions}
& \multicolumn{3}{c}{f.\ \patternedgridicon~Patterned Grid} \\
\cmidrule(lr){2-4}  \cmidrule(lr){5-7}  \cmidrule(lr){8-10}
Model
& Chess & Xiangqi & Mean
& Original & Modified & Mean
& Remove & Rep/Add & Mean \\
\midrule
\geminilogo~\geminifull~~~
& 17.36 & \textbf{36.11} & 26.74
& 73.16 & 26.52 & 49.81
& 13.10 & 28.57 & 20.83 \\
\newsonnetlogo~\newsonnet~~~
& 7.64 & 10.42 & 9.03
& 42.68 & \textbf{65.91} & \textbf{54.29}
& \textbf{35.71} & \textbf{33.33} & \textbf{34.52} \\
\gptfourlogo~\gptfour~~~
& 11.81 & 5.56 & 8.68
& \textbf{92.17} & 5.05 & 48.61
& 10.12 & 27.38 & 18.75 \\
\gptfulllogo~\gptfull~~~
& \textbf{56.94} & 28.47 & 42.71
& 91.67 & 9.09 & 50.38
& 14.88 & 26.19 & 20.54 \\
\gptminilogo~\gptmini~~~
& 55.56 & 32.64 & \textbf{44.10}
& 90.40 & 12.12 & 51.26
& 12.50 & 22.62 & 17.56 \\
\midrule
Mean
& 29.86 & 22.64 & \hallucinatepanel{26.25}
& 78.02 & 23.74 & \hallucinatepanel{50.87}
& 17.26 & 27.62 & \hallucinatepanel{22.44} \\
\bottomrule
\end{tabular}}
\label{tab:full_results}
\end{table}

\subsec{Results} 
On average, VLMs perform poorly (2.12\% accuracy) at counting legs of 3-legged and 5-legged counterfactual animals (\cref{tab:overall_performance}\animalicon, \cref{appfig:qualitative_results_animals}).
Furthermore, 94.14\% of the wrong answers match the original, well-known leg counts (\cref{fig:bias-errors}, \cref{fig:teaser}a, and \cref{tab:full_bias_aligned_rate}), demonstrating that VLMs rely mostly on memorized prior knowledge to answer rather than inspecting the legs in the image (see \cref{appfig:sanity_check}c, and \cref{sec:vision_encoder_analysis}).

VLMs are slightly worse at counting the legs of birds than counting the legs of mammals (1.01\% vs. 2.50\%;~\cref{tab:full_results}\animalicon). 
Bird legs (\cref{fig:teaser}a) are typically thinner, which may make it harder to detect than mammals' legs (\cref{appfig:sanity_check}b). 
On birds, except for \gptfourlogo~\gptfour, all VLMs score 0\% accuracy (\cref{tab:full_results}\animalicon).


\subsection{VLMs struggle to detect logo modifications, often relying on context rather than visual detail~\logoicon}\label{sec:result_logos}

\subsec{Experiments} We replicate the experiment settings from \cref{sec:animal_result_main} on our \logoicon~logo task, evaluating five VLMs on modified shoe and car logo images.

\subsec{Results} VLM performance on car logos (0.44\%;~\cref{tab:full_results}\logoicon) is significantly worse than on shoe logos (17.57\%;~\cref{tab:full_results}\logoicon), as the emblem is small relative to the vehicle (see~\cref{fig:teaser}b). In contrast, shoe logos occupy more image area (see~\cref{appfig:sanity_check}e) and involve only a few simple curves or stripes (\ie, one extra curve for Nike, one added stripe for Adidas). These results highlight two key limitations: VLMs fail to attend to small, context-embedded visual changes and instead rely on memorization, without visually verifying the \logoicon~logo itself (\eg, by zooming in~\citep{imagenet-hard}).

\subsection{VLMs fail to count visual elements in modified flags~\flagicon} 
\label{sec:result_flags}

\subsec{Experiments}  
We follow the procedure from \cref{sec:animal_result_main} on our \flagicon~flag tasks. Five VLMs are prompted to count either the number of stars or the number of stripes in original and modified versions of national flags. Modifications consist of adding or removing a single star or stripe, and each model uses its default settings.

\subsec{Results}
VLMs achieve higher mean accuracy on star modifications (11.79\%;~\cref{tab:full_results}\flagicon) than on stripe modifications (4.52\%;~\cref{tab:full_results}\flagicon). This pattern indicates that models are somewhat more attuned to discrete symbol changes (missing or extra stars; see~\cref{fig:abstract_images_flags_chess}d) than to subtle structural alterations (extra or missing stripes; see~\cref{fig:abstract_images_flags_chess}b), yet overall sensitivity to flag modifications is extremely limited (\hallucinatepanel{9.25\%};~\cref{tab:full_results}\flagicon).

\subsection{Thinking models better detect chess piece changes in modified chess starting positions~\chessicon}
\label{sec:result_chess_pieces}

\subsec{Experiments}  We evaluate five VLMs on a \chessicon~chess‐piece counting task using standard starting positions for both Western chess and xiangqi. For each board type, we generate images in which exactly one piece is either removed or replaced by another piece of the same color. All models use their default settings and are prompted to report the total number of pieces or number of a certain piece (\eg, Knights) on the board.

\subsec{Results} VLMs perform significantly better on Western chess (see~\cref{fig:teaser}\chessicon) than on xiangqi (see~\cref{fig:abstract_images_flags_chess}a) in terms of mean accuracy (29.86 \% vs. 22.64\%; \cref{tab:full_results}\chessicon). 
Thinking models (\geminilogo~\geminifull, \gptfulllogo~\gptfull, and \gptminilogo~\gptmini) all exceed 26\% accuracy, whereas non-thinking models (\gptfourlogo~\gptfour and \newsonnetlogo~\newsonnet) remain below 10\% (\cref{tab:full_results}\chessicon). This suggests that on well-structured abstract images, models with explicit reasoning capabilities are better able to detect anomalies.

\subsection{VLMs cannot  count rows and columns in simple game boards~\boardgameicon}
\label{sec:res_boardgame}

\subsec{Experiments} Following our previous tasks, we evaluate five VLMs on counting tasks in four \boardgameicon~grid‐based game boards: chess (8×8), Go (19×19), Sudoku (9×9), and xiangqi (10×9). For chess (see~\cref{fig:abstract_images_flags_chess}e) and Sudoku (see~\cref{fig:abstract_images_flags_chess}c), models are asked to report the number of rows and columns. For Go and xiangqi (see~\cref{appfig:sanity_check}f), they report the counts of horizontal and vertical lines. 

\begin{table}[h!]
\caption{All VLMs' performance is extremely low (\hallucinatepanel{2.26\%}) across \boardgameicon~game boards, confirming that current VLMs are largely unable to perform simple counting operations in structured visual settings}
\centering
\resizebox{0.6\textwidth}{!}{
\begin{tabular}{lcccccc}
\toprule
Model & Chess & Go & Sudoku & Xiangqi & Mean \\
\midrule
\geminilogo~\geminifull 
& 2.08
& 0.00
& 0.00
& 6.25
& 2.38
\\
\newsonnetlogo~\newsonnet 
& 0.00
& 0.00
& 0.00
& 6.25
& 1.79
\\
\gptfourlogo~\gptfour 
& 0.00
& 0.00
& 0.00
& 0.00
& 0.00
\\
\gptfulllogo~\gptfull 
& 0.00
& 0.00
& 0.00
& \textbf{8.33}
& 2.38
\\
\gptminilogo~\gptmini 
& \textbf{16.67}
& 0.00
& 0.00
& 0.00
& \textbf{4.76}
\\
\midrule
Mean 
& 3.75
& 0.00
& 0.00
& 4.17
& \hallucinatepanel{2.26}

\\
\bottomrule
\end{tabular}
}
\label{tab:boardgame_results}
\end{table}

\subsec{Results} All VLMs perform extremely poorly on \boardgameicon, (\hallucinatepanel{2.26\%} mean accuracy; \cref{tab:boardgame_results}). The models even failed to answer any counting questions correctly on Sudoku (see~\cref{fig:abstract_images_flags_chess}c) and Go (0\%;~\cref{tab:boardgame_results}). These findings confirm that current VLMs are unable to execute basic visual counting tasks in structured settings and instead default to overconfident but incorrect guesses.  

\subsection{VLMs are biased towards the known illusions and fail to recognize the changes in the counterfactual, modified versions~\illusionicon}
\label{sec:res_illusion}

\subsec{Experiment} We test five VLMs on 6 classic optical illusions, \ie, Müller-Lyer, Zöllner, Ebbinghaus, Vertical-Horizontal, Pogendorff, and Ponzo (\cref{fig:appendix_optical_illusion_qualitative,fig:appendix_optical_illusion_qualitative_2}). 
Each illusion is presented in two versions: (a) its original form and (b) a counterfactual, modified version where the groundtruth answer is reversed (\cref{fig:main_optical_illusion}). 
For both versions per illusion, we ask VLMs the same Y/N question
(see~\cref{sec:appendix_illusion}).  

\subsec{Results} 
On average, over original and CF versions, all 5 VLMs perform around the random chance (mean accuracy of 50.87\%; \cref{tab:full_results}\illusionicon).
78.02\% of the time, VLMs provide answers that are well-known (corresponding to the prior knowledge) but \emph{false} given our CF images (23.74\% accuracy).


4 out of 5 VLMs perform well on the original versions of the illusions but poorly on the CF versions, exhibiting \textbf{a strong bias to the well-known answers}.
Notably, \newsonnetlogo~\newsonnet performs only slightly above the random chance (54.29\% accuracy).
However, it behaves differently from 4 other VLMs, performing much better on the CF versions than on the original illusions (65.91\% vs. 42.68\% accuracy; \cref{tab:full_results}\illusionicon).
In sum, our results support the findings that VLMs have a poor, low-level vision capability \citep{vlms-are-blind} and that they are \emph{over}confident.


\subsection{VLMs are biased towards the global pattern in a grid~\patternedgridicon}
\label{sec:res_patternedgrid}

\subsec{Experiments} We test VLMs on counting the shapes or tally marks inside an anomaly cell where the total number of shapes or marks do not follow the patterns in the surrounding cells  (\cref{fig:teaser}g).

\subsec{Results} 
Overall, VLMs perform poorly at 22.44\% accuracy.
43.45\% of all count predictions, both correct and incorrect, match the biased answers (\cref{fig:bias-errors}\patternedgridicon) that correspond to the surrounding cells.
In other words, when VLMs make a \emph{wrong} counting predictions, more than half (\ie, 56.02\%) of the time, their answers match the \textbf{global pattern of most cells in the grid} rather than the target anomaly cell in question (\cref{fig:patterned_grid}).
Our results confirm a striking influence of the background pattern on VLMs' assessment on a small local region.
Here, our patterns in the grids are created from scratch and, therefore, do not represent a pattern memorized from the Internet.


\subsection{Linear probing: The vision encoders of VLMs actually extract sufficient leg count information from animal images \animalicon}
\label{sec:vision_encoder_analysis}

\cref{sec:res-sanity-check} demonstrates that VLMs exhibit visual bias, defaulting to memorized answers \sethlcolor{biascolor}\hl{75.70\%} of the time across all models. Here, we investigate whether this failure stems from vision encoders' inability to detect fine-grained modifications or from language models overriding visual evidence with prior knowledge. This experiment is crucial for understanding the source of VLM biases.

\begin{table}[h]
\centering
\caption{Vision encoder features contain sufficient information to distinguish 4-leg from 5-leg animals (95.26\% accuracy before projection), but the complete VLM fails dramatically (49.71\%), defaulting to biased answers 99.43\% of the time. On abstract images, both linear probing (99.42\%) and VLM (65.52\%) perform substantially better.}

\label{tab:linear_probing_results}
\resizebox{\textwidth}{!}{%
\begin{tabular}{lccc}
\toprule
 & \multicolumn{2}{c}{\textbf{Animals (5-leg vs 4-leg)}} & \textbf{Rectangles (5 vs 4)} \\
 & Full image & Background removal & Abstract \\
\cmidrule(lr){1-1}\cmidrule(lr){2-3}\cmidrule(lr){4-4}

\multicolumn{4}{c}{\textit{Accuracy (\%) $\uparrow$}}\\
Linear probing (before projection) & \textbf{95.26} & \textbf{95.98} & 99.42\\
Linear probing (after projection) & 91.24 & 93.39 & 98.41 \\
Linear probing (last LLM layer) & 89.08 & 95.40 & \textbf{100.00} \\

\midrule
\rowcolor{gray!15} \llavalogosmall~\llavaonevision\ (full VLM) & 49.71 & 41.95 & 65.52 \\
Random baseline & 50.00 & 50.00 & 50.00 \\

\midrule
\multicolumn{4}{c}{\textit{Bias rate (\%) $\downarrow$}} \\
\rowcolor{gray!15} \llavalogosmall~\llavaonevision\ (full VLM) & 99.43 & 78.30 & — \\
\bottomrule
\end{tabular}
}
\end{table}

\begin{figure}[h]
    \centering

    \begin{subfigure}{0.44\linewidth}
        \centering
        \includegraphics[width=\linewidth]{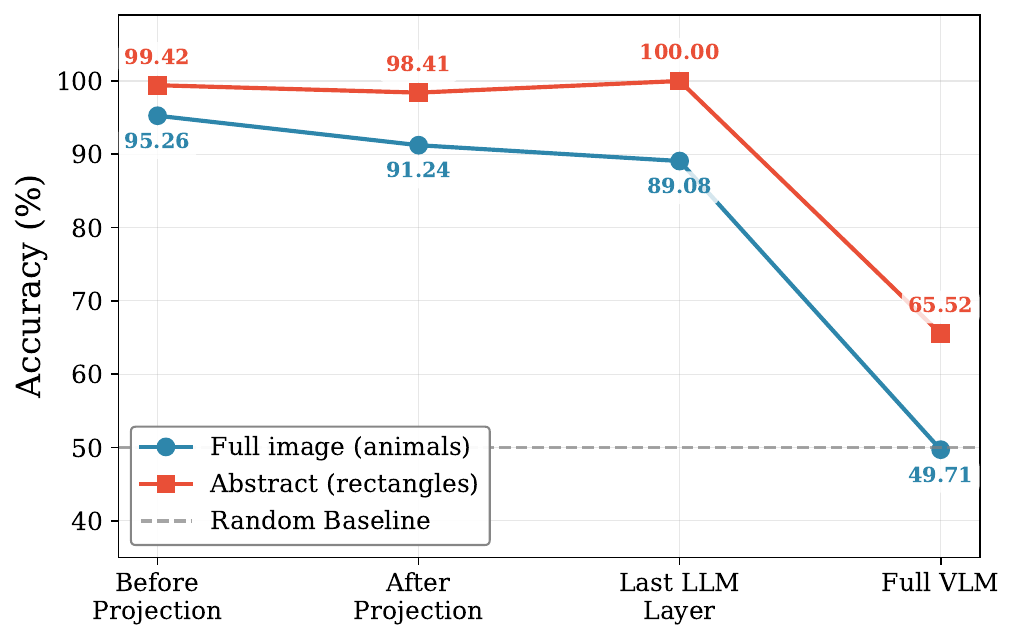}
        \caption{Linear probe across processing stages}
        \label{fig:linear_probe_acc}
    \end{subfigure}
    \hfill
    \begin{subfigure}{0.27\linewidth}
        \centering
        \includegraphics[width=\linewidth]{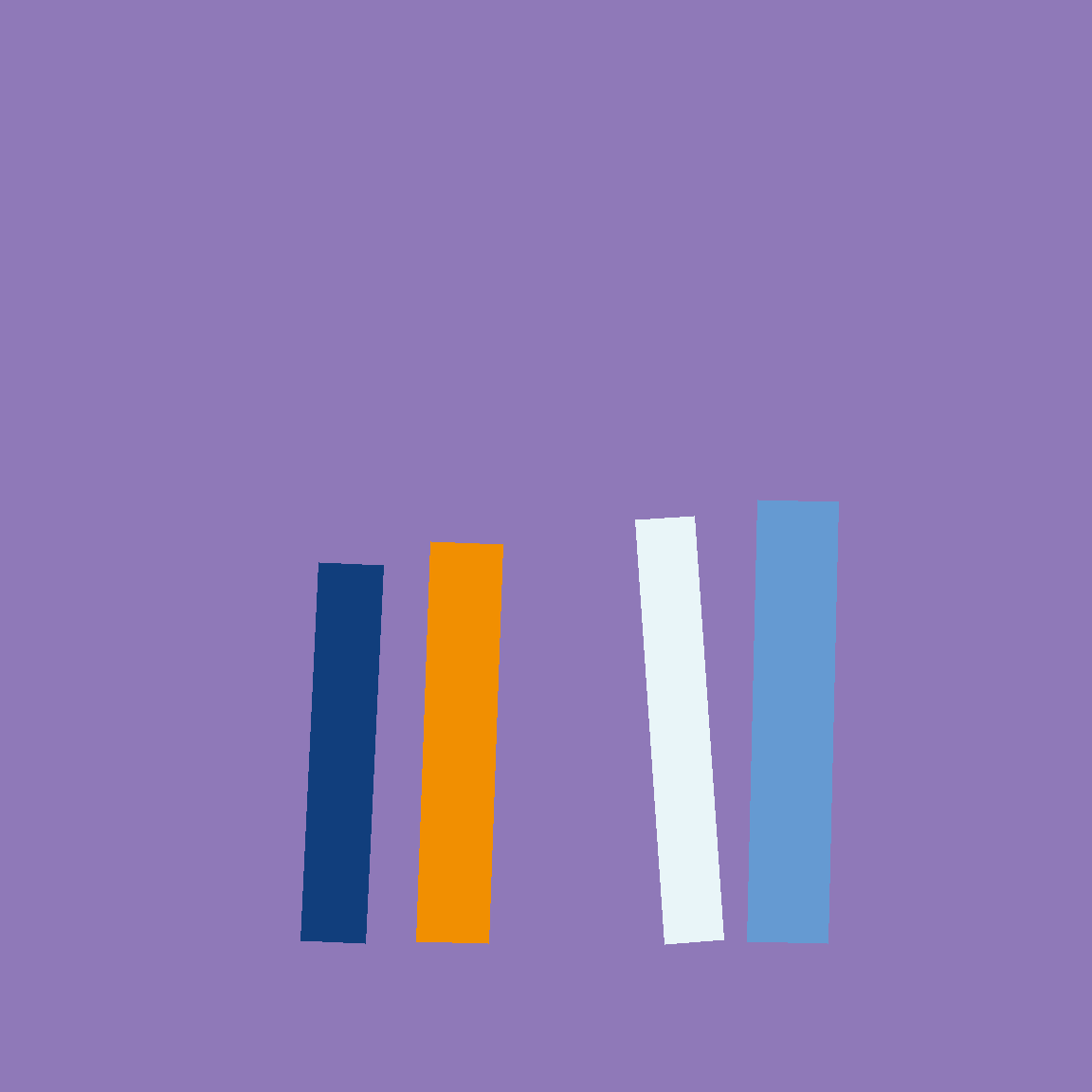}
        \caption{4-rectangle abstract image}
        \label{fig:4-rectangle_abstract_image}
    \end{subfigure}
    \hfill
    \begin{subfigure}{0.27\linewidth}
        \centering
        \includegraphics[width=\linewidth]{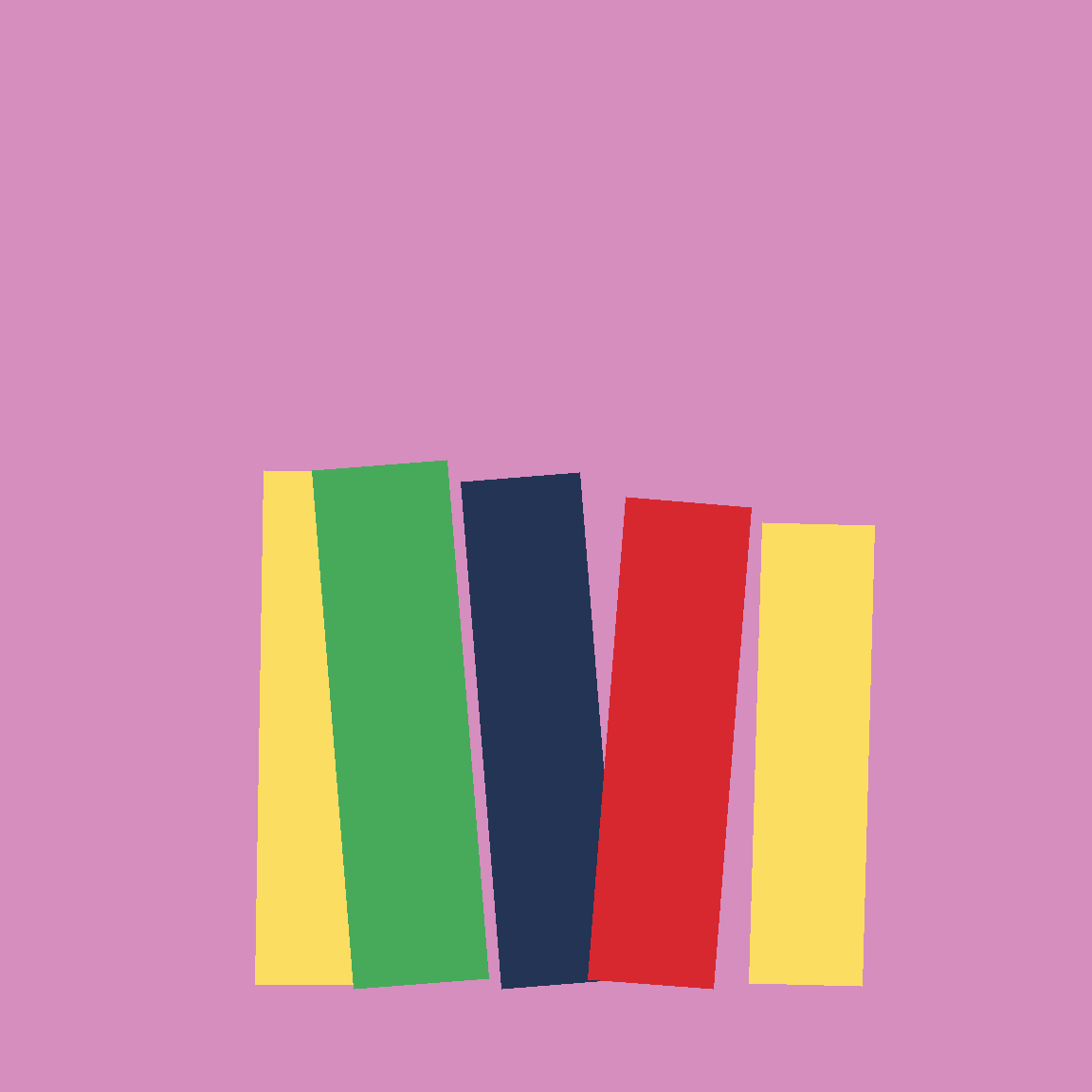}
        \caption{5-rectangle abstract image}
        \label{fig:5_rectangle_abstract_image}
    \end{subfigure}

    \caption{Accuracy degradation across VLM processing stages reveals where bias emerges (a). Vision encoder features maintain high accuracy for both animals (95.26\%) and abstract rectangles (99.42\%) before projection. As information flows through the LLM, animal counting accuracy collapses to 49.71\% while abstract counting degrades less severely to 65.52\%, demonstrating that prior knowledge in language models selectively overrides visual evidence. (b–c) Examples of abstract images.}
    \label{fig:linear_probing_threepanel}
\end{figure}

\textbf{Experiments} We conduct linear probing experiments using features from the vision encoder (\siglip 400M~\citep{siglip}) and the language model (Qwen2 0.5B~\citep{qwen2}) of \llavalogosmall~\llavaonevision~\citep{llavaonevision} on the \animalicon~animal leg counting task. Following~\citet{vlms-are-blind}, we extract features from three processing stages: (1) before projection (vision encoder output, average-pooled to $1 \times 1152$ dimensions), (2) after projection, and (3) the last LLM layer (both average-pooled to $1 \times 896$ dimensions). We train a logistic regression classifier on these frozen features to distinguish 4-legged from 5-legged an

To do this, we create a dataset of 6,594 mammal images (5,598/300/696 train/val/test split) using the same \geminiflash-based generation procedure (\cref{sec:appendix_animals}). We restrict this to mammals only, as they have more diverse species appearances, allowing us to scale up our datasets. We evaluate under two conditions: (1) full image: full images with backgrounds and (2) background removal: cropped images showing only the lower half containing legs (similar to \cref{sec:background_removal}). To isolate the effect of memorized knowledge, we also generate an abstract dataset of rectangles (4 vs. 5 rectangles; \cref{fig:4-rectangle_abstract_image,fig:5_rectangle_abstract_image}) with the same size data split.

\textbf{Results} The \siglip vision encoder successfully distinguishes 4-legged from 5-legged animals and 4-rectangles from 5-rectangles (95.26\%; \cref{tab:linear_probing_results}). In contrast, \llavalogosmall~\llavaonevision, which uses the same \siglip encoder paired with \texttt{Qwen2-0.5B} LLM performing at random chance (49.71\%; \cref{tab:linear_probing_results}). Most striking, it outputs ``4 legs'' for 99.43\% of all images (\ie, bias rate) of all 5-legged animal images.  Removing backgrounds by cropping to legs maintains high linear classifier accuracy (95.26\% $\rightarrow$ 95.98\%) while reducing the VLM's bias rate (99.43\% $\rightarrow$ 78.30\%), though performance of \llavalogosmall~\llavaonevision remains poor (41.95\%; \cref{tab:linear_probing_results}). 

On abstract rectangles with no counterfactuals, linear probing achieves almost perfect accuracy before projection (99.42\%), and \llavalogosmall~\llavaonevision performs substantially better compared to itself on animals (66.52\% vs. 49.71\%). Across processing stages, linear probing accuracy degrades slightly on animals (95.26\% $\rightarrow$ 91.24\% $\rightarrow$ 89.08\%; see~\cref{fig:linear_probe_acc}) but remains near perfect on abstract images (99.42\% $\rightarrow$ 98.41\% $\rightarrow$ 100.00\%; see~\cref{fig:linear_probe_acc}). This suggests that the language model increasingly biases representations toward memorized answers. These results confirm that \textbf{vision encoders successfully detect visual modifications, but language models override this evidence with memorized knowledge}.

\subsection{VLMs are even more biased when the subject name is inserted into the image}
\label{sec:res_title}

Prior sections have shown that VLMs perform poorly on the objective task of counting when the background contains \textbf{visual} cues that strongly correlate with well-known subjects.
As VLM outputs may be influenced by adversarial or distracting text in the image \citep{goh2021multimodal}, here, we test how in-image \textbf{textual} cues about the subjects (\eg, ``Ebbinghaus illusion'') influence VLMs on the same counting questions.

\begin{figure}[h!]
    \centering
    \includegraphics[width=0.6\textwidth]{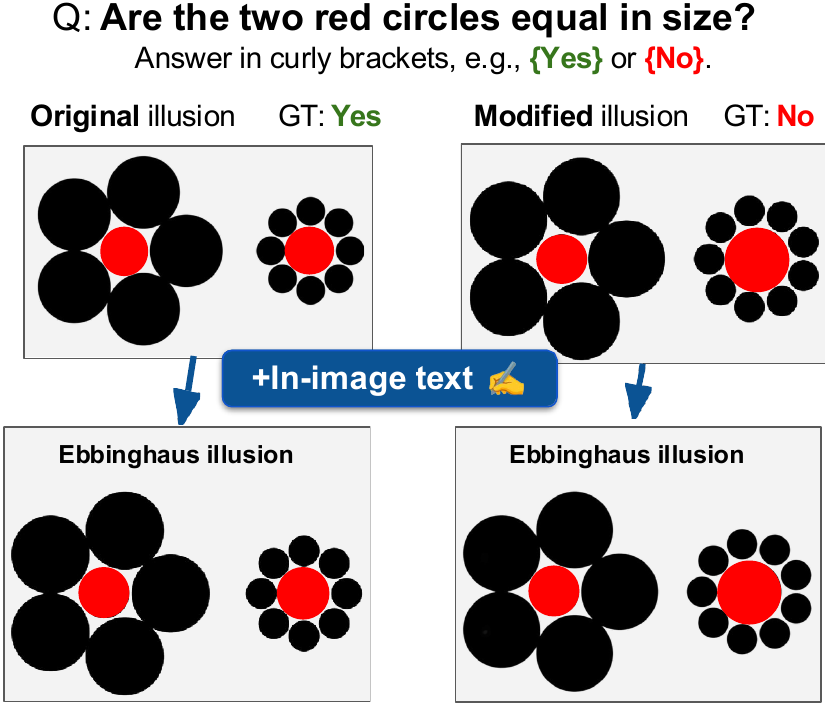}
    \caption{Original vs. modified versions without (top) and with (bottom) the in-image text (``Ebbinghaus illusion'').}
    \label{fig:main_optical_illusion}
\end{figure}

\subsec{Experiments} 
We insert the subject name (\eg, ``Adidas'' or ``Ebbinghaus illusion''; \cref{fig:main_optical_illusion}) into the top of all original and CF images, extending the image vertically but keeping the original content unchanged.
We repeat previous experiments asking VLMs the two counting questions (\qone \& \qtwo).

\subsec{Results} 
All VLMs perform worse when an in-image text is added (\decreasenoparent{4.49};~\cref{tab:prompt_results}). 
Interestingly, the decrease is more pronounced for thinking models (\cref{tab:prompt_results}), such as \gptminilogo~\gptmini (\decreasenoparent{6.56}), \gptfulllogo~\gptfull (\decreasenoparent{6.41}), than for non-thinking ones such as \newsonnetlogo~\newsonnet (\decreasenoparent{2.81}) and \gptfourlogo~\gptfour (\decreasenoparent{2.67}).
This result is consistent with recent findings that thinking models tend to hallucinate more \citep{openai2025o3o4mini,zhang2024snowball}, here more biased toward the text in the image despite contradictory visuals.

\begin{table}[h]
\caption{
Adding adversarial, in-image textual cues that state the subject name (\eg, ``Adidas'') cause VLMs to decrease their accuracy (\decreasenoparent{4.49}) on counterfactual images (b).
In contrast, instructing VLMs to rely exclusively on the image details to answer questions (Debiased) or to double-check its answers (Double-Check) only slightly improves accuracy, by \increasenoparent{1.87} and \increasenoparent{2.70}, respectively (c).
}
\centering
\resizebox{0.95\textwidth}{!}{
\begin{tabular}{lccccc}
\toprule
Model
  & a. Baseline
  & \multicolumn{1}{c}{b. Adversarial}
  & \multicolumn{2}{c}{c. Helpful textual prompt} \\
\cmidrule(lr){3-3}\cmidrule(lr){4-5}
  &  
  & w/ In-image text
  & w/ Debiased Prompt
  & w/ Double-Check
\\
\midrule
\geminilogo~\geminifull 
& 16.02 
& 12.04 \decrease{3.98}
& 19.72 \increase{3.70} 
& 20.22 \increase{4.20}
 \\
\newsonnetlogo~\newsonnet 
& 16.59 
& 13.78 \decrease{2.81} 
& 19.29 \increase{2.70}
& 20.86 \increase{4.27}
\\

\gptfourlogo~\gptfour 
& 13.88 
& 11.21 \decrease{2.67} 
& 14.38 \increase{0.50} 
& 16.00 \increase{2.12}

\\
\gptfulllogo~\gptfull 
& 18.50 
& 12.09 \decrease{6.41} 
& 18.94 \increase{0.44} 
& 21.02 \increase{2.52} \\
\gptminilogo~\gptmini 
& 20.25 
& 13.69 \decrease{6.56} 
& 22.25 \increase{2.00} 
& 20.61 \increase{0.36} \\
\midrule
Mean 
& 17.05 
& 12.56 \decrease{4.49} 
& 18.92 \increase{1.87}
& 19.75 \increase{2.70} \\
\bottomrule
\end{tabular}
}

\label{tab:prompt_results}
\end{table}


\subsection{Helpful prompts do not ameliorate the bias issues in VLMs}
\label{sec:helpful_prompts}

Previous results show that VLMs rely heavily on prior knowledge to answer objective counting questions.
Here, we test how incorporating \emph{helpful} instructions in the prompts may help VLMs become less biased.

\subsec{Experiments} 
We apply two prompting strategies across all \biastest tasks:

(1) {Debiased Prompt}: We prepend the original question (\qone and \qtwo) with ``\emph{Do not assume from prior knowledge and answer only based on what is visible in the image.}'' to encourage models to rely exclusively on image contents. 

(2) {Double-Check}: After VLMs answer the original question, we add a follow-up prompt of ``\emph{Please double-check your answer and give your final answer in curly brackets, following the format above.}'' 

These prompts are designed to encourage VLMs to examine the image more carefully.
All experiments use the same images and default model settings as in the baseline setup.

\subsec{Results}  
Both helpful prompting strategies improve VLM accuracy but only slightly over the baseline,  \increasenoparent{1.87} for Debiased and \increasenoparent{2.70} for Double-Check (\cref{tab:prompt_results}c).
That is, explicitly instructing models to rely on image contents or verify their answer helps to some extent but does not address the core issue of bias (\cref{appsec:qualitative-helpful-prompts}).

\subsection{Re-running experiments multiple times yields consistent results}
\label{appsec:multi_runs}

To ensure the robustness of our findings and provide richer evaluation metrics, we investigate whether VLM performance varies significantly across multiple runs and examine other aspects beyond accuracy and bias rate.

\subsec{Experiments} We conduct 5-run experiments on our top-performing thinking and non-thinking VLMs (\ie, \gptmini and \newsonnet) across all \biastest tasks. For each run, we measure:
\begin{enumerate}
    \item Mean accuracy across 5 runs: average percentage of correct answers when the model is evaluated 5 times on the same dataset
    \item  Pass@5 rate: the frequency that at least 1 of 5 outputs is correct
    \item Bias rate: probability of biased answers across runs
    \item Agreement-based consistency: probability of the most frequent answer
    \item Model self-reported confidence scores: we ask VLMs in a second turn to provide confidence scores for their first-turn answers.
\end{enumerate}

\begin{table}[h!]
\centering
\caption{VLMs demonstrate consistently poor performance (17.89\% mean accuracy, 26.55\% pass@5 rate) yet exhibit severe overconfidence (91.25\% self-reported confidence score), with high agreement-based consistency (92.79\%) indicating they reliably produce the same incorrect answers across 5 runs.}
\label{tab:r1}
\resizebox{0.75\textwidth}{!}{%
\begin{tabular}{lccc}
\toprule
Metric & \gptminilogo~\gptmini & \newsonnetlogo~\newsonnet & Mean \\
\midrule
Mean accuracy $\uparrow$ & $19.54 \pm 0.68$ & $16.23 \pm 0.36$ & $17.89$ \\
Pass@5 rate $\uparrow$ & 30.16 & 22.93 & 26.55 \\
Bias rate $\downarrow$ & 73.66 & 77.27 & 75.47 \\
Agreement-based consistency $\uparrow$ & 90.82 & 94.75 & 92.79 \\
Model self-reported confidence score $\uparrow$ & 84.73 & 97.77 & 91.25 \\
\bottomrule
\end{tabular}
}
\end{table}

\subsec{Results} Mean accuracy scores remain stable across 5 runs (17.89\%). Pass@5 rates provide only modest improvement (26.55\%), indicating that even with multiple attempts, VLMs cannot effectively solve counterfactual problems in \biastest. Most striking is the high agreement-based consistency (92.79\% model mean), showing VLMs consistently produce identical answers across runs. Despite poor performance, VLMs exhibit severe overconfidence with self-reported confidence scores (91.25\% model mean). The bias rate remains consistently high (75.47\% model mean), confirming that VLMs persistently default to memorized patterns regardless of multiple attempts for correction.

\subsection{Providing in-context examples of animals with abnormal legs fails for \gptmini as it sometimes distrusts the provided labels}
\label{sec:few_shot_distrust}

Few-shot prompting typically improves VLM performance by providing in-context learning examples that help models adapt to specific tasks. 
Here, we investigate whether visual demonstrations of counterfactual animals~\animalicon can help VLMs overcome their systematic biases and improve counting accuracy.

\begin{table}[h!]
\centering
\caption{\gptminilogo~\gptmini (thinking model) exhibit strong resistance to few-shot examples and distrusts visual evidence (\increasenoparent{1.66}--\increasenoparent{8.12}), while \gptfourlogo~\gptfour (non-thinking model) responds effectively to few-shot prompting (\increasenoparent{15.75}--\increasenoparent{51.29}).}
\label{tab:few_shot_results}
\resizebox{0.9\linewidth}{!}{%
\begin{tabular}{lcccc}
\toprule
\multirow{2}{*}{Configuration} & \multicolumn{2}{c}{Accuracy (\%)} & \multicolumn{2}{c}{Bias rate (\%)} \\
\cmidrule(lr){2-3}\cmidrule(lr){4-5}
 & \gptminilogo~\gptmini & \gptfourlogo~\gptfour & \gptminilogo~\gptmini & \gptfourlogo~\gptfour \\
\midrule
Zero-shot & 0.18 & 9.52 & 97.25 & 79.67 \\
Few-shot & 1.84 (\increasenoparent{1.66}) & 25.27 (\increasenoparent{15.75}) & 80.51 (\decreasenoparentgreen{16.74}) & 70.70 (\decreasenoparentgreen{8.97}) \\
Few-shot + strong labels & 2.57 (\increasenoparent{2.39}) & 23.81 (\increasenoparent{14.29}) & 77.94 (\decreasenoparentgreen{19.31}) & 72.53 (\decreasenoparentgreen{7.14}) \\
Few-shot + strong labels + hint & \textbf{8.30} (\increasenoparent{8.12}) & \textbf{60.81} (\increasenoparent{51.29}) & \textbf{13.04} (\decreasenoparentgreen{84.21}) & \textbf{30.40} (\decreasenoparentgreen{49.27}) \\
\bottomrule
\end{tabular}%
}
\end{table}


\subsec{Experiments}
We design three few-shot prompting strategies to test on the \animalicon~animal counting task:

\begin{enumerate}
    \item \textbf{Few-shot}: Provide one normal (4-legged) and one counterfactual (5-legged) example, each clearly labeled as ``This is an x-legged animal.'' This establishes the basic task format and demonstrates that animals can have non-standard leg counts.
    
    \item \textbf{Few-shot + strong labels}: Use the same visual examples but reinforce with stronger verification language: ``This is an x-legged animal, which has been verified.'' This approach tests whether stronger language can override model biases.
    
    \item \textbf{Few-shot + strong labels + hint}: Build upon the previous strategy by adding an explicit warning to the test question: ``HINT: This is an animal with an unusual number of legs.'' This directly alerts the model to expect counterfactual cases.
\end{enumerate}

To ensure robust evaluation, we randomize the order of few-shot examples across questions and vary the animal species used in demonstrations (\eg, dogs, cats, lions). We evaluate these strategies on two models with different capabilities: \gptmini (\ie, thinking) and \gptfour (\ie, non-thinking).

\subsec{Results}
\gptminilogo~\gptmini demonstrates strong resistance to few-shot examples, showing only minimal improvement (\increasenoparent{1.66}; \cref{tab:few_shot_results}) over zero-shot performance. Qualitative analysis reveals active distrust of provided labels (\cref{fig:few_shot}), persisting even with strong verification language (\cref{fig:few_shot_strong_label,fig:few_shot_strong_label_2}). This distrust causes the model to rely on knowledge priors rather than visual and few-shot evidence. Even with explicit hints, \gptmini reluctantly acknowledges counterfactual cases but continues miscounting (\cref{fig:few_shot_strong_label_hint}). While this significantly reduces bias-aligned errors (\decreasenoparentgreen{84.21}; \cref{tab:few_shot_results}), accuracy improvement remains modest (\increasenoparent{8.12}; \cref{tab:few_shot_results}) compared to zero-shot. 

In contrast, \gptfourlogo~\gptfour responds effectively to few-shot prompting  (\increasenoparent{14.29}--\increasenoparent{51.29}; \cref{tab:few_shot_results}). 
This finding aligns with recent observations that thinking models exhibit increased hallucination rates \citep{openai2025o3o4mini}, here manifesting as rejection of accurate visual information.

\begin{table}[ht!]
    \centering
    \caption{Full results across proprietary SOTA VLMs (\cref{sec:result_combined}), open-source VLMs (\cref{appsec:inverse_scaling}), pointing VLMs (\cref{sec:pointing_vlms}) and tool-using VLMs (\cref{appsec:tool_use_vlms}). Latest thinking VLMs show mixed results on canonical answer bias: \gptfivelogo~\gptfive achieves modest improvement (30.72\%) while \groklogo~\grok underperforms older VLMs (17.71\% vs. \gptminilogo~\gptmini's 20.25\%).}
    \label{tab:overall_performance_latest_vlms}
    \resizebox{1.0\textwidth}{!}{%
    \begin{tabular}{lccccccccc}
\toprule
Model & \multicolumn{8}{c}{Accuracy (\%) in counting questions (\qone\&\qtwo) on counterfactual images} & \multicolumn{1}{c}{Bias rate (\%)} \\ 
\cmidrule(lr){2-9}\cmidrule(lr){10-10}

 & a.\animalicon & b.\logoicon & c. \flagicon & d. \chessicon & e. \boardgameicon & f. \illusionicon  & g. \patternedgridicon & Task mean & Task mean\\

\midrule
\multicolumn{10}{c}{\textit{Proprietary SOTA VLMs (\cref{sec:result_combined})}}\\
\raisebox{-.1\height}\geminilogo~\geminifull~~~   
& 0.00
& 1.96
& 10.42
& 26.74
& 2.38
& 49.81
& 20.83
& 16.02
& 76.79
\\

\raisebox{-.2\height}\newsonnetlogo~\newsonnet~~~
& 0.00
& 2.72
& 13.75
& 9.03
& 1.79
& \textbf{54.29}
& \textbf{34.52}
& 16.59
& 76.63
\\

\raisebox{-.2\height}\gptfourlogo~\gptfour~~~
& \textbf{9.52}
& 9.07
& 2.50
& 8.68
& 0.00
& 48.61
& 18.75
& 13.88
& 76.62
\\

\raisebox{-.2\height}\gptfulllogo~\gptfull~~~
& 0.92
& 7.60
& 5.00
& 42.71
& 2.38
& 50.38
& 20.54
& 18.50
& 74.81
\\

\raisebox{-.2\height}\gptminilogo~\gptmini~~~
& 0.18
& \textbf{9.31}
& \textbf{14.58}
& \textbf{44.10}
& \textbf{4.76}
& 51.26
& 17.56
& \textbf{20.25}
& 73.66
\\
\raisebox{-.2\height}\groklogo~\grok~~~ 
& 2.56
& 7.84 
& 9.58
& 34.72
& 8.93
& 51.39
& 8.93
& 17.71
& 54.32
\\
\raisebox{-.2\height}\gptfivelogo~\gptfive~~~
& 4.76
& \textbf{14.95}
& \textbf{25.83}
& \textbf{84.72}
& \textbf{18.15}
& 48.48
& 18.15
& \textbf{30.72}
& 57.36
\\
\midrule
Mean 
& 2.56
& 7.64
& 11.67
& 35.81
& 5.48
& 50.60
& 19.90
& \hallucinatepanel{19.10}
& \sethlcolor{biascolor}\hl{70.03}

\\
\midrule
\midrule
\multicolumn{10}{c}{\textit{Open-source VLMs (\cref{appsec:inverse_scaling})}}\\
\pixtralsmalllogo~\pixtralsmall & 0.00 &  1.47 & 18.52 & 1.02 & \textbf{10.13} & 50.94 &  2.99 & 12.15 & 58.96 \\
\pixtrallargelogo~\pixtrallarge & 0.00 &  8.09 &  7.66 & 1.39 &  7.83 & 51.77 & 18.45 & 13.60 & 72.31 \\
\gwensmalllogo~\gwensmall & \textbf{0.18} & \textbf{13.48} & \textbf{23.75} & 0.70 &  9.58 & \textbf{55.19} & 13.43 & \textbf{16.62} & \textbf{52.56} \\
\gwenlargelogo~\gwenlarge  & 0.00 &  7.84 & 11.25 & \textbf{1.74} &  2.98 & 53.03 & \textbf{20.24} & 13.87 & 67.94 \\
\midrule
 Mean                 & 0.05 &  7.72 & 15.29 & 1.21 &  7.63 & 52.73 & 13.78 & \hallucinatepanel{14.06} & \sethlcolor{biascolor}\hl{62.94} \\

\midrule
\midrule
\multicolumn{10}{c}{\textit{Pointing VLMs (\cref{sec:pointing_vlms})}}\\
\moondreamlogo~\moondream & \textbf{74.36} & 16.91 & 55.00 & 35.07 & 1.79 & 49.75 & 0.00 & 33.27 & 46.78 \\
\molmosmalllogo~\molmosmall & 45.79 & \textbf{19.57} & \textbf{59.58} & 24.31 & \textbf{60.71} & \textbf{54.29} & 4.46  & \textbf{38.39} & 32.80 \\
\molmolargelogo~\molmolarge & 48.90 & 9.18  & 36.25 & \textbf{36.81} & 53.57 & 56.06 & 13.99 & 36.39 & \textbf{23.92} \\
\midrule
Mean         & 56.35 & 15.22 & 50.28 & 32.06 & 38.69 & 53.37 & 6.15 & \hallucinatepanel{36.02} & \sethlcolor{biascolor}\hl{34.50} \\

\midrule
\midrule
\multicolumn{10}{c}{\textit{Tool-using VLMs (\cref{appsec:tool_use_vlms})}}\\
\gptminilogo~\gptmini\ (chat w/ tools) 
& 3.30 & 15.63 & 21.57 & 51.04 & 14.06 & 52.08 & 17.86 & \hallucinatepanel{25.08} & \sethlcolor{biascolor}\hl{68.92} \\
\bottomrule
\end{tabular}
}
\end{table}


\subsection{Thinking VLMs show limited improved accuracy}\label{appsec:latest_vlms_results}
Recently, newer thinking VLMs have been released, which need to be evaluated on our benchmark to provide a complete view of current models' capabilities.

\subsec{Experiments}
We replicate the previous experiments on \qone and \qtwo on our 7 tasks of \biastest on the latest notable VLMs: \gptfivelogo~\gptfive~\citep{openai2025gpt5}, \groklogo~\grok~\citep{grok_4}.

\subsec{Results} \groklogo~\grok does not surpass \gptminilogo~\gptmini and \gptfulllogo~\gptfull (17.71\% vs. 20.25\% and 18.50\%, \cref{tab:overall_performance_latest_vlms}). Meanwhile, \gptfivelogo~\gptfive outperforms \gptminilogo~\gptmini and \gptfulllogo~\gptfull (30.72\% vs. 20.25\% and 18.50\%, \cref{tab:overall_performance_latest_vlms}), particularly excelling on the \chessicon~chess pieces (84.72\%). However, \gptfivelogo~\gptfive still falls far short of expectations, and these latest results do not change our conclusions that VLMs remain biased toward canonical answers on our \biastest.


\subsection{Larger open-source VLMs are more biased}\label{appsec:inverse_scaling}

The prevailing assumption in the field is that larger models with more parameters should perform better on visual reasoning tasks due to increased representational capacity. However, it remains unclear whether this scaling benefit holds for tasks requiring models to override strong prior knowledge, as larger models may suffer from inverse scaling \citep{inverse-scaling} having memorized more biased associations from training data.

\textbf{Experiments} We evaluate four open-source VLMs of varying sizes on all \biastest tasks: \pixtralsmalllogo~\pixtralsmall \citep{pixtral_small}, \pixtrallargelogo~\pixtrallarge \citep{pixtral_large}, \gwensmalllogo~\gwensmall, and \gwenlargelogo~\gwenlarge \citep{gwen_2-5} (\cref{tab:opensource_model_details}). We use the same experimental setup as previous sections, asking counting questions (\qone~and \qtwo) on counterfactual images across all 7 domains and measuring both accuracy and bias rates.

\textbf{Results} Larger models do not consistently outperform smaller variants and often exhibit increased bias. The mean accuracy across all open-source VLMs is remarkably low (14.06\%; \cref{tab:opensource_overall}), with the smallest model (\gwensmalllogo~\gwensmall) achieving the highest accuracy (16.62\%), which is comparable to SOTA closed-source models (17.05\% mean accuracy). More concerningly, larger VLMs demonstrate substantially higher bias rates (72.31\% for \pixtrallargelogo~\pixtrallarge vs. 58.96\% for \pixtralsmalllogo~\pixtralsmall; 67.94\% for \gwenlargelogo~\gwenlarge vs. 52.56\% for \gwensmalllogo~\gwensmall; \cref{tab:opensource_overall}). This pattern suggests that increased model size may actually reinforce memorized biased associations rather than improve visual reasoning capabilities. Moreover, since open-source VLMs are much smaller than closed-source ones, they contain less knowledge and consequently show lower bias rates compared to closed-source models (\sethlcolor{biascolor}\hl{62.94\%} vs. \sethlcolor{biascolor}\hl{75.70\%}). These findings support the hypothesis that \emph{more knowledge leads to more bias} in counterfactual scenarios.

\begin{table}[h!]
\centering
\caption{Larger open-source VLMs do not outperform smaller variants and exhibit higher bias rates. The smallest VLM (\gwensmalllogo~\gwensmall with 7B parameters) achieves the highest accuracy (16.62\%) while larger VLMs show substantially increased bias rates (72.31\% for \pixtrallargelogo~\pixtrallarge vs. 58.96\% for \pixtralsmalllogo~\pixtralsmall), supporting the hypothesis that more knowledge leads to more bias.}
\label{tab:opensource_overall}
\resizebox{1.0\textwidth}{!}{
\begin{tabular}{lccccccccc}
\toprule
Model & \multicolumn{8}{c}{Accuracy $\uparrow$ in counting questions (\qone\&\qtwo) on counterfactual images} & \multicolumn{1}{c}{Bias rate $\downarrow$} \\
\cmidrule(lr){2-9}\cmidrule(lr){10-10}
& a.\animalicon & b.\logoicon & c.\flagicon & d.\chessicon & e.\boardgameicon & f.\illusionicon & g.\patternedgridicon & Task mean & Task mean \\
\midrule
\pixtralsmalllogo~\pixtralsmall & 0.00 &  1.47 & 18.52 & 1.02 & \textbf{10.13} & 50.94 &  2.99 & 12.15 & 58.96 \\
\pixtrallargelogo~\pixtrallarge & 0.00 &  8.09 &  7.66 & 1.39 &  7.83 & 51.77 & 18.45 & 13.60 & 72.31 \\
\gwensmalllogo~\gwensmall & \textbf{0.18} & \textbf{13.48} & \textbf{23.75} & 0.70 &  9.58 & \textbf{55.19} & 13.43 & \textbf{16.62} & \textbf{52.56} \\
\gwenlargelogo~\gwenlarge  & 0.00 &  7.84 & 11.25 & \textbf{1.74} &  2.98 & 53.03 & \textbf{20.24} & 13.87 & 67.94 \\
\midrule
 Mean                 & 0.05 &  7.72 & 15.29 & 1.21 &  7.63 & 52.73 & 13.78 & \hallucinatepanel{14.06} & \sethlcolor{biascolor}\hl{62.94} \\
\bottomrule
\end{tabular}
}
\end{table}

\subsection{\gptmini uses tools to analyze images only $\sim$30\% of the time and mostly outputs directly biased answers}
\label{appsec:tool_use_vlms}

Previous experiments evaluate VLMs through API access without tool capabilities. By leveraging tools such as zooming and localization, VLMs can potentially improve their counting accuracy by examining visual details more carefully. However, it remains unclear whether VLMs recognize when visual reasoning is needed for familiar subjects familiar subjects with strong bias cues.

\textbf{Experiments} We compare \gptminilogo~\gptmini in two configurations: (1) standard API access without tools, and (2) ChatGPT interface \citep{chatgpt_o4mini} with full Python tool access (\eg, zoom, crop images). We evaluate both versions on counting questions (\qone~and \qtwo) in \biastest tasks. For the ChatGPT interface, we access it via Puppeteer and measure tool usage frequency through the \texttt{tool} tag in the JSON provided by OpenAI's Data Export, and record thinking time to assess computational effort. Due to the rate limit of the ChatGPT interface, we evaluate these two configurations on the 1152px resolution subset.

\begin{table}[h]
\centering
\caption{\gptminilogo~\gptmini with tool access shows only modest improvements (\increasenoparent{1.9} accuracy, \decreasenoparentgreen{2.11} bias rate) despite having access to python tools (\eg, zooming, cropping). The limited gains suggest that Python tools cannot overcome deep-seated biases as effectively as specialized built-in counting mechanisms (\cref{sec:pointing_vlms}).}
\label{tab:o4mini_tool_accuracy}
\resizebox{1.0\textwidth}{!}{%
\begin{tabular}{lcccccccccc}
\toprule
Model & \multicolumn{8}{c}{Accuracy in counting questions (\qone\&\qtwo) on counterfactual images} & \multicolumn{1}{c}{Bias rate} \\ 
\cmidrule(lr){2-9}\cmidrule(lr){10-10}
 & a.\animalicon & b.\logoicon & c.\flagicon & d.\chessicon & e.\boardgameicon & f.\illusionicon  & g.\patternedgridicon & Task mean & Overall \\
\midrule
\gptminilogo~\gptmini\ (API w/o tools)           
& 0.0 & 13.24 & 18.75 & 53.12 & 5.36 & 50.38 & 21.43 & 23.18 & \textbf{71.03} \\
\gptminilogo~\gptmini\ (chat w/ tools) 
& 3.30 & 15.63 & 21.57 & 51.04 & 14.06 & 52.08 & 17.86 & \textbf{25.08} & 68.92 \\
\midrule
$\Delta$ (tools $-$ API)               
& \increasenoparent{3.30} & \increasenoparent{2.39} & \increasenoparent{2.82} & \decreasenoparent{2.08} & \increasenoparent{10.49} & \increasenoparent{1.7} & \decreasenoparent{3.57} & \increasenoparent{1.9} & \decreasenoparentgreen{2.11} \\
\bottomrule
\end{tabular}}
\end{table}



\begin{table}[h!]
\centering
\begin{minipage}[t]{0.49\textwidth}
    \centering
    \vspace{0pt}
    \caption{ Performance when \gptminilogo~\gptmini \textbf{activates tool-using capabilities} (29.66\% of queries; \cref{tab:o4mini_thinking_tools}). Tool use substantially improves accuracy (\increasenoparent{26.51}) and reduces bias rate (\decreasenoparentgreen{33.81}) for task mean. Yet, the low tool usage rate driven by overconfidence in memorized knowledge limits overall performance.}    
    \label{tab:o4mini_tools_performance}
    \scriptsize
    \begin{tabular}{lcc}
    \toprule
    Task & Accuracy (\%) & Bias rate (\%) \\
    \midrule
    a.\animalicon~Animals & 8.33 & 79.17 \\
    b.\logoicon~Logos & 68.75 & 18.75  \\
    c.\flagicon~Flags & 60.0 & 36.0  \\
    d.\chessicon~Chess Pieces & 75.0 & 25.0  \\
    e.\boardgameicon~Game Boards & 53.33 & 13.33  \\
    f.\illusionicon~Optical Illusions & 76.47 & 23.53  \\
    g.\patternedgridicon~Patterned Grid & 19.23 & 50.0  \\
    \midrule
    Task mean & 51.59 (\increasenoparent{26.51}) & 35.11 (\decreasenoparentgreen{33.81}) \\
    \bottomrule
    \end{tabular}
\end{minipage}%
\hfill
\begin{minipage}[t]{0.49\textwidth}
    \centering
    \vspace{10pt}
    \caption{\gptminilogo~\gptmini uses available tools in only 29.66\% of queries on average. The low usage rate indicates that overconfidence in memorized knowledge prevents recognition of when visual reasoning is needed.}
    \label{tab:o4mini_thinking_tools}
    \scriptsize
    \begin{tabular}{lcc}
    \toprule
    Task & Avg. time (s) & Tool use (\%) \\
    \midrule
    a.\animalicon~Animals & 9.89 & 39.56 \\
    b.\logoicon~Logos & 5.46 & 10.87 \\
    c.\flagicon~Flags & 14.00 & 38.75 \\
    d.\chessicon~Chess Pieces & 16.59 & 37.50 \\
    e.\boardgameicon~Game Boards & 10.69 & 26.79 \\
    f.\illusionicon~Optical Illusions & 2.27 & 6.82 \\
    g.\patternedgridicon~Patterned Grid & 16.55 & 47.32 \\
    \midrule
    Task mean & 10.78 & 29.66 \\
    \bottomrule
    \end{tabular}
\end{minipage}
\end{table}

\textbf{Results} Tool access provides modest improvement, increasing accuracy from 20.25\% to 25.08\% (\increasenoparent{4.83}; \cref{tab:o4mini_tool_accuracy}). Similarly, the bias rate decreases slightly from 72.41\% to 68.92\% (\decreasenoparentgreen{3.49}; \cref{tab:o4mini_tool_accuracy}), indicating marginal improvement in avoiding memorized answers. However, despite having access to zooming and localization tools, \gptminilogo~\gptmini employs them in only 29.66\% of queries on average (\cref{tab:o4mini_thinking_tools}). That is, the model defaults to direct visual assessment 70.34\% of the time, suggesting overconfidence in memorized knowledge prevents recognition of when visual reasoning is needed. 
{Importantly, when tools \emph{are} activated, performance improves noticeably: accuracy increases by \increasenoparent{26.51} and bias rate decreases by \decreasenoparentgreen{33.81} on average compared to baseline (\cref{tab:o4mini_tools_performance}). This demonstrates that tools are highly effective when used, but the low activation rate (29.66\%) severely limits their overall impact on model performance. 

\subsection{Small VLMs trained explicitly on counting significantly outperform proprietary SOTA VLMs}
\label{sec:pointing_vlms}

Fine-tuning VLMs with specialized capabilities may help overcome counting biases. Pointing VLMs~\citep{moondream,molmo}---models specifically trained to output coordinate locations---can potentially force visual reasoning rather than relying on memorized patterns. We investigate whether VLMs with explicit pointing abilities can better handle \biastest compared to larger, general-purpose VLMs without pointing capabilities.

\textbf{Experiments} We test three pointing VLMs: \moondreamlogo~\moondream~\citep{moondream}, \molmosmalllogo~\molmosmall, and \molmolargelogo~\molmolarge~\citep{molmo} (\cref{sec:pointing_vlms}). For \moondreamlogo~\moondream, we use its counting API, which outputs only coordinate lists, and count the array length as the final answer---this ensures 100\% pointing capability usage. The exception is \illusionicon~optical illusion task, which requires Y/N responses rather than counting, so we use standard reasoning APIs. For \molmosmalllogo and \molmolargelogo, they autonomously decide whether to invoke pointing capabilities. When used, the coordinate outputs assist in subsequent counting.  We evaluate them on the same counting questions (\qone~and \qtwo) across all \biastest tasks using identical experimental setups as in the previous sections.

\begin{table}[h]
\centering
\caption{Pointing VLMs substantially outperform commercial VLMs across all domains (\hallucinatepanel{36.02\%} vs. \hallucinatepanel{17.05\%} mean accuracy; \cref{tab:full_results}) and regular open-source VLMs (\hallucinatepanel{36.02\%} vs. \hallucinatepanel{14.06\%} mean accuracy; \cref{tab:opensource_overall}). Even the smallest model (\moondreamlogo~\moondream with 2B parameters) achieves 33.27\% accuracy, exceeding most commercial VLMs despite being orders of magnitude smaller.
}
\label{tab:pointing_vlms_results}
\resizebox{1.0\textwidth}{!}{
\begin{tabular}{lccccccccc}
\toprule
Model & \multicolumn{8}{c}{Accuracy in counting questions (\qone\ \&\ \qtwo) on counterfactual images} & \multicolumn{1}{c}{Bias rate} \\
\cmidrule(lr){2-9}\cmidrule(lr){10-10}
& a.\animalicon & b.\logoicon & c.\flagicon & d.\chessicon & e.\boardgameicon & f.\illusionicon & g.\patternedgridicon & Task mean & Task mean \\
\midrule
\moondreamlogo~\moondream & \textbf{74.36} & 16.91 & 55.00 & 35.07 & 1.79 & 49.75 & 0.00 & 33.27 & 46.78 \\
\molmosmalllogo~\molmosmall & 45.79 & \textbf{19.57} & \textbf{59.58} & 24.31 & \textbf{60.71} & \textbf{54.29} & 4.46  & \textbf{38.39} & 32.80 \\
\molmolargelogo~\molmolarge & 48.90 & 9.18  & 36.25 & \textbf{36.81} & 53.57 & 56.06 & 13.99 & 36.39 & \textbf{23.92} \\
\midrule
Mean         & 56.35 & 15.22 & 50.28 & 32.06 & 38.69 & 53.37 & 6.15 & \hallucinatepanel{36.02} & \sethlcolor{biascolor}\hl{34.50} \\
\bottomrule
\end{tabular}
}
\end{table}



\begin{table}[h]
\centering
\caption{Pointing-capable VLMs: frequency of \emph{tool use} (“pointing use”) and counting performance. 
These models are trained to activate the pointing tool only when prompts contain specific trigger patterns; without those triggers they often do not invoke the tool, even when doing so could improve accuracy.}
\label{tab:pointing_use_perf}
\resizebox{0.8\textwidth}{!}{
\begin{tabular}{lcccccc}
\toprule
\multirow{2}{*}{Model} & \multicolumn{3}{c}{Pointing use (\%)} & \multicolumn{3}{c}{Accuracy (\%)} \\
\cmidrule(lr){2-4}\cmidrule(lr){5-7}
 & \qone & \qtwo ($\Delta$) & Avg & \qone & \qtwo ($\Delta$) & Avg \\
\midrule
\molmosmalllogo~\molmosmall & 41.59 & 63.00 & 52.30 & 32.16 & 44.61 & 38.39 \\
\molmolargelogo~\molmolarge & 36.14 & 63.36 & 49.75 & 26.82 & 45.97 & 36.39 \\
\midrule
Mean & 38.87 & 63.18 (\increasenoparent{24.31}) & 51.03 & 29.49 & 45.29 (\increasenoparent{15.80}) & 37.39 \\
\midrule
\geminilogo~\geminifull & -- & -- & -- & 15.59 & 16.45 & 16.02 \\
\newsonnetlogo~\newsonnet & -- & -- & -- & 16.81 & 16.36 & 16.59 \\
\gptfourlogo~\gptfour & -- & -- & -- & 12.55 & 15.20 & 13.88 \\
\gptfulllogo~\gptfull & -- & -- & -- & 17.33 & 19.67 & 18.50 \\
\gptminilogo~\gptmini & -- & -- & -- & 19.96 & 20.55 & 20.25 \\
\midrule
Mean & -- & -- & -- & 16.45 & 17.65 (\increasenoparent{1.20}) & 17.05 \\
\bottomrule
\end{tabular}}
\end{table}
\begin{table}[h]
\centering
\caption{
Ablation study comparing Molmo models' overall performance versus performance when \textbf{pointing capabilities are activated}. On both models, pointing improves counting accuracy (\increasenoparent{5.90} mean across tasks) while reducing bias rates (\decreasenoparentgreen{8.86} mean). The most notable performance gain occurs in \animalicon~animals (\increasenoparent{41.21} for \molmolargelogo~\molmolarge, \increasenoparent{37.00} for \molmosmalllogo~\molmosmall).
}
\label{tab:pointing_with_without}
\resizebox{\textwidth}{!}{%
\begin{tabular}{lccccccccc}
\toprule
Model 
& a.\animalicon 
& b.\logoicon 
& c.\flagicon 
& d.\chessicon 
& e.\boardgameicon 
& g.\patternedgridicon 
& Task mean 
& Bias rate
\\
\midrule
\molmosmalllogo~\molmosmall
& 45.79
& 19.57
& 59.58
& 24.31
& 60.71
& 4.46
& 38.39
& 32.80
\\
\molmosmalllogo~\molmosmall (w/ pointing)
& 82.78 \increase{37.0}
& 20.45 \increase{0.88}
& 56.57 \decrease{3.01}
& 7.83 \decrease{16.48}
& 55.36 \decrease{5.35}
& 4.46 \increase{0.00}
& 41.83 \increase{3.44}
& 14.36 (\decreasenoparentgreen{18.44})
\\
\midrule
\molmolargelogo~\molmolarge
& 48.90
& 9.18
& 36.25
& 36.81
& 53.57
& 13.99
& 36.39
& 23.92
\\
\molmolargelogo~\molmolarge (w/ pointing)
& 90.11 \increase{41.21}
& 9.46 \increase{0.28}
& 44.44 \increase{8.19}
& 53.54 \increase{16.73}
& 55.06 \increase{1.49}
& 14.16 \increase{0.17}
& 45.85 \increase{9.46}
& 23.01 (\decreasenoparentgreen{0.91})
\\
\bottomrule
\end{tabular}
}
\end{table}



\textbf{Results} Pointing VLMs significantly outperform commercial VLMs (\hallucinatepanel{36.02\%} vs. \hallucinatepanel{17.05\%} mean accuracy; \cref{tab:pointing_vlms_results,tab:full_results}) and regular open-source VLMs (\hallucinatepanel{36.02\%} vs. \hallucinatepanel{14.06\%} mean accuracy; \cref{tab:pointing_vlms_results,tab:opensource_overall}). Most remarkably, \moondreamlogo~\moondream with only 2B parameters substantially outperforms \gptminilogo~\gptmini (33.27\% vs. 20.25\%; \cref{tab:pointing_vlms_results}) despite being orders of magnitude smaller. This suggests that training objectives matter more than model scale for overcoming biases in \biastest. Qualitative results can be found in \cref{appsec:qualitative_pointing_vlms}.

However, pointing capabilities remain significantly underutilized. \molmosmalllogo~\molmosmall and \molmolargelogo~\molmolarge even achieve better performance (38.39\% vs. 36.39\%; \cref{tab:pointing_vlms_results}) but only use pointing 51.03\% of the time (\cref{tab:pointing_use_perf}). This could be due to overconfidence, defaulting to direct answers without utilizing their pointing capabilities. One interesting finding is that on \qtwo (\eg, ``Count the legs''), \molmosmalllogo and \molmolargelogo use pointing capabilities much more than on \qone (\eg, ``How many legs'') (63.18\% vs. 38.87\%; \cref{tab:pointing_use_perf}). This leads to a much higher $\Delta$ between \qone and \qtwo for \molmosmalllogo and \molmolargelogo compared to commercial VLMs, which show negligible differences (\increasenoparent{15.80} vs. \increasenoparent{1.20}; \cref{tab:pointing_use_perf}). 
This pattern suggests that explicit counting prompts (\ie, \qtwo) better trigger pointing verification than implicit counting questions (\ie, \qone), though the underutilization indicates that even specialized VLMs struggle to recognize when their memorized knowledge might be misleading.

When pointing is activated, both Molmo models' performance noticeably improves (\cref{tab:pointing_with_without}): \molmolargelogo~\molmolarge gains \increasenoparent{9.46} accuracy with \decreasenoparentgreen{8.09} bias reduction, while \molmosmalllogo~\molmosmall achieves \increasenoparent{3.44} accuracy and \decreasenoparentgreen{18.44} bias reduction. Most notably, on \animalicon~animals, pointing achieves 82.78--90.11\% accuracy, demonstrating that localization overcomes memorized priors.

\subsection{Same failures across model families rule out image generation bias}\label{appsec:image_generation_pipeline_bias}
A potential concern is that bias could arise from generating and evaluating images with the same model families.

\subsec{Experiments} We analyze the results on \animalicon~animals (generated by \geminiflashlogo~\geminiflash) and \logoicon~logos (generated by \gptfulllogo~\gptfouro) from \cref{sec:result_combined} to investigate whether generation bias affects our findings. We examine performance differences between model families on images generated by their own family versus images generated by other families or created programmatically.

\begin{table}[h]
    \centering
    \caption{When presented with modified, counterfactual images in \biastest, VLMs exhibit substantial bias alignment in their counting responses.
    The \sethlcolor{biascolor}\hl{mean bias rate} of five state-of-the-art VLMs across our seven tasks is \sethlcolor{biascolor}\hl{75.70\%}. 
    \gptminilogo~\gptmini shows the lowest bias alignment (\textbf{73.66\%}) indicating relatively better resistance to visual biases.
    VLMs with thinking capabilities (\gptminilogo~\gptmini, \gptfulllogo~\gptfull, \geminilogo~\geminifull) demonstrate bias susceptibility similar to non-thinking models (\newsonnetlogo~\newsonnet, \gptfourlogo~\gptfour).
    }
    \label{tab:full_bias_aligned_rate}
    \resizebox{1.0\textwidth}{!}{%
    \begin{tabular}{lcccccccc}
\toprule
Model & \multicolumn{8}{c}{Bias rate $\downarrow$ in counting questions (\qone\&\qtwo) on counterfactual images} \\ 
\cmidrule(lr){2-9}

 & a.\animalicon & b.\logoicon & c. \flagicon & d. \chessicon & e. \boardgameicon & f. \illusionicon  & g. \patternedgridicon & Task mean \\

\midrule
\raisebox{-.1\height}\geminilogo~\geminifull~~~   
& 100.00
& 98.04
& 89.58
& 70.83
& \textbf{83.93}
& 50.19
& 44.94
& 76.79
\\

\raisebox{-.2\height}\newsonnetlogo~\newsonnet~~~
& 100.00
& 96.79
& 82.50
& 84.72
& 97.62
& \textbf{45.33}
& \textbf{29.46}
& 76.63
\\

\raisebox{-.2\height}\gptfourlogo~\gptfour~~~
& \textbf{79.67}
& \textbf{88.73}
& 97.08
& 80.21
& 98.81
& 51.39
& 40.48
& 76.62
\\

\raisebox{-.2\height}\gptfulllogo~\gptfull~~~
& 93.77
& 91.18
& 93.33
& \textbf{49.65}
& 95.24
& 49.62
& 50.89
& 74.81
\\

\raisebox{-.2\height}\gptminilogo~\gptmini~~~
& 97.25
& 90.20
& \textbf{82.08}
& 54.17
& 91.67
& 48.74
& 51.49
& \textbf{73.66}
\\

\midrule
Mean 
& 94.14
& 92.99
& 88.92
& 67.92
& 93.45
& 49.05
& 43.45
& \sethlcolor{biascolor}\hl{75.70}
\\
\bottomrule
\end{tabular}
}
\end{table}
\subsec{Results} GPT-family models show no substantial advantage on \gptfouro generated \logoicon images (bias rates of 88.73\% for \gptfourlogo~\gptfour vs. 98.04\% for \geminilogo~\geminifull and 96.79\% for \newsonnetlogo~\newsonnet; \cref{tab:full_bias_aligned_rate}). Similar results are shown on \geminiflash generated \animalicon images (100\% bias rate for \geminilogo~\geminifull vs. 97.25\% for \gptminilogo~\gptmini and 100\% for \newsonnetlogo~\newsonnet; \cref{tab:full_bias_aligned_rate}). All VLMs consistently achieve 100\% accuracy on unmodified images but fail dramatically on counterfactual versions (\hallucinatepanel{17.05\%} mean accuracy; \cref{tab:overall_performance}) regardless of image generation source. \textbf{This confirms that the observed bias stems from models' inherent preferences for canonical answers rather than artifacts of the image generation process}.

\subsection{Image resolution has minimal impact on VLM performance across \biastest tasks}\label{appsec:resolution_impact}
Since our VLMBias dataset contains images rendered at multiple resolutions (384px, 768px, 1152px) as part of our generation process, we analyze whether performance varies across these different image sizes to understand if resolution affects bias-driven failures in counting tasks.

\begin{table}[h]
\centering
\caption{VLM accuracy (\%) across different image resolutions shows minimal variation, with only a 2.85-point mean difference between lowest and highest resolutions.}
\label{tab:resolution_analysis}
\resizebox{0.7\textwidth}{!}{%
\begin{tabular}{lccccc}
\toprule
Model & 384px & 768px & 1152px & Mean & $\Delta$ (1152-384) \\
\midrule
\gptminilogo~\gptmini & 17.27 & 20.30 & 23.18 & 20.25 & \increasenoparent{5.91} \\
\gptfulllogo~\gptfull & 16.67 & 17.90 & 20.94 & 18.50 & \increasenoparent{4.27} \\
\newsonnetlogo~\newsonnet & 14.36 & 17.79 & 17.60 & 16.59 & \increasenoparent{3.24} \\
\gptfourlogo~\gptfour & 13.71 & 13.43 & 14.49 & 13.88 & \increasenoparent{0.78} \\
\geminilogo~\geminifull & 15.13 & 17.76 & 15.17 & 16.02 & \increasenoparent{0.04} \\
\midrule
Mean & 15.43 & 17.43 & 18.28 & 17.05 & \increasenoparent{2.85} \\
\bottomrule
\end{tabular}}
\end{table}

\subsec{Experiments} We break down the accuracy results from our main experiments by the three resolutions present in our dataset: 384px, 768px, and 1152px. Each image was originally generated and tested at these different resolutions, allowing us to examine whether VLM performance on counterfactual counting questions (\qone \& \qtwo) varies with image size across all 7 domains.

\subsec{Results}  Performance remains remarkably consistent across resolutions (15.43\% at 384px $\rightarrow$ 18.28\% at 1152px; \cref{tab:resolution_analysis}). These consistent patterns across resolutions reinforce that VLM failures stem from memorized knowledge overriding visual analysis rather than insufficient image detail.

\subsection{Humans~\humanlogo can count animal legs almost perfectly after 2 seconds of analyzing the image}
\label{sec:human_study}

To establish performance baselines and validate that our counterfactual images are not inherently ambiguous, we investigate human performance on \biastest under various time constraints. Understanding human capabilities provides crucial context for interpreting VLM failures and confirms whether the visual modifications are perceivable given sufficient examination time.

\subsec{Experiments} We conduct a \emph{anonymous} human study (consent obtained, no PII collected, minimal risk) with 78 participants (mean age 24.4 years, 82.1\% with Bachelor's degree or higher, men 51.6\%, women 46.2\%) who completed the \animalicon~animal leg counting task through our project website. Each participant is randomly assigned to one image viewing time condition (see~\cref{fig:human_study_instrution_page}) throughout the session to answer 10 randomly selected questions (5 original, 5 counterfactual) from our 91-image dataset on \animalicon~animals (\cref{sec:animal_add_legs}). We vary image viewing times (\cref{fig:human_study_image_viewing_page}) across four conditions: 0.2, 0.5, 1.0, and 2.0 seconds, while allowing unlimited time for reading questions (\cref{fig:human_study_question_page}) and responding (\cref{fig:human_study_answer_page}).

\begin{figure}[h!]
    \centering
    \includegraphics[width=0.5\textwidth]{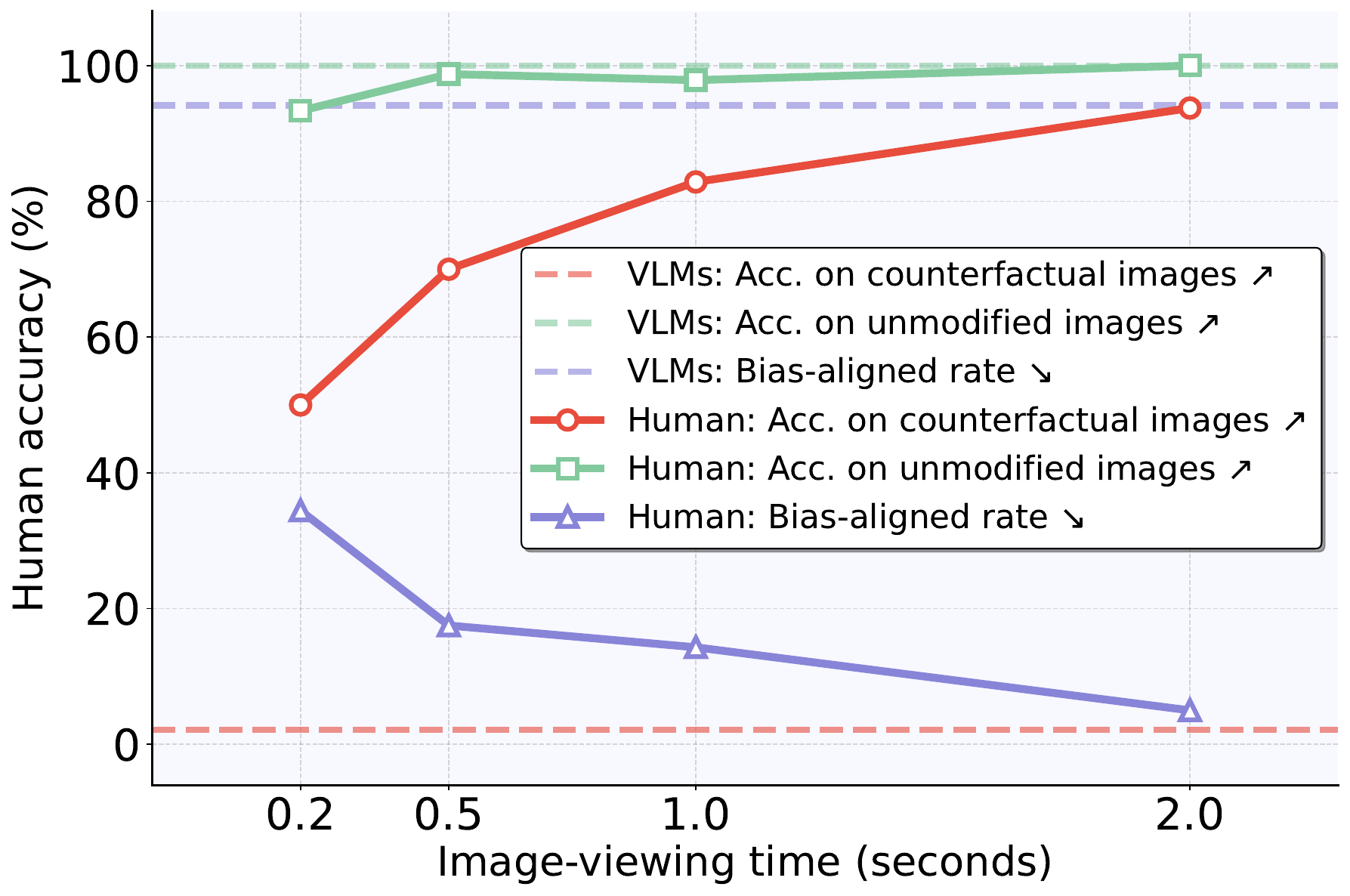}
    \caption{Human accuracy on counterfactual images significantly outperforms VLMs (reaching 93.75\% vs 2.12\%) on \animalicon~animals with longer image-viewing times, while bias-aligned responses decrease substantially with extended exposure.}
    \label{fig:human_study}
\end{figure}


\begin{table}[h!]
    \centering
    \caption{Human accuracy increases with image-viewing time, reaching 96.88\% at 2.0 seconds on \animalicon~animals. Even under severe time pressure (0.2s), humans achieve 50.00\% on counterfactual images significantly better than VLMs (mean accuracy of 2.21\%) on \animalicon.}
    \resizebox{\textwidth}{!}{%
    \begin{tabular}{lcccccc}
    \toprule
     & Image-viewing time & Participants & \multicolumn{2}{c}{Accuracy (\%)} & Overall & Bias \\
    \cmidrule(lr){4-5}
     & (seconds) & (\#) & Counterfactual & Original & accuracy (\%) & rate (\%) \\
    \midrule
    \multirow{4}{*}{\humanlogo~Human} & 0.2 & 18 & 50.00 & 93.33 & 71.67 & 34.44 \\
     & 0.5 & 16 & 70.00 & 98.75 & 84.38 & 17.50 \\
     & 1.0 & 28 & 82.86 & 97.86 & 90.36 & 14.29 \\
     & 2.0 & 16 & \textbf{93.75} & \textbf{100.00} & \textbf{96.88} & \textbf{5.00} \\
    \midrule
    SOTA VLMs (\gptfulllogo\gptminilogo\gptfourlogo\geminilogo\newsonnetlogo) & -- & -- & 2.12 & \textbf{100.00} & 51.06 & 94.14 \\
    \bottomrule
    \end{tabular}%
    }
    \label{tab:human_study}
\end{table}


\subsec{Results} 
Human~\humanlogo counting accuracy on \animalicon~animals improves dramatically with increased image-viewing time  (71.67\% at 0.2 seconds $\rightarrow$ 96.88\% at 2.0 seconds; \cref{tab:human_study,fig:human_study}). On counterfactual images specifically, accuracy also rises from 50.00\%  at 0.2s to 93.75\% at 2.0s. 

Under extreme time pressure (0.2 seconds), humans exhibit higher bias-aligned responses (34.44\%; \cref{tab:human_study,fig:human_study}) compared to longer viewing times. But even in this challenging condition, humans still outperform SOTA VLMs (50.00\% vs. 2.12\% counterfactual accuracy; \cref{tab:human_study}). 
This confirms that our counterfactual images are not inherently ambiguous and complex for humans.

\subsection{Locate-then-count  prompting does not significantly improve counting accuracy}
While simple counting prompts prove ineffective (\qone \& \qtwo; \cref{appsec:appendix_questions}). the strong performance of pointing VLMs like \moondream, \molmosmall (\hallucinatepanel{36.02\%} accuracy; \cref{sec:pointing_vlms}) suggests that forcing explicit localization might be key. We investigate whether more descriptive, step-by-step prompts that first locate, then count (\ie, explicitly instruct VLMs first to locate each element, then count one by one) can help VLMs overcome their bias and improve counting accuracy.

\subsec{Experiments} We replicate the \animalicon~animal leg counting experiment from \cref{sec:result_combined} but modify the prompt to encourage a more procedural approach, using the following enhanced prompt: ``\emph{First, locate each leg individually, count them one by one, and then state the final number in curly brackets, e.g., \{9\}}.'' This prompt explicitly guides the model through a localization-then-counting workflow rather than asking for a direct count.

\begin{table}[h]
\caption{Locate-then-count prompting yields only marginal improvements over \qone \& \qtwo prompts (\increasenoparent{0.67\%} accuracy, \decreasenoparentgreen{1.09\%} bias rate).}
\centering
\resizebox{0.9\textwidth}{!}{
\begin{tabular}{lccccc}
\toprule
& \multicolumn{2}{c}{Accuracy} & \multicolumn{2}{c}{Bias rate} \\
\cmidrule(lr){2-3}\cmidrule(lr){4-5}
Model
  & \qone \& \qtwo
  & Locate-then-count prompt ($\Delta$)
  & \qone \& \qtwo
  & Locate-then-count prompt ($\Delta$) \\
\midrule
\geminilogo~\geminifull 
  & 0.00 
  & 0.00 \increase{0.00}
  & 100.00 
  & 96.70 \decreasegreen{3.30} \\
\newsonnetlogo~\newsonnet 
  & 0.00  
  & 1.83 \increase{1.83}
  & 100.00 
  & 98.17 \decreasegreen{1.83} \\
\gptfourlogo~\gptfour 
  & 9.52  
  & 10.62 \increase{1.10}
  & 79.67 
  & 82.78 \increasered{3.11} \\
\gptfulllogo~\gptfull 
  & 0.92  
  & 1.54 \increase{0.62}
  & 93.77 
  & 93.08 \decreasegreen{0.69} \\
\gptminilogo~\gptmini
  & 0.18  
  & 0.00 \decrease{0.18}
  & 97.25 
  & 94.51 \decreasegreen{2.74} \\
\midrule
Mean 
  & 2.12  
  & 2.80 \increase{0.67}
  & 94.14 
  & 93.05 \decreasegreen{1.09} \\
\bottomrule
\end{tabular}
}
\label{tab:descriptive_cot_res}
\end{table}

\subsec{Results} Locate-then-count prompting yields only marginal improvements over the original simple prompts \qone \& \qtwo (\increasenoparent{0.67}; \cref{tab:descriptive_cot_res}), while the bias rate remains high (\decreasenoparentgreen{1.09}; \cref{tab:descriptive_cot_res}). \textbf{These results indicate that explicitly locate-then-count instructions are insufficient to overcome VLMs' strong visual bias} (see \cref{appfig:procedural_prompt_stork_qualitative,appfig:procedural_prompt_lion_qualitative}), consistent with findings that prompting-based interventions provide only limited improvements (\cref{sec:helpful_prompts}). Instead, the correct way to help is by providing tools for VLMs and ensuring that VLMs themselves know when to use them (see \cref{appsec:tool_use_vlms,sec:pointing_vlms}).

\subsection{Adding subject name to text prompts further decreases VLM accuracy}

Our \biastest uses neutral prompts (\eg, ``\emph{Count the legs of this animal.}'') to isolate visual bias from prompt bias. However, a key question remains: \emph{does this neutral framing actually matter?} To address this, we test whether injecting object-specific names into our prompts (\ie, non-neutral; similar to \cref{sec:res_title}) affects VLM counting performance and bias rates.

\subsec{Experiments} We modify our neutral prompts (\qone \& \qtwo) from previous experiments by replacing generic descriptors with specific object names. For example, ``the left shoe'' becomes ``the left \emph{Nike} shoe'' (\logoicon~logos), ``this puzzle'' becomes ``this \emph{Sudoku} puzzle'' (\boardgameicon~game boards). We evaluate \newsonnetlogo~\newsonnet (best non-thinking) and \gptminilogo~\gptmini (best thinking) using these non-neutral prompts on the same counterfactual images across all 7 tasks.

\begin{table}[h]
    \centering
    \caption{Non-neutral prompts substantially reduce counting \textbf{accuracy} (\decreasenoparent{4.75}), with \gptminilogo~\gptmini experiencing 3$\times$ larger degradation than \newsonnetlogo~\newsonnet (\decreasenoparent{7.09} vs. \decreasenoparent{2.41} points) across 7 tasks.}
    \label{tab:acc_neutral_vs_non_neutral}
    \resizebox{\textwidth}{!}{%
    \begin{tabular}{lcccccccc}
        \toprule
        Model 
        & a.\animalicon 
        & b.\logoicon 
        & c.\flagicon 
        & d.\chessicon 
        & e.\boardgameicon 
        & f.\illusionicon 
        & g.\patternedgridicon 
        & Task mean \\
        \midrule
        \newsonnetlogo~\newsonnet  (Neutral)
        & 0.00
        & 2.72
        & 13.75
        & 9.03
        & 1.79
        & 54.29
        & 34.52
        & 16.59
        \\
       \newsonnetlogo~\newsonnet  (Non-neutral)
        & 0.00 \increase{0.00}
        & 1.98 \decrease{0.74}
        & 9.58 \decrease{4.17}
        & 2.43 \decrease{6.60}
        & 1.79 \increase{0.00}
        & 49.87 \decrease{4.42}
        & 33.63 \decrease{0.89}
        & 14.18 \decrease{2.41}
        \\
        \midrule
        \gptminilogo~\gptmini (Neutral)
        & 0.18
        & 9.31
        & 14.58
        & 44.10
        & 4.76
        & 51.26
        & 17.56
        & 20.25
        \\
        \gptminilogo~\gptmini (Non-neutral)
        & 0.18 \increase{0.00}
        & 8.09 \decrease{1.23}
        & 5.42 \decrease{9.17}
        & 15.62 \decrease{28.47}
        & 0.00 \decrease{4.76}
        & 50.00 \decrease{1.26}
        & 12.80 \decrease{4.76}
        & 13.16 \decrease{7.09}
        \\

        \midrule
        Model Mean (Neutral)
        & 0.09
        & 6.01
        & 14.17
        & 26.56
        & 3.27
        & 52.78
        & 26.04
        & 18.42
        \\
        Model Mean (Non-neutral)
        & 0.09 \increase{0.00}
        & 5.03 \decrease{0.98}
        & 7.50 \decrease{6.67}
        & 9.03 \decrease{17.53}
        & 0.89 \decrease{2.38}
        & 49.94 \decrease{2.84}
        & 23.21 \decrease{2.83}
        & 13.67 \decrease{4.75}
        \\
        \bottomrule
    \end{tabular}%
    }
\end{table}


\begin{table}[h]
    \centering
    \caption{Non-neutral prompts increase \textbf{bias rates} across all tasks (\increasenoparentred{5.32}), demonstrating that object-specific names strongly activate textual priors.}
    \label{tab:bias_neutral_vs_non_neutral}
    \resizebox{\textwidth}{!}{%
    \begin{tabular}{lcccccccc}
        \toprule
        Model 
        & a.\animalicon 
        & b.\logoicon 
        & c.\flagicon 
        & d.\chessicon 
        & e.\boardgameicon 
        & f.\illusionicon 
        & g.\patternedgridicon 
        & Task mean \\
        \midrule
        \newsonnetlogo~\newsonnet (Neutral)
        & 100.0
        & 96.79
        & 82.5
        & 84.72
        & 97.62
        & 45.33
        & 29.46
        & 76.63
        \\
        \newsonnetlogo~\newsonnet (Non-neutral)
        & 99.82 \decreasegreen{0.18}
        & 97.77 \increasered{0.98}
        & 88.33 \increasered{5.83}
        & 97.57 \increasered{12.85}
        & 98.21 \increasered{0.60}
        & 47.22 \increasered{1.89}
        & 31.25 \increasered{1.79}
        & 80.03 \increasered{3.40}
        \\
        \midrule
        \gptminilogo~\gptmini (Neutral)
        & 97.25
        & 90.20
        & 82.08
        & 54.17
        & 91.67
        & 48.74
        & 51.49
        & 73.66
        \\
        \gptminilogo~\gptmini (Non-neutral)
        & 97.25 \increasered{0.00}
        & 88.24 \decreasegreen{1.96}
        & 89.58 \increasered{7.50}
        & 84.38 \increasered{30.21}
        & 98.81 \increasered{7.14}
        & 50.00 \increasered{1.26}
        & 58.04 \increasered{6.55}
        & 80.90 \increasered{7.24}
        \\
        \midrule
        Model Mean (Neutral)
        & 98.63
        & 93.49
        & 82.29
        & 69.44
        & 94.64
        & 47.03
        & 40.48
        & 75.14
        \\
        Model Mean (Non-neutral)
        & 98.53 \decreasegreen{0.09}
        & 93.00 \decreasegreen{0.49}
        & 88.96 \increasered{6.67}
        & 90.97 \increasered{21.53}
        & 98.51 \increasered{3.87}
        & 48.61 \increasered{1.58}
        & 44.64 \increasered{4.17}
        & 80.46 \increasered{5.32}
        \\
        \bottomrule
    \end{tabular}%
    }
\end{table}

\subsec{Results} Adding object names to prompts significantly degrades performance (\decreasenoparent{4.75\%} mean accuracy; \cref{tab:acc_neutral_vs_non_neutral}) and increases bias rates (\increasenoparentred{5.32\%}; \cref{tab:bias_neutral_vs_non_neutral}) for both \newsonnetlogo~\newsonnet and \gptminilogo~\gptmini. Notably, the thinking model \gptminilogo suffers nearly 3$\times$ larger accuracy degradation than the non-thinking \newsonnetlogo when exposed to non-neutral prompts (\decreasenoparent{7.09} vs. \decreasenoparent{2.41} points). \textbf{These results demonstrate that non-neutral prompts invoke stronger textual priors that override visual information, and even extended reasoning capabilities overcome this bias.} This confirms that neutral prompting is essential for fairly assessing whether VLMs can overcome their language bias when analyzing counterfactual images.

\subsection{VLMs fail to detect modifications even with side-by-side comparison}
Prior sections show that VLMs struggle to count legs correctly in counterfactual images. Here, we test whether providing explicit side-by-side comparisons with original images helps VLMs detect the modifications, as the reference image may make the differences more noticeable.

\subsec{Experiments} We present VLMs with two images simultaneously: the original animal image (with canonical leg count) and its modified counterfactual version (with one extra leg). We prompt models with: \textit{``Compare the two images side by side. Do the animals in image 1 and image 2 have the same number of legs? Return the final Yes/No answer in curly brackets (\eg, \{Yes\} or \{No\}).''}. Here, we expect the VLMs to always answer \{No\} if they can distinguish the differences.

\begin{figure}[h!]
\centering
\begin{minipage}[t]{0.44\textwidth}
    \centering
    \vspace{0pt}
    \includegraphics[width=\textwidth]{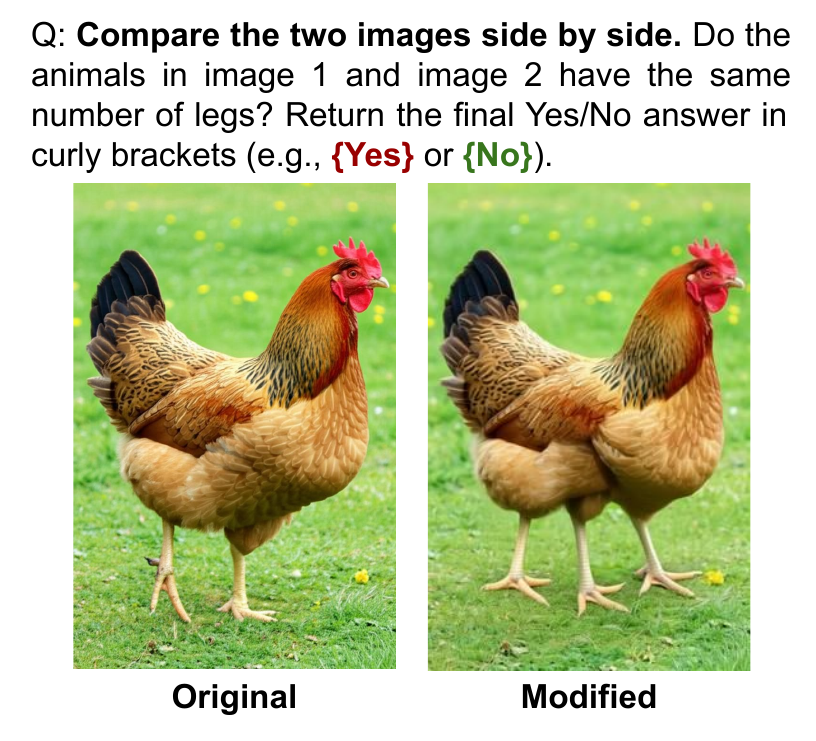}
    \caption{The side-by-side comparison prompt and an example input image pair.}
    \label{fig:compare_prompt}
\end{minipage}
\hfill
\begin{minipage}[t]{0.50\textwidth}
    \centering
    \vspace{10pt}
    \small
    \begin{tabular}{lc}
    \toprule
    Model & Percentage of \{No\} (\%) \\
    Random baseline & 50 \\
    \midrule
    \geminilogo~\geminifull & 9.89 \\
    \newsonnetlogo~\newsonnet & 9.89 \\
    \gptfourlogo~\gptfour & 10.99 \\
    \gptfulllogo~\gptfull & 10.99 \\
    \gptminilogo~\gptmini & \textbf{15.38} \\
    \midrule
    Mean & \hallucinatepanel{11.76}\\
    \bottomrule
    \end{tabular}
    \captionof{table}{VLMs fail to detect leg count differences when comparing original and counterfactual images side-by-side. The ground truth is ``No'', but models output ``No'' only \hallucinatepanel{11.76\%} of the time, far below the 50\% expected from random guessing.}
    \label{tab:comparison_results}
\end{minipage}
\end{figure}

\subsec{Results} \textbf{VLMs are so biased that even when counterfactual and original images are placed side by side, they still cannot detect the modifications.} The mean percentage of \{No\} across 5 SOTA VLMs is only \hallucinatepanel{11.76\%} (\cref{tab:comparison_results}). 
This demonstrates that VLMs' bias toward prior knowledge is so strong that even direct visual comparison fails to surpass the random-guessing (50\%) baseline.


\subsection{Even when attending to correct regions, VLMs still fail to generate correct answers}
Prior sections demonstrate that VLMs fail at counting counterfactual elements despite vision encoders successfully encoding visual information (\cref{sec:vision_encoder_analysis}). Here, we investigate whether VLMs attend to the correct visual regions during inference by analyzing attention patterns when generating answers.

\subsec{Experiments} Our preliminary analysis of attention patterns evolution throughout the layers (\cref{fig:attention_visualization_layers}) reveals that \gwensmalllogo~\gwensmall progressively localizes relevant objects (\eg, legs, logo elements) in later layers. Following this, we compute the final layer's attention mapping of the answer token over the image tokens for \gwensmalllogo~\gwensmall. For example, when the model outputs ``\{3\}'' in response to counting a dog's legs, we extract the attention weights of the token ``3'' across the image tokens of the last layer. We visualize the attention by mapping each image token to its corresponding image patch and overlay the attention heatmap on the original image.

\begin{figure}[ht!]
\centering
\scriptsize
\setlength{\tabcolsep}{2pt}
\renewcommand{\arraystretch}{1.0}
\begin{tabular}{@{}ccccc@{}}
\multicolumn{1}{c}{\textbf{Original}} &
\multicolumn{1}{c}{Layer 7} &
\multicolumn{1}{c}{Layer 14} &
\multicolumn{1}{c}{Layer 21} &
\multicolumn{1}{c}{Layer 28} \\

\includegraphics[width=0.18\textwidth]{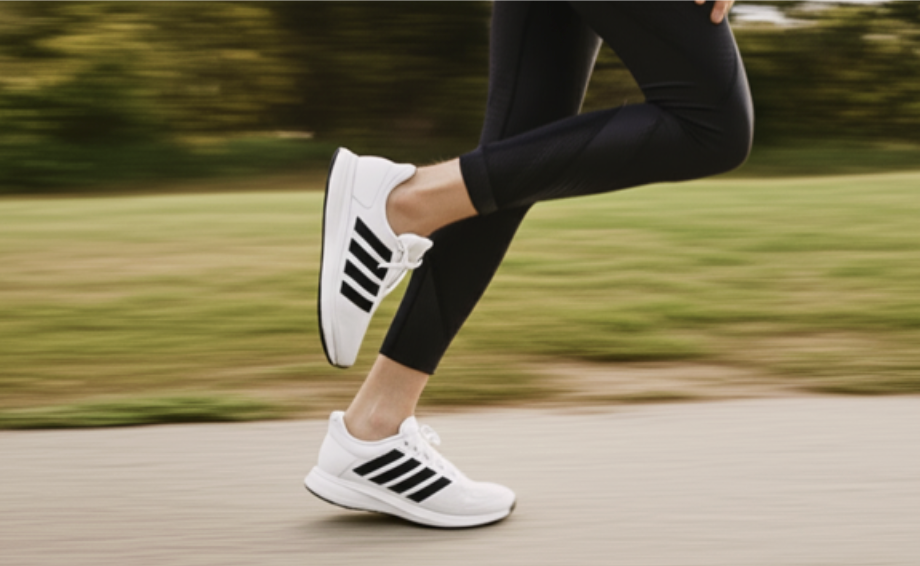} &
\includegraphics[width=0.18\textwidth]{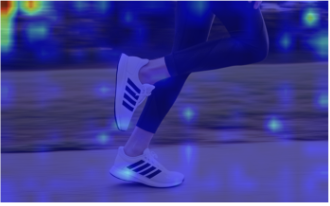} &
\includegraphics[width=0.18\textwidth]{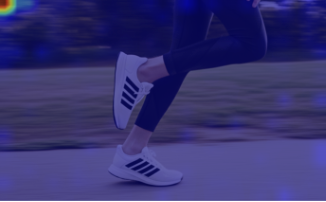} &
\includegraphics[width=0.18\textwidth]{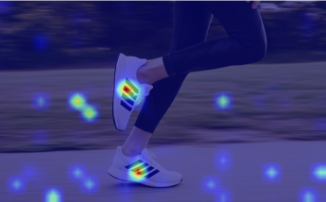} &
\includegraphics[width=0.18\textwidth]{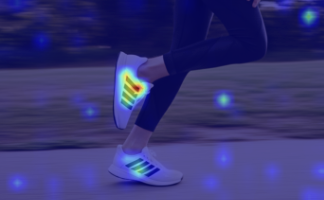} \\
\multicolumn{5}{l}{\logoicon~\textit{Q: Count the visible black stripes in the logo on the left shoe.} {Prediction = 3 \textcolor{red}{\xmark}, Ground truth = 4 \textcolor{ForestGreen}{\cmark}}} \\[0.5em]

\includegraphics[width=0.18\textwidth]{images/attention/flags/original.png} &
\includegraphics[width=0.18\textwidth]{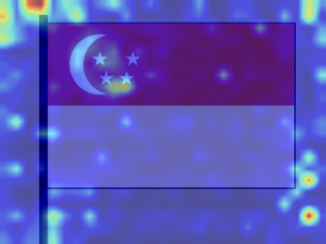} &
\includegraphics[width=0.18\textwidth]{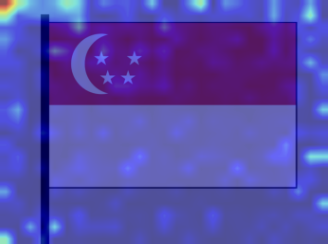} &
\includegraphics[width=0.18\textwidth]{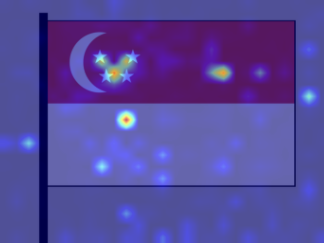} &
\includegraphics[width=0.18\textwidth]{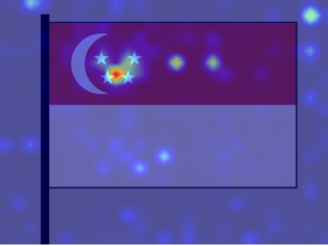} \\
\multicolumn{5}{l}{\flagicon~\textit{Q: Count the stars in this flag.} {Prediction = 5 \textcolor{red}{\xmark}, Ground truth = 4 \textcolor{ForestGreen}{\cmark}}} \\[0.5em]

\includegraphics[width=0.18\textwidth]{images/attention/animals/original.png} &
\includegraphics[width=0.18\textwidth]{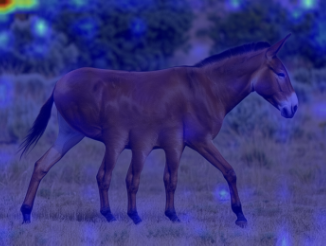} &
\includegraphics[width=0.18\textwidth]{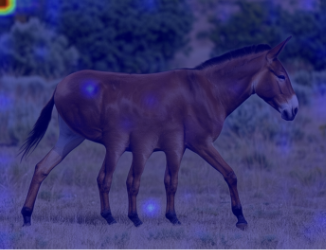} &
\includegraphics[width=0.18\textwidth]{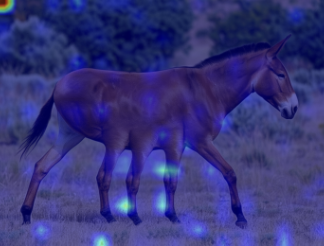} &
\includegraphics[width=0.18\textwidth]{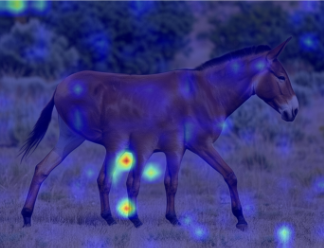} \\
\multicolumn{5}{l}{\animalicon~\textit{Q: Count the legs of this animal.}  {Prediction = 4 \textcolor{red}{\xmark}, Ground truth = 5 \textcolor{ForestGreen}{\cmark}}}

\end{tabular}

\caption{
Attention heatmaps across layers for \gwensmalllogo~\gwensmall, revealing that it progressively localizes relevant regions in later layers. 
\textbf{Original}: Input image without attention overlay. 
\textbf{Layers 7-28}: Attention heatmaps overlaid on images, with warmer colors indicating higher attention weights. 
}
\label{fig:attention_visualization_layers}
\end{figure}

\begin{figure}[h]
\centering
\scriptsize
\setlength{\tabcolsep}{4pt}
\renewcommand{\arraystretch}{1.0}

\begin{tabular}{@{}ccc@{}}
\includegraphics[width=0.31\textwidth]{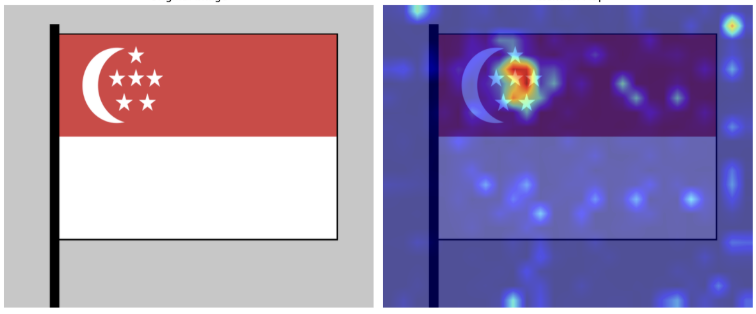} &
\includegraphics[width=0.31\textwidth]{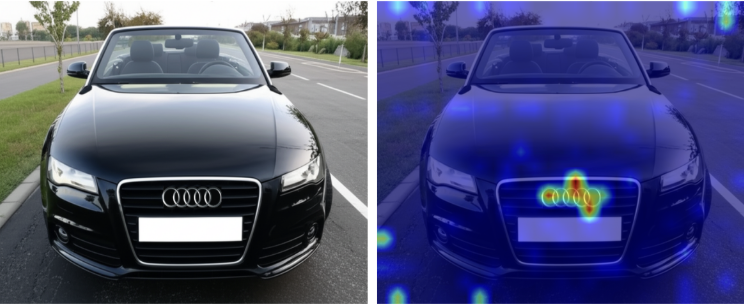} &
\includegraphics[width=0.31\textwidth]{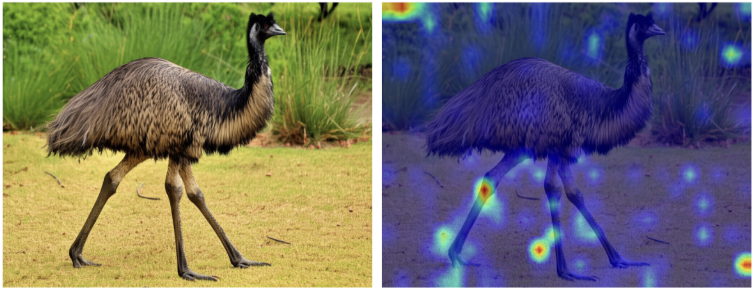} \\
Prediction = 5 \textcolor{red}{\xmark}, Ground truth = 6 \textcolor{ForestGreen}{\cmark} &
Prediction = 4 \textcolor{red}{\xmark}, Ground truth = 5 \textcolor{ForestGreen}{\cmark} &
Prediction = 2 \textcolor{red}{\xmark}, Ground truth = 3 \textcolor{ForestGreen}{\cmark} \\[1em]

\includegraphics[width=0.31\textwidth]{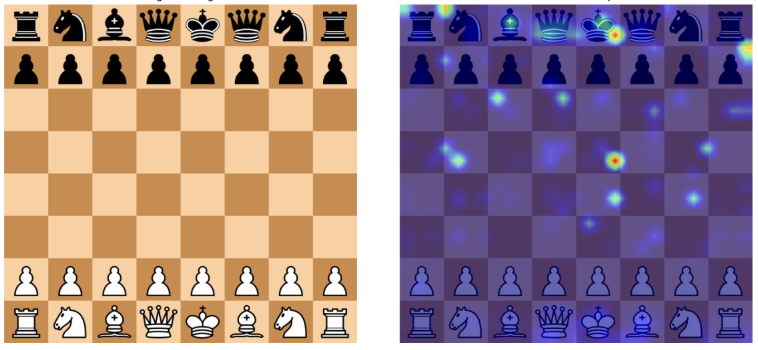} &
\includegraphics[width=0.31\textwidth]{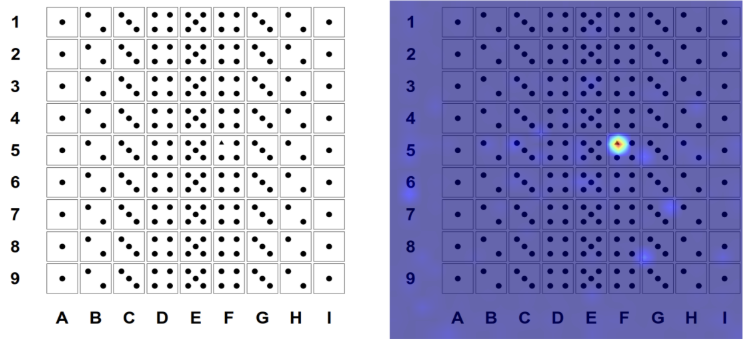} &
\includegraphics[width=0.31\textwidth]{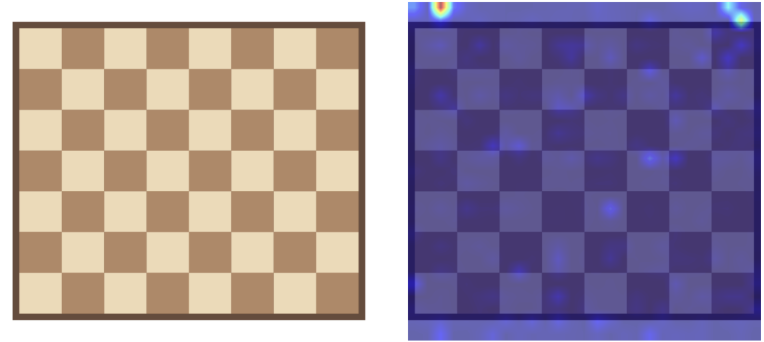} \\
Prediction = 2 \textcolor{red}{\xmark}, Ground truth = 3 \textcolor{ForestGreen}{\cmark} &
Prediction = 7 \textcolor{red}{\xmark}, Ground truth = 3 \textcolor{ForestGreen}{\cmark} &
Prediction = 8 \textcolor{red}{\xmark}, Ground truth = 7 \textcolor{ForestGreen}{\cmark} \\

\end{tabular}
\caption{Attention heatmaps from the final layer of the prediction token of \gwensmalllogo~\gwensmall. The model correctly attends to the visual details for flags, car logos, patterned grids, and three-legged animals when generating its answer token. However, it still outputs incorrect or biased answers instead of the correct count.}
\label{fig:attention_visualization}
\end{figure}

\subsec{Results} Interestingly, even when \textbf{the model correctly attend to the regions of interest, it often produces incorrect or biased answers}. For instance, when counting overlapping circles on a modified Audi logo, \gwensmalllogo~\gwensmall attends strongly to all five circles in the final layer yet outputs ``4''. This finding is consistent with prior work showing disconnects between visual attention and final model outputs~\citep{liu2025seeingbelievingprobingdisconnect, zhang2025mllms}. Combined with our linear probing results (\cref{sec:vision_encoder_analysis}), this provides strong evidence that VLMs can \emph{see} the correct visual information but are highly influenced by memorized knowledge priors during answer generation.



\clearpage
\section{Human study details}\label{appsec:appendix_human_study}

\begin{figure}[h!]
    \centering
    \includegraphics[width=0.8\textwidth]{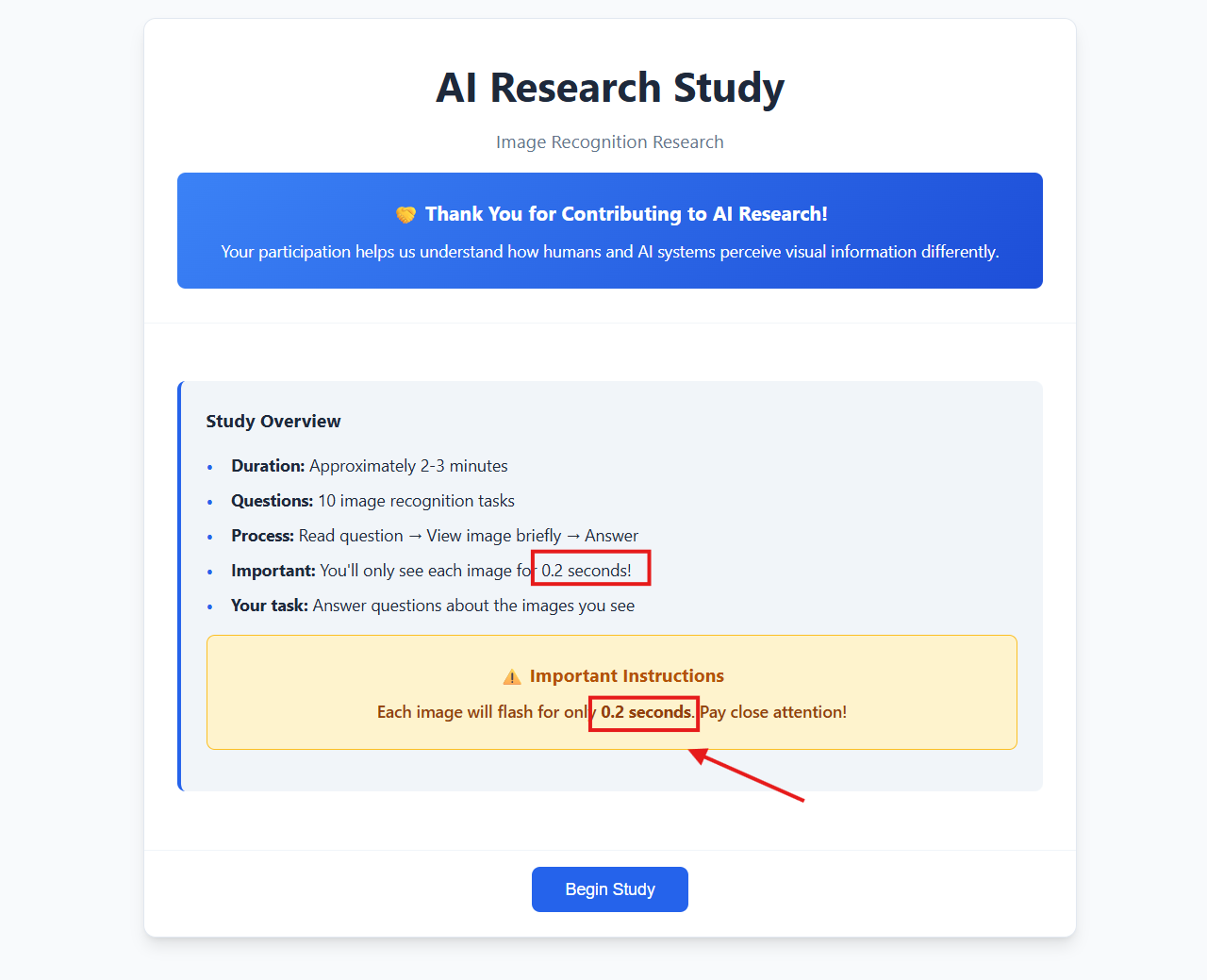}
    \caption{Participants are informed about the task and their randomly assigned image viewing duration (0.2, 0.5, 1, or 2 seconds).}
    \label{fig:human_study_instrution_page}
\end{figure}



\begin{figure}[h!]
    \centering
    \includegraphics[width=0.8\textwidth]{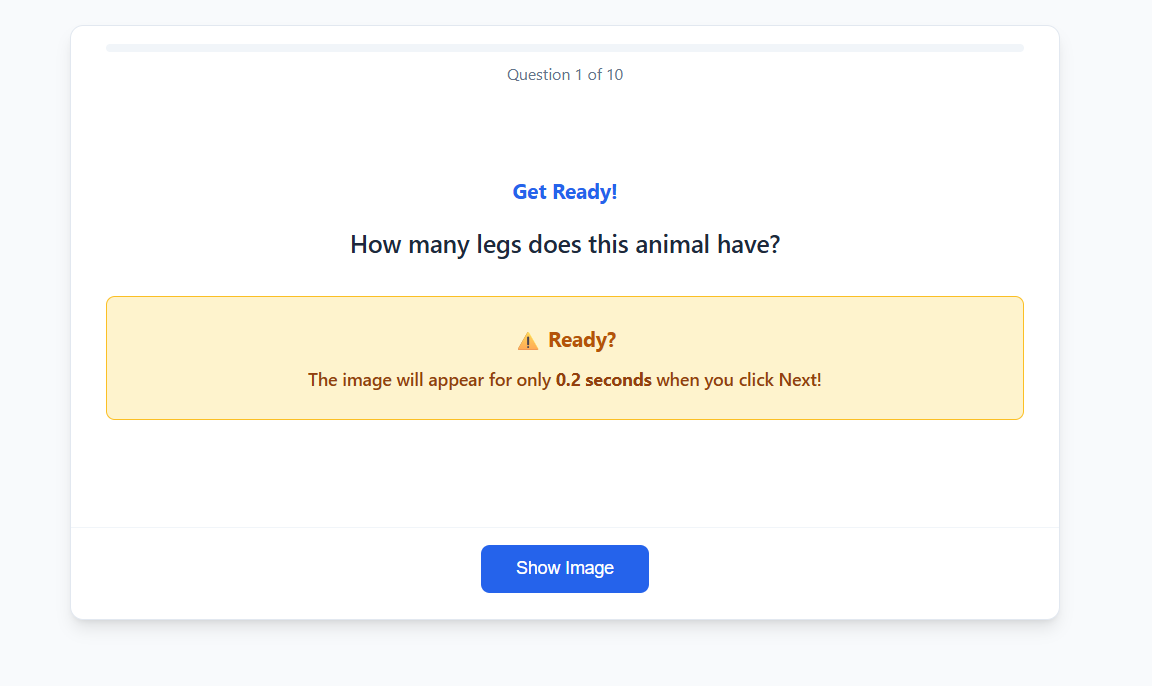}
    \caption{Participants read the question with unlimited time before viewing the image.}
    \label{fig:human_study_question_page}
\end{figure}



\begin{figure}[h!]
    \centering
    \includegraphics[width=0.8\textwidth]{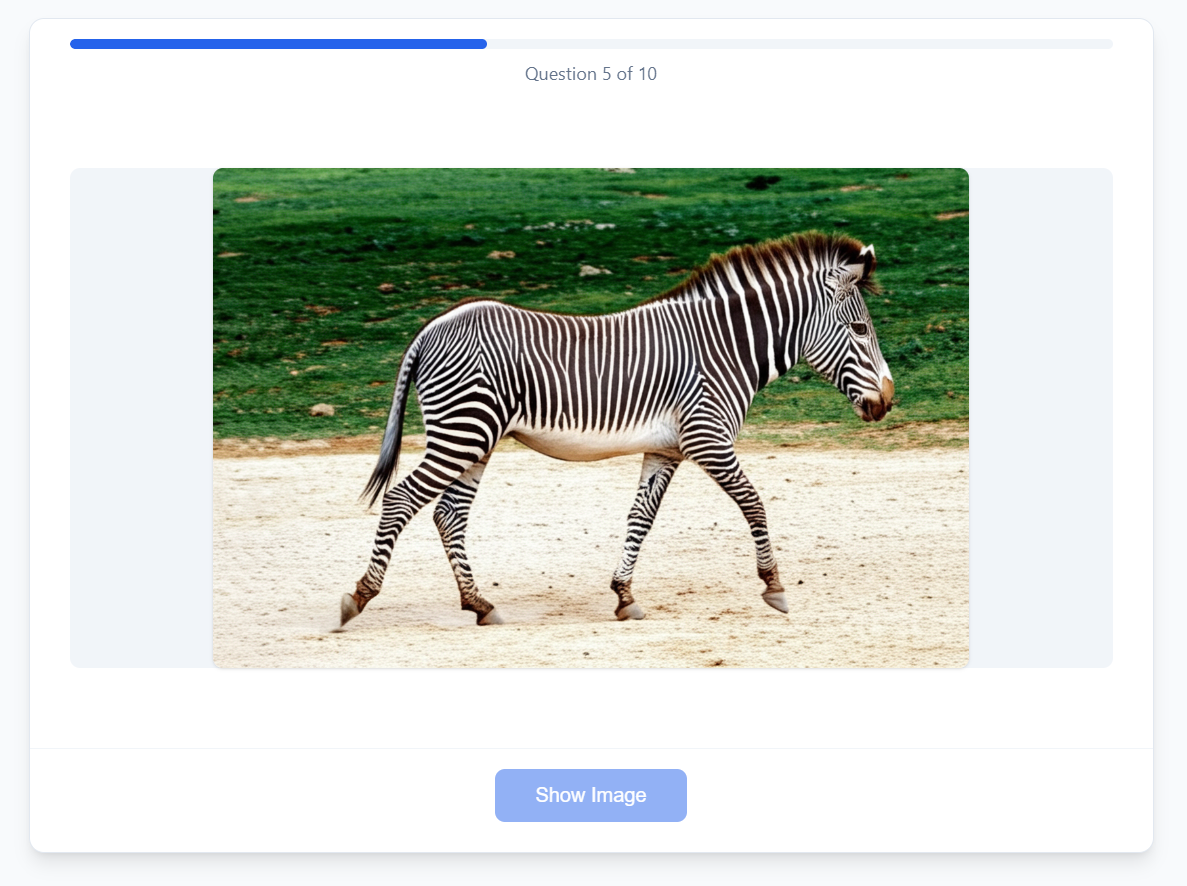}
    \caption{The target image is displayed for the assigned duration (0.2, 0.5, 1, or 2 seconds).}
    \label{fig:human_study_image_viewing_page}
\end{figure}



\begin{figure}[h!]
    \centering
    \includegraphics[width=0.8\textwidth]{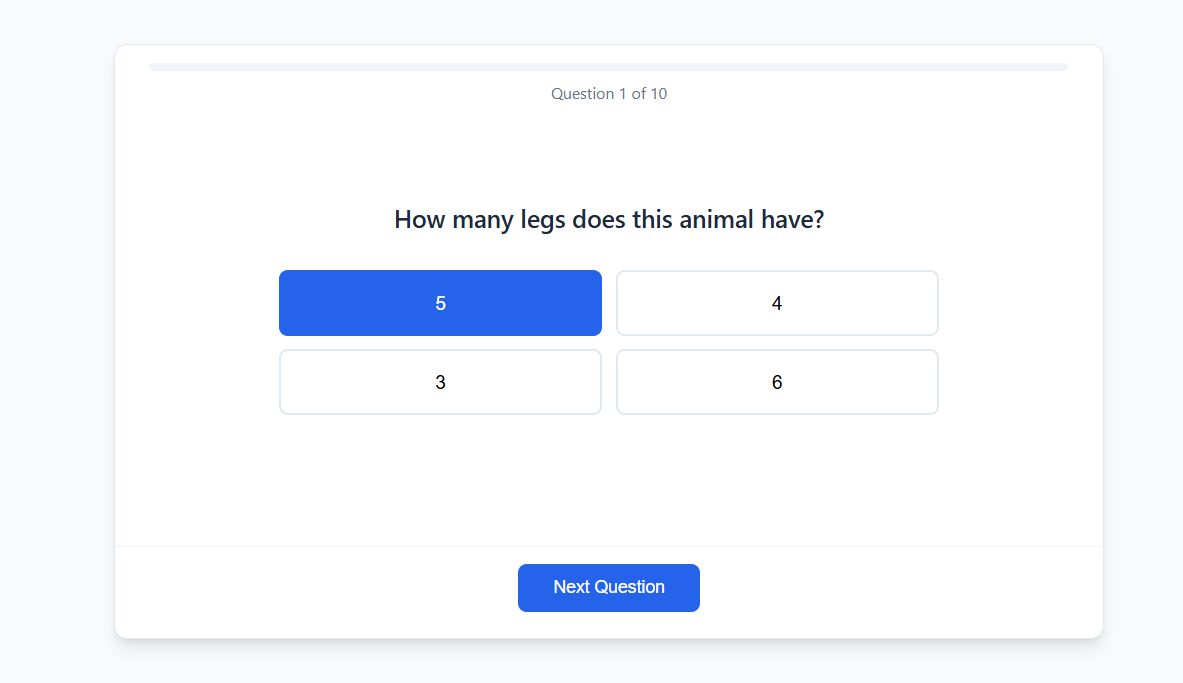}
    \caption{Participants have unlimited time to choose their response from multiple-choice options.}
    \label{fig:human_study_answer_page}
\end{figure}


\clearpage
\section{Detailed comparison with existing VLM bias benchmarks}
\label{appsec:related_works_in_details}

This section provides additional detailed comparison between \biastest and related benchmarks (\cref{tab:compare_datasets}) discussed in \cref{sec:related_work}, organized by key evaluation criteria.

\subsection{Source of bias}
\biastest isolates visual bias through neutral prompts and objective counting, while other benchmarks introduce bias through their question formulations. Specifically,
PhD-ccs~\citep{liu2024phd}, VLind-Bench~\citep{lee2024vlind}, and HallusionBench~\citep{liu2023hallusionbench} explicitly mention objects in prompts (\eg, ``\textit{Does the car have square wheels?}''), priming models toward knowledge priors.
ViLP~\citep{luo2024probing} contains two subsets: ViLP$^{F}$ includes distractor facts that bias responses, while ViLP$^{P}$ omits distractors.
Though ViLP$^{P}$ is more relevant to our work, it doesn't directly address visual bias like \biastest. It either uses identification questions (\eg, \textit{``Which animal in the image stores fat in its humps?''}) on modified subjects (horses with humps), which are inherently ambiguous as a horse with humps arguably ceases to be a ``horse''; or questions that explicitly mention the object (\eg, \textit{``From the image, in which city is the Red Square located?''}), priming models toward prior knowledge of these named entities (\eg, \textit{the Red Square}). While there are also counting questions in ViLP, they take up only 4\% of the questions (12/300), compared to our benchmark which fully focused on counting.

In contrast, \biastest uses neutral language (\eg, ``\textit{How many legs does this animal have?}'') with objective counting that results in unambiguous numerical answers. This design ensures that failures indicate memorized knowledge overriding visual evidence, not susceptibility to cues in the questions.

\subsection{Benchmark scale}
The main dataset of \biastest provides 1,392 counterfactual images across 7 diverse tasks, exceeding most related benchmarks in scale.
Specifically, our main dataset is 1.9 times larger than PhD-ccs (750 images), 2.3 times larger than ViLP (600 images), and 7.7 times larger than HallusionBench (181 images).
While VLind-Bench (2,576 images) is larger than our main dataset, \biastest's full evaluation suite which includes the background removal subset and in-image text injection subset totals 4,176 images, surpassing the scale of VLind-Bench.
This scale enables more robust evaluation of VLMs, covering a broad range of scenarios from photo-realistic animals to abstract patterns.

\subsection{Image generation method}
\biastest systematically generates photo-realistic, subtly modified versions of familiar subjects using state-of-the-art models, while other benchmarks (1) use older image generators producing surreal-looking images or (2) manually collect images. Specifically, PhD-ccs and VLind-Bench rely on DALL-E, while ViLP uses DALL-E and FLUX to create artificial and surreal scenes. Meanwhile,
HallusionBench manually curates counterfactual images, achieving high-quality but lacking scalability.
In contrast, \biastest employs state-of-the-art generators (\geminiflashlogo~\geminiflash, \gptfourlogo~\gptfouro) to create subtle modifications of highly familiar subjects (\eg a 5-legged dog) that looks highly realistic.


\clearpage
\section{Models and access details} \label{appsec:appendix_model_settings}

\begin{table}[h!]
\centering
\caption{Model specifications and access details for evaluated commercial VLMs}
\label{tab:commercial_model_details}

\resizebox{1.0\textwidth}{!}{%
\begin{tabular}{llcll}
\toprule
\textbf{Model} & \textbf{Model ID} & \textbf{Thinking} & \textbf{Platform} & \textbf{Settings} \\
\midrule
\geminilogo~\geminifull & \model{gemini-2.5-pro-preview-05-06} & \greentick  & \href{aistudio.google.com}{Google AI Studio} & \texttt{temperature}=1.0 \\
\newsonnetlogo~\newsonnet & \model{claude-3-7-sonnet} & \redxmark & \href{console.anthropic.com}{Anthropic} & \texttt{temperature}=1.0 \\
\gptfourlogo~\gptfour & \model{gpt-4.1}
 & \redxmark & \href{platform.openai.com}{OpenAI} & \texttt{temperature}=1.0 \\
\gptfulllogo~\gptfull & \model{o3} & \greentick & \href{platform.openai.com}{OpenAI} & \texttt{reasoning\_effort}=medium \\
\gptminilogo~\gptmini & \model{o4-mini} & \greentick & \href{platform.openai.com}{OpenAI} & \texttt{reasoning\_effort}=medium \\
\groklogo~\grok & \model{grok-4} & \greentick & \href{https://console.x.ai}{xAI} & -- \\
\gptfivelogo~\gptfive & \model{gpt-5} & \greentick & \href{platform.openai.com}{OpenAI} & \texttt{reasoning\_effort}=medium \\

\bottomrule
\end{tabular}
}
\end{table}


\begin{table}[h!]
\centering
\caption{Model specifications and access details for evaluated open-source VLMs}
\label{tab:opensource_model_details}

\resizebox{1.0\textwidth}{!}{%
\begin{tabular}{llcll}
\toprule
\textbf{Model} & \textbf{Model ID} & \textbf{Thinking} & \textbf{Platform} & \textbf{Settings} \\
\midrule
\pixtralsmalllogo~\pixtralsmall & \model{pixtral-12b} & \redxmark  & \href{https://openrouter.ai/}{OpenRouter} & \texttt{temperature}=1.0 \\
\pixtrallargelogo~\pixtrallarge & \model{pixtral-large-2411} & \redxmark  & \href{https://openrouter.ai/}{OpenRouter} & \texttt{temperature}=1.0 \\
\gwensmalllogo~\gwensmall & \model{qwen-2.5-vl-7b-instruct} & \redxmark  & \href{https://openrouter.ai/}{OpenRouter} & \texttt{temperature}=1.0 \\
\gwenlargelogo~\gwenlarge & \model{qwen2.5-vl-72b-instruct} & \redxmark  & \href{https://openrouter.ai/}{OpenRouter} & \texttt{temperature}=1.0 \\
\bottomrule
\end{tabular}
}
\end{table}

\begin{table}[h!]
\centering
\caption{Model specifications and access details for evaluated open-source counting VLMs}
\label{tab:pointing_model_details}

\resizebox{\textwidth}{!}{%
\begin{tabular}{llcll}
\toprule
\textbf{Model} & \textbf{Model ID} & \textbf{Text output} & \textbf{Platform} & \textbf{Settings} \\
\midrule
\molmosmalllogo~\molmosmall & \model{allenai/Molmo-7B-D-0924} & \greentick & \href{https://huggingface.co}{HuggingFace} & \texttt{temperature}=1.0 \\
\molmolargelogo~\molmolarge & \model{allenai/Molmo-72B-0924} & \greentick & \href{https://huggingface.co}{HuggingFace} & \texttt{temperature}=1.0 \\
 \moondreamlogo~\moondream & \model{vikhyatk/moondream2} & \redxmark & \href{https://huggingface.co}{HuggingFace} & \texttt{--} \\
\bottomrule
\end{tabular}
}
\end{table}





\clearpage

\section{Task 1: Counting legs with added limb \animalicon}\label{sec:appendix_animals}
\subsection{Task design}
\label{app:animal-leg-count}

\begin{figure}[h!]
    \centering
    \includegraphics[width=0.9\textwidth]{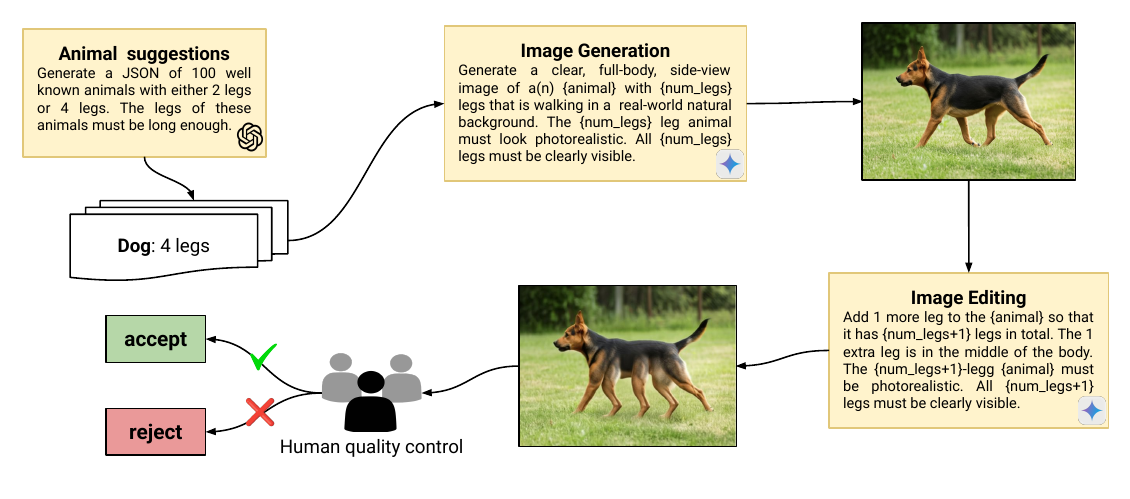}
    \caption{Data generation pipeline for Task 1: Counting legs with added limb.
    }
    \label{fig:animal_diagram}
\end{figure}

Pretrained on the Internet data, VLMs must have colossal prior knowledge of the count of \animalicon animal legs from both textual and image data. Following this hypothesis, we generate images of usual animals with \emph{one additional leg} (\eg, 3-legged birds or 5-legged dogs) and ask VLMs to count legs to evaluate if these models are biased toward their prior knowledge. 

\begin{itemize} 
\item \textbf{\colorbox{blue!15}{Animal types}}: We modify the legs of \textcolor{blue!80}{\textbf{2}} types of animals: birds and mammals.

\item \textbf{\colorbox{green!15}{Modification types}}: Each animal is modified to have \textcolor{green!80}{\textbf{1}} additional leg.
    
\item \textbf{\colorbox{orange!15}{Target animals}}: We select \textcolor{orange!80}{\textbf{91}} well-known animals, consisting of 23 two-legged birds and 68 four-legged mammals.

\item \textbf{\colorbox{purple!15}{Image resolutions}}: We generate each animal image and rescale them at \textcolor{purple!80}{\textbf{3}} different pixel sizes \{384, 768, 1152\}px using the scaling factor in \cref{sec:animal_add_legs} to test resolution sensitivity
\end{itemize}

This approach generates a total of \textcolor{orange!80}{\textbf{91}} animals $\times$ \textcolor{green!80}{\textbf{1}} modification type $\times$ \textcolor{purple!80}{\textbf{3}} resolutions = \textbf{273} total images.

\subsection{Implementation and image generation}
\subsec{Implementation details}
Our image generation pipeline follows this sequence:
\begin{enumerate}
\item Use \gptminilogo~\gptmini to collect a list of well-known animals with clearly visible legs
\item Generate full-body and side-view images of these animals using \geminiflashlogo~\geminiflash 
\item For each animal image, use \geminiflashlogo~\geminiflash to add one extra leg to the animal. Each animal image is edited over 4 independent trials.
\item Manually inspect and filter out unsatisfactory images
\item Render each approved image at three different resolutions
\end{enumerate}

\subsec{Quality control} We manually inspect the images to ensure that each modified animal image has exactly one additional leg. For cases that fail (\eg, more than one added leg), we remove them from our dataset.

\subsec{Prompt} We use the following prompts to test the VLMs:
\begin{itemize}
    \item \textbf{Q1:} \emph{How many legs does this animal have? Answer with a number in curly brackets, e.g., \{9\}.}
    \item \textbf{Q2:} \emph{Count the legs of this animal. Answer with a number in curly brackets, e.g., \{9\}.}
    \item \textbf{Q3:} \emph{Is this an animal with [NumModifiedLegs] legs? Answer in curly brackets, e.g., \{Yes\} or \{No\}.}
\end{itemize}

\subsec{Ground truth calculation} The ground truth answers are as follow:
\begin{itemize}
    \item \textbf{Birds leg counting (Q1\&Q2)}:
    \begin{itemize}
        \item Correct answer: 3 (one additional leg)
        \item Expected bias: 2
    \end{itemize}
    \item \textbf{Mammals leg counting (Q1\&Q2)}:
    \begin{itemize}
        \item Correct answer: 5 (one additional leg)
        \item Expected bias: 4
    \end{itemize}
    \item \textbf{Animal leg identification question (Q3)}:
    \begin{itemize}
        \item Correct answer: ``No'' (always, since each animal has one additional leg)
        \item Expected bias: ``Yes''
    \end{itemize}
\end{itemize}

\subsection{Qualitative results}
\begin{figure}[h]
\centering
\small
\begin{AIbox}{How many legs does this animal have? Answer with a number in curly brackets{,} e.g.{,} \{9\}.}

    \raggedright
\footnotesize
      \animalicon~\textbf{(a)-(e)} How many legs does this animal have? Answer with a number in curly brackets, e.g., \{9\}. \\
     \hfill \break
\footnotesize
\centering

    \begin{tabular}{lp{0.2cm}c|p{0.2cm}c|p{0.2cm}c|p{0.2cm}c|p{0.2cm}c}
    &\multicolumn{2}{c}{\makecell{\textbf{(a) Lion}}} 
    &\multicolumn{2}{c}{\makecell{\textbf{(b) Stork}}}
    &\multicolumn{2}{c}{\makecell{\textbf{(c) Elephant}}}
    &\multicolumn{2}{c}{\makecell{\textbf{(d) Duck}}} 
    &\multicolumn{2}{c}{\makecell{\textbf{(e) Dog}}}
    \\
     &\multicolumn{2}{c}{\includegraphics[width=0.15\textwidth]{images/animal_examples/lion_0_0_768.png}}
     &\multicolumn{2}{c}{\includegraphics[width=0.15\textwidth]{images/animal_examples/stork_2_1_768.png}}
     &\multicolumn{2}{c}{\includegraphics[width=0.15\textwidth]{images/animal_examples/african_elephant_2_0_768.png}}
     &\multicolumn{2}{c}{\includegraphics[width=0.15\textwidth]{images/animal_examples/duck_0_3_768.png}}
     &\multicolumn{2}{c}{\includegraphics[width=0.15\textwidth]{images/animal_examples/domestic_dog_4_3_768.png}}
     \\ 

\rowcolor{lightgray}
\raisebox{-0.2\height}\geminilogo & \centering 4
& \textcolor{red}{\xmark} & \centering 2
& \textcolor{red}{\xmark} & \centering 4
& \textcolor{red}{\xmark} & \centering 2
& \textcolor{red}{\xmark} & \centering 4
& \textcolor{red}{\xmark} \\

\raisebox{-0.2\height}\newsonnetlogo & \centering 4
& \textcolor{red}{\xmark} & \centering 2
& \textcolor{red}{\xmark} & \centering 4
& \textcolor{red}{\xmark} & \centering 2
& \textcolor{red}{\xmark} & \centering 4
& \textcolor{red}{\xmark} \\

\rowcolor{lightgray}
\raisebox{-0.2\height}\gptfourlogo & \centering 5
& \textcolor{ForestGreen}{\cmark} & \centering 2
& \textcolor{red}{\xmark} & \centering 6
& \textcolor{red}{\xmark} & \centering 2
& \textcolor{red}{\xmark} & \centering 6
& \textcolor{red}{\xmark} \\

\raisebox{-0.2\height}\gptfulllogo & \centering 4
& \textcolor{red}{\xmark} & \centering 2
& \textcolor{red}{\xmark} & \centering 4
& \textcolor{red}{\xmark} & \centering 2
& \textcolor{red}{\xmark} & \centering 6
& \textcolor{red}{\xmark} \\

\rowcolor{lightgray}
\raisebox{-0.2\height}\gptminilogo & \centering 4
& \textcolor{red}{\xmark} & \centering 2
& \textcolor{red}{\xmark} & \centering 4
& \textcolor{red}{\xmark} & \centering 2
& \textcolor{red}{\xmark} & \centering 4
& \textcolor{red}{\xmark} \\ 

\midrule
\textbf{Bias} & \centering 4
& \textcolor{red}{\xmark} & \centering 2
& \textcolor{red}{\xmark} & \centering 4
& \textcolor{red}{\xmark} & \centering 2
& \textcolor{red}{\xmark} & \centering 4
& \textcolor{red}{\xmark} \\

\rowcolor{lightgray}
\textbf{GT} & \centering 5
& \textcolor{ForestGreen}{\cmark} & \centering 3
& \textcolor{ForestGreen}{\cmark} & \centering 5
& \textcolor{ForestGreen}{\cmark} & \centering 3
& \textcolor{ForestGreen}{\cmark} & \centering 5
& \textcolor{ForestGreen}{\cmark} \\

     \end{tabular}
    \vspace{4pt}
    \centering
    \begin{tabular}{cccccccccccccc}
     \raisebox{-0.1\height}\geminilogo~\gemini 
     & \raisebox{-0.1\height}\newsonnetlogo~\newsonnet 
     & \raisebox{-0.1\height}\gptfourlogo~\gptfour 
     & \raisebox{-0.1\height}\gptfulllogo~\gptfull 
     & \raisebox{-0.1\height}\gptminilogo~\gptmini
     \\
    \end{tabular}

\end{AIbox}
\caption{VLMs are often biased toward the original number of legs \animalicon~animals have, and they tend to answer based on prior knowledge rather than by analyzing the image.}
\label{appfig:qualitative_results_animals}
\end{figure}

\subsection{List of animals}
\begin{AIbox}{Mammals: Four-legged animals}
horse, zebra, donkey, mule, cow, buffalo, yak, water buffalo, deer, elk, moose, reindeer, caribou, gazelle, giraffe, camel, dromedary camel, bactrian camel, llama, alpaca, goat, ibex, mountain goat, pronghorn, bighorn sheep, wild boar, pig, warthog, coyote, lynx, bobcat, leopard, tiger, lion, jaguar, puma, ocelot, caracal, hyena, rabbit, impala, springbok, kudu, eland, wildebeest, okapi, hippopotamus, african elephant, asian elephant, indian rhinoceros, gnu, maned wolf, arctic fox, red fox, fennec fox, red wolf, domestic dog, domestic cat, african wilddog, dingo, jackal, gray wolf, hare, cheetah, antelope, bison, sheep, serval
\end{AIbox}

\begin{AIbox}{Birds: Two-legged animals}
ostrich, emu, rhea, cassowary, heron, stork, crane, egret, ibis, spoonbill, turkey, chicken, rooster, duck, swan, peacock, sandpiper, avocet, stilt, plover, lapwing, oystercatcher, secretary bird
\end{AIbox}

\clearpage
\section{Task 2: Counting elements in modified brand logos \logoicon}\label{sec:appendix_logos}

\begin{figure}[h!]
    \centering
    \includegraphics[width=0.9\textwidth]{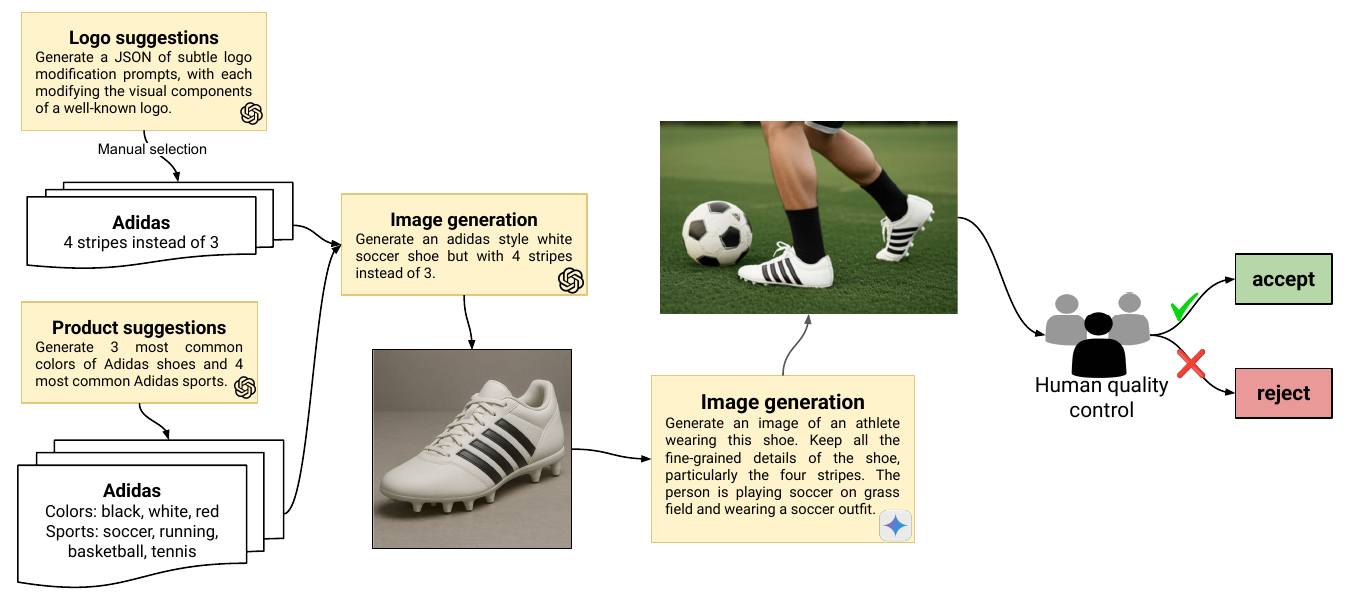}
    \caption{Data generation pipeline of shoe logos for Task 2: Counting elements in modified brand logos
    }
    \label{fig:shoelogo_diagram_updated}
\end{figure}

\begin{figure}[h!]
    \centering
    \includegraphics[width=0.9\textwidth]{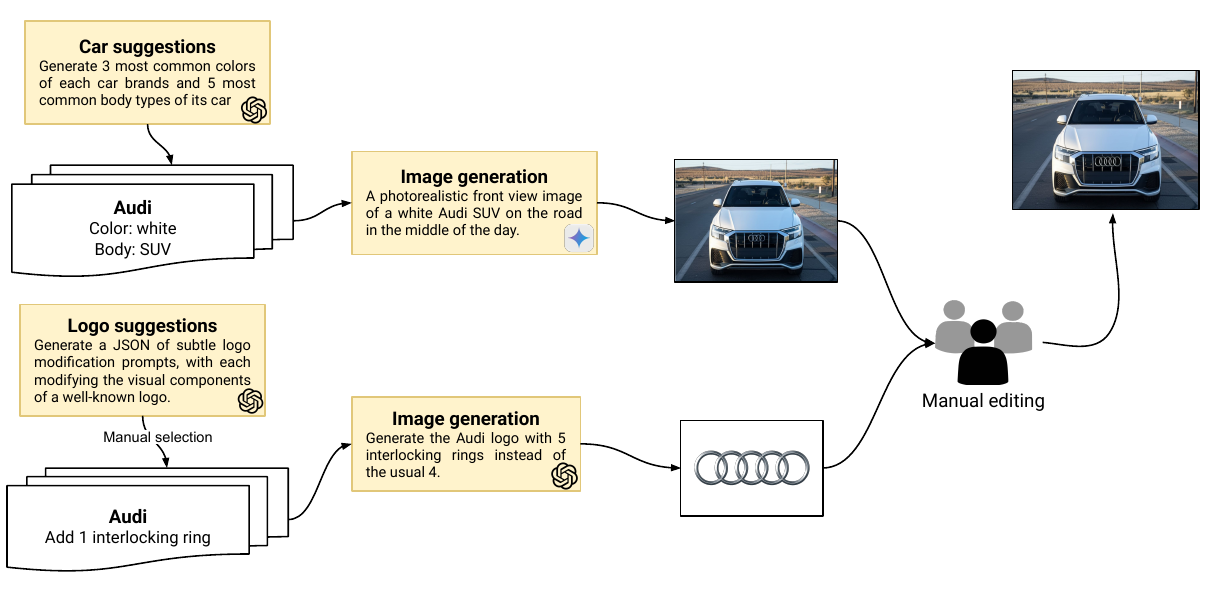}
    \caption{Data generation pipeline of car logos for Task 2: Counting elements in modified brand logos
    }
    \label{fig:carlogo_diagram_updated}
\end{figure}

\subsection{Task design}
Our initial evaluation show that some VLMs, such as \gptminilogo~\gptmini, can accurately count the four stripes on modified Adidas logo on white background. As such, to increase the task difficulty, we hypothesize that VLMs strongly associate \logoicon~logos with the background they typically appear on. Subsequently, we examine if the visual cues from the background would be strong enough to suppress counting the elements in the logos. Our task is designed as follow:
\begin{itemize}
\item \textbf{\colorbox{blue!15}{Brand types}}: We use \textcolor{blue!80}{\textbf{2}} different brand types: \emph{cars} and \emph{shoes}

\item \textbf{\colorbox{green!15}{Target brands}}: We select \textcolor{green!80}{\textbf{5}} well-known brands with quantifiable graphical elements:
    \begin{itemize}
        \item \emph{Car brands}: Mercedes-Benz, Maserati, and Audi (\textcolor{green!80}{\textbf{3}} brands)
        \item \emph{Shoe brands}: Adidas and Nike (\textcolor{green!80}{\textbf{2}} brands)
    \end{itemize}
    
\item \textbf{\colorbox{orange!15}{Background variations}}: Each brand logo has specific background settings:
    \begin{itemize}
        \item \emph{Car logo background}: Car logos always appear on cars. For each logo, we collect \textcolor{orange!80}{\textbf{5}} car body types $\times$ \textcolor{orange!80}{\textbf{3}} colors (white, grey, black)
        \item \emph{Shoe logo background}: Shoe logos are often seen on the footwear of athletes. For each logo, we collect a list of \textcolor{orange!80}{\textbf{4}} relevant sports (tennis, running, basketball, soccer) $\times$ \textcolor{orange!80}{\textbf{3}} colors (black, red, white)
    \end{itemize}

\item \textbf{\colorbox{purple!15}{Image resolutions}}: We generate each image and rescale them at \textcolor{purple!80}{\textbf{3}} different pixel sizes \{384, 768, 1152\}px using the scaling factor in \cref{sec:animal_add_legs} to test resolution sensitivity
\end{itemize}

This systematic approach generates a total of [\textcolor{green!80}{\textbf{3}} car brands $\times$ (\textcolor{orange!80}{\textbf{5}} $\times$ \textcolor{orange!80}{\textbf{3}}) $\times$ \textcolor{purple!80}{\textbf{3}} resolutions] + [\textcolor{green!80}{\textbf{2}} shoe brands $\times$ (\textcolor{orange!80}{\textbf{4}} $\times$ \textcolor{orange!80}{\textbf{3}}) $\times$ \textcolor{purple!80}{\textbf{3}} resolutions] = \textbf{135} + \textbf{72} = \textbf{207} total images.


\subsection{Implementation and prompts}
\label{app:logos-task-implementation}
\subsec{Implementation details}
We employ the following process to generate logo modification images:

\begin{enumerate}
\item Use \gptminilogo\gptmini to suggest graphical modifications for each logo (e.g., increasing Adidas' three stripes to four). We then select the most relevant suggestions for our benchmark.
\item Generate modified logo versions using \gptfulllogo~\gptfouro.
\item Create background images:
\begin{itemize}
\item \emph{Background images for car logos}: 
\begin{itemize}
    \item Use \gptminilogo\gptmini to suggest popular colors and body types of each car logo. 
    \item For each logo, generate and select relevant images of cars from the logo brand with the determined body types and colors.
    \item Manually place modified logos in typical car logo positions.
\end{itemize}
\item \emph{Background images for shoe logos}:
\begin{itemize}
    \item Use \gptminilogo\gptmini to suggest popular shoe colors and sports affiliated with each shoe logo. 
    \item For each logo, generate and select relevant images of athletes wearing shoes with the modified logo for each determined color and sport.
\end{itemize}

\end{itemize}
\item Render each image at three different resolutions.
\end{enumerate}
\subsec{Quality control}
To ensure high-quality images, we manually review to make sure that: (1) each generated logo has the correct number of modified elements; (2) each product is clearly visible and oriented correctly; and (3) the position of the logos on the products are natural-looking.

\subsec{Prompts} We use the following prompts
\begin{enumerate}
    \item \textbf{Counting questions (Q1 \& Q2)}:
    \begin{itemize}
        \item \textbf{Q1 (Adidas):} \emph{How many visible [StripeColor] stripes are there in the logo of the left shoe? Answer with a number in curly brackets, e.g., \{9\}.}
        \item \textbf{Q1 (Nike):} \emph{How many visible [CurveColor] stylized curves are there in the logo of the left shoe? Answer with a number in curly brackets, e.g., \{9\}}
        \item \textbf{Q1 (Audi):} \emph{How many overlapping circles are there in the logo of this car? Answer with a number in curly brackets, e.g., \{9\}.}
     
        \item \textbf{Q1 (Mercedes):} \emph{How many points are there on the star in the logo of this car? Answer with a number in curly brackets, e.g., \{9\}.}
        \item \textbf{Q1 (Maserati):} \emph{How many prongs are there in the logo of this car? Answer with a number in curly brackets, e.g., \{9\}}

        \item \textbf{Q2 (Adidas):} \emph{Count the visible [StripeColor] stripes in the logo of the left shoe. Answer with a number in curly brackets, e.g., \{9\}.}
        \item \textbf{Q2 (Nike):} \emph{Count the visible [CurveColor] stylized curves in the logo of the left shoe. Answer with a number in curly brackets, e.g., \{9\}}
        \item \textbf{Q2 (Audi):} \emph{Count the overlapping circles in the logo of this car. Answer with a number in curly brackets, e.g., \{9\}.}
     
        \item \textbf{Q2 (Mercedes):} \emph{Count the points on the star in the logo of this car. Answer with a number in curly brackets, e.g., \{9\}.}
        \item \textbf{Q2 (Maserati):} \emph{Count the prongs in the logo of this car. Answer with a number in curly brackets, e.g., \{9\}}
    \end{itemize}
    
    \item \textbf{Y/N identification questions (Q3)}:
    \begin{itemize}
        \item \textbf{Q3 (Adidas):} \emph{Are the logos on these shoes Adidas logos? Answer in curly brackets, e.g., \{Yes\} or \{No\}. }
        \item \textbf{Q3 (Nike):} \emph{Are the logos on these shoes Nike logos? Answer in curly brackets, e.g., \{Yes\} or \{No\}.}
        \item \textbf{Q3 (Audi):} \emph{Is the logo on this car Audi logo? Answer in curly brackets, e.g., \{Yes\} or \{No\}.} 
        \item \textbf{Q3 (Mercedes):} \emph{Is the logo on this car Mercedes-Benz logo? Answer in curly brackets, e.g., \{Yes\} or \{No\}.}
        \item \textbf{Q3 (Maserati):} \emph{Is the logo on this car Maserati logo? Answer in curly brackets, e.g., \{Yes\} or \{No\}.}

    \end{itemize}
\end{enumerate}

\subsec{Ground truth calculation} The ground truth answers are as follow:
\begin{itemize}
    \item \textbf{Adidas stripes counting (Q1\&Q2)}:
    \begin{itemize}
        \item Correct answer: 4
        \item Expected bias: 3
    \end{itemize}
    \item \textbf{Nike stylized curves counting (Q1\&Q2)}:
    \begin{itemize}
        \item Correct answer: 2
        \item Expected bias: 1
    \end{itemize}
    \item \textbf{Audi overlapping circles counting (Q1\&Q2)}:
    \begin{itemize}
        \item Correct answer: 5
        \item Expected bias: 4
    \end{itemize}
    \item \textbf{Mercedes-Benz points on the star counting (Q1\&Q2)}:
    \begin{itemize}
        \item Correct answer: 4
        \item Expected bias: 3
    \end{itemize}
    \item \textbf{Maserati prongs counting (Q1\&Q2)}:
    \begin{itemize}
        \item Correct answer: 5
        \item Expected bias: 3
    \end{itemize}
    \item \textbf{Logo identification question (Q3)}:
    \begin{itemize}
        \item Correct answer: ``No'' (all logos are modified)
        \item Expected bias: ``Yes''
    \end{itemize}
\end{itemize}

\subsection{Qualitative results}
\begin{figure}[h]
\centering
\small
\begin{AIbox}{Car logos}

    \raggedright
\footnotesize
      \logoicon~\textbf{(a), (d):} How many overlapping circles are there in the logo of this car? Answer with a number in curly brackets, e.g., \{9\}. \\
      \logoicon~\textbf{(b), (e):} How many points are there on the star in the logo of this car? Answer with a number in curly brackets, e.g., \{9\}. \\
      \logoicon~\textbf{c:} How many prongs are there in the logo of this car? Answer with a number in curly brackets, e.g., \{9\}. \\
     \hfill \break
\footnotesize
\centering

    \begin{tabular}{lp{0.2cm}c|p{0.2cm}c|p{0.2cm}c|p{0.2cm}c|p{0.2cm}c}
    &\multicolumn{2}{c}{\makecell{\textbf{(a) Audi}}} 
    &\multicolumn{2}{c}{\makecell{\textbf{(b) Mercedes}}}
    &\multicolumn{2}{c}{\makecell{\textbf{(c) Maserati}}}
    &\multicolumn{2}{c}{\makecell{\textbf{(d) Audi}}} 
    &\multicolumn{2}{c}{\makecell{\textbf{(e) Mercedes}}}
    \\
     &\multicolumn{2}{c}{\includegraphics[width=0.15\textwidth]{images/car_logo_examples/audi-gray-convertible_1152.png}}
     &\multicolumn{2}{c}{\includegraphics[width=0.15\textwidth]{images/car_logo_examples/mercedes_benz-white-coupe_1152.png}}
     &\multicolumn{2}{c}{\includegraphics[width=0.15\textwidth]{images/car_logo_examples/maserati-black-coupe_1152.png}}
     &\multicolumn{2}{c}{\includegraphics[width=0.15\textwidth]{images/car_logo_examples/audi-black-convertible_1152.png}}
     &\multicolumn{2}{c}{\includegraphics[width=0.15\textwidth]{images/car_logo_examples/mercedes_benz-gray-convertible_1152.png}}
     \\ 

\rowcolor{lightgray}
\raisebox{-0.2\height}\geminilogo & \centering 4
& \textcolor{red}{\xmark} & \centering 3
& \textcolor{red}{\xmark} & \centering 3
& \textcolor{red}{\xmark} & \centering 4
& \textcolor{red}{\xmark} & \centering 3
& \textcolor{red}{\xmark} \\

\raisebox{-0.2\height}\newsonnetlogo & \centering 4
& \textcolor{red}{\xmark} & \centering 3
& \textcolor{red}{\xmark} & \centering 3
& \textcolor{red}{\xmark} & \centering 4
& \textcolor{red}{\xmark} & \centering 3
& \textcolor{red}{\xmark} \\

\rowcolor{lightgray}
\raisebox{-0.2\height}\gptfourlogo & \centering 4
& \textcolor{red}{\xmark} & \centering 3
& \textcolor{red}{\xmark} & \centering 3
& \textcolor{red}{\xmark} & \centering 4
& \textcolor{red}{\xmark} & \centering 3
& \textcolor{red}{\xmark} \\

\raisebox{-0.2\height}\gptfulllogo & \centering 4
& \textcolor{red}{\xmark} & \centering 3
& \textcolor{red}{\xmark} & \centering 3
& \textcolor{red}{\xmark} & \centering 4
& \textcolor{red}{\xmark} & \centering 3
& \textcolor{red}{\xmark} \\

\rowcolor{lightgray}
\raisebox{-0.2\height}\gptminilogo & \centering 4
& \textcolor{red}{\xmark} & \centering 3
& \textcolor{red}{\xmark} & \centering 3
& \textcolor{red}{\xmark} & \centering 4
& \textcolor{red}{\xmark} & \centering 3
& \textcolor{red}{\xmark} \\ 

\midrule
\textbf{Bias} & \centering 4
& \textcolor{red}{\xmark} & \centering 3
& \textcolor{red}{\xmark} & \centering 3
& \textcolor{red}{\xmark} & \centering 4
& \textcolor{red}{\xmark} & \centering 3
& \textcolor{red}{\xmark} \\

\rowcolor{lightgray}
\textbf{GT} & \centering 5
& \textcolor{ForestGreen}{\cmark} & \centering 4
& \textcolor{ForestGreen}{\cmark} & \centering 5
& \textcolor{ForestGreen}{\cmark} & \centering 5
& \textcolor{ForestGreen}{\cmark} & \centering 4
& \textcolor{ForestGreen}{\cmark} \\

     \end{tabular}
    \vspace{4pt}
    \centering
    \begin{tabular}{cccccccccccccc}
     \raisebox{-0.1\height}\geminilogo~\gemini 
     & \raisebox{-0.1\height}\newsonnetlogo~\newsonnet 
     & \raisebox{-0.1\height}\gptfourlogo~\gptfour 
     & \raisebox{-0.1\height}\gptfulllogo~\gptfull 
     & \raisebox{-0.1\height}\gptminilogo~\gptmini
     \\
    \end{tabular}

\end{AIbox}
\caption{VLMs are completely biased and rely entirely on prior knowledge when answering questions about \logoicon~brand logos. Please zoom in to see the logo clearly.}
\label{appfig:qualitative_results_car_logos}
\end{figure}


\begin{figure}[h]
\centering
\small
\begin{AIbox}{Shoe logos}

    \raggedright
\footnotesize
      \logoicon~\textbf{(a), (c):} How many visible \textbf{white stripes} are there in the logo of the left shoe? Answer with a number in curly brackets, e.g., \{9\}. \\
      \logoicon~\textbf{(b):} How many visible \textbf{white stylized curves} are there in the logo of the left shoe? Answer with a number in curly brackets, e.g., \{9\}. \\
      \logoicon~\textbf{(d):} How many visible \textbf{black stylized curves} are there in the logo of the left shoe? Answer with a number in curly brackets, e.g., \{9\}. \\
      \logoicon~\textbf{e):} How many visible \textbf{black stripes} are there in the logo of the left shoe? Answer with a number in curly brackets, e.g., \{9\}. \\
     \hfill \break
\footnotesize
\centering

    \begin{tabular}{lp{0.2cm}c|p{0.2cm}c|p{0.2cm}c|p{0.2cm}c|p{0.2cm}c}
    &\multicolumn{2}{c}{\makecell{\textbf{(a) Adidas}}} 
    &\multicolumn{2}{c}{\makecell{\textbf{(b) Nike}}}
    &\multicolumn{2}{c}{\makecell{\textbf{(c) Adidas}}}
    &\multicolumn{2}{c}{\makecell{\textbf{(d) Nike}}} 
    &\multicolumn{2}{c}{\makecell{\textbf{(e) Adidas}}}
    \\
     &\multicolumn{2}{c}{\includegraphics[width=0.15\textwidth]{images/shoe_logos/adidas-red-running-5_1152.png}}
     &\multicolumn{2}{c}{\includegraphics[width=0.15\textwidth]{images/shoe_logos/nike-black-soccer-1_1152.png}}
     &\multicolumn{2}{c}{\includegraphics[width=0.15\textwidth]{images/shoe_logos/adidas-black-tennis-1_1152.png}}
     &\multicolumn{2}{c}{\includegraphics[width=0.15\textwidth]{images/shoe_logos/nike-white-basketball-1_1152.png}}
     &\multicolumn{2}{c}{\includegraphics[width=0.15\textwidth]{images/shoe_logos/adidas-white-soccer-4_1152.png}}
     \\ 

\rowcolor{lightgray}
\raisebox{-0.2\height}\geminilogo & \centering 3
& \textcolor{red}{\xmark} & \centering 1
& \textcolor{red}{\xmark} & \centering 3
& \textcolor{red}{\xmark} & \centering 2
& \textcolor{ForestGreen}{\cmark} & \centering 3
& \textcolor{red}{\xmark} \\

\raisebox{-0.2\height}\newsonnetlogo & \centering 3
& \textcolor{red}{\xmark} & \centering 1
& \textcolor{red}{\xmark} & \centering 3
& \textcolor{red}{\xmark} & \centering 1
& \textcolor{red}{\xmark} & \centering 3
& \textcolor{red}{\xmark} \\

\rowcolor{lightgray}
\raisebox{-0.2\height}\gptfourlogo & \centering 3
& \textcolor{red}{\xmark} & \centering 1
& \textcolor{red}{\xmark} & \centering 3
& \textcolor{red}{\xmark} & \centering 1
& \textcolor{red}{\xmark} & \centering 3
& \textcolor{red}{\xmark} \\

\raisebox{-0.2\height}\gptfulllogo & \centering 3
& \textcolor{red}{\xmark} & \centering 1
& \textcolor{red}{\xmark} & \centering 3
& \textcolor{red}{\xmark} & \centering 1
& \textcolor{red}{\xmark} & \centering 4
& \textcolor{ForestGreen}{\cmark} \\

\rowcolor{lightgray}
\raisebox{-0.2\height}\gptminilogo & \centering 3
& \textcolor{red}{\xmark} & \centering 1
& \textcolor{red}{\xmark} & \centering 3
& \textcolor{red}{\xmark} & \centering 1
& \textcolor{red}{\xmark} & \centering 4
& \textcolor{ForestGreen}{\cmark} \\ 

\midrule
\textbf{Bias} & \centering 3
& \textcolor{red}{\xmark} & \centering 1
& \textcolor{red}{\xmark} & \centering 3
& \textcolor{red}{\xmark} & \centering 1
& \textcolor{red}{\xmark} & \centering 3
& \textcolor{red}{\xmark} \\

\rowcolor{lightgray}
\textbf{GT} & \centering 4
& \textcolor{ForestGreen}{\cmark} & \centering 2
& \textcolor{ForestGreen}{\cmark} & \centering 4
& \textcolor{ForestGreen}{\cmark} & \centering 2
& \textcolor{ForestGreen}{\cmark} & \centering 4
& \textcolor{ForestGreen}{\cmark} \\

     \end{tabular}
    \vspace{4pt}
    \centering
    \begin{tabular}{cccccccccccccc}
     \raisebox{-0.1\height}\geminilogo~\gemini 
     & \raisebox{-0.1\height}\newsonnetlogo~\newsonnet 
     & \raisebox{-0.1\height}\gptfourlogo~\gptfour 
     & \raisebox{-0.1\height}\gptfulllogo~\gptfull 
     & \raisebox{-0.1\height}\gptminilogo~\gptmini
     \\
    \end{tabular}

\end{AIbox}
\caption{VLMs are often biased and rely on prior knowledge when answering questions about \logoicon~shoe logos, even with simple ones like the Nike Swoosh. Please zoom in to see the logo clearly.}
\label{appfig:qualitative_results_shoe_logos}
\end{figure}

\clearpage
\section{Task 3: Counting stripes/stars in modified national flags \flagicon}\label{sec:appendix_flags}
\begin{figure}[t]
    \centering
    \includegraphics[width=0.9\textwidth]{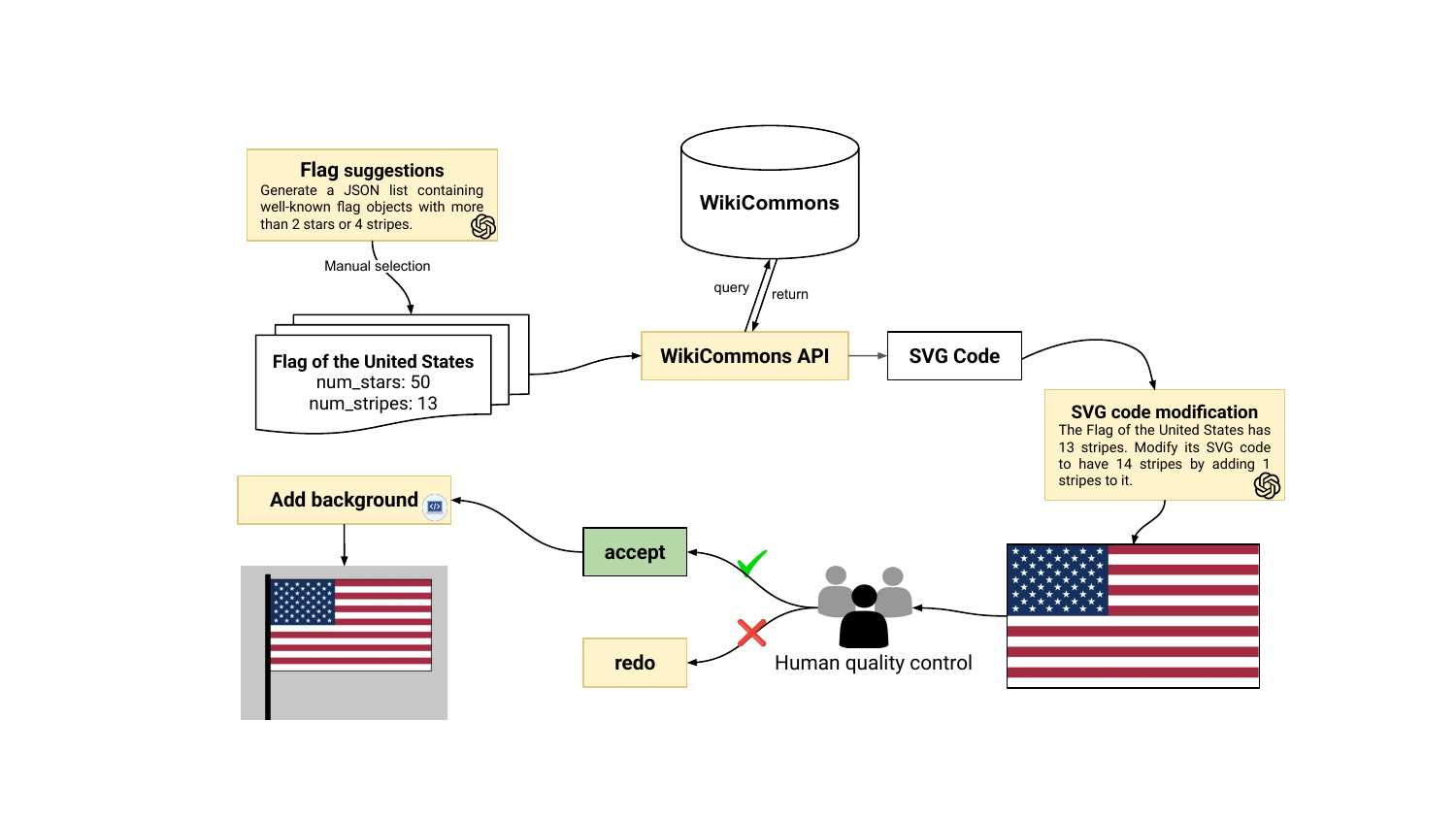}
    \caption{Data generation pipeline for Task 3: Counting stripes/stars in modified national flags.
    }
    \label{fig:flag_diagram_updated}
\end{figure}


\subsection{Task design} \label{app:flags-task-design} Flags of countries contain easily recognizable patterns. To evaluate if existing VLMs overly rely on their knowledge of these \flagicon flags to count a certain element, we design the task as follow:

\begin{itemize} 
\item \textbf{\colorbox{blue!15}{Flag types}}: We modify \textcolor{blue!80}{\textbf{2}} commonly used elements across different flags: \emph{stars} and \emph{stripes}

\item \textbf{\colorbox{green!15}{Modification types}}: Each flag has \textcolor{green!80}{\textbf{2}} types of modifications:
    \begin{itemize}
        \item \emph{Add}: We add an additional element (star or stripe) to a chosen flag
        \item \emph{Remove}: We remove one element (star or stripe) from a chosen flag
    \end{itemize}
    
\item \textbf{\colorbox{orange!15}{Target flags}}: We select \textcolor{orange!80}{\textbf{20}} well-known country flags with either 3+ stars or 5+ stripes (a total of 13 star-typed flags and 7 stripe-typed flags) to ensure the modified flags retain recognizable traits to test visual bias.

\item \textbf{\colorbox{purple!15}{Image resolutions}}: We generate each flag and rescale them at \textcolor{purple!80}{\textbf{3}} different pixel sizes \{384, 768, 1152\}px using the scaling factor in \cref{sec:animal_add_legs} to test resolution sensitivity
\end{itemize}

This systematic approach generates a total of \textcolor{orange!80}{\textbf{20}} target flags $\times$ \textcolor{green!80}{\textbf{2}} modification types $\times$ \textcolor{purple!80}{\textbf{3}} resolutions = \textbf{120} total images.

\subsection{Implementation and image generation} \label{app:flags-task-implementation}

\subsec{Implementation details} We modify the SVG code of a chosen flag to create new variants following this sequence: 
\begin{enumerate} 
\item Identify 20 well-known country flags (13 with 3+ stars, 7 with 5+ stripes) based on the suggestions from \gptminilogo\gptmini.
\item Retrieve original SVG code from WikiCommons for each flag.
\item Use \gptminilogo\gptmini to modify each SVG to create two variants: 

\begin{itemize} 
\item An ``Add'' variant with one additional element.
\item A ``Remove'' variant with one fewer element.
\end{itemize} 
\item Render each modified flag at three different resolutions.
\end{enumerate}


\subsec{Quality control}
We employ the following steps to ensure high-quality and consistent images:
\begin{itemize}
\item \textbf{Manual inspection}: We manually review each generated sample to verify modification quality and visual consistency
\item \textbf{Filtering}: We remove unsatisfactory samples from the benchmark and rerun the pipeline on these cases to obtain new samples.
\item \textbf{Fallback}: For rare cases (3 in total) that consistently fail automated generation, we manual modify the flags to ensure they strictly follow the modification rules.
\end{itemize}

\subsec{Prompts} We use the following prompts:
\begin{enumerate}
    \item \textbf{Counting questions (Q1 \& Q2)}:
    \begin{itemize}
        \item \textbf{Q1 (Star-typed flags):} \emph{How many stars are there on this flag? Answer with a number in curly brackets, e.g., \{9\}.}
        \item \textbf{Q1 (Stripe-typed flags):} \emph{How many stripes are there on this flag? Answer with a number in curly brackets, e.g., \{9\}.}
        \item \textbf{Q2 (Star-typed flags):} \emph{Count the stars on this flag. Answer with a number in curly brackets, e.g., \{9\}.}
        \item \textbf{Q2 (Stripe-typed flags):} \emph{Count the stripes on this flag. Answer with a number in curly brackets, e.g., \{9\}.}
    \end{itemize}
    
    \item \textbf{Y/N identification questions (Q3)}:
    \begin{itemize}
        \item \emph{Is this the flag of [CountryName]? Answer in curly brackets, e.g., \{Yes\} or \{No\}.} 
    \end{itemize}
\end{enumerate}

\subsec{Ground truth calculation} We calculate the ground truth as follow:
\begin{itemize}
    \item \textbf{Direct counting questions (Q1 \& Q2)}:
    \begin{itemize}
        \item \textbf{Correct answer}: The actual count of the elements (stars or stripes) on the flag after modification
        \begin{itemize}
            \item For \emph{Remove modifications}: Standard element count minus 1
            \item For \emph{Add modifications}: Standard element count plus 1  
        \end{itemize}
        \item \textbf{Expected bias}: The standard element count
    \end{itemize}
    
    \item \textbf{Flag verification question (Q3)}:
    \begin{itemize}
        \item \textbf{Correct answer}: ``No'' (since the flag's element has been modified)
        \item \textbf{Expected bias}: ``Yes''
    \end{itemize}
\end{itemize}


\subsection{Qualitative results}
\begin{figure}[h]
\centering
\small
\begin{AIbox}{National Flag}

    \raggedright
\footnotesize
      \flagicon~\textbf{(a), (b), (e):} How many \textbf{stripes} are there in this flag? Answer with a number in curly brackets, e.g., \{9\}. \\
      \flagicon~\textbf{(b), (c):} How many \textbf{stars} are there in this flag? Answer with a number in curly brackets, e.g., \{9\}. \\
     \hfill \break
\footnotesize
\centering

    \begin{tabular}{lp{0.2cm}c|p{0.2cm}c|p{0.2cm}c|p{0.2cm}c|p{0.2cm}c}
    &\multicolumn{2}{c}{\makecell{\textbf{(a) US}}} 
    &\multicolumn{2}{c}{\makecell{\textbf{(b) US}}}
    &\multicolumn{2}{c}{\makecell{\textbf{(c) EU}}}
    &\multicolumn{2}{c}{\makecell{\textbf{(d) EU}}} 
    &\multicolumn{2}{c}{\makecell{\textbf{(e) Zimbabwe}}}
    \\
     &\multicolumn{2}{c}{\includegraphics[width=0.15\textwidth]{images/flag_examples/Flag_of_the_United_States-stripes=12_1152.png}}
     &\multicolumn{2}{c}{\includegraphics[width=0.15\textwidth]{images/flag_examples/Flag_of_the_United_States-stripes=14_1152.png}}
     &\multicolumn{2}{c}{\includegraphics[width=0.15\textwidth]{images/flag_examples/Flag_of_Europe-stars=11_1152.png}}
     &\multicolumn{2}{c}{\includegraphics[width=0.15\textwidth]{images/flag_examples/Flag_of_Europe-stars=13_1152.png}}
     &\multicolumn{2}{c}{\includegraphics[width=0.15\textwidth]{images/flag_examples/Flag_of_Zimbabwe-stripes=6_1152.png}}
     \\ 

\rowcolor{lightgray}
\raisebox{-0.2\height}\geminilogo & \centering 13
& \textcolor{red}{\xmark} & \centering 13
& \textcolor{red}{\xmark} & \centering 12
& \textcolor{red}{\xmark} & \centering 13
& \textcolor{ForestGreen}{\cmark} & \centering 7
& \textcolor{red}{\xmark} \\

\raisebox{-0.2\height}\newsonnetlogo & \centering 13
& \textcolor{red}{\xmark} & \centering 13
& \textcolor{red}{\xmark} & \centering 12
& \textcolor{red}{\xmark} & \centering 12
& \textcolor{red}{\xmark} & \centering 7
& \textcolor{red}{\xmark} \\

\rowcolor{lightgray}
\raisebox{-0.2\height}\gptfourlogo & \centering 13
& \textcolor{red}{\xmark} & \centering 13
& \textcolor{red}{\xmark} & \centering 12
& \textcolor{red}{\xmark} & \centering 12
& \textcolor{red}{\xmark} & \centering 7
& \textcolor{red}{\xmark} \\

\raisebox{-0.2\height}\gptfulllogo & \centering 13
& \textcolor{red}{\xmark} & \centering 13
& \textcolor{red}{\xmark} & \centering 12
& \textcolor{red}{\xmark} & \centering 12
& \textcolor{red}{\xmark} & \centering 7
& \textcolor{red}{\xmark} \\

\rowcolor{lightgray}
\raisebox{-0.2\height}\gptminilogo & \centering 13
& \textcolor{red}{\xmark} & \centering 13
& \textcolor{red}{\xmark} & \centering 12
& \textcolor{red}{\xmark} & \centering 13
& \textcolor{ForestGreen}{\cmark} & \centering 7
& \textcolor{red}{\xmark} \\ 

\midrule
\textbf{Bias} & \centering 13
& \textcolor{red}{\xmark} & \centering 13
& \textcolor{red}{\xmark} & \centering 12
& \textcolor{red}{\xmark} & \centering 12
& \textcolor{red}{\xmark} & \centering 7
& \textcolor{red}{\xmark} \\

\rowcolor{lightgray}
\textbf{GT} & \centering 12
& \textcolor{ForestGreen}{\cmark} & \centering 14
& \textcolor{ForestGreen}{\cmark} & \centering 11
& \textcolor{ForestGreen}{\cmark} & \centering 13
& \textcolor{ForestGreen}{\cmark} & \centering 6
& \textcolor{ForestGreen}{\cmark} \\

     \end{tabular}
    \vspace{4pt}
    \centering
    \begin{tabular}{cccccccccccccc}
     \raisebox{-0.1\height}\geminilogo~\gemini 
     & \raisebox{-0.1\height}\newsonnetlogo~\newsonnet 
     & \raisebox{-0.1\height}\gptfourlogo~\gptfour 
     & \raisebox{-0.1\height}\gptfulllogo~\gptfull 
     & \raisebox{-0.1\height}\gptminilogo~\gptmini
     \\
    \end{tabular}

\end{AIbox}
\caption{VLMs are biased when counting the stars and stripes on \flagicon~national flags.}
\label{appfig:qualitative_results_flags}
\end{figure}


\clearpage
\section{Task 4: Counting chess pieces on modified starting position \chessicon}\label{sec:appendix_chess_pieces}
\begin{figure}[h!]
    \centering
    \includegraphics[width=0.9\textwidth]{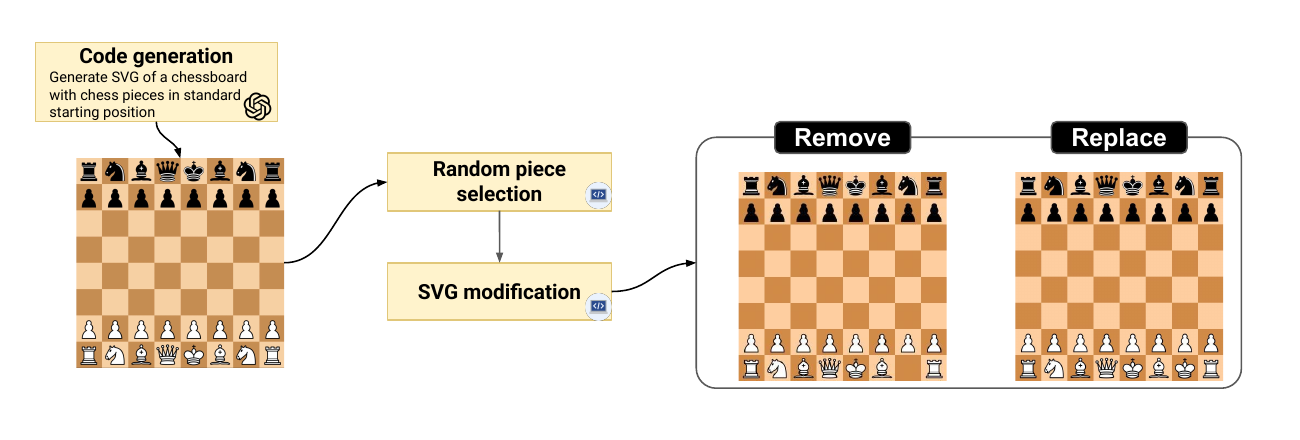}
    \caption{Data generation pipeline for Task 4: Counting chess pieces on modified starting position
    }
    \label{fig:chesspieces_diagram}
\end{figure}

\subsection{Task design}
\label{app:chess-task-design}
To evaluate if VLMs rely on expected structure or attend to actual pieces, we test their ability to count pieces on boards with subtle modifications. We design our task with careful control of visual parameters to ensure systematic evaluation:

\begin{itemize}
    \item \textbf{\colorbox{blue!15}{Board types}}: We use \textcolor{blue!80}{\textbf{2}} different game boards: \{\emph{chess (Western chess)}, \emph{xiangqi} (Chinese chess)\}
    
    \item \textbf{\colorbox{green!15}{Modification types}}: Each board has \textcolor{green!80}{\textbf{2}} types of modifications:
        \begin{itemize}
            \item \emph{Remove}: We remove exactly one piece from the standard starting position.
            \item \emph{Replace}: We replace exactly one piece with a different piece of the same color.
        \end{itemize}
        
    \item \textbf{\colorbox{orange!15}{Target squares}}: We select \textcolor{orange!80}{\textbf{12}} unique occupied squares per board type, maintaining the same target squares across the Remove and Replace modifications to ensure controlled comparison.
    
    \item \textbf{\colorbox{purple!15}{Image resolutions}}: We generate each board at \textcolor{purple!80}{\textbf{3}} different pixel sizes \{384, 768, 1152\}px to test resolution sensitivity.
\end{itemize}

This systematic approach generates a total of \textcolor{blue!80}{\textbf{2}} board types $\times$ \textcolor{green!80}{\textbf{2}} modification types $\times$ \textcolor{orange!80}{\textbf{12}} target squares $\times$ \textcolor{purple!80}{\textbf{3}} resolutions = \textbf{144} total images.

\subsection{Implementation and prompts}
\label{app:chess-task-implementation}

\subsec{Implementation details} 
Our implementation utilizes specialized libraries for each board type. For chess, we leverage the Python \texttt{chess} library to manipulate board states and \texttt{chess.svg} for rendering. For xiangqi (Chinese chess), we created a custom implementation using \texttt{svgwrite} for rendering.

The algorithm for both board types follows the same sequence:
\begin{enumerate}
    \item Create a standard board with all 32 pieces in their starting positions
    \item Randomly select 12 target squares from the occupied squares
    \item For each target square, create (a) a Remove variant and (b) a Replace variant
    \item Render each modified board at three different resolutions
\end{enumerate}

The xiangqi implementation required special handling for:
\begin{itemize}
    \item The traditional 9×10 board layout with the central river and two palaces
    \item Chinese character rendering for pieces, which requires detecting appropriate CJK fonts
    \item Different piece distribution (Chariots, Knights, Elephants, Advisors, General, Cannons, and Soldiers)
\end{itemize}

\subsec{Quality control}
To ensure consistent image quality across all variants, we implemente several technical measures:
\begin{itemize}
    \item \textbf{SVG to PNG conversion}: We used direct SVG rendering with adjustable scaling factors based on target resolution
    \item \textbf{Quality scaling}: We applied a quality multiplier ($5.0 \times$ base resolution factor) to ensure clear piece visibility
\end{itemize}

\subsec{Prompts}    
We use different prompts for each modification type to test VLMs' visual attention:

\begin{enumerate}
    \item \textbf{Remove modifications}: 
    \begin{itemize}
        \item \textbf{Q1:} \emph{How many [chess/xiangqi] pieces are there on this board? Answer with a number in curly brackets, e.g., \{9\}.}
        \item \textbf{Q2:} \emph{Count the [chess/xiangqi] pieces on this board. Answer with a number in curly brackets, e.g., \{9\}.}
    \end{itemize}
    
    \item \textbf{Replace modifications}: 
    \begin{itemize}
        \item \textbf{Q1:} \emph{How many [Added Piece Type] pieces are there on this board? Answer with a number in curly brackets, e.g., \{9\}.}
        \item \textbf{Q2:} \emph{Count the [Added Piece Type] pieces on this board? Answer with a number in curly brackets, e.g., \{9\}.}
    \end{itemize}
    
    \item \textbf{Both modification types}:
    \begin{itemize}
        \item \textbf{Q3:} \emph{Is this the [chess/xiangqi] starting position? Answer in curly brackets, e.g., \{Yes\} or \{No\}.}
    \end{itemize}
\end{enumerate}

For Replace modifications, [Added Piece Type] refers to the specific piece type that is added to the board through replacement, chosen from:
\begin{itemize}
    \item For chess: Pawn, Knight, Bishop, Rook, Queen, or King
    \item For xiangqi: Soldier, Horse, Elephant, Chariot, Cannon, Advisor, or General
\end{itemize}

For Replace modifications, we ask about the added piece type rather than total count because this more effectively tests whether VLMs rely on prior knowledge of standard piece distributions or actually inspect the board carefully.

\subsec{Ground truth calculation}
We calculate the ground truth answers for each prompt type:

\begin{itemize}
    \item \textbf{Total piece count (Remove modifications only)}:
    \begin{itemize}
        \item Correct answer: 31 (one fewer than the standard 32 pieces)
        \item Expected bias: 32 (the standard piece count)
    \end{itemize}
    
    \item \textbf{Added piece type count (Replace modifications only)}:
    \begin{itemize}
        \item Correct answer: The standard count for that piece type plus one
        \item For example, if a Knight is replaced with a Bishop in chess, the Bishop count would be 3 (standard 2 + 1 added)
        \item Expected bias: The standard count for that piece type (e.g., 2 for Bishops in chess)
        \item This tests if VLMs rely on their knowledge of standard piece counts or actually inspect the board
    \end{itemize}
    
    \item \textbf{Starting position question (Both modification types)}:
    \begin{itemize}
        \item Correct answer: Always ``No'' (since the board has been modified)
        \item Expected bias: ``Yes'' (since the board closely resembles the starting position)
    \end{itemize}
\end{itemize}

\subsection{Qualitative results}

\begin{figure}[h]
\centering
\small
\begin{AIbox}{Chess Pieces}

    \raggedright
\footnotesize
      \textbf{(a):} How many \textbf{chess pieces} are there on this board? Answer with a number in curly brackets, e.g., \{9\}. \\
      \textbf{(b):} How many \textbf{Pawn pieces} are there on this board? Answer with a number in curly brackets, e.g., \{9\}. \\
      \textbf{(c):} How many \textbf{xiangqi pieces} are there on this board? Answer with a number in curly brackets, e.g., \{9\}. \\
      \textbf{(d):} How many \textbf{General pieces} are there on this board? Answer with a number in curly brackets, e.g., \{9\}. \\
     \hfill \break
\footnotesize
\centering

    \begin{tabular}{lp{0.2cm}c|p{0.2cm}c|p{0.2cm}c|p{0.2cm}c}
    &\multicolumn{2}{c}{\makecell{\textbf{(a) Chess}}} 
    &\multicolumn{2}{c}{\makecell{\textbf{(b) Chess}}}
    &\multicolumn{2}{c}{\makecell{\textbf{(c) Xiangqi}}}
    &\multicolumn{2}{c}{\makecell{\textbf{(d) Xiangqi}}} 
    \\
     &\multicolumn{2}{c}{\includegraphics[width=0.20\textwidth]{images/chess_pieces_examples/chess_pieces_004_remove_blackbishop_at_c8_notitle_px1152.png}}
     &\multicolumn{2}{c}{\includegraphics[width=0.20\textwidth]{images/chess_pieces_examples/chess_pieces_004_replace_blackbishop_to_blackpawn_at_c8_notitle_px1152.png}}
     &\multicolumn{2}{c}{\includegraphics[width=0.20\textwidth]{images/chess_pieces_examples/xiangqi_pieces_006_remove_red_chariot_at_i10_notitle_px1152.png}}
     &\multicolumn{2}{c}{\includegraphics[width=0.20\textwidth]{images/chess_pieces_examples/xiangqi_pieces_006_replace_red_chariot_to_red_general_at_i10_notitle_px1152.png}}
     \\ 

\rowcolor{lightgray}
\raisebox{-0.2\height}\geminilogo & \centering 32
& \textcolor{red}{\xmark} & \centering 16
& \textcolor{red}{\xmark} & \centering 32
& \textcolor{red}{\xmark} & \centering 2
& \textcolor{red}{\xmark} \\

\raisebox{-0.2\height}\newsonnetlogo & \centering 32
& \textcolor{red}{\xmark} & \centering 16
& \textcolor{red}{\xmark} & \centering 32
& \textcolor{red}{\xmark} & \centering 2
& \textcolor{red}{\xmark} \\

\rowcolor{lightgray}
\raisebox{-0.2\height}\gptfourlogo & \centering 28
& \textcolor{red}{\xmark} & \centering 16
& \textcolor{red}{\xmark} & \centering 32
& \textcolor{red}{\xmark} & \centering 2
& \textcolor{red}{\xmark} \\

\raisebox{-0.2\height}\gptfulllogo & \centering 31
& \textcolor{ForestGreen}{\cmark} & \centering 17
& \textcolor{ForestGreen}{\cmark} & \centering 32
& \textcolor{red}{\xmark} & \centering 2
& \textcolor{red}{\xmark} \\

\rowcolor{lightgray}
\raisebox{-0.2\height}\gptminilogo & \centering 32
& \textcolor{red}{\xmark} & \centering 17
& \textcolor{ForestGreen}{\cmark} & \centering 32
& \textcolor{red}{\xmark} & \centering 2
& \textcolor{red}{\xmark} \\ 

\midrule
\textbf{Bias} & \centering 32
& \textcolor{red}{\xmark} & \centering 16
& \textcolor{red}{\xmark} & \centering 32
& \textcolor{red}{\xmark} & \centering 2
& \textcolor{red}{\xmark} \\

\rowcolor{lightgray}
\textbf{GT} & \centering 31
& \textcolor{ForestGreen}{\cmark} & \centering 17
& \textcolor{ForestGreen}{\cmark} & \centering 31
& \textcolor{ForestGreen}{\cmark} & \centering 3
& \textcolor{ForestGreen}{\cmark} \\

     \end{tabular}
    \vspace{4pt}
    \centering
    \begin{tabular}{cccccccccccccc}
     \raisebox{-0.1\height}\geminilogo~\gemini 
     & \raisebox{-0.1\height}\newsonnetlogo~\newsonnet 
     & \raisebox{-0.1\height}\gptfourlogo~\gptfour 
     & \raisebox{-0.1\height}\gptfulllogo~\gptfull 
     & \raisebox{-0.1\height}\gptminilogo~\gptmini
     \\
    \end{tabular}

\end{AIbox}
\caption{VLMs are biased when counting the pieces on \chessicon~chess and xiangqi.}
\label{appfig:qualitative_results_chess_pieces}
\end{figure}

\clearpage
\section{Task 5: Counting rows and columns of game boards \boardgameicon}\label{sec:appendix_boardgame_grid}
\begin{figure}[h!]
    \centering
    \includegraphics[width=0.9\textwidth]{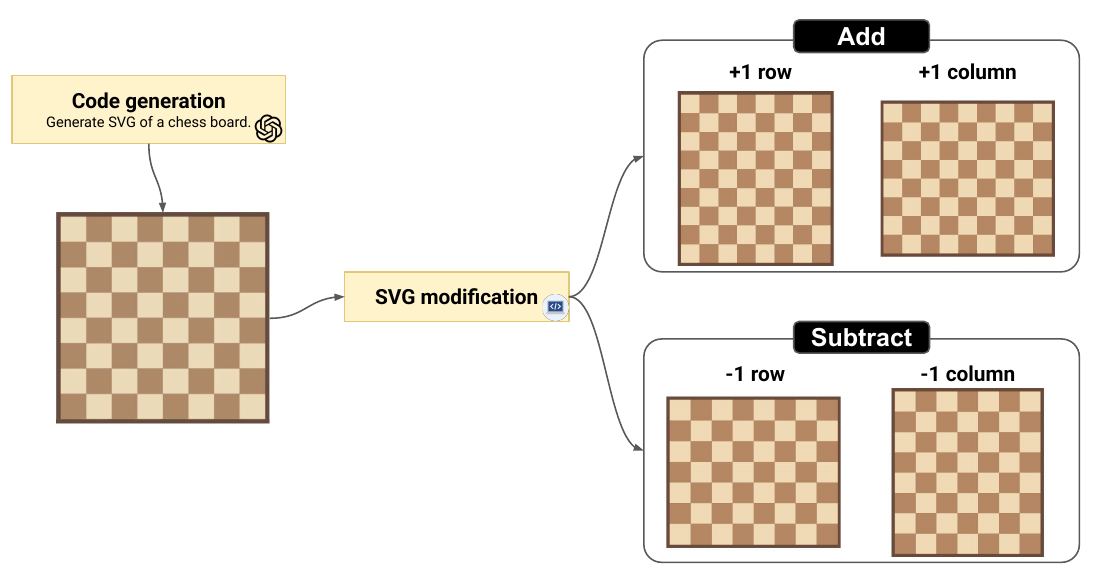}
    \caption{Data generation pipeline for Task 5: Counting rows and columns of board game
    }
    \label{fig:board_diagram}
\end{figure}

\subsection{Task design}
\label{app:grid-task-design}
To evaluate VLMs' over-reliance on visual bias versus actual counting, we adapted the row and column counting task from BlindTest~\citep{vlms-are-blind} where \texttt{Claude-3.5-Sonnet} achieved 74.26\% accuracy. Instead of simple grids, we leverage modified versions of well-known game boards to test whether VLMs rely on prior knowledge or perform actual visual counting. We design our task with careful control of visual parameters to ensure systematic evaluation:

\begin{itemize}
    \item \textbf{\colorbox{blue!15}{Board types}}: We use \textcolor{blue!80}{\textbf{4}} different grid-based game boards: \{\emph{Chess} (8×8), \emph{Xiangqi} (Chinese chess, 10×9), \emph{Sudoku} (9×9), \emph{Go} (19×19)\}
    
    \item \textbf{\colorbox{green!15}{Modification types}}: Each board has up to \textcolor{green!80}{\textbf{4}} types of modifications:
        \begin{itemize}
            \item \emph{Remove row}: We remove exactly one row from the grid.
            \item \emph{Remove column}: We remove exactly one column from the grid.
            \item \emph{Add row}: We add exactly one row to the grid.
            \item \emph{Add column}: We add exactly one column to the grid.
        \end{itemize}
        
    \item \textbf{\colorbox{orange!15}{Board-specific variations}}: For Chess, Xiangqi, and Sudoku boards, all four modifications (remove/add row, remove/add column) are visually distinct, with additional positional variations (first/last), resulting in \textcolor{orange!80}{\textbf{8}} variants per board. Go boards have uniform grid structure, so we produce only \textcolor{orange!80}{\textbf{4}} variations.
    
    \item \textbf{\colorbox{purple!15}{Image resolutions}}: We generate each board at \textcolor{purple!80}{\textbf{3}} different pixel sizes \{384, 768, 1152\}px to test resolution sensitivity.
\end{itemize}

This systematic approach generates a total of (\textcolor{orange!80}{\textbf{8}} variants × \textcolor{blue!80}{\textbf{3}} board types (Xiangqi/Chess/Sudoku) + \textcolor{orange!80}{\textbf{4}} Go variants) × \textcolor{purple!80}{\textbf{3}} resolutions = \textbf{84} total images.

\subsection{Implementation and prompts}
\label{app:grid-task-implementation}

\subsec{Implementation details} 
Our implementation utilizes specialized drawing libraries for each board type. For Chess, we use standard 8×8 chessboard grid generation with alternating square colors. For Xiangqi, we implement the traditional 10×9 board layout with river gap and palace diagonal lines. For Sudoku, we create 9×9 grids with bold 3×3 block boundaries and sample numbers. For Go, we generate uniform line grids with traditional star points.

The algorithm for all board types follows the same sequence:
\begin{enumerate}
    \item Create a standard board with correct dimensions and visual elements
    \item Apply systematic modifications (add/remove rows/columns at specific positions)
    \item Maintain visual consistency of special elements
    \item Render each modified board at three different resolutions
\end{enumerate}

The board-specific implementations required special handling for:
\begin{itemize}
    \item \textbf{Chess}: Alternating light/dark square pattern preservation across dimension changes
    \item \textbf{Xiangqi}: River gap positioning and palace diagonal lines adjustment for row modifications
    \item \textbf{Sudoku}: Bold 3×3 block boundary lines based on original 9×9 grid structure
    \item \textbf{Go}: Uniform line spacing and star point positioning for various board sizes
\end{itemize}

\subsec{Quality control}
To ensure consistent image quality across all variants, we implemented several technical measures:
\begin{itemize}
    \item \textbf{SVG to PNG conversion}: We used direct SVG rendering with adjustable scaling factors based on target resolution
    \item \textbf{Quality scaling}: We applied a quality multiplier (5.0× base resolution factor) to ensure clear structural visibility
    \item \textbf{Font and layout fidelity}: Automatic detection and usage of appropriate fonts, particularly critical for Xiangqi (Chinese characters) and Sudoku (numbers)
\end{itemize}

\subsec{Prompts}    
We use different prompts for different question types to test VLMs' visual counting versus prior knowledge:

\begin{enumerate}
    \item \textbf{Counting questions (Q1 \& Q2)}:
    \begin{itemize}
        \item \textbf{Q1 (Chess):} \emph{How many [rows/columns] are there on this board? Answer with a number in curly brackets, e.g., \{9\}.}
        \item \textbf{Q1 (Xiangqi, Go):} \emph{How many [horizontal/vertical] are there on this board? Answer with a number in curly brackets, e.g., \{9\}.}
        \item \textbf{Q1 (Sudoku):} \emph{How many [rows/columns] are there on this puzzle? Answer with a number in curly brackets, e.g., \{9\}.}

        \item \textbf{Q2 (Chess):} \emph{Count the [rows/columns] on this board. Answer with a number in curly brackets, e.g., \{9\}.}
        \item \textbf{Q2 (Xiangqi, Go):} \emph{Count the [horizontal/vertical] lines on this board. Answer with a number in curly brackets, e.g., \{9\}.}
        \item \textbf{Q2 (Sudoku):} \emph{Count the [rows/columns] on this puzzle. Answer with a number in curly brackets, e.g., \{9\}.}
    \end{itemize}
    
    \item \textbf{Y/N identification questions (Q3)}:
    \begin{itemize}
        \item \textbf{Q3 (Chess):} \emph{Is this a 8×8 Chessboard? Answer in curly brackets, e.g., \{Yes\} or \{No\}.}
        \item \textbf{Q3 (Xiangqi):} \emph{Is this a 10×9 Xiangqi board? Answer in curly brackets, e.g., \{Yes\} or \{No\}.}
        \item \textbf{Q3 (Sudoku):} \emph{Is this a 9×9 Sudoku puzzle? Answer in curly brackets, e.g., \{Yes\} or \{No\}.}
        \item \textbf{Q3 (Go):} \emph{Is this a 19×19 Go board? Answer in curly brackets, e.g., \{Yes\} or \{No\}.}
    \end{itemize}
\end{enumerate}


\subsec{Ground truth calculation}
We calculate the ground truth answers for each prompt type:
\begin{itemize}
    \item \textbf{Row/Column count (Q1 \& Q2)}:
    \begin{itemize}
        \item \textbf{Correct answer}: The actual number of rows/columns after modification. For example, if one row is removed from a 9×9 Sudoku, the row count is 8.
        \item \textbf{Expected bias}: The standard count for that board type (e.g., 8 for Chess rows, 10 for Xiangqi horizontal lines, 9 for Sudoku rows, 19 for Go horizontal lines)
    \end{itemize}
    
    \item \textbf{Standard layout question (Q3)}:
    \begin{itemize}
        \item \textbf{Correct answer}: Always ``No'' (since all boards have been modified from standard dimensions)
        \item \textbf{Expected bias}: ``Yes'' (since the boards closely resemble their standard counterparts)
    \end{itemize}
\end{itemize}


\subsection{Qualitative results}

\begin{figure}[h]
\centering
\small
\begin{AIbox}{Game Boards}

    \raggedright
\footnotesize
      \textbf{(a):} How many \textbf{columns} are there on this puzzle? Answer with a number in curly brackets, e.g., \{9\}. \\
      \textbf{(b), (c):} How many \textbf{horizontal lines} are there on this board? Answer with a number in curly brackets, e.g., \{9\}. \\
      \textbf{(d):} How many \textbf{rows} are there on this board? Answer with a number in curly brackets, e.g., \{9\}. \\
     \hfill \break
\footnotesize
\centering

    \begin{tabular}{lp{0.2cm}c|p{0.2cm}c|p{0.2cm}c|p{0.2cm}c}
    &\multicolumn{2}{c}{\makecell{\textbf{(a) Sudoku}}} 
    &\multicolumn{2}{c}{\makecell{\textbf{(b) Go}}}
    &\multicolumn{2}{c}{\makecell{\textbf{(c) Xiangqi}}}
    &\multicolumn{2}{c}{\makecell{\textbf{(d) Chess}}} 
    \\
     &\multicolumn{2}{c}{\includegraphics[width=0.20\textwidth]{images/game_board_examples/sudoku_grid_07_col_remove_first_col_notitle_px1152.png}}
     &\multicolumn{2}{c}{\includegraphics[width=0.20\textwidth]{images/game_board_examples/go_grid_03_add_row_px1152.png}}
     &\multicolumn{2}{c}{\includegraphics[width=0.20\textwidth]{images/game_board_examples/xiangqi_grid_04_row_add_row_in_river_bottom_notitle_px1152.png}}
     &\multicolumn{2}{c}{\includegraphics[width=0.20\textwidth]{images/game_board_examples/chess_grid_02_row_remove_last_row_notitle_px1152.png}}
     \\ 

\rowcolor{lightgray}
\raisebox{-0.2\height}\geminilogo & \centering 9
& \textcolor{red}{\xmark} & \centering 13
& \textcolor{red}{\xmark} & \centering 10
& \textcolor{red}{\xmark} & \centering 6
& \textcolor{red}{\xmark} \\

\raisebox{-0.2\height}\newsonnetlogo & \centering 9
& \textcolor{red}{\xmark} & \centering 19
& \textcolor{red}{\xmark} & \centering 10
& \textcolor{red}{\xmark} & \centering 8
& \textcolor{red}{\xmark} \\

\rowcolor{lightgray}
\raisebox{-0.2\height}\gptfourlogo & \centering 9
& \textcolor{red}{\xmark} & \centering 19
& \textcolor{red}{\xmark} & \centering 10
& \textcolor{red}{\xmark} & \centering 8
& \textcolor{red}{\xmark} \\

\raisebox{-0.2\height}\gptfulllogo & \centering 9
& \textcolor{red}{\xmark} & \centering 19
& \textcolor{red}{\xmark} & \centering 10
& \textcolor{red}{\xmark} & \centering 8
& \textcolor{red}{\xmark} \\

\rowcolor{lightgray}
\raisebox{-0.2\height}\gptminilogo & \centering 9
& \textcolor{red}{\xmark} & \centering 19
& \textcolor{red}{\xmark} & \centering 12
& \textcolor{red}{\xmark} & \centering 8
& \textcolor{red}{\xmark} \\ 

\midrule
\textbf{Bias} & \centering 9
& \textcolor{red}{\xmark} & \centering 19
& \textcolor{red}{\xmark} & \centering 10
& \textcolor{red}{\xmark} & \centering 8
& \textcolor{red}{\xmark} \\

\rowcolor{lightgray}
\textbf{GT} & \centering 8
& \textcolor{ForestGreen}{\cmark} & \centering 20
& \textcolor{ForestGreen}{\cmark} & \centering 11
& \textcolor{ForestGreen}{\cmark} & \centering 7
& \textcolor{ForestGreen}{\cmark} \\

     \end{tabular}
    \vspace{4pt}
    \centering
    \begin{tabular}{cccccccccccccc}
     \raisebox{-0.1\height}\geminilogo~\gemini 
     & \raisebox{-0.1\height}\newsonnetlogo~\newsonnet 
     & \raisebox{-0.1\height}\gptfourlogo~\gptfour 
     & \raisebox{-0.1\height}\gptfulllogo~\gptfull 
     & \raisebox{-0.1\height}\gptminilogo~\gptmini
     \\
    \end{tabular}

\end{AIbox}
\caption{VLMs are biased when counting the rows and columns on \boardgameicon~game boards.}
\label{appfig:qualitative_results_game_board}
\end{figure}



\clearpage
\section{Task 6: Visual testing with both original and modified optical illusion \illusionicon}\label{sec:appendix_illusion}

\begin{figure}[h]
    \centering
    \includegraphics[width=0.9\textwidth]{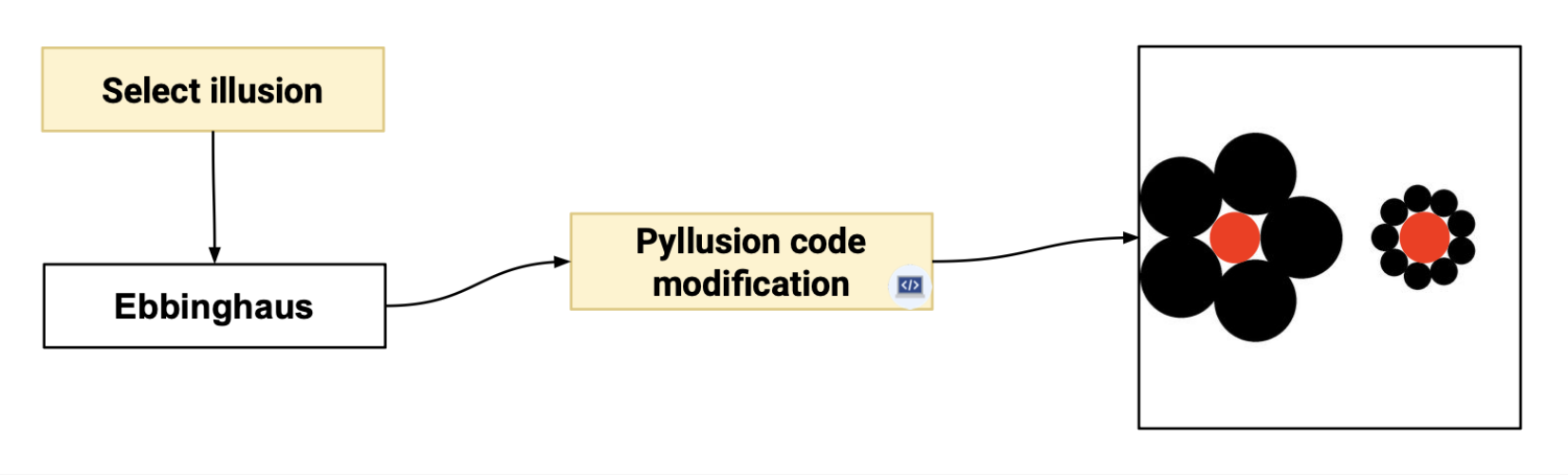}
    \caption{Data generation pipeline for Task 6: Visual testing with both original and modified optical illusion
    }
    \label{fig:illusion_diagram}
\end{figure}

\subsection{Task design}
\label{app:illusion-task-design}
Recent VLMs show improved performance on optical illusion tasks, with \texttt{o4-mini} achieving 71.49\% accuracy on IllusionVQA. However, these VLMs might have merely memorized the common optical illusions rather than truly perceiving visual information. To investigate this hypothesis, we test their ability to correctly identify illusion effects on both original and strategically modified versions. We design our task with careful control of visual parameters to ensure systematic evaluation:

\begin{itemize}
    \item \textbf{\colorbox{blue!15}{Illusion types}}: We use \textcolor{blue!80}{\textbf{6}} different classical optical illusions: \{\emph{Ebbinghaus}, \emph{Müller-Lyer}, \emph{Ponzo}, \emph{Vertical-Horizontal}, \emph{Zöllner}, \emph{Poggendorff}\}
    
    \item \textbf{\colorbox{green!15}{Condition types}}: Each illusion has \textcolor{green!80}{\textbf{2}} conditions:
        \begin{itemize}
            \item \emph{Original}: Standard illusion where the visual effect should occur (e.g., two identical circles appearing different sizes).
            \item \emph{Modified}: Reversed version where the actual measurements contradict the typical illusion effect (e.g., circles that are genuinely different sizes).
        \end{itemize}
        
    \item \textbf{\colorbox{orange!15}{Parameter variations}}: We generate \textcolor{orange!80}{\textbf{multiple combinations}} of illusion parameters:
        \begin{itemize}
            \item Most illusions: 12 original + 12 modified versions with varying illusion strength and difference
            \item Vertical-Horizontal: 6 original + 6 modified versions (fixed T-shape structure)
        \end{itemize}
    
    \item \textbf{\colorbox{purple!15}{Image resolutions}}: We generate each illusion at \textcolor{purple!80}{\textbf{3}} different pixel sizes \{384, 768, 1152\}px to test resolution sensitivity.
\end{itemize}

This systematic approach generates a total of (12 original + 12 modified) $\times$ \textcolor{blue!80}{\textbf{5}} illusion types + (6 original + 6 modified) $\times$ \textcolor{blue!80}{\textbf{1}} Vertical-Horizontal illusion) $\times$ \textcolor{purple!80}{\textbf{3}} resolutions = \textbf{396} total images.

\subsection{Implementation and prompts}
\label{app:illusion-task-implementation}

\subsec{Implementation details} 
Our implementation adapts code from Pyllusion (\url{https://github.com/RealityBending/Pyllusion}) to generate consistent, parametrically controlled optical illusions. We systematically vary two key parameters: \emph{illusion strength} (which controls the intensity of contextual elements that create the illusion effect, representing how strongly the surrounding context biases perceptual experience) and \emph{difference} (which controls the objective, actual difference between target elements being compared, where 0 means identical elements and non-zero values create genuine physical differences).

The algorithm for all illusion types follows the same sequence:
\begin{enumerate}
    \item Define parameter ranges for each illusion type (strength values, difference values).
    \item Generate original versions with standard illusion parameters (diff=0 for equal elements).
    \item Generate modified versions with reversed parameters (diff$\neq$0 for unequal elements).
    \item Render each illusion variant at three different resolutions.
\end{enumerate}

The illusion-specific implementations required special parameter handling for:
\begin{itemize}
    \item \textbf{Ebbinghaus}: Varying surrounding circle sizes (strength) and central circle differences (difference).
    \item \textbf{Müller-Lyer}: Different arrowhead angles (strength) and line length differences (difference).
    \item \textbf{Ponzo}: Perspective line angles (strength) and horizontal bar length differences (difference).
    \item \textbf{Vertical-Horizontal}: Fixed T-shape with varying line length ratios (difference).
    \item \textbf{Zöllner}: Background line angles (strength) and main line parallelism differences (difference).
    \item \textbf{Poggendorff}: Interrupting rectangle positions (strength) and diagonal line alignments (difference).
\end{itemize}

\subsec{Quality control}
To ensure consistent image quality and valid illusion effects across all variants, we implemented several technical measures:
\begin{itemize}
    \item \textbf{Parameter validation}: Ensured all strength and difference values produce visually meaningful illusions, with diff$\neq$ 0 cases design to be easily recognizable by humans to distinguish actual physical differences from perceptual biases clearly.
    \item \textbf{Balanced generation}: Equal numbers of diff=0 (original) and diff$\neq$0 (modified) cases per illusion type
\end{itemize}

\subsec{Prompts}    
We use consistent prompts across illusion types to test VLMs' visual perception versus memorized knowledge:

\begin{enumerate}
    \item \textbf{Main questions (Q1 \& Q2)}:
    \begin{itemize}
        \item \textbf{Q1 (Ebbinghaus):} \emph{Are the two red circles equal in size? Answer in curly brackets, e.g., \{Yes\} or \{No\}.}
        \item \textbf{Q1 (Müller-Lyer, Ponzo):} \emph{Are the two horizontal lines equal in length? Answer in curly brackets, e.g., \{Yes\} or \{No\}.}
        \item \textbf{Q1 (Vertical-Horizontal):} \emph{Are the horizontal and vertical lines equal in length? Answer in curly brackets, e.g., \{Yes\} or \{No\}.}
        \item \textbf{Q1 (Zöllner):} \emph{Are the two horizontal lines parallel? Answer in curly brackets, e.g., \{Yes\} or \{No\}.}
        \item \textbf{Q1 (Poggendorff):} \emph{Are the two diagonal line segments aligned? Answer in curly brackets, e.g., \{Yes\} or \{No\}.}
        
        \item \textbf{Q2 (Ebbinghaus):} \emph{Do the two red circles have the same size? Answer in curly brackets, e.g., \{Yes\} or \{No\}.}
        \item \textbf{Q2 (Müller-Lyer):} \emph{Do the two horizontal lines have the same length? Answer in curly brackets, e.g., \{Yes\} or \{No\}.}
        \item \textbf{Q2 (Ponzo):} \emph{Do the two horizontal lines have the same length? Answer in curly brackets, e.g., \{Yes\} or \{No\}.}
        \item \textbf{Q2 (Vertical-Horizontal):} \emph{Do the horizontal and vertical lines have the same length? Answer in curly brackets, e.g., \{Yes\} or \{No\}.}
        \item \textbf{Q2 (Zöllner):} \emph{Do the two horizontal lines run parallel? Answer in curly brackets, e.g., \{Yes\} or \{No\}.}
        \item \textbf{Q2 (Poggendorff):} \emph{Do the two diagonal lines form a straight line? Answer in curly brackets, e.g., \{Yes\} or \{No\}.}
    \end{itemize}
    
    \item \textbf{Y/N identification questions (Q3)}:
    \begin{itemize}
        \item \textbf{Q3:} \emph{Is this an example of the [Ebbinghaus/Müller-Lyer/Ponzo/Vertical-Horizontal/Zöllner/Poggendorff] illusion? Answer in curly brackets, e.g., \{Yes\} or \{No\}.}
    \end{itemize}
\end{enumerate}

\subsec{Ground truth calculation}
We calculate the ground truth answers based on the actual measurements in each image:

\begin{itemize}
    \item \textbf{Counting questions (Q1 \& Q2)}:
    \begin{itemize}
        \item \textbf{Correct answer}:
        \begin{itemize}
            \item \textbf{Original illusions (diff=0)}: Elements are actually equal, so the correct answer is ``Yes''
            \item \textbf{Modified illusions (diff$\neq$0)}: Elements are actually different, so the correct answer is ``No''
        \end{itemize}
        \item \textbf{Expected bias}:
        \begin{itemize}
            \item \textbf{Original illusions}: VLMs might incorrectly say ``No'' expecting the illusion effect to make equal elements appear different
            \item \textbf{Modified illusions}: VLMs might incorrectly say ``Yes'' expecting the illusion to make genuinely different elements appear equal
        \end{itemize}
    \end{itemize}
    
    \item \textbf{Y/N identification questions (Q3)}:
    \begin{itemize}
        \item \textbf{Correct answer}:
        \begin{itemize}
            \item \textbf{Original illusions}: ``Yes'' (standard examples of the specified illusion type).
            \item \textbf{Modified illusions}: ``No'' (modified versions that contradict typical illusion effects).
        \end{itemize}
        \item \textbf{Expected bias}:
        \begin{itemize}
            \item \textbf{Original illusions}: VLMs likely correctly identify as ``Yes'' since they match memorized illusion patterns
            \item \textbf{Modified illusions}: VLMs may incorrectly say ``Yes'' if they rely on visual similarity rather than recognizing the effect contradiction
        \end{itemize}
    \end{itemize}
\end{itemize}


\subsection{Qualitative results}
\begin{figure}[h]
\centering
\small
\begin{AIbox}{Abstract images: Optical Illusions}
\centering

    \begin{tabular}{lp{0.6cm}c|p{0.6cm}c|p{0.6cm}c|p{0.6cm}c|p{0.6cm}c|p{0.6cm}c}
    &\multicolumn{2}{c}{\makecell{\textbf{(a) Original}\\Müller-Lyer}}
    &\multicolumn{2}{c}{\makecell{\textbf{(b) Modified}\\Müller-Lyer}}
    &\multicolumn{2}{c}{\makecell{\textbf{(c) Original}\\Zöllner}}
    &\multicolumn{2}{c}{\makecell{\textbf{(d) Modified}\\Zöllner}} 
    &\multicolumn{2}{c}{\makecell{\textbf{(e) Original}\\Ebbinghaus}}
    &\multicolumn{2}{c}{\makecell{\textbf{(f) Modified}\\Ebbinghaus}} 
    \\
    
     &\multicolumn{2}{c}{\includegraphics[width=0.12\textwidth]{images/illusion_example/MullerLyer_007_str45_diff0_notitle_px1152.png}} 
     & \multicolumn{2}{c}{\includegraphics[width=0.12\textwidth]{images/illusion_example/MullerLyer_023_str50_diff0p5_notitle_px1152.png}} 
     & \multicolumn{2}{c}{\includegraphics[width=0.12\textwidth]{images/illusion_example/Zollner_007_str65_diff0_notitle_px1152.png}} 
     & \multicolumn{2}{c}{\includegraphics[width=0.12\textwidth]{images/illusion_example/Zollner_021_str75_diff2p5_notitle_px1152.png}}
     & \multicolumn{2}{c}{\includegraphics[width=0.12\textwidth]{images/illusion_example/Ebbinghaus_002_strneg3_diff0_notitle_px1152.png}} 
     & \multicolumn{2}{c}{\includegraphics[width=0.12\textwidth]{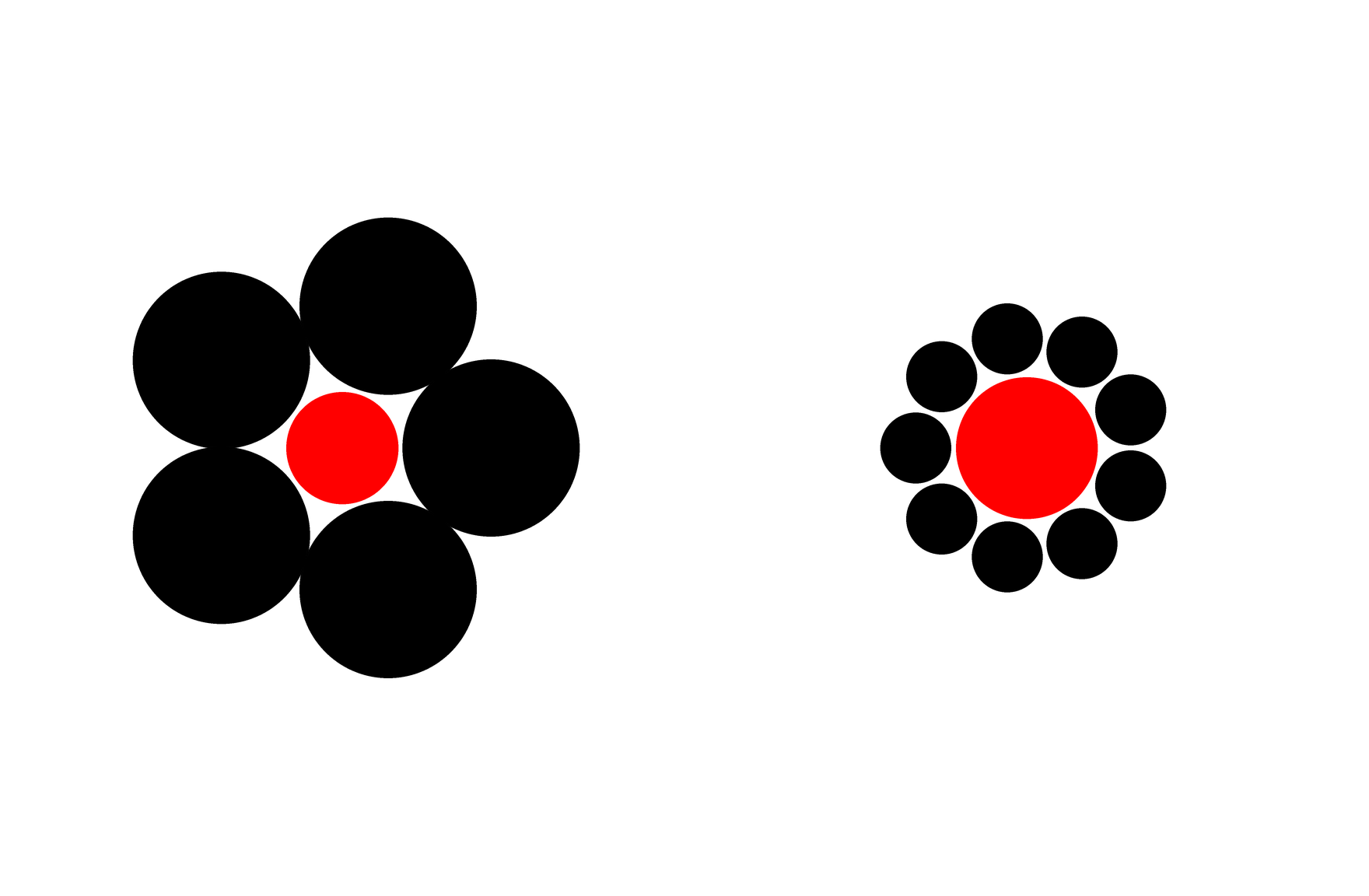}}
     \\ 
\rowcolor{lightgray}
\raisebox{-0.2\height}\geminilogo & \centering Yes
& \textcolor{ForestGreen}{\cmark} & \centering Yes
& \textcolor{red}{\xmark} & \centering Yes
& \textcolor{ForestGreen}{\cmark} & \centering Yes
& \textcolor{red}{\xmark} & \centering Yes
& \textcolor{ForestGreen}{\cmark} & \centering Yes
& \textcolor{red}{\xmark} \\

\raisebox{-0.2\height}\newsonnetlogo & \centering Yes
& \textcolor{ForestGreen}{\cmark} & \centering Yes
& \textcolor{red}{\xmark} & \centering Yes
& \textcolor{ForestGreen}{\cmark} & \centering Yes
& \textcolor{red}{\xmark} & \centering No
& \textcolor{red}{\xmark} & \centering No
& \textcolor{ForestGreen}{\cmark} \\

\rowcolor{lightgray}
\raisebox{-0.2\height}\gptfourlogo & \centering Yes
& \textcolor{ForestGreen}{\cmark} & \centering Yes
& \textcolor{red}{\xmark} & \centering Yes
& \textcolor{ForestGreen}{\cmark} & \centering Yes
& \textcolor{red}{\xmark} & \centering Yes
& \textcolor{ForestGreen}{\cmark} & \centering Yes
& \textcolor{red}{\xmark} \\

\raisebox{-0.2\height}\gptfulllogo & \centering Yes
& \textcolor{ForestGreen}{\cmark} & \centering Yes
& \textcolor{red}{\xmark} & \centering Yes
& \textcolor{ForestGreen}{\cmark} & \centering Yes
& \textcolor{red}{\xmark} & \centering Yes
& \textcolor{ForestGreen}{\cmark} & \centering Yes
& \textcolor{red}{\xmark} \\

\rowcolor{lightgray}
\raisebox{-0.2\height}\gptminilogo & \centering Yes
& \textcolor{ForestGreen}{\cmark} & \centering Yes
& \textcolor{red}{\xmark} & \centering Yes
& \textcolor{ForestGreen}{\cmark} & \centering Yes
& \textcolor{red}{\xmark} & \centering No
& \textcolor{red}{\xmark} & \centering Yes
& \textcolor{red}{\xmark} \\ 

\midrule
\textbf{Bias} & \centering No
& \textcolor{red}{\xmark} & \centering Yes
& \textcolor{red}{\xmark} & \centering No
& \textcolor{red}{\xmark} & \centering Yes
& \textcolor{red}{\xmark} & \centering No
& \textcolor{red}{\xmark} & \centering Yes
& \textcolor{red}{\xmark} \\

\rowcolor{lightgray}
\textbf{GT} & \centering Yes
& \textcolor{ForestGreen}{\cmark} & \centering No
& \textcolor{ForestGreen}{\cmark} & \centering Yes
& \textcolor{ForestGreen}{\cmark} & \centering No
& \textcolor{ForestGreen}{\cmark} & \centering Yes
& \textcolor{ForestGreen}{\cmark} & \centering No
& \textcolor{ForestGreen}{\cmark} \\ 
     \end{tabular}
    \vspace{4pt}
    \centering
    \begin{tabular}{cccccccccccccc}
     \raisebox{-0.1\height}\geminilogo~\gemini 
     & \raisebox{-0.1\height}\newsonnetlogo~\newsonnet 
     & \raisebox{-0.1\height}\gptfourlogo~\gptfour 
     & \raisebox{-0.1\height}\gptfulllogo~\gptfull 
     & \raisebox{-0.1\height}\gptminilogo~\gptmini
     \\
    \end{tabular}
     \hfill \break
    \hfill \break
    \raggedright
    \scriptsize
      \textbf{(a), (b):} Are the two horizontal lines \textbf{equal} in length? Answer in curly brackets, e.g., \{Yes\} or \{No\}. \\
      \textbf{(c), (d):} Are the two horizontal lines \textbf{parallel}? Answer in curly brackets, e.g., \{Yes\} or \{No\}. \\
      \textbf{(e), (f):} Are the two red circles \textbf{equal} in size? Answer in curly brackets, e.g., \{Yes\} or \{No\}. \\
      
\end{AIbox}
\caption{VLMs show systematic biases, often relying on prior knowledge of \illusionicon~optical illusions rather than directly interpreting the image.}
\label{fig:appendix_optical_illusion_qualitative}
\end{figure}

\begin{figure}[h]
\centering
\small
\begin{AIbox}{Abstract images: Optical Illusions}
\centering

    \begin{tabular}{lp{0.6cm}c|p{0.6cm}c|p{0.6cm}c|p{0.6cm}c|p{0.6cm}c|p{0.6cm}c}
    &\multicolumn{2}{c}{\makecell{\textbf{(a) Original}\\Ponzo}}
    &\multicolumn{2}{c}{\makecell{\textbf{(b) Modified}\\Ponzo}}
    &\multicolumn{2}{c}{\makecell{\textbf{(c) Original}\\V-H}}
    &\multicolumn{2}{c}{\makecell{\textbf{(d) Modified}\\V-H}} 
    &\multicolumn{2}{c}{\makecell{\textbf{(e) Original}\\Poggendorff}}
    &\multicolumn{2}{c}{\makecell{\textbf{(f) Modified}\\Poggendorff}} 
    \\
    
     &\multicolumn{2}{c}{\includegraphics[width=0.11\textwidth]{images/illusion_example/Ponzo_001_str15_diff0_notitle_px1152.png}} 
     & \multicolumn{2}{c}{\includegraphics[width=0.11\textwidth]{images/illusion_example/Ponzo_017_str15_diff0p25_notitle_px1152.png}} 
     & \multicolumn{2}{c}{\includegraphics[width=0.11\textwidth]{images/illusion_example/VerticalHorizontal_001_min0p5_diff0_notitle_px1152.png}} 
     & \multicolumn{2}{c}{\includegraphics[width=0.11\textwidth]{images/illusion_example/VerticalHorizontal_007_min0p5_diff0p3_notitle_px1152.png}}
     & \multicolumn{2}{c}{\includegraphics[width=0.11\textwidth]{images/illusion_example/Poggendorff_001_str30_diff0_notitle_px1152.png}} 
     & \multicolumn{2}{c}{\includegraphics[width=0.11\textwidth]{images/illusion_example/Poggendorff_013_str40_diff0p1_notitle_px1152.png}}
     \\ 
\rowcolor{lightgray}
\raisebox{-0.2\height}\geminilogo & \centering Yes
& \textcolor{ForestGreen}{\cmark} & \centering No
& \textcolor{ForestGreen}{\cmark} & \centering No
& \textcolor{red}{\xmark} & \centering No
& \textcolor{ForestGreen}{\cmark} & \centering Yes
& \textcolor{ForestGreen}{\cmark} & \centering Yes
& \textcolor{red}{\xmark} \\

\raisebox{-0.2\height}\newsonnetlogo & \centering Yes
& \textcolor{ForestGreen}{\cmark} & \centering Yes
& \textcolor{red}{\xmark} & \centering No
& \textcolor{red}{\xmark} & \centering No
& \textcolor{ForestGreen}{\cmark} & \centering No
& \textcolor{red}{\xmark} & \centering No
& \textcolor{ForestGreen}{\cmark} \\

\rowcolor{lightgray}
\raisebox{-0.2\height}\gptfourlogo & \centering Yes
& \textcolor{ForestGreen}{\cmark} & \centering Yes
& \textcolor{red}{\xmark} & \centering No
& \textcolor{red}{\xmark} & \centering No
& \textcolor{ForestGreen}{\cmark} & \centering Yes
& \textcolor{ForestGreen}{\cmark} & \centering Yes
& \textcolor{red}{\xmark} \\

\raisebox{-0.2\height}\gptfulllogo & \centering Yes
& \textcolor{ForestGreen}{\cmark} & \centering Yes
& \textcolor{red}{\xmark} & \centering No
& \textcolor{red}{\xmark} & \centering No
& \textcolor{ForestGreen}{\cmark} & \centering Yes
& \textcolor{ForestGreen}{\cmark} & \centering Yes
& \textcolor{red}{\xmark} \\

\rowcolor{lightgray}
\raisebox{-0.2\height}\gptminilogo & \centering Yes
& \textcolor{ForestGreen}{\cmark} & \centering Yes
& \textcolor{red}{\xmark} & \centering No
& \textcolor{red}{\xmark} & \centering No
& \textcolor{ForestGreen}{\cmark} & \centering Yes
& \textcolor{ForestGreen}{\cmark} & \centering Yes
& \textcolor{red}{\xmark} \\ 

\midrule
\textbf{Bias} & \centering No
& \textcolor{red}{\xmark} & \centering Yes
& \textcolor{red}{\xmark} & \centering No
& \textcolor{red}{\xmark} & \centering Yes
& \textcolor{red}{\xmark} & \centering No
& \textcolor{red}{\xmark} & \centering Yes
& \textcolor{red}{\xmark} \\

\rowcolor{lightgray}
\textbf{GT} & \centering Yes
& \textcolor{ForestGreen}{\cmark} & \centering No
& \textcolor{ForestGreen}{\cmark} & \centering Yes
& \textcolor{ForestGreen}{\cmark} & \centering No
& \textcolor{ForestGreen}{\cmark} & \centering Yes
& \textcolor{ForestGreen}{\cmark} & \centering No
& \textcolor{ForestGreen}{\cmark} \\ 
     \end{tabular}
    \vspace{4pt}
    \centering
    \begin{tabular}{cccccccccccccc}
     \raisebox{-0.1\height}\geminilogo~\gemini 
     & \raisebox{-0.1\height}\newsonnetlogo~\newsonnet 
     & \raisebox{-0.1\height}\gptfourlogo~\gptfour 
     & \raisebox{-0.1\height}\gptfulllogo~\gptfull 
     & \raisebox{-0.1\height}\gptminilogo~\gptmini
     \\
    \end{tabular}
     \hfill \break
    \hfill \break
    \raggedright
    \scriptsize
      \textbf{(a), (b):} Are the two horizontal lines \textbf{equal} in length? Answer in curly brackets, e.g., \{Yes\} or \{No\}. \\
      \textbf{(c), (d):} Are the horizontal and vertical lines \textbf{equal} in length? Answer in curly brackets, e.g., \{Yes\} or \{No\}. \\
      \textbf{(e), (f):} Are the two diagonal line segments \textbf{aligned}? Answer in curly brackets, e.g., \{Yes\} or \{No\}. \\
      
\end{AIbox}
\caption{VLMs show systematic biases, often relying on prior knowledge of optical illusions (\eg, Ponzo and Poggendorff illusions) rather than directly interpreting the image. In contrast, in the vertical–horizontal illusion, VLMs respond like humans. They are misled by the illusion itself, leading them to answer the original question incorrectly rather than the counterfactual ones.}
\label{fig:appendix_optical_illusion_qualitative_2}
\end{figure}


\clearpage
\section{Task 7: Counting circles or lines in an anomaly cell within a patterned grid~\patternedgridicon}\label{sec:appendix_patterned_grid}

\begin{figure}[h]
    \centering
    \includegraphics[width=0.9\textwidth]{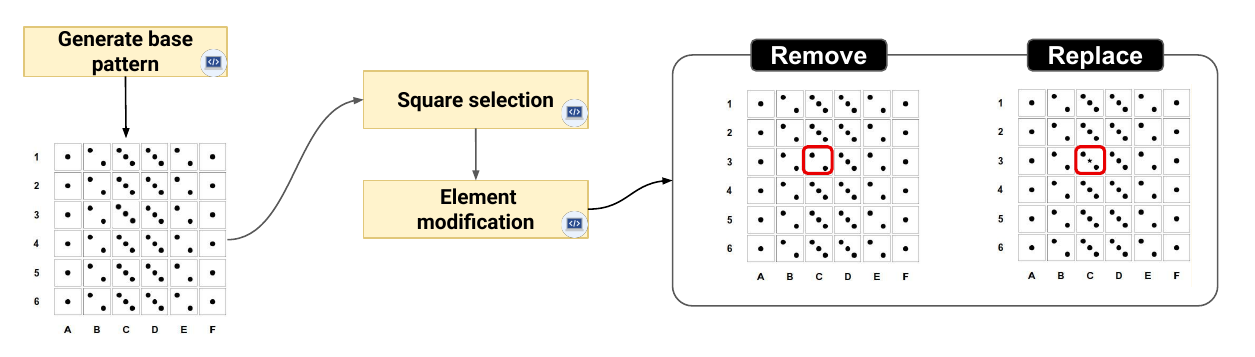}
    \caption{Data generation pipeline for Task 7: Counting circles or lines in an anomaly cell within a patterned grid}
    \label{fig:pattern_diagram}
\end{figure}


\subsection{Task design}
\label{app:grid-pattern-task-design}
VLMs can infer patterns from nearby visual elements to answer visual questions~\citep{huang2024visual}. To evaluate whether VLMs rely on pattern recognition over actual visual counting, we create square grids with systematic numerical patterns (represented visually by dice faces or tally marks) where exactly one cell violates the expected pattern. We hypothesize that VLMs will prioritize the inferred pattern over the actual visual information and report the expected pattern-completing value instead of the true count. We design our task with careful control of visual parameters to ensure systematic evaluation:

\begin{itemize}
    \item \textbf{\colorbox{blue!15}{Grid types}}: We use \textcolor{blue!80}{\textbf{2}} different visual representation types: \{\emph{dice} (circular dots in dice-face patterns), \emph{tally} (traditional tally mark lines)\}.

    \item \textbf{\colorbox{green!15}{Modification types per grid type}}: For each grid type, we apply \textcolor{green!80}{\textbf{2}} distinct types of cell-level modifications:
        \begin{itemize}
            \item \emph{Dice grids}: Remove (one dot is removed from a cell) and Replace (one dot is replaced with a different shape, like a square or star, within a cell).
            \item \emph{Tally grids}: Remove (one tally line is removed from a cell) and Add (one extra tally line is added to a cell).
        \end{itemize}

    \item \textbf{\colorbox{orange!15}{Grid Dimensions}}: We generate grids of \textcolor{orange!80}{\textbf{7}} different dimensions, ranging from 6$\times$6 to 12$\times$12 cells.

    \item \textbf{\colorbox{purple!15}{Unique scenarios for anomaly placement (single anomaly per grid image)}}: To create \textcolor{purple!80}{\textbf{14}} distinct base settings for placing anomalies, \textbf{where each final grid image will feature only a single modified cell}. We proceed as follows: for each of the \textcolor{orange!80}{\textbf{7}} grid dimensions, we define \underline{two} separate base settings. Each of these two settings for a given grid dimension involves selecting a \emph{different}, unique cell location to be the \emph{sole} anomaly cell for images generated under that specific setting. These potential anomaly cell locations are carefully chosen to avoid edges and corners. This gives us (7 grid dimensions $\times$ 2 distinct choices of a single anomaly cell location per dimension) = 14 distinct base settings. For each of these 14 base settings (defined by a grid dimension and the location of its single anomaly cell), we then apply all combinations of grid types and their respective modifications to generate the final images, each still containing only that one pre-determined anomaly.

    \item \textbf{\colorbox{red!15}{Image resolutions}}: Each generated grid image is rendered at \textcolor{red!80}{\textbf{3}} different pixel sizes \{384, 768, 1152\}px to assess sensitivity to image resolution.
\end{itemize}
This systematic generation process yields a total of \textcolor{blue!80}{\textbf{2}} (grid types) $\times$ \textcolor{green!80}{\textbf{2}} (modification types) $\times$ \textcolor{purple!80}{\textbf{14}} (unique scenarios) $\times$ \textcolor{red!80}{\textbf{3}} (resolutions) = \textbf{168} distinct images.

\subsection{Implementation and prompts}
\label{app:grid-pattern-task-implementation}

\subsec{Implementation details} 
Our implementation generates systematic pattern grids using a distance-from-edge algorithm to create naturally increasing-then-decreasing numerical patterns. For dice grids, we use circular dots arranged in traditional dice-face configurations (1-6 dots per cell). For tally grids, we render authentic tally marks with proper grouping (four vertical lines crossed by a diagonal fifth line).

The algorithm for both grid types follows the same sequence:
\begin{enumerate}
    \item Generate base grid with pattern-consistent cell counts using distance-from-edge calculation
    \item Organize target positions across 14 groups, with each group containing both dice and tally variants
    \item For each target cell, create modification variants:
        \begin{itemize}
            \item \textbf{Dice}: Remove one dot OR replace one dot with alternative shape (triangle, square, star)
            \item \textbf{Tally}: Remove one line OR add one extra line
        \end{itemize}
    \item Render each modified grid at three different resolutions with consistent visual quality
\end{enumerate}

The grid-specific implementations required special handling for:
\begin{itemize}
    \item \textbf{Dice pattern consistency}: Maintaining standard dice-face arrangements (1-6 dots) while allowing single-dot modifications
    \item \textbf{Tally mark authenticity}: Proper grouping of marks with diagonal crosses for every fifth line
    \item \textbf{Pattern calculation}: Distance-from-edge algorithm ensuring natural numerical progression across grid cells
    \item \textbf{Cell positioning}: Strategic selection of anomaly cells away from edges to preserve pattern context
\end{itemize}

\subsec{Quality control}
To ensure consistent image quality and valid pattern recognition challenges across all variants, we implemented several technical measures:
\begin{itemize}
    \item \textbf{SVG to PNG conversion}: We used direct SVG rendering with adjustable scaling factors based on target resolution
    \item \textbf{Quality scaling}: We applied a quality multiplier (5.0× base resolution factor) to ensure clear shape and line visibility
\end{itemize}

\subsec{Prompts}    
We use consistent prompts across both grid types to test VLMs' pattern recognition versus actual visual counting:

\begin{enumerate}
    \item \textbf{Counting questions (Q1 \& Q2)}:
    \begin{itemize}
        \item \textbf{Q1 (Dice):} \emph{How many circles are there in cell [CellID]? Answer with a number in curly brackets, e.g., \{9\}.}
        \item \textbf{Q1 (Tally):} \emph{How many lines are there in cell [CellID]? Answer with a number in curly brackets, e.g., \{9\}.}
        
        \item \textbf{Q2 (Dice):} \emph{Count the circles in cell [CellID]. Answer with a number in curly brackets, e.g., \{9\}.}
        \item \textbf{Q2 (Tally):} \emph{Count the lines in cell [CellID]. Answer with a number in curly brackets, e.g., \{9\}.}
    \end{itemize}
    
    \item \textbf{Y/N identification questions (Q3)}:
    \begin{itemize}
        \item \textbf{Q3 (Dice):} \emph{Does cell [CellID] contain [ExpectedCount] circles? Answer in curly brackets, e.g., \{Yes\} or \{No\}.}
        \item \textbf{Q3 (Tally):} \emph{Does cell [CellID] contain [ExpectedCount] lines? Answer in curly brackets, e.g., \{Yes\} or \{No\}.}
    \end{itemize}
\end{enumerate}

For all prompts, [CellID] refers to the specific anomaly cell using standard spreadsheet notation (e.g., C3, F7), and [ExpectedCount] represents the pattern-consistent count that would be expected based on surrounding cells.

\subsec{Ground truth calculation}
We calculate the ground truth answers based on the actual visual content in each modified cell:

\begin{itemize}
    \item \textbf{Direct counting questions (Q1 \& Q2)}:
    \begin{itemize}
        \item \textbf{Correct answer}: The actual count of visual elements in the target cell after modification
        \begin{itemize}
            \item For \emph{Remove modifications}: Standard pattern count minus 1
            \item For \emph{Add modifications}: Standard pattern count plus 1  
            \item For \emph{Replace modifications}: Standard pattern count minus 1 (since one circle is replaced with a different shape)
        \end{itemize}
        \item \textbf{Expected bias}: The pattern-consistent count that VLMs might infer from surrounding cells, ignoring the actual modification
    \end{itemize}
    
    \item \textbf{Pattern-based verification question (Q3)}:
    \begin{itemize}
        \item \textbf{Correct answer}: Always ``No'' (since the target cell has been modified to break the pattern)
        \item \textbf{Expected bias}: ``Yes'' (if VLMs rely on pattern inference rather than direct visual inspection)
    \end{itemize}
\end{itemize}

\subsection{Qualitative results}
\begin{figure}[h!]
\centering
\small
\begin{AIbox}{Abstract Images: Patterned Grid}
\centering

    \begin{tabular}{lp{0.8cm}c|p{0.8cm}c|p{0.8cm}c|p{0.8cm}c}
    &\multicolumn{2}{c}{\makecell{\textbf{(a) Dice}\\Remove}}
    &\multicolumn{2}{c}{\makecell{\textbf{(b) Dice}\\Replace}}
    &\multicolumn{2}{c}{\makecell{\textbf{(c) Tally}\\Remove}}
    &\multicolumn{2}{c}{\makecell{\textbf{(d) Tally}\\Add}} 

    \\
    
     &\multicolumn{2}{c}{\includegraphics[width=0.2\textwidth]{images/patterned_grid_examples/dice_001_remove_C3_notitle_px1152.png}} 
     & \multicolumn{2}{c}{\includegraphics[width=0.2\textwidth]{images/patterned_grid_examples/dice_001_replace_C3_notitle_px1152.png}} 
     & \multicolumn{2}{c}{\includegraphics[width=0.2\textwidth]{images/patterned_grid_examples/tally_001_remove_C3_notitle_px1152.png}} 
     & \multicolumn{2}{c}{\includegraphics[width=0.2\textwidth]{images/patterned_grid_examples/tally_001_add_C3_notitle_px1152.png}}
     \\

\rowcolor{lightgray}
\raisebox{-0.2\height}\geminilogo & \centering 3
& \textcolor{red}{\xmark} & \centering 3
& \textcolor{red}{\xmark} & \centering 2
& \textcolor{ForestGreen}{\cmark} & \centering 3
& \textcolor{red}{\xmark} \\

\raisebox{-0.2\height}\newsonnetlogo & \centering 2
& \textcolor{ForestGreen}{\cmark} & \centering 2
& \textcolor{ForestGreen}{\cmark} & \centering 2
& \textcolor{ForestGreen}{\cmark} & \centering 4
& \textcolor{ForestGreen}{\cmark} \\

\rowcolor{lightgray}
\raisebox{-0.2\height}\gptfourlogo & \centering 3
& \textcolor{red}{\xmark} & \centering 3
& \textcolor{red}{\xmark} & \centering 3
& \textcolor{red}{\xmark} & \centering 4
& \textcolor{ForestGreen}{\cmark} \\

\raisebox{-0.2\height}\gptfulllogo & \centering 3
& \textcolor{red}{\xmark} & \centering 2
& \textcolor{ForestGreen}{\cmark} & \centering 3
& \textcolor{red}{\xmark} & \centering 4
& \textcolor{ForestGreen}{\cmark} \\

\rowcolor{lightgray}
\raisebox{-0.2\height}\gptminilogo & \centering 3
& \textcolor{red}{\xmark} & \centering 3
& \textcolor{red}{\xmark} & \centering 2
& \textcolor{ForestGreen}{\cmark} & \centering 3
& \textcolor{red}{\xmark} \\ 

\midrule
\textbf{Bias} & \centering 3
& \textcolor{red}{\xmark} & \centering 3
& \textcolor{red}{\xmark} & \centering 3
& \textcolor{red}{\xmark} & \centering 3
& \textcolor{red}{\xmark} \\

\rowcolor{lightgray}
\textbf{GT} & \centering 2
& \textcolor{ForestGreen}{\cmark} & \centering 2
& \textcolor{ForestGreen}{\cmark} & \centering 2
& \textcolor{ForestGreen}{\cmark} & \centering 4
& \textcolor{ForestGreen}{\cmark} \\ 

     \end{tabular}
    \vspace{4pt}
    \centering
    \begin{tabular}{cccccccccccccc}
     \raisebox{-0.1\height}\geminilogo~\gemini 
     & \raisebox{-0.1\height}\newsonnetlogo~\newsonnet 
     & \raisebox{-0.1\height}\gptfourlogo~\gptfour 
     & \raisebox{-0.1\height}\gptfulllogo~\gptfull 
     & \raisebox{-0.1\height}\gptminilogo~\gptmini
     \\
    \end{tabular}
     \hfill \break
    \hfill \break
    \raggedright
    \scriptsize
      \textbf{(a), (b):} How many \textbf{circles} are there in cell \textbf{C3}? Answer with a number in curly brackets, e.g., \{9\}. \\
      \textbf{(c), (d):} How many \textbf{lines} are there in cell \textbf{C3}? Answer with a number in curly brackets, e.g., \{9\}. \\
      
\end{AIbox}
\caption{All VLMs, except \newsonnetlogo~\newsonnet, fail to correctly identify the abnormal cell (C3) in both the \patternedgridicon~patterned grids.}
\label{fig:patterned_grid}
\end{figure}


\clearpage
\section{Details of prompts}\label{appsec:appendix_prompt_details}
\subsection{Examples of \qone, \qtwo and \qthree}\label{appsec:appendix_questions}

\begin{table}[h!]
\caption{Some examples of questions on \animalicon~animal, \logoicon~brand logos, and \flagicon~flags}
\small
\centering
\setlength{\tabcolsep}{4pt}
\renewcommand{\arraystretch}{1.2}
\begin{tabular}{ccp{3.3cm}p{3.3cm}p{3.3cm}}
\toprule
\rowcolor{headerblue!20}
\multicolumn{1}{c}{\textbf{Topic}} & \multicolumn{1}{c}{\textbf{Subtopic}} & \multicolumn{1}{c}{\textbf{Q1}} & \multicolumn{1}{c}{\textbf{Q2}} & \multicolumn{1}{c}{\textbf{Q3}} \\
\midrule
\multirow{1}{*}{\textbf{Animal}} & {\cellcolor{animal!15}} & 
{\cellcolor{animal!15}}How many legs does this animal have? Answer with a number in curly brackets, e.g., \{9\}. & {\cellcolor{animal!15}}Count the legs of this animal. Answer with a number in curly brackets, e.g., \{9\}. & {\cellcolor{animal!15}}Is this an animal with 4 legs? Answer in curly brackets, e.g., \{Yes\} or \{No\}. \\
\midrule
\multirow{5}{*}{\textbf{Logo}} & {\cellcolor{shoelogo!15}}Adidas & 
{\cellcolor{shoelogo!15}}How many visible stripes are there in the logo of the left shoe? Answer with a number in curly brackets, e.g., \{9\}. & {\cellcolor{shoelogo!15}}Count the visible stripes in the logo on the left shoe. Answer with a number in curly brackets, e.g., \{9\}. & {\cellcolor{shoelogo!15}}Are the logos on these shoes Adidas logos? Answer in curly brackets, e.g., \{Yes\} or \{No\}. \\
\cmidrule{2-5}
& {\cellcolor{shoelogo!25}}Nike & 
{\cellcolor{shoelogo!25}}How many visible white stylized curves are there in the logo of the left shoe? Answer with a number in curly brackets, e.g., \{9\}. & {\cellcolor{shoelogo!25}}Count the visible white stylized curves in the logo on the left shoe. Answer with a number in curly brackets, e.g., \{9\}. & {\cellcolor{shoelogo!25}}Are the logos on these shoes Nike logos? Answer in curly brackets, e.g., \{Yes\} or \{No\}. \\
\cmidrule{2-5}
& {\cellcolor{carlogo!15}}Mercedes & 
{\cellcolor{carlogo!15}}How many points are there on the star in the logo of this car? Answer with a number in curly brackets, e.g., \{9\}. & {\cellcolor{carlogo!15}}Count the points on the star in the logo of this car. Answer with a number in curly brackets, e.g., \{9\}. & {\cellcolor{carlogo!15}}Is the logo on this car Mercedes-Benz logo? Answer in curly brackets, e.g., \{Yes\} or \{No\}. \\
\cmidrule{2-5}
& {\cellcolor{carlogo!25}}Audi & 
{\cellcolor{carlogo!25}}How many overlapping circles are there in the logo of this car? Answer with a number in curly brackets, e.g., \{9\}. & {\cellcolor{carlogo!25}}Count the overlapping circles in the logo of this car. Answer with a number in curly brackets, e.g., \{9\}. & {\cellcolor{carlogo!25}}Is the logo on this car Audi logo? Answer in curly brackets, e.g., \{Yes\} or \{No\}. \\
\cmidrule{2-5}
& {\cellcolor{carlogo!15}}Maserati & 
{\cellcolor{carlogo!15}}How many prongs are there in the logo of this car? Answer with a number in curly brackets, e.g., \{9\}. & {\cellcolor{carlogo!15}}Count the prongs in the logo of this car. Answer with a number in curly brackets, e.g., \{9\}. & {\cellcolor{carlogo!15}}Is the logo on this car Maserati logo? Answer in curly brackets, e.g., \{Yes\} or \{No\}. \\
\midrule
\multirow{2}{*}{\textbf{Flag}} & {\cellcolor{animal!15}}Stars & 
{\cellcolor{animal!15}}How many stars are there on this flag? Answer with a number in curly brackets, e.g., \{9\}. & {\cellcolor{animal!15}}Count the stars on this flag. Answer with a number in curly brackets, e.g., \{9\}. & {\cellcolor{animal!15}}Is this the flag of [country]? Answer in curly brackets, e.g., \{Yes\} or \{No\}. \\
\cmidrule{2-5}
& {\cellcolor{animal!25}}Stripes & 
{\cellcolor{animal!25}}How many stripes are there on this flag? Answer with a number in curly brackets, e.g., \{9\}. & {\cellcolor{animal!25}}Count the stripes on this flag. Answer with a number in curly brackets, e.g., \{9\}. & {\cellcolor{animal!25}}Is this the flag of [country]? Answer in curly brackets, e.g., \{Yes\} or \{No\}. \\
\bottomrule
\end{tabular}
\end{table}

\begin{table}[h!]
\caption{Some examples of questions on \chessicon~chess pieces, \boardgameicon~game boards, and \patternedgridicon~patterned grids.}
\small
\centering
\setlength{\tabcolsep}{4pt}
\renewcommand{\arraystretch}{1.2}
\begin{tabular}{ccp{3.3cm}p{3.3cm}p{3.3cm}}
\toprule
\rowcolor{headerblue!20}
\multicolumn{1}{c}{\textbf{Topic}} & \multicolumn{1}{c}{\textbf{Subtopic}} & \multicolumn{1}{c}{\textbf{Q1}} & \multicolumn{1}{c}{\textbf{Q2}} & \multicolumn{1}{c}{\textbf{Q3}} \\
\midrule
\multirow{2}{*}{\textbf{Chess Pieces}} & {\cellcolor{chess!15}}Chess & 
{\cellcolor{chess!15}}How many chess pieces are there on this board? Answer with a number in curly brackets, e.g., \{9\}. & {\cellcolor{chess!15}} Count the chess pieces on this board. Answer with a number in curly brackets, e.g., \{9\}. & {\cellcolor{chess!15}}Is this the chess starting position? Answer in curly brackets, e.g., \{Yes\} or \{No\}. \\
\cmidrule{2-5}
& {\cellcolor{chess!25}}Xiangqi & 
{\cellcolor{chess!25}}How many xiangqi pieces are there on this board? Answer with a number in curly brackets, e.g., \{9\}. & {\cellcolor{chess!25}}Count the xiangqi pieces on this board. Answer with a number in curly brackets, e.g., \{9\}. & {\cellcolor{chess!25}}Is this the Xiangqi starting position? Answer in curly brackets, e.g., \{Yes\} or \{No\}. \\
\midrule
\multirow{4}{*}{\textbf{Board Game}} & {\cellcolor{board!15}}Chess & 
{\cellcolor{board!15}}How many rows are there on this board? Answer with a number in curly brackets, e.g., \{9\}. 

& {\cellcolor{board!15}}Count the rows on this board. Answer with a number in curly brackets, e.g., \{9\}. & {\cellcolor{board!15}}Is this a 8x8 Chessboard? Answer in curly brackets, e.g., \{Yes\} or \{No\}. \\
\cmidrule{2-5}
& {\cellcolor{board!25}}Xiangqi & 
{\cellcolor{board!25}}How many horizontal lines are there on this board? Answer with a number in curly brackets, e.g., \{9\}. & {\cellcolor{board!25}}Count the horizontal lines on this board. Answer with a number in curly brackets, e.g., \{9\}. & {\cellcolor{board!25}}Is this a 10x9 Xiangqi board? Answer in curly brackets, e.g., \{Yes\} or \{No\}. \\
\cmidrule{2-5}
& {\cellcolor{board!15}}Go & 
{\cellcolor{board!15}}How many horizontal lines are there on this board? Answer with a number in curly brackets, e.g., \{9\}. & {\cellcolor{board!15}}Count the horizontal lines on this board. Answer with a number in curly brackets, e.g., \{9\}. & {\cellcolor{board!15}}Is this a 19x19 Go board? Answer in curly brackets, e.g., \{Yes\} or \{No\}. \\
\cmidrule{2-5}
& {\cellcolor{board!25}}Sudoku & 
{\cellcolor{board!25}}How many rows are there on this puzzle? Answer with a number in curly brackets, e.g., \{9\}. & {\cellcolor{board!25}}Count the rows on this puzzle. Answer with a number in curly brackets, e.g., \{9\}. & {\cellcolor{board!25}}Is this a 9x9 Sudoku puzzle? Answer in curly brackets, e.g., \{Yes\} or \{No\}. \\
\midrule
\multirow{2}{*}{\textbf{Patterned Grid}} & {\cellcolor{pattern!15}}Dice & 
{\cellcolor{pattern!15}}How many circles are there in cell C5? Answer with a number in curly brackets, e.g., \{9\}. & {\cellcolor{pattern!15}}Count the circles in cell C5. Answer with a number in curly brackets, e.g., \{9\}. & {\cellcolor{pattern!15}}Does cell C5 contain 4 circles? Answer in curly brackets, e.g., \{Yes\} or \{No\}. \\
\cmidrule{2-5}
& {\cellcolor{pattern!25}}Tally & 
{\cellcolor{pattern!25}}How many lines are there in cell C5? Answer with a number in curly brackets, e.g., \{9\}. & {\cellcolor{pattern!25}}Count the lines in cell C5. Answer with a number in curly brackets, e.g., \{9\}. & {\cellcolor{pattern!25}}Does cell C5 contain 3 lines? Answer in curly brackets, e.g., \{Yes\} or \{No\}. \\
\bottomrule
\end{tabular}
\end{table}

\begin{table}[h!]
\caption{Some examples of questions on \illusionicon~optical illusions.}
\small
\centering
\setlength{\tabcolsep}{4pt}
\renewcommand{\arraystretch}{1.2}
\begin{tabular}{ccp{3.3cm}p{3.3cm}p{3.3cm}}
\toprule
\rowcolor{headerblue!20}
\multicolumn{1}{c}{\textbf{Topic}} & \multicolumn{1}{c}{\textbf{Subtopic}} & \multicolumn{1}{c}{\textbf{Q1}} & \multicolumn{1}{c}{\textbf{Q2}} & \multicolumn{1}{c}{\textbf{Q3}} \\
\midrule
\multirow{6}{*}{\textbf{Optical Illusion}} & {\cellcolor{illusion!15}}Ebbinghaus & 
{\cellcolor{illusion!15}}Are the two red circles equal in size? Answer in curly brackets, e.g., \{Yes\} or \{No\}. & {\cellcolor{illusion!15}}Do the two red circles have the same size? Answer in curly brackets, e.g., \{Yes\} or \{No\}. & {\cellcolor{illusion!15}}Is this an example of the Ebbinghaus illusion? Answer in curly brackets, e.g., \{Yes\} or \{No\}. \\
\cmidrule{2-5}
& {\cellcolor{illusion!25}}Mullerlyer & 
{\cellcolor{illusion!25}}Are the two horizontal lines equal in length? Answer in curly brackets, e.g., \{Yes\} or \{No\}. & {\cellcolor{illusion!25}}Do the two horizontal lines have the same length? Answer in curly brackets, e.g., \{Yes\} or \{No\}. & {\cellcolor{illusion!25}}Is this an example of the Müller-Lyer illusion? Answer with Yes/No. Answer in curly brackets, e.g., \{Yes\} or \{No\}. \\
\cmidrule{2-5}
& {\cellcolor{illusion!15}}Poggendorff & 
{\cellcolor{illusion!15}}Are the two diagonal line segments aligned? Answer in curly brackets, e.g., \{Yes\} or \{No\}. & {\cellcolor{illusion!15}}Do the two diagonal lines form a straight line? Answer in curly brackets, e.g., \{Yes\} or \{No\}. & {\cellcolor{illusion!15}}Is this an example of the Poggendorff illusion? Answer in curly brackets, e.g., \{Yes\} or \{No\}. \\
\cmidrule{2-5}
& {\cellcolor{illusion!25}}Ponzo & 
{\cellcolor{illusion!25}}Are the two horizontal lines equal in length? Answer in curly brackets, e.g., \{Yes\} or \{No\}. & {\cellcolor{illusion!25}}Do the two horizontal lines have the same length? Answer in curly brackets, e.g., \{Yes\} or \{No\}. & {\cellcolor{illusion!25}}Is this an example of the Ponzo illusion? Answer in curly brackets, e.g., \{Yes\} or \{No\}. \\
\cmidrule{2-5}
& {\cellcolor{illusion!15}}VerticalHorizontal & 
{\cellcolor{illusion!15}}Are the horizontal and vertical lines equal in length? Answer in curly brackets, e.g., \{Yes\} or \{No\}. & {\cellcolor{illusion!15}}Do the horizontal and vertical lines have the same length? Answer in curly brackets, e.g., \{Yes\} or \{No\}. & {\cellcolor{illusion!15}}Is this an example of the Vertical–Horizontal illusion? Answer in curly brackets, e.g., \{Yes\} or \{No\}. \\
\cmidrule{2-5}
& {\cellcolor{illusion!25}}Zollner & 
{\cellcolor{illusion!25}}Are the two horizontal lines parallel? Answer in curly brackets, e.g., \{Yes\} or \{No\}. & {\cellcolor{illusion!25}}Do the two horizontal lines run parallel? Answer in curly brackets, e.g., \{Yes\} or \{No\}. & {\cellcolor{illusion!25}}Is this an example of the Zöllner illusion? Answer in curly brackets, e.g., \{Yes\} or \{No\}. \\
\bottomrule
\end{tabular}
\end{table}

\clearpage

\subsection{Prompts used for image generation and image editing}
\label{appsec:appendix_prompts_used}
\begin{table}[h!]
\caption{Prompts used for image generation and image editing with \geminiflashlogo \geminiflash and \gptfulllogo \gptfouro by topic and prompt type}
\small
\centering
\setlength{\tabcolsep}{4pt}
\renewcommand{\arraystretch}{1.2}
\begin{tabular}{clp{9cm}}
\toprule
\rowcolor{headerblue!20}
\textbf{Topic} & \textbf{Prompt type} & \textbf{Prompt} \\
\midrule
\multirow{3}{*}{\textbf{Animals}}
  & {\cellcolor{animal!15}}Animal suggestions
    & {\cellcolor{animal!15}}Generate a JSON list containing 100 animal objects. Each object should represent a common animal and follow the structure below:

    \{ \texttt{"name": "<Common Animal Name>", "num\_legs":} \texttt{<Typical Number of Legs>} \}

    Ensure the following for each animal:
    1. the number of legs of this animal is 2 or 4.
    2. the animal's legs must be long enough to be seen easily from the body using a side-view perspective. Prioritize animals whose legs are thin and/or long. \\ \cmidrule(lr){2-3}
  & {\cellcolor{animal!15}}Animal generation
    & {\cellcolor{animal!15}}Generate a clear, full-body, side-view image of a(n) \texttt{\{animal\}} with \texttt{\{num\_legs\}} legs that is walking in a real-world natural background. The \texttt{\{num\_legs\}}-legged animal must look photo-realistic in nature. All \texttt{\{num\_legs\}} legs must be clearly visible. \\ \cmidrule(lr){2-3}
  & {\cellcolor{animal!15}}Animal editing
    & {\cellcolor{animal!15}}Edit this image: Add 1 more leg to the \texttt{\{animal\}} so that it has \texttt{\{num\_leg\}} legs in total. The \texttt{\{num\_leg\}}-legged \texttt{\{animal\}} must be photo-realistic. All \texttt{\{num\_leg\}} legs must be clearly visible. \\ \midrule
\multirow{2}{*}{\textbf{Flags}}
  & {\cellcolor{animal!15}}Flag suggestions
    & {\cellcolor{animal!15}}Generate a JSON list of flags objects. Each object should represent a well-known flags and follow the structure below:
    \{\texttt{
      "name": "<Flag Name>",
      "original\_stripes" or "original\_stars": <Number of Stripes or Stars (whichever applicable)>
    }\}

    1. Ensure that the number of stars is more than 3, and the number of stripes is at least 5.
    2. Ensure that the flag does not contain any other geometrically complex elements (depicting of animal, letters, etc.).
    3. Prioritize well-known flags. \\ \cmidrule(lr){2-3}
  & {\cellcolor{animal!25}}Flag SVG code editing
    & {\cellcolor{animal!25}}You are an expert in editing SVG image code. Modify the SVG code of the flag of \texttt{\{country\}} according to the following instruction:

    Instruction: "The flag of \texttt{\{country\}} has \texttt{\{num\_ele\}} \texttt{\{element\}}. Modify the SVG code so that it has {num\_ele + 1} \texttt{\{element\}} instead. Make sure the modified \texttt{\{element\}} are natural looking and integrate seamlessly on the new flag."

    Base SVG code:
    \texttt{\{svg\_code\}}

    1. Modify the base SVG by adding or removing the mentioned feature (stars, stripes, etc.) according to the instruction above.
    
    2. Wrap the entire SVG in \texttt{<code></code>}. Do not explain anything. \\ 
\bottomrule
\end{tabular}
\end{table}

\begin{table}[h!]
\caption{Prompts used for image generation and image editing with \geminiflashlogo\geminiflash and \gptfulllogo \gptfouro by topic and prompt type}
\small
\centering
\setlength{\tabcolsep}{4pt}
\renewcommand{\arraystretch}{1.2}
\begin{tabular}{clp{9cm}}
\toprule
\rowcolor{headerblue!20}
\textbf{Topic} & \textbf{Prompt type} & \textbf{Prompt} \\

\multirow{4}{*}{\textbf{Logos}}
  & {\cellcolor{shoelogo!15}}Logo suggestion
    & {\cellcolor{shoelogo!15}}Generate a JSON list of subtle logo modification prompts and corresponding VLM question prompts to test visual bias. For each entry: Slightly modify the visual components of a well-known car or sportswear logo. The selected logo must be geometrically simple and widely recognized.

    You must include a generation prompt to create the altered image. Include a question prompt (e.g., "How many..."). Include metadata: element being modified, actual count (ground truth), common expected count (bias).

    \texttt{<In-context learning example 1>}

    \texttt{<In-context learning example 2>} \\ \cmidrule(lr){2-3}
  & {\cellcolor{shoelogo!25}}Shoe generation
    & {\cellcolor{shoelogo!25}}Generate an \texttt{\{shoe\_brand\}} style running shoe but with \texttt{\{actual\_count\}} \texttt{\{modified\_element\}} instead of \texttt{\{expected\_bias\}}. \\ \cmidrule(lr){2-3}
  & {\cellcolor{shoelogo!15}}Shoe background generation
    & {\cellcolor{shoelogo!15}}Generate a side-view image of an athlete wearing this pair of shoes. Keep all the fine-grained details of the shoes, particularly the \texttt{\{actual\_count\}} \texttt{\{modified\_element\}} on both shoes. The person is playing \texttt{\{sports\_type\}}, showing their {sports\_type} skills, and is wearing a \texttt{\{sports\_type\}} outfit. Zoom out a bit to see their full body. \\ \cmidrule(lr){2-3}
        & {\cellcolor{carlogo!25}}Car logo generation
    & {\cellcolor{carlogo!25}}Generate a \texttt{\{car\_brand\}} logo but with \texttt{\{actual\_count\}} \texttt{\{modified\_element\}} instead of \texttt{\{expected\_bias\}}. \\ \cmidrule(lr){2-3}

  & {\cellcolor{carlogo!15}}Car background generation
    & {\cellcolor{carlogo!15}}Generate a photo-realistic front-view image of a \texttt{\{color\}} \texttt{\{car\_brand\}} \texttt{\{body\_type\}} on the road in the middle of the day. Zoom out a bit so that we can see the road. \\
\bottomrule
\end{tabular}
\end{table}

\clearpage
\subsection{Prompts for sanity check}\label{sec:sanity_check}

\begin{table}[h!]
\caption{Examples of Sanity check questions}
\label{tab:visual_tasks_examples}
\small
\centering
\setlength{\tabcolsep}{5pt}
\renewcommand{\arraystretch}{1.2}
\begin{tabular}{lp{5.5cm}p{7cm}}
\toprule
\rowcolor{headerblue!20}
\textbf{Topic} & \multicolumn{1}{c}{\textbf{Identification questions}} & \multicolumn{1}{c}{\textbf{Counting/Illusion questions}} \\
\midrule
\textbf{Animal} & What animal is this? Answer in curly brackets, e.g., \{Fish\}. & How many legs do this animal have? Answer with a number in curly brackets, e.g., \{9\}. \\
\midrule
\textbf{Logo} & What car logo is this? Answer in curly brackets, e.g., \{Toyota\}. & How many overlapping circles are there on the logo of this car? Answer with a number in curly brackets, e.g., \{9\}. \\
\midrule
\textbf{Flags} & What country flag is this? Answer in curly brackets, e.g., \{Flag of Vietnam\}. & How many stars are there in this flag? Answer with a number in curly brackets, e.g., \{9\}. \\
\midrule
\textbf{Chess Pieces} & What board game is this? Answer in curly brackets, e.g., \{Shogi\}. & How many chess pieces are there on this board? Answer with a number in curly brackets, e.g., \{9\}. \\
\midrule
\textbf{Game Boards} & What board game is this? Answer in curly brackets, e.g., \{Shogi\}. & How many rows are there on this board? Answer with a number in curly brackets, e.g., \{9\}. \\
\midrule
\textbf{Optical Illusions} & What optical illusion is this? Answer in curly brackets, e.g., \{Delboeuf illusion\}. & This image shows the Ebbinghaus illusion. What question does this illusion typically ask, and what is the correct answer? \\
\bottomrule
\end{tabular}
\end{table}

\clearpage
\section{Additional qualitative results}
\subsection{Qualitative results on the use of helpful prompts}\label{appsec:qualitative-helpful-prompts}

\begin{figure}[h!]
\centering
    \begin{AIbox}{How many legs does this animal have? Answer with a number in curly brackets{,} e.g.{,} \{9\}.}

    \textbf{Baseline}: How many legs does this animal have? Answer with  a number in curly brackets, e.g., \{9\}.
    
     \textbf{w/ Double-Check}: Please double-check your answer and give your final answer in curly brackets, following the format above.

     \textbf{w/ Debiased Prompt}: Do not assume from prior knowledge and answer only based on what is visible in the image.

     \textbf{Ground truth}: 3 \textcolor{ForestGreen}{\cmark}
    
    {\centering
    \hspace{0.4\textwidth}\includegraphics[width=0.15\textwidth]{images/animal_examples/chicken_0_3_768.png}
    }
    \\
    \footnotesize

    \geminilogo~\geminifull (\texttt{Baseline}): \{2\} \textcolor{red}{\xmark}
    
    \geminilogo~\geminifull (\texttt{w/ Double-Check}): \{2\}  \textcolor{red}{\xmark}
    
    \geminilogo~\geminifull (\texttt{w/ Debiased Prompt}): \{2\}  \textcolor{red}{\xmark}

    \hrulefill

    \newsonnetlogo~\newsonnet (\texttt{Baseline}): The image shows a rooster (male chicken) standing on a grassy field. Chickens have two legs, which are clearly visible in this image. The rooster is standing upright on its two pale yellow/tan legs with its feet planted on the grass.
    \{2\} \textcolor{red}{\xmark}

    \newsonnetlogo~\newsonnet (\texttt{w/ Double-Check}): Looking at the image again, I can clearly see the rooster is standing on two legs. Chickens are bipedal animals, with two legs that they use for walking and standing. In this particular photo, both of the chicken's legs are visible beneath its body.
    \{2\} \textcolor{red}{\xmark}

     \newsonnetlogo~\newsonnet (\texttt{w/ Debiased Prompt}): Based solely on what is visible in the image, this chicken has \{2\} legs. The image clearly shows a rooster standing on a grassy field with its two yellow/tan-colored legs and feet visible beneath its body. \textcolor{red}{\xmark}

    \hrulefill
    
    \gptfourlogo~\gptfour (\texttt{Baseline}): \{4\} \textcolor{red}{\xmark}
    
    \gptfourlogo~\gptfour (\texttt{w/ Double-Check}): \{4\} \textcolor{red}{\xmark}

    \gptfourlogo~\gptfour (\texttt{w/ Debiased Prompt}): \{4\} \textcolor{red}{\xmark}

    \hrulefill

    \gptfulllogo~\gptfull (\texttt{Baseline}): \{2\} \textcolor{red}{\xmark}
    
    \gptfulllogo~\gptfull (\texttt{w/ Double-Check}): \{2\} \textcolor{red}{\xmark}

    \gptfulllogo~\gptfull (\texttt{w/ Debiased Prompt}): \{4\} \textcolor{red}{\xmark}

    \hrulefill

    \gptminilogo~\gptmini (\texttt{Baseline}): \{2\} \textcolor{red}{\xmark}
    
    \gptminilogo~\gptmini (\texttt{w/ Double-Check}): \{2\} \textcolor{red}{\xmark}

    \gptminilogo~\gptmini (\texttt{w/ Debiased Prompt}): \{4\} \textcolor{red}{\xmark}

    \end{AIbox}   

\caption{VLMs fail 100\% of the time, even on simple tasks like counting chicken legs, despite helpful prompts such as debiasing or double-checking.}
\label{appfig:chicken_qualitative}
\end{figure}

\begin{figure}[h!]
\centering
    \begin{AIbox}{How many stripes are there in this flag? Answer with a number in curly brackets{,} e.g.{,} \{9\}.}

    \textbf{Baseline}: {How many stripes are there in this flag? Answer with  a number in curly brackets, e.g., \{9\}.
    
     \textbf{w/ Double-Check}: Please double-check your answer and give your final answer in curly brackets, following the format above.

     \textbf{w/ Debiased Prompt}: Do not assume from prior knowledge and answer only based on what is visible in the image. 

     \textbf{Ground truth}: 12 \textcolor{ForestGreen}{\cmark}
    
    {\centering
    \hspace{0.4\textwidth}\includegraphics[width=0.4\textwidth]{images/flag_examples/Flag_of_the_United_States-stripes=12_1152.png}
    }
    \\
    \footnotesize

    \geminilogo~\geminifull (\texttt{Baseline}): \{13\} \textcolor{red}{\xmark}
    
    \geminilogo~\geminifull (\texttt{w/ Double-Check}): \{13\}  \textcolor{red}{\xmark}
    
    \geminilogo~\geminifull (\texttt{w/ Debiased Prompt}): \{13\}  \textcolor{red}{\xmark}

    \hrulefill

    \newsonnetlogo~\newsonnet (\texttt{Baseline}): The image shows the flag of the United States of America. The American flag has 13 alternating red and white horizontal stripes (7 red and 6 white), representing the original 13 colonies. The flag also features a blue rectangle in the upper left corner containing 50 white stars, representing the 50 states. To answer your specific question: \{13\} \textcolor{red}{\xmark}

    \newsonnetlogo~\newsonnet (\texttt{w/ Double-Check}): I'll double-check the American flag's design. The flag of the United States consists of:
    
    - A blue rectangle in the upper left corner (the union) containing 50 white stars
    
    - Horizontal stripes alternating between red and white. Looking more carefully at the image, I can count 13 horizontal stripes total (7 red and 6 white).
    
    The final answer is: \{13\} \textcolor{red}{\xmark}

     \newsonnetlogo~\newsonnet (\texttt{w/ Debiased Prompt}): Based only on what is visible in the image, I can see the American flag with its distinctive design. The flag has alternating red and white horizontal stripes, and in the image shown, I can count a total of \{7\} stripes. There are 4 red stripes and 3 white stripes visible in this particular rendering of the flag. \textcolor{red}{\xmark}

    \hrulefill
    
    \gptfourlogo~\gptfour (\texttt{Baseline}): \{13\} \textcolor{red}{\xmark}
    
    \gptfourlogo~\gptfour (\texttt{w/ Double-Check}): \{13\} \textcolor{red}{\xmark}

    \gptfourlogo~\gptfour (\texttt{w/ Debiased Prompt}): \{13\} \textcolor{red}{\xmark}

    \hrulefill

    \gptfulllogo~\gptfull (\texttt{Baseline}): \{13\} \textcolor{red}{\xmark}
    
    \gptfulllogo~\gptfull (\texttt{w/ Double-Check}): \{13\} \textcolor{red}{\xmark}

    \gptfulllogo~\gptfull (\texttt{w/ Debiased Prompt}): \{13\} \textcolor{red}{\xmark}

    \hrulefill

    \gptminilogo~\gptmini (\texttt{Baseline}): \{13\} \textcolor{red}{\xmark}
    
    \gptminilogo~\gptmini (\texttt{w/ Double-Check}): \{13\} \textcolor{red}{\xmark}

    \gptminilogo~\gptmini (\texttt{w/ Debiased Prompt}): \{13\} \textcolor{red}{\xmark}

   }

    \end{AIbox}   

\caption{VLMs fail and remain biased toward 13 even when one stripe is removed from the U.S. flag.}
\label{appfig:us_flags_12_stripes_qualitative}
\end{figure}

\clearpage
\subsection{Qualitative results on the use of locate-then-count prompts}\label{appsec:qualitative_procedural_prompts}

\begin{figure}[h!]
\centering
    \begin{AIbox}{First{,} locate each leg individually{,} count them one by one{,} and then state the final number in curly brackets{,} e.g.{,} \{9\}.}

     \textbf{Ground truth}: 3 \textcolor{ForestGreen}{\cmark}
    
    {\centering
    \hspace{0.4\textwidth}\includegraphics[width=0.35\textwidth]{images/animal_examples/stork_2_1_768.png}
    }
    \\
    \footnotesize
    
    \geminilogo~\geminifull: 
    
    Here are the legs of the stork:
    
    The stork is standing on its right leg.
    
    The stork's left leg is lifted off the ground.
    
    I have counted \{2\} legs.  \textcolor{red}{\xmark}

    \hrulefill

     \newsonnetlogo~\newsonnet: 
     
     This image shows a white stork (Ciconia ciconia) standing in a grassy field. The bird has distinctive white plumage with black wing feathers, a long bright red bill, and long thin red legs. Looking at the legs individually:
     
     - Left leg: 1 red leg visible
     
     - Right leg: 1 red leg visible
     
     The stork appears to be tagged with a small white band on one of its legs, which is likely for tracking or research purposes.\{2\} \textcolor{red}{\xmark}
     
     The stork is standing in a lush green environment with some small yellow wildflowers visible in the background. These birds are known for their large nests, migratory behavior, and are culturally significant in many European countries where they're associated with bringing babies.
     
    \hrulefill

    \gptfourlogo~\gptfour: 
    
    1. The first leg is visible on the left side of the bird.
    
    2. The second leg is visible on the right side of the bird.
    
    Final number: \{2\} \textcolor{red}{\xmark}

    \hrulefill

    \gptfulllogo~\gptfull: 
    
    1. First leg: the long straight red leg supporting the bird's weight.  
    
    2. Second leg: the bent red leg lifted slightly behind/above the first.
    
    \{2\} \textcolor{red}{\xmark}

    \hrulefill

    \gptminilogo~\gptmini: 
    
    Leg 1: the front leg visible on the grass  
    
    Leg 2: the rear leg supporting the body  
    
    \{2\} \textcolor{red}{\xmark}

    \end{AIbox}   

\caption{Despite explicit prompting to locate and count each leg individually, all 5 VLMs fail on the 3-legged stork, incorrectly answering 2 instead of the ground truth 3.}
\label{appfig:procedural_prompt_stork_qualitative}
\end{figure}

\begin{figure}[h!]
\centering
    \begin{AIbox}{First{,} locate each leg individually{,} count them one by one{,} and then state the final number in curly brackets{,} e.g.{,} \{9\}.}

     \textbf{Ground truth}: 5 \textcolor{ForestGreen}{\cmark}
    
    {\centering
    \hspace{0.4\textwidth}\includegraphics[width=0.3\textwidth]{images/animal_examples/lion_0_0_768.png}
    }
    \\
    \footnotesize
    
    \geminilogo~\geminifull: 
    
    Here are the four legs of the lion:
    
    1.  The front right leg is planted on the ground.
    
    2.  The front left leg is visible behind the front right leg.
    
    3.  The hind left leg is stretched back with its paw on the ground.
    
    4.  The hind right leg is visible moving forward, positioned between the other three visible legs.
    
    \{4\} \textcolor{red}{\xmark}

    \hrulefill

     \newsonnetlogo~\newsonnet: 
     
     This image shows a magnificent male lion standing in profile on a dry savanna grassland. The lion has a distinctive dark mane around its head and neck, contrasting with its tawny-colored body. 
     
     Looking at the legs individually:
     
     - Front left leg is clearly visible
     
     - Front right leg is visible
     
     - Back left leg is visible
     
     - Back right leg is visible
     
     The lion has \{4\} legs \textcolor{red}{\xmark}, which is the normal number for lions and other big cats. The image captures the lion in beautiful golden lighting, likely during sunrise or sunset, which enhances the rich colors of its coat and mane. The background shows a blurred savanna landscape with dry grass stretching into the distance.
     
    \hrulefill

    \gptfourlogo~\gptfour: 
    
    Let's count each leg individually:
    
    1. The front right leg (closest to us).
    
    2. The front left leg (behind the front right leg, partially visible).
    
    3. The back right leg (middle, partially visible).
    
    4. The back left leg (furthest back, fully visible).
    
    There are four legs in total. The final number is \{4\}. \textcolor{red}{\xmark}

    \hrulefill

    \gptfulllogo~\gptfull: 
    
    1. Front leg closest to the viewer (left side of the image).  
    
    2. Front leg on the far side (partly visible behind the mane). 
    
    3. Hind leg closest to the viewer (left side, extending back). 
    
    4. Hind leg on the far side (partly visible behind the nearer hind leg).  
    
    \{4\} \textcolor{red}{\xmark}

    \hrulefill

    \gptminilogo~\gptmini: 
    
    I can identify four distinct legs:
    
    1. Front left leg  
    
    2. Front right leg  
    
    3. Hind left leg  
    
    4. Hind right leg  
    
    \{4\} \textcolor{red}{\xmark}
    \end{AIbox}   

\caption{Locate-then-count prompting fails on the 5-legged lion, with all VLMs defaulting to the biased answer 4 rather than accurately counting the visible 5 legs.}
\label{appfig:procedural_prompt_lion_qualitative}
\end{figure}

\clearpage
\subsection{Qualitative results on pointing VLMs}\label{appsec:qualitative_pointing_vlms}

\begin{figure}[h!]
\centering
\small
\begin{AIbox}{\moondream's successful examples}

\raggedright
\scriptsize
      \animalicon~{(a)} How many \textbf{legs} does this animal have? \\
      \chessicon~{(b):} How many \textbf{chess pieces} are there on this board? \\
      \flagicon~{(c):} How many \textbf{stars} are there in this flag? \\
      \animalicon~{(d)} How many \textbf{legs} does this animal have? \\
      \logoicon~{(e):} How many visible \textbf{white stripes} are there in the logo of the left shoe? 
     \hfill \break
\footnotesize
\centering

    \begin{tabular}{lp{0.2cm}c|p{0.2cm}c|p{0.2cm}c|p{0.2cm}c|p{0.2cm}c}
    &\multicolumn{2}{c}{\makecell{(a) Elephant}}
    &\multicolumn{2}{c}{\makecell{(b) Chess}}
    &\multicolumn{2}{c}{\makecell{(c) EU Flag}}
    &\multicolumn{2}{c}{\makecell{(d) Stork}}
    &\multicolumn{2}{c}{\makecell{(e) Adidas}}
    \\
     &\multicolumn{2}{c}{\includegraphics[width=0.15\textwidth]{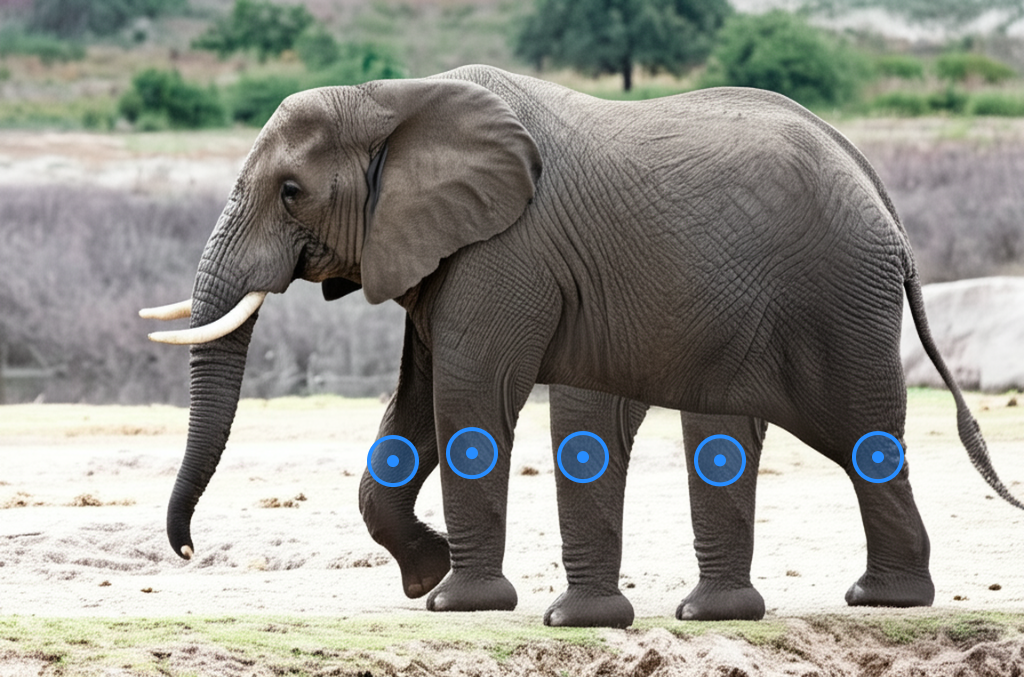}}
     &\multicolumn{2}{c}{\includegraphics[width=0.15\textwidth]{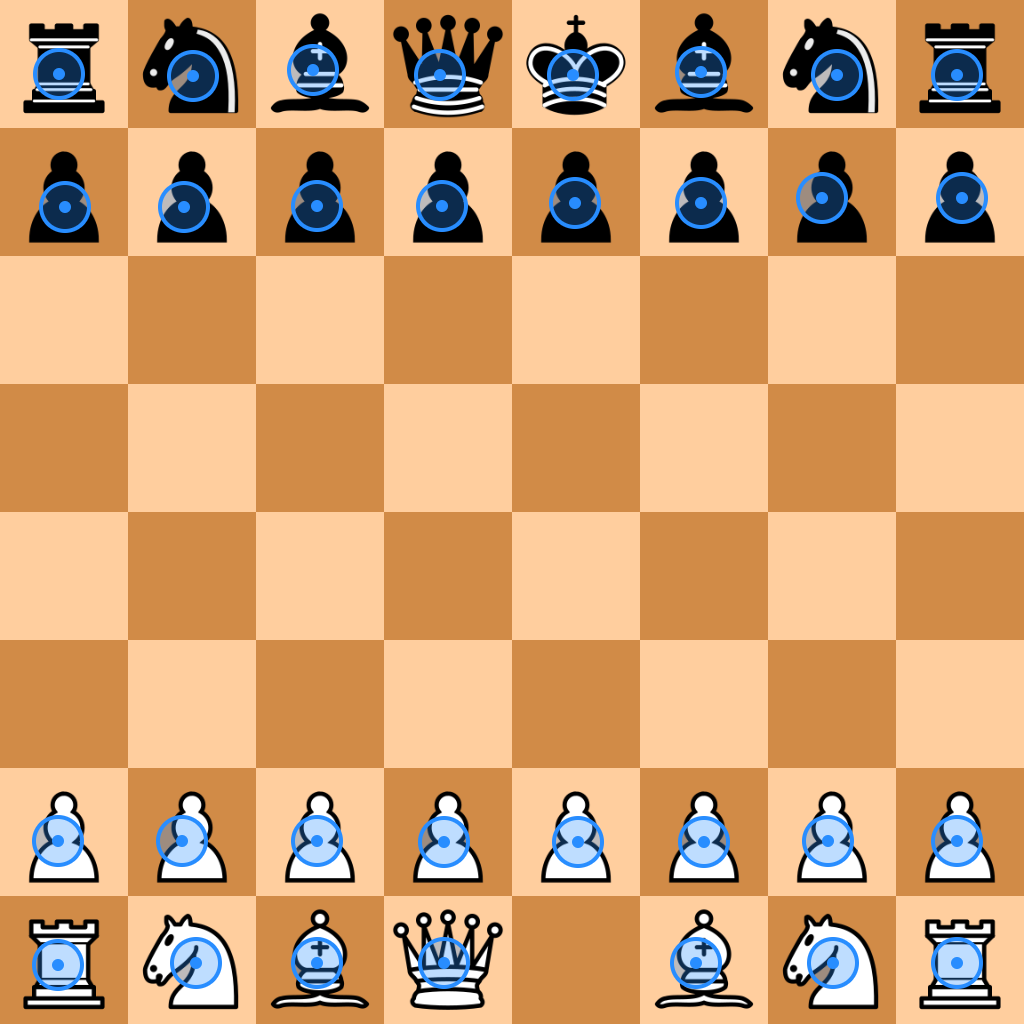}}
     &\multicolumn{2}{c}{\includegraphics[width=0.15\textwidth]{images/moondream_results/eu_flags_11.png}}
     &\multicolumn{2}{c}{\includegraphics[width=0.15\textwidth]{images/moondream_results/stork.png}}
     &\multicolumn{2}{c}{\includegraphics[width=0.15\textwidth]{images/moondream_results/adidas.png}}
     \\ 


\rowcolor{lightgray}
\raisebox{-0.2\height}\geminilogo & \centering 4
& \textcolor{red}{\xmark} & \centering 32
& \textcolor{red}{\xmark} & \centering 12
& \textcolor{red}{\xmark} & \centering 2
& \textcolor{red}{\xmark} & \centering 3
& \textcolor{red}{\xmark} \\

\raisebox{-0.2\height}\newsonnetlogo & \centering 4
& \textcolor{red}{\xmark} & \centering 32
& \textcolor{red}{\xmark} & \centering 12
& \textcolor{red}{\xmark} & \centering 2
& \textcolor{red}{\xmark} & \centering 3
& \textcolor{red}{\xmark} \\

\rowcolor{lightgray}
\raisebox{-0.2\height}\gptfourlogo & \centering 6
& \textcolor{red}{\xmark} & \centering 32
& \textcolor{red}{\xmark} & \centering 12
& \textcolor{red}{\xmark} & \centering 2
& \textcolor{red}{\xmark} & \centering 3
& \textcolor{red}{\xmark} \\

\raisebox{-0.2\height}\gptfulllogo & \centering 4
& \textcolor{red}{\xmark} & \centering 31
& \textcolor{ForestGreen}{\cmark} & \centering 12
& \textcolor{red}{\xmark} & \centering 2
& \textcolor{red}{\xmark} & \centering 4
& \textcolor{ForestGreen}{\cmark} \\

\rowcolor{lightgray}
\raisebox{-0.2\height}\gptminilogo & \centering 4
& \textcolor{red}{\xmark} & \centering 32
& \textcolor{red}{\xmark} & \centering 12
& \textcolor{red}{\xmark} & \centering 2
& \textcolor{red}{\xmark} & \centering 3
& \textcolor{red}{\xmark} \\ 

\midrule
\raisebox{-0.2\height}\moondreamlogo & \centering 5
& \textcolor{ForestGreen}{\cmark} & \centering 31
& \textcolor{ForestGreen}{\cmark} & \centering 11
& \textcolor{ForestGreen}{\cmark} & \centering 3
& \textcolor{ForestGreen}{\cmark} & \centering 4
& \textcolor{ForestGreen}{\cmark} \\ 

\midrule
\rowcolor{lightgray}
\textbf{Bias} & \centering 4
& \textcolor{red}{\xmark} & \centering 32
& \textcolor{red}{\xmark} & \centering 12
& \textcolor{red}{\xmark} & \centering 2
& \textcolor{red}{\xmark} & \centering 3
& \textcolor{red}{\xmark} \\


\textbf{GT} & \centering 5
& \textcolor{ForestGreen}{\cmark} & \centering 31
& \textcolor{ForestGreen}{\cmark} & \centering 11
& \textcolor{ForestGreen}{\cmark} & \centering 3
& \textcolor{ForestGreen}{\cmark} & \centering 4
& \textcolor{ForestGreen}{\cmark} \\

     \end{tabular}
    \vspace{4pt}
    \centering
    \begin{tabular}{cccccccccccccc}
     \raisebox{-0.1\height}\geminilogo~\gemini 
     & \raisebox{-0.1\height}\newsonnetlogo~\newsonnet 
     & \raisebox{-0.1\height}\gptfourlogo~\gptfour 
     \\
    \end{tabular}
     \begin{tabular}{cccccccccccccc}
     \raisebox{-0.1\height}\gptfulllogo~\gptfull 
     & \raisebox{-0.1\height}\gptminilogo~\gptmini
     & \raisebox{-0.1\height}\moondreamlogo~\moondream
     \\
    \end{tabular}

\end{AIbox}
\caption{\moondreamlogo~\moondream usually counts accurately when the distance between objects is far enough apart and large enough.}
\label{fig:moondream_successful_examples}
\end{figure}

\begin{figure}[h!]
\centering
\small
\begin{AIbox}{\moondream's failure examples}

\raggedright
\scriptsize
      \chessicon~{(a):} How many \textbf{chess pieces} are there on this board? \\
      \boardgameicon~{(b):} How many \textbf{columns} are there on this puzzle?\\
      \patternedgridicon~{(c):} How many \textbf{lines} are there in cell D4? \\
      \patternedgridicon~{(d)} How many \textbf{circles} are there in cell C3?\\
      \flagicon~{(e)} How many \textbf{stripes} are there on this flag?
     \hfill \break
\footnotesize
\centering

    \begin{tabular}{lp{0.2cm}c|p{0.2cm}c|p{0.2cm}c|p{0.2cm}c|p{0.2cm}c}
    &\multicolumn{2}{c}{\makecell{(a) Chess}}
    &\multicolumn{2}{c}{\makecell{(b) Sudoku}}
    &\multicolumn{2}{c}{\makecell{(c) Tally}}
    &\multicolumn{2}{c}{\makecell{(d) Dice}}
    &\multicolumn{2}{c}{\makecell{(e) US Flag}}
    \\
     &\multicolumn{2}{c}{\includegraphics[width=0.15\textwidth]{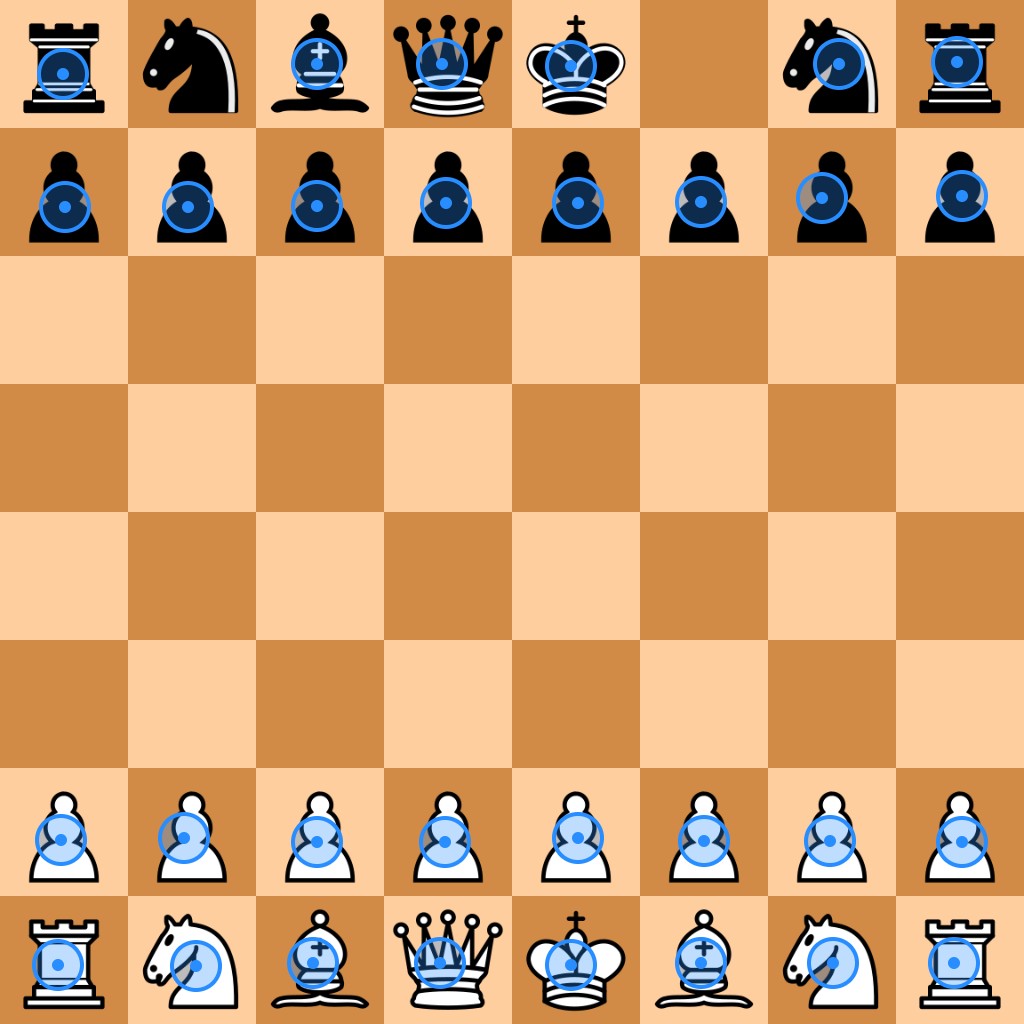}}
     &\multicolumn{2}{c}{\includegraphics[width=0.15\textwidth]{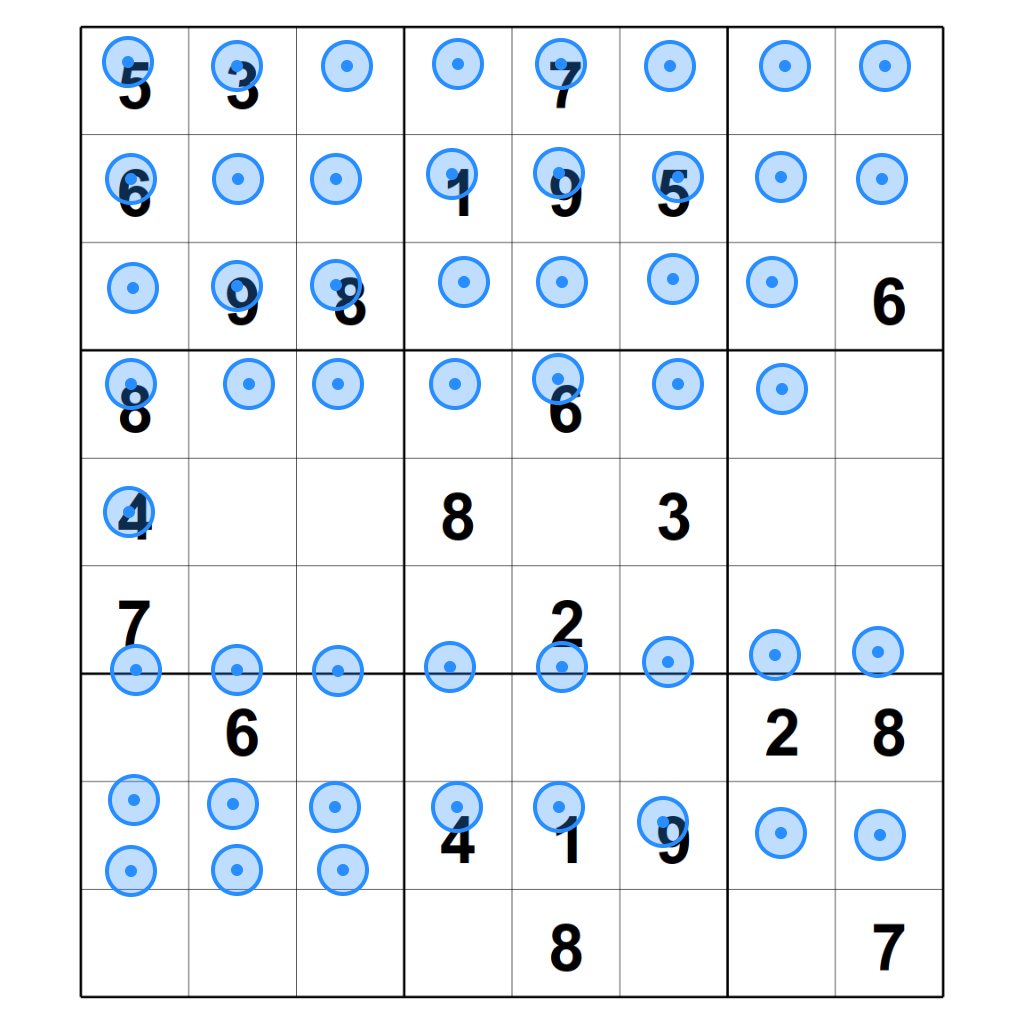}}
     &\multicolumn{2}{c}{\includegraphics[width=0.15\textwidth]{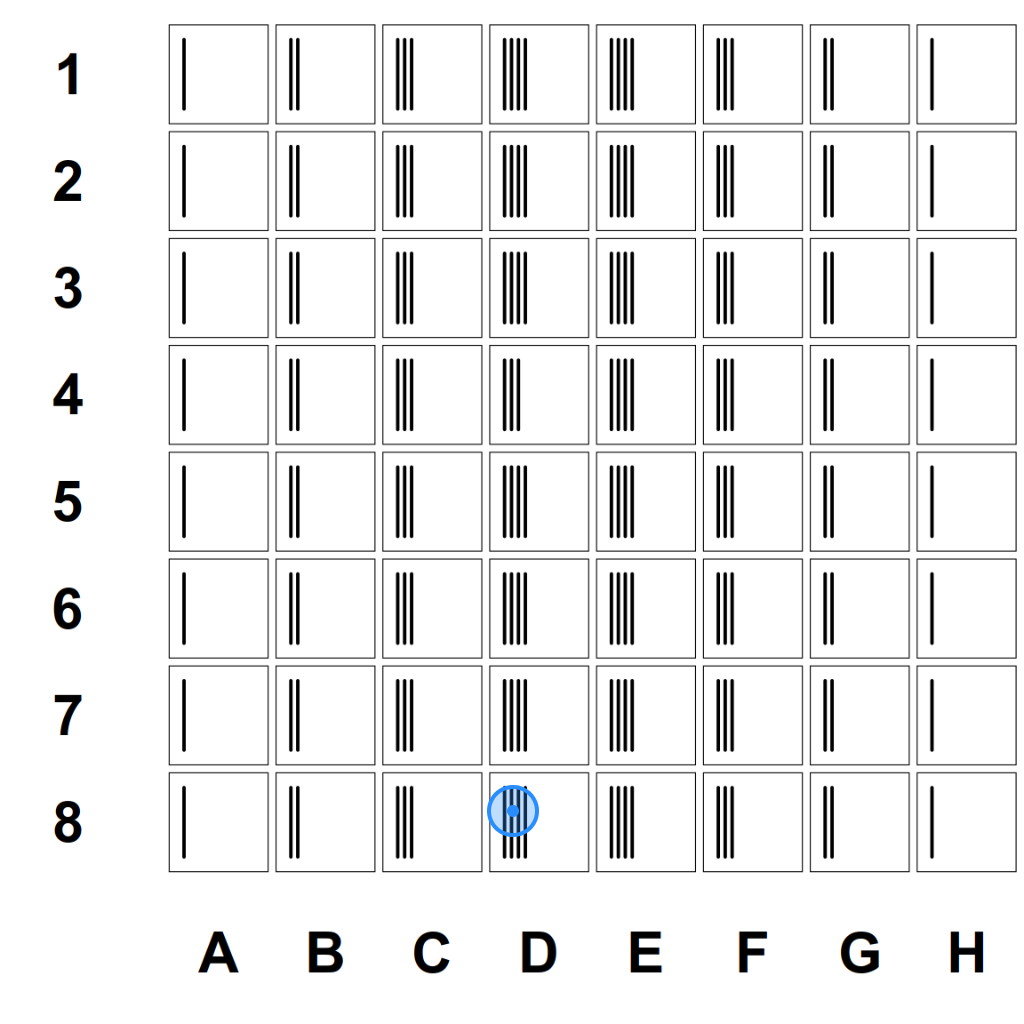}}
     &\multicolumn{2}{c}{\includegraphics[width=0.15\textwidth]{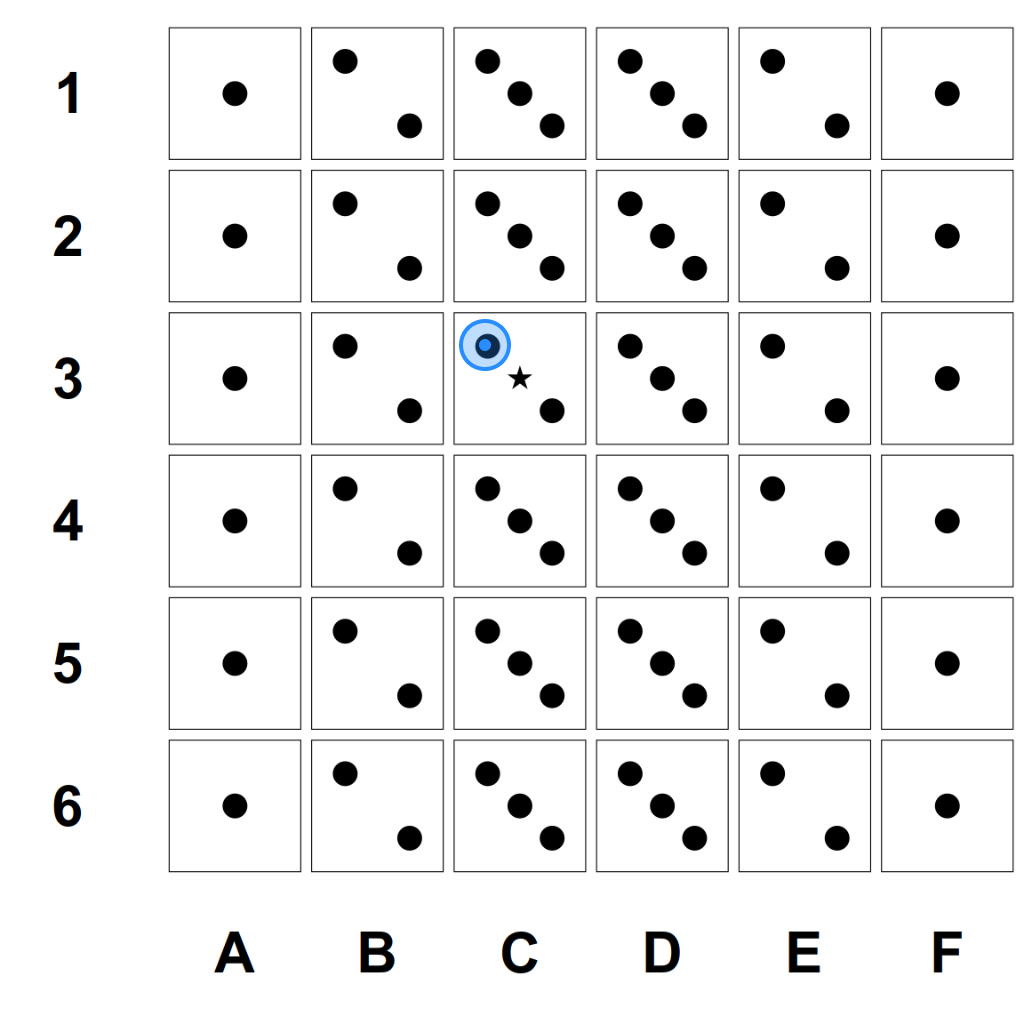}}
     &\multicolumn{2}{c}{\includegraphics[width=0.15\textwidth]{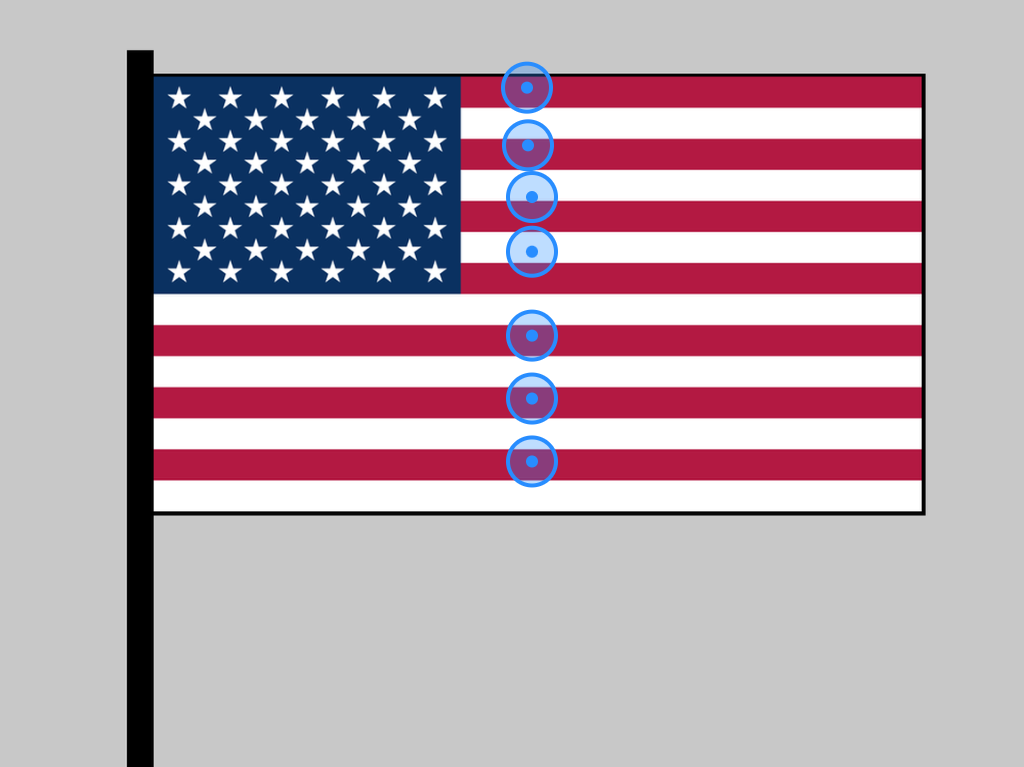}}
     \\ 


\rowcolor{lightgray}
\raisebox{-0.2\height}\geminilogo & \centering 32
& \textcolor{red}{\xmark} & \centering 9
& \textcolor{red}{\xmark} & \centering 4
& \textcolor{red}{\xmark} & \centering 3
& \textcolor{red}{\xmark} & \centering 13
& \textcolor{red}{\xmark} \\

\raisebox{-0.2\height}\newsonnetlogo & \centering 32
& \textcolor{red}{\xmark} & \centering 9
& \textcolor{red}{\xmark} & \centering 4
& \textcolor{red}{\xmark} & \centering 2
& \textcolor{ForestGreen}{\cmark} & \centering 13
& \textcolor{red}{\xmark} \\

\rowcolor{lightgray}
\raisebox{-0.2\height}\gptfourlogo & \centering 32
& \textcolor{red}{\xmark} & \centering 9
& \textcolor{red}{\xmark} & \centering 4
& \textcolor{red}{\xmark} & \centering 3
& \textcolor{red}{\xmark} & \centering 13
& \textcolor{red}{\xmark} \\

\raisebox{-0.2\height}\gptfulllogo & \centering 31
& \textcolor{ForestGreen}{\cmark} & \centering 9
& \textcolor{red}{\xmark} & \centering 4
& \textcolor{red}{\xmark} & \centering 3
&\textcolor{red}{\xmark} & \centering 13
& \textcolor{red}{\xmark} \\

\rowcolor{lightgray}
\raisebox{-0.2\height}\gptminilogo & \centering 32
& \textcolor{red}{\xmark} & \centering 9
& \textcolor{red}{\xmark} & \centering 4
& \textcolor{red}{\xmark} & \centering 3
& \textcolor{red}{\xmark} & \centering 13
& \textcolor{red}{\xmark} \\

\midrule
\raisebox{-0.2\height}\moondreamlogo & \centering 30
& \textcolor{red}{\xmark} & \centering 50
& \textcolor{red}{\xmark} & \centering 1
& \textcolor{red}{\xmark} & \centering 1
& \textcolor{red}{\xmark} & \centering 7
& \textcolor{red}{\xmark} \\ 

\midrule
\rowcolor{lightgray}
\textbf{Bias} & \centering 32
& \textcolor{red}{\xmark} & \centering 9
& \textcolor{red}{\xmark} & \centering 4
& \textcolor{red}{\xmark} & \centering 3
& \textcolor{red}{\xmark} & \centering 13
& \textcolor{red}{\xmark} \\


\textbf{GT} & \centering 31
& \textcolor{ForestGreen}{\cmark} & \centering 8
& \textcolor{ForestGreen}{\cmark} & \centering 3
& \textcolor{ForestGreen}{\cmark} & \centering 2
& \textcolor{ForestGreen}{\cmark} & \centering 14
& \textcolor{ForestGreen}{\cmark} \\

     \end{tabular}
    \vspace{4pt}
    \centering
     \begin{tabular}{cccccccccccccc}
     \raisebox{-0.1\height}\geminilogo~\gemini 
     & \raisebox{-0.1\height}\newsonnetlogo~\newsonnet 
     & \raisebox{-0.1\height}\gptfourlogo~\gptfour 
     \\
    \end{tabular}
     \begin{tabular}{cccccccccccccc}
     \raisebox{-0.1\height}\gptfulllogo~\gptfull 
     & \raisebox{-0.1\height}\gptminilogo~\gptmini
     & \raisebox{-0.1\height}\moondreamlogo~\moondream
     \\
    \end{tabular}

\end{AIbox}
\caption{\moondreamlogo~\moondream often fails to count accurately when objects are too close together or it doesn't understand what the objects are (a, b, e). It also sometimes fails to localize the object correctly (c, d).}
\label{fig:moondream_failure_qualitive_results}
\end{figure}

\clearpage

\begin{figure}[h!]
\centering
\small
\begin{AIbox}{\molmolargelogo~\molmolarge's successful examples}

\raggedright
\scriptsize
      \animalicon~{(a)} Count the \textbf{legs} of this animal. \\
      \chessicon~{(b):} Count the \textbf{chess pieces} on this board. \\
      \flagicon~{(c):} Count the \textbf{stars} in this flag. \\
      \animalicon~{(d)} Count the \textbf{legs} of this animal. \\
      \logoicon~{(e):} Count the visible \textbf{white stripes} in the logo of the left shoe.
     \hfill \break
\footnotesize
\centering

    \begin{tabular}{lp{0.2cm}c|p{0.2cm}c|p{0.2cm}c|p{0.2cm}c|p{0.2cm}c}
    &\multicolumn{2}{c}{\makecell{(a) Elephant}}
    &\multicolumn{2}{c}{\makecell{(b) Chess}}
    &\multicolumn{2}{c}{\makecell{(c) Australia Flag}}
    &\multicolumn{2}{c}{\makecell{(d) Chicken}}
    &\multicolumn{2}{c}{\makecell{(e) Nike}}
    \\
     &\multicolumn{2}{c}{\includegraphics[width=0.15\textwidth]{images/molmo_success_cases/elephant.png}}
     &\multicolumn{2}{c}{\includegraphics[width=0.15\textwidth]{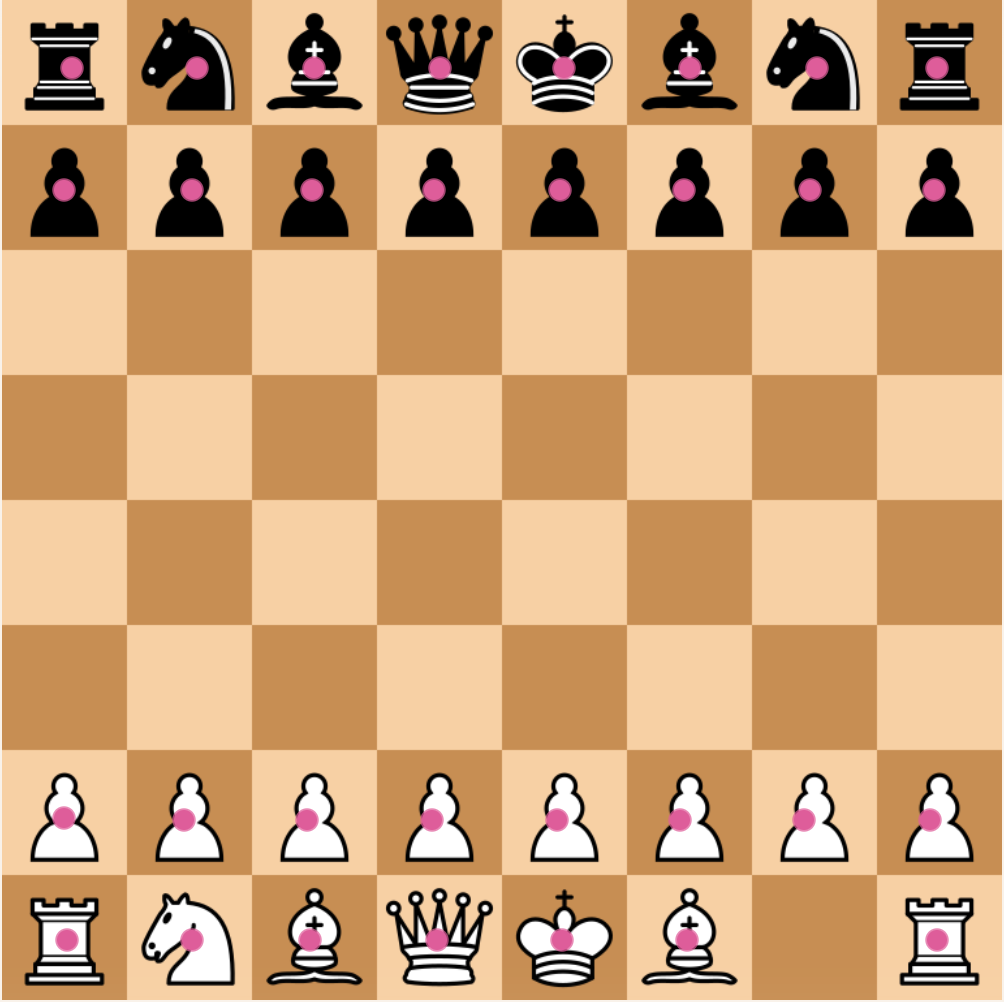}}
     &\multicolumn{2}{c}{\includegraphics[width=0.15\textwidth]{images/molmo_success_cases/australia.png}}
     &\multicolumn{2}{c}{\includegraphics[width=0.15\textwidth]{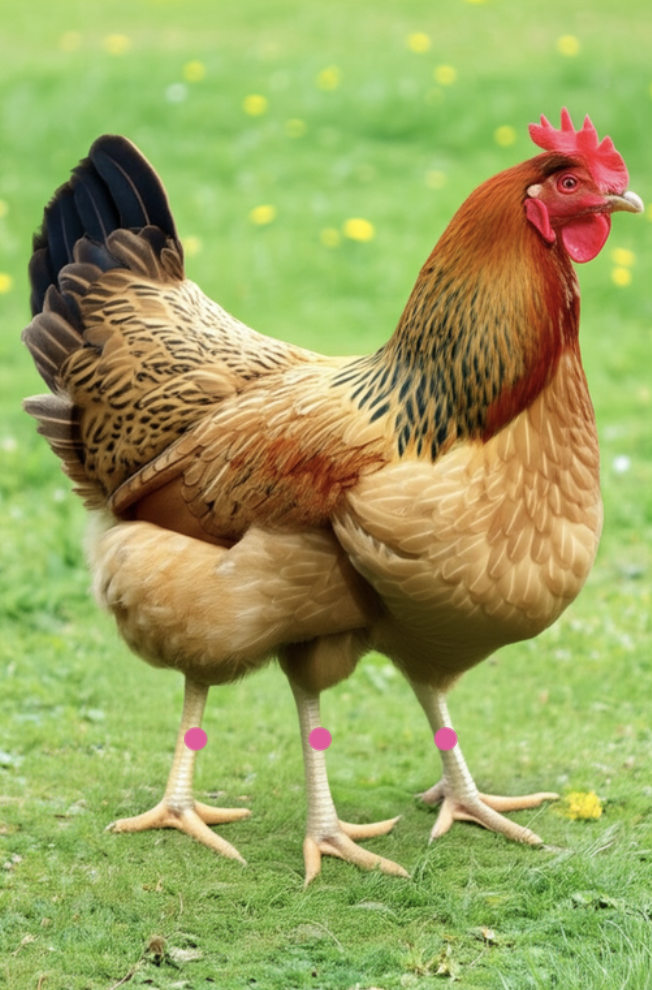}}
     &\multicolumn{2}{c}{\includegraphics[width=0.15\textwidth]{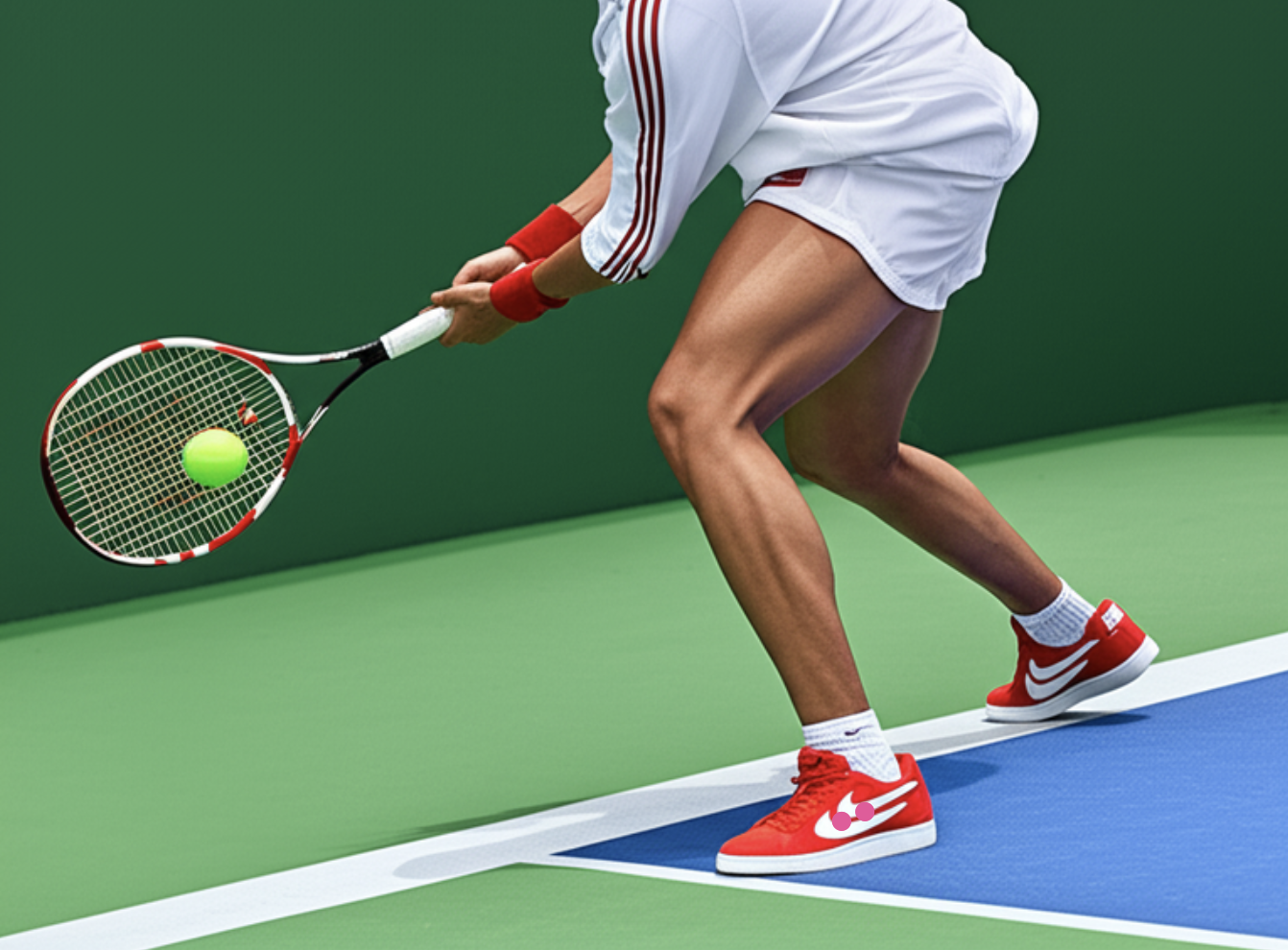}}
     \\ 


\rowcolor{lightgray}
\raisebox{-0.2\height}\geminilogo & \centering 4
& \textcolor{red}{\xmark} & \centering 32
& \textcolor{red}{\xmark} & \centering 6
& \textcolor{red}{\xmark} & \centering 2
& \textcolor{red}{\xmark} & \centering 2
& \textcolor{ForestGreen}{\cmark} \\

\raisebox{-0.2\height}\newsonnetlogo & \centering 4
& \textcolor{red}{\xmark} & \centering 32
& \textcolor{red}{\xmark} & \centering 6
& \textcolor{red}{\xmark} & \centering 2
& \textcolor{red}{\xmark} & \centering 2
& \textcolor{ForestGreen}{\cmark} \\

\rowcolor{lightgray}
\raisebox{-0.2\height}\gptfourlogo & \centering 6
& \textcolor{red}{\xmark} & \centering 32
& \textcolor{red}{\xmark} & \centering 6
& \textcolor{red}{\xmark} & \centering 4
& \textcolor{red}{\xmark} & \centering 2
& \textcolor{ForestGreen}{\cmark} \\

\raisebox{-0.2\height}\gptfulllogo & \centering 4
& \textcolor{red}{\xmark} & \centering 32
& \textcolor{red}{\xmark} & \centering 6
& \textcolor{red}{\xmark} & \centering 2
& \textcolor{red}{\xmark} & \centering 2
& \textcolor{ForestGreen}{\cmark} \\

\rowcolor{lightgray}
\raisebox{-0.2\height}\gptminilogo & \centering 4
& \textcolor{red}{\xmark} & \centering 32
& \textcolor{red}{\xmark} & \centering 6
& \textcolor{red}{\xmark} & \centering 4
& \textcolor{red}{\xmark} & \centering 2
& \textcolor{ForestGreen}{\cmark} \\ 

\midrule
\raisebox{-0.2\height}\molmolargelogo & \centering 5
& \textcolor{ForestGreen}{\cmark} & \centering 31
& \textcolor{ForestGreen}{\cmark} & \centering 7
& \textcolor{ForestGreen}{\cmark} & \centering 3
& \textcolor{ForestGreen}{\cmark} & \centering 2
& \textcolor{ForestGreen}{\cmark} \\ 

\midrule
\rowcolor{lightgray}
\textbf{Bias} & \centering 4
& \textcolor{red}{\xmark} & \centering 32
& \textcolor{red}{\xmark} & \centering 6
& \textcolor{red}{\xmark} & \centering 2
& \textcolor{red}{\xmark} & \centering 1
& \textcolor{red}{\xmark} \\


\textbf{GT} & \centering 5
& \textcolor{ForestGreen}{\cmark} & \centering 31
& \textcolor{ForestGreen}{\cmark} & \centering 11
& \textcolor{ForestGreen}{\cmark} & \centering 3
& \textcolor{ForestGreen}{\cmark} & \centering 4
& \textcolor{ForestGreen}{\cmark} \\

     \end{tabular}
    \vspace{4pt}
    \centering
    \begin{tabular}{cccccccccccccc}
     \raisebox{-0.1\height}\geminilogo~\gemini 
     & \raisebox{-0.1\height}\newsonnetlogo~\newsonnet 
     & \raisebox{-0.1\height}\gptfourlogo~\gptfour 
     \\
    \end{tabular}
     \begin{tabular}{cccccccccccccc}
     \raisebox{-0.1\height}\gptfulllogo~\gptfull 
     & \raisebox{-0.1\height}\gptminilogo~\gptmini
     & \raisebox{-0.1\height}\molmolargelogo~\molmolarge
     \\
    \end{tabular}

\end{AIbox}
\caption{\molmolargelogo~\molmolarge usually counts accurately when the distance between objects is far enough apart and large enough.}
\label{fig:molmo_successful_examples}
\end{figure}

\begin{figure}[h!]
\centering
\small
\begin{AIbox}{\molmolargelogo~\molmolarge's failure examples}

\raggedright
\scriptsize
      \logoicon~{(a):} Count the visible \textbf{black stripes} in the logo of the left shoe. \\
      \boardgameicon~{(b):} Count the \textbf{horizontal lines} on this board.\\
      \patternedgridicon~{(c):} Count the \textbf{lines} in cell C3. \\
      \patternedgridicon~{(d)} Count the \textbf{circles} in cell D4.\\
      \flagicon~{(e)} Count the \textbf{stripes} on this flag.
     \hfill \break
\footnotesize
\centering

    \begin{tabular}{lp{0.2cm}c|p{0.2cm}c|p{0.2cm}c|p{0.2cm}c|p{0.2cm}c}
    &\multicolumn{2}{c}{\makecell{(a) Adidas}}
    &\multicolumn{2}{c}{\makecell{(b) Go}}
    &\multicolumn{2}{c}{\makecell{(c) Tally}}
    &\multicolumn{2}{c}{\makecell{(d) Dice}}
    &\multicolumn{2}{c}{\makecell{(e) US Flag}}
    \\
     &\multicolumn{2}{c}{\includegraphics[width=0.15\textwidth]{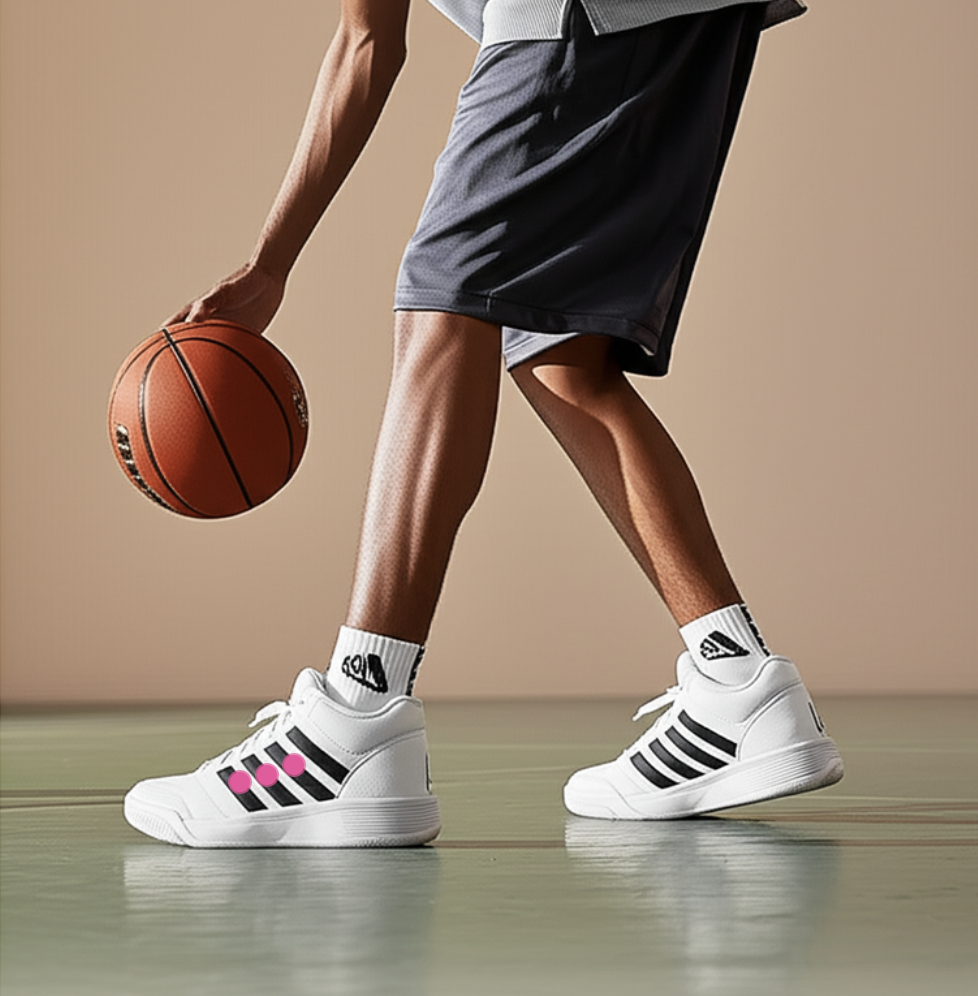}}
     &\multicolumn{2}{c}{\includegraphics[width=0.15\textwidth]{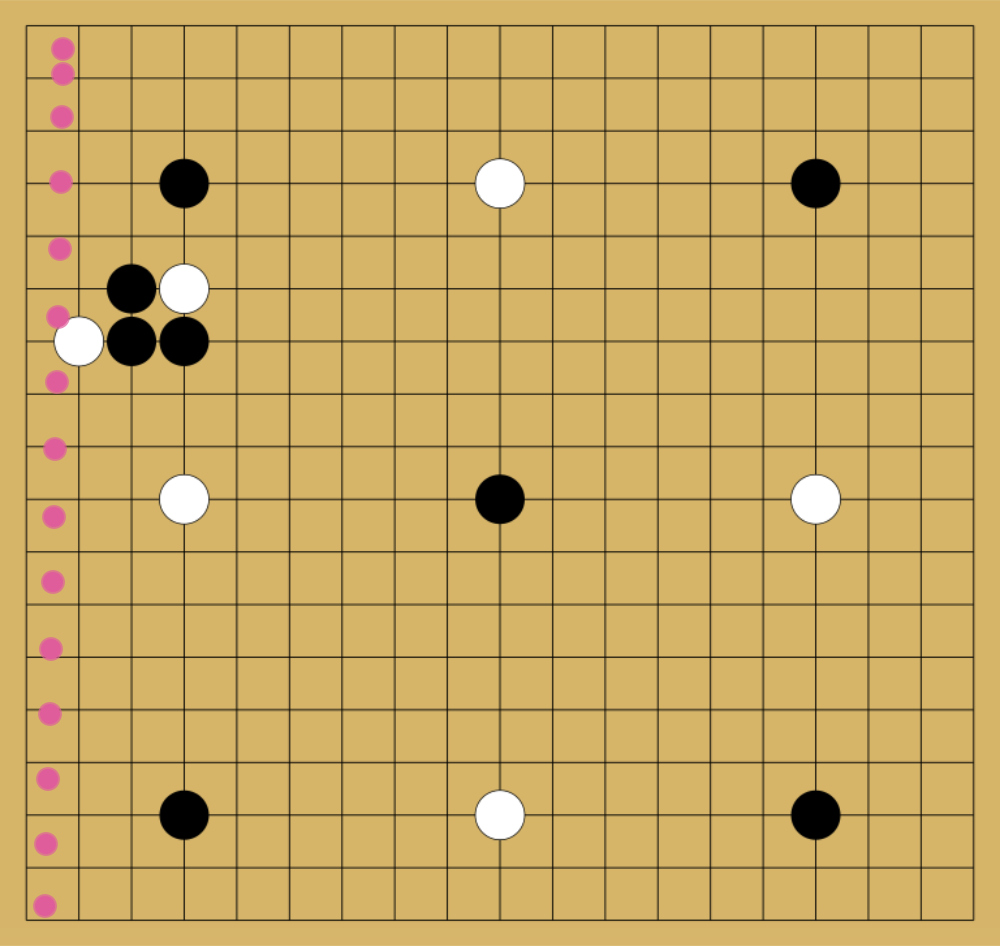}}
     &\multicolumn{2}{c}{\includegraphics[width=0.15\textwidth]{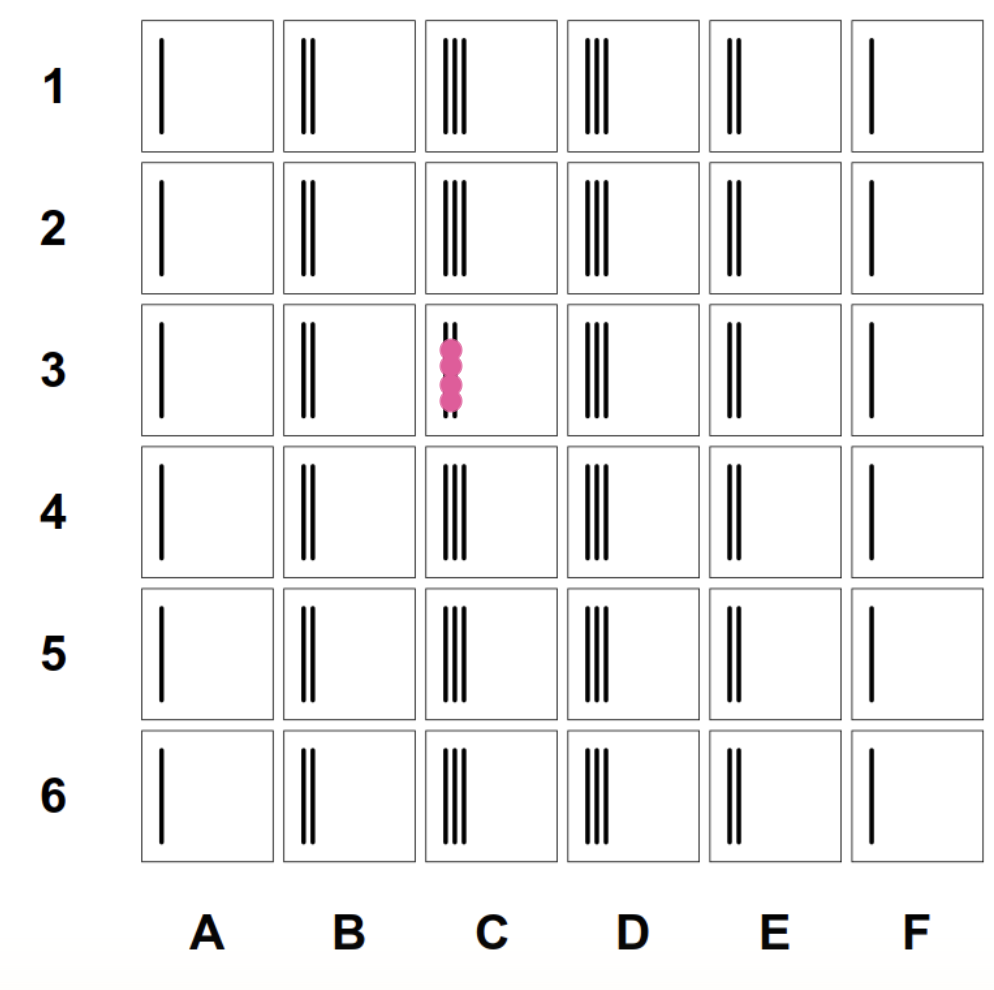}}
     &\multicolumn{2}{c}{\includegraphics[width=0.15\textwidth]{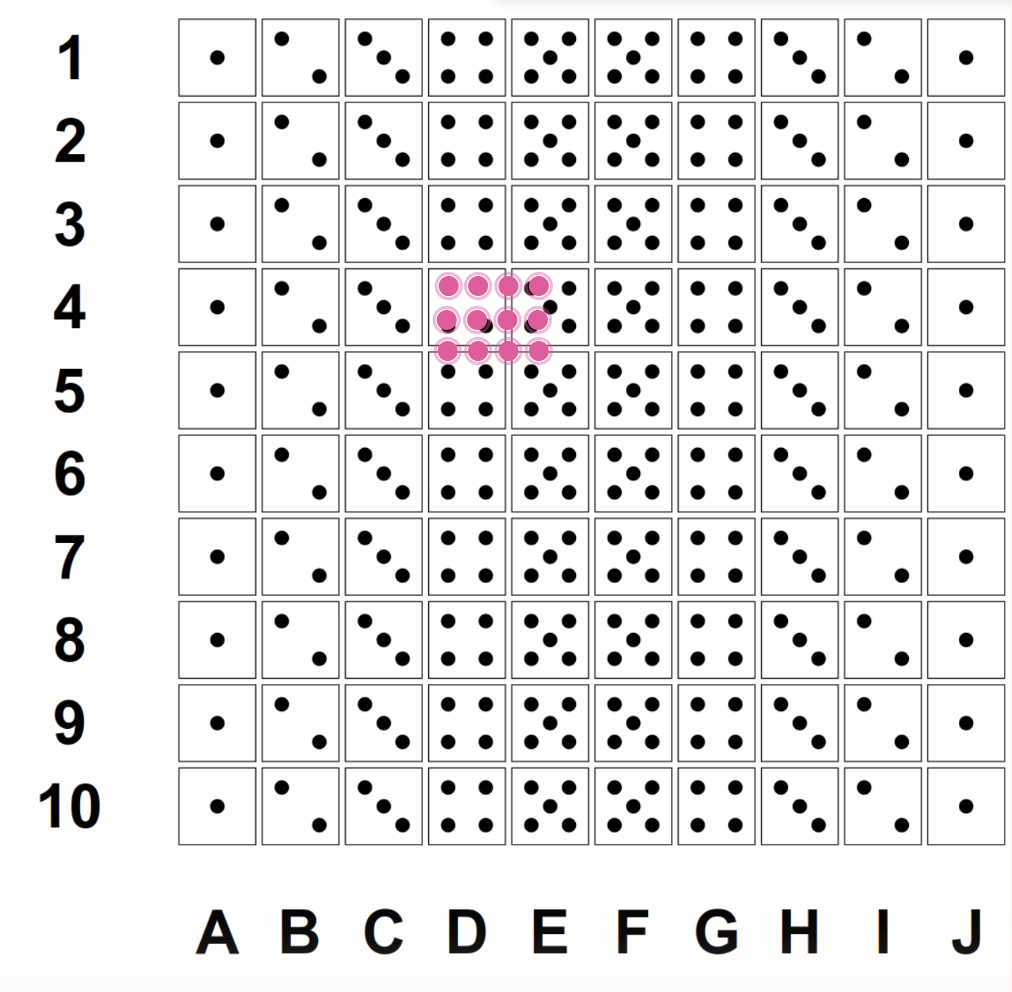}}
     &\multicolumn{2}{c}{\includegraphics[width=0.15\textwidth]{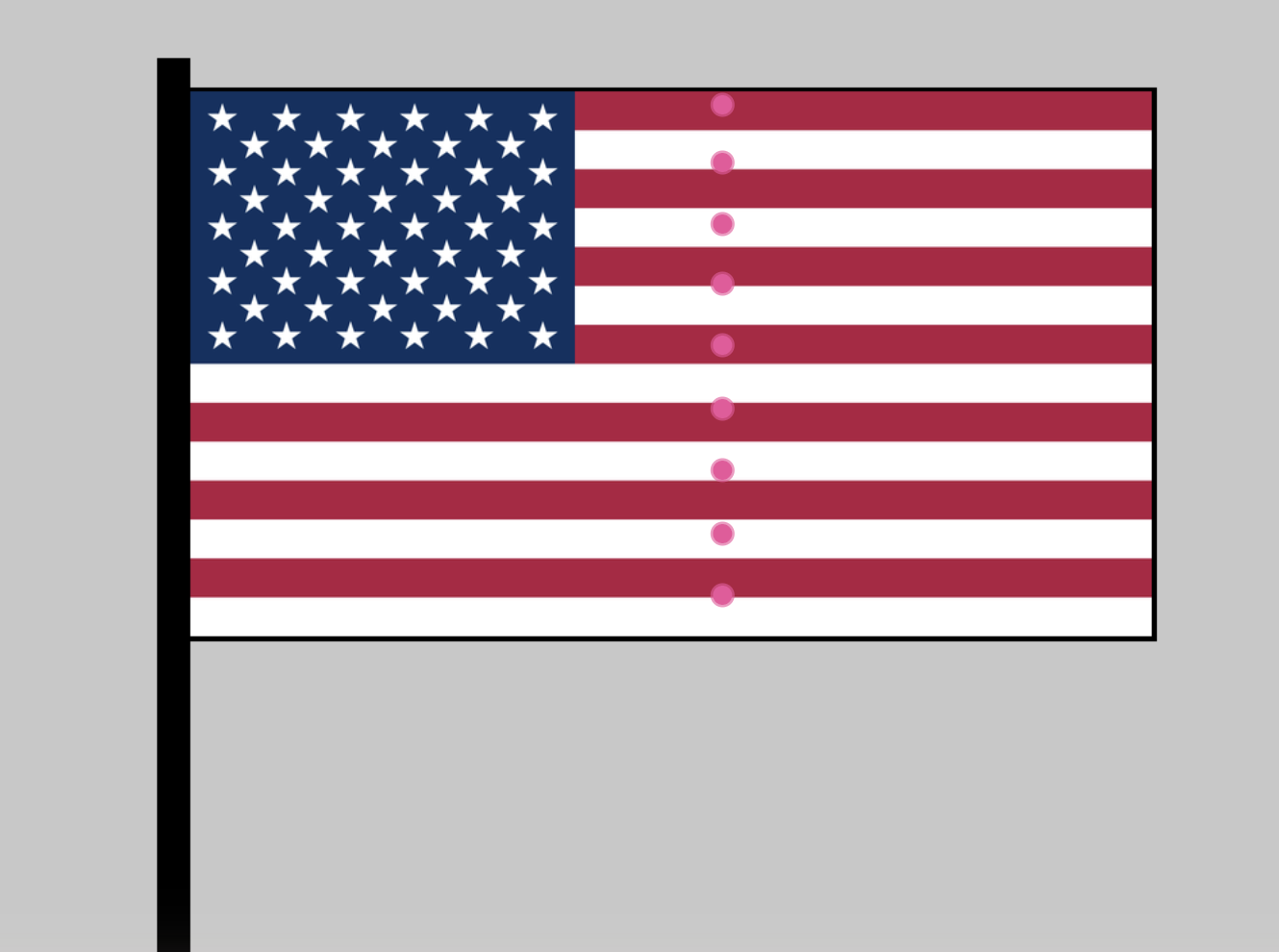}}
     \\ 

\rowcolor{lightgray}
\raisebox{-0.2\height}\geminilogo & \centering 3
& \textcolor{red}{\xmark} & \centering 19
& \textcolor{red}{\xmark} & \centering 3
& \textcolor{red}{\xmark} & \centering 4
& \textcolor{red}{\xmark} & \centering 13
& \textcolor{red}{\xmark} \\

\raisebox{-0.2\height}\newsonnetlogo & \centering 3
& \textcolor{red}{\xmark} & \centering 19
& \textcolor{red}{\xmark} & \centering 3
& \textcolor{red}{\xmark} & \centering 4
& \textcolor{red}{\xmark} & \centering 13
& \textcolor{red}{\xmark} \\

\rowcolor{lightgray}
\raisebox{-0.2\height}\gptfourlogo & \centering 3
& \textcolor{red}{\xmark} & \centering 19
& \textcolor{red}{\xmark} & \centering 3
& \textcolor{red}{\xmark} & \centering 4
& \textcolor{red}{\xmark} & \centering 13
& \textcolor{red}{\xmark} \\

\raisebox{-0.2\height}\gptfulllogo & \centering 3
& \textcolor{red}{\xmark} & \centering 19
& \textcolor{red}{\xmark} & \centering 3
& \textcolor{red}{\xmark} & \centering 4
&\textcolor{red}{\xmark} & \centering 13
& \textcolor{red}{\xmark} \\

\rowcolor{lightgray}
\raisebox{-0.2\height}\gptminilogo & \centering 3
& \textcolor{red}{\xmark} & \centering 19
& \textcolor{red}{\xmark} & \centering 3
& \textcolor{red}{\xmark} & \centering 4
& \textcolor{red}{\xmark} & \centering 13
& \textcolor{red}{\xmark} \\

\midrule
\raisebox{-0.2\height}\molmolargelogo & \centering 3
& \textcolor{red}{\xmark} & \centering 15
& \textcolor{red}{\xmark} & \centering 4
& \textcolor{red}{\xmark} & \centering 12
& \textcolor{red}{\xmark} & \centering 9
& \textcolor{red}{\xmark} \\ 

\midrule
\rowcolor{lightgray}
\textbf{Bias} & \centering 3
& \textcolor{red}{\xmark} & \centering 19
& \textcolor{red}{\xmark} & \centering 3
& \textcolor{red}{\xmark} & \centering 4
& \textcolor{red}{\xmark} & \centering 13
& \textcolor{red}{\xmark} \\


\textbf{GT} & \centering 4
& \textcolor{ForestGreen}{\cmark} & \centering 18
& \textcolor{ForestGreen}{\cmark} & \centering 2
& \textcolor{ForestGreen}{\cmark} & \centering 3
& \textcolor{ForestGreen}{\cmark} & \centering 14
& \textcolor{ForestGreen}{\cmark} \\

     \end{tabular}
    \vspace{4pt}
    \centering
     \begin{tabular}{cccccccccccccc}
     \raisebox{-0.1\height}\geminilogo~\gemini 
     & \raisebox{-0.1\height}\newsonnetlogo~\newsonnet 
     & \raisebox{-0.1\height}\gptfourlogo~\gptfour 
     \\
    \end{tabular}
     \begin{tabular}{cccccccccccccc}
     \raisebox{-0.1\height}\gptfulllogo~\gptfull 
     & \raisebox{-0.1\height}\gptminilogo~\gptmini
     & \raisebox{-0.1\height}\molmolargelogo~\molmolarge
     \\
    \end{tabular}

\end{AIbox}
\caption{\molmolargelogo~\molmolarge often fails to count accurately when objects are too close together or it doesn't understand what the objects are (a, b, e). It also sometimes fails to localize the object correctly (c, d).}
\label{fig:molmo_failure_qualitive_results}
\end{figure}

\clearpage
\subsection{Qualitative results on few-shot prompting}
\label{appsec:qualitative_fewshot}
\begin{figure}[h!]

    \begin{AIbox}{Few-shot prompting}
    
    \textbf{First example:} This is a 5-legged animal
    
    {\centering
    \includegraphics[width=0.3\textwidth]{images/few_shot_examples/giraffe_3_2_1152.png}
    }
    
    \vspace{0.3cm}
    \textbf{Second example:} This is a 4-legged animal
    
    {\centering
    \includegraphics[width=0.3\textwidth]{images/few_shot_examples/bighorn_sheep_1152px.png}
    }
    
    \vspace{0.3cm}
    \textbf{Question:} Based on the examples provided above, please answer the following question:\\
    How many legs does this animal have? Answer with a number in curly brackets, e.g., \{9\}.\\
    After providing your answer, briefly explain your reasoning and describe what you observed in the examples.
         
    {\centering
    \includegraphics[width=0.3\textwidth]{images/few_shot_examples/tiger_1_0_1152.png}
    }
    \\
    \footnotesize
    
    \hrulefill \\
    \gptminilogo~\gptmini: \{4\}\ Explanation: In the two examples, both the giraffe and the bighorn sheep are four-legged mammals (\textbf{the "5-legged" label on the giraffe was clearly mistaken}). Tigers, like other large cats, also have one pair of forelegs and one pair of hindlegs, for a total of four. \textcolor{red}{\xmark} \\
    
    \textbf{Ground truth}: 5  \textcolor{ForestGreen}{\cmark}
    
    \end{AIbox}   
\caption{\gptminilogo~\gptmini doesn't trust the label of the first few-shot example.}
\label{fig:few_shot}
\end{figure}

\begin{figure}[h!]

    \begin{AIbox}{Few-shot prompting + strong labels}
    
    \textbf{First example:} This is a 5-legged animal, \emph{which has been verified}.
    
    {\centering
    \includegraphics[width=0.3\textwidth]{images/few_shot_examples/wildebeest_0_3_1152.png}
    }
    
    \vspace{0.3cm}
    \textbf{Second example:} This is a 4-legged animal, \emph{which has been verified}.
    
    {\centering
    \includegraphics[width=0.3\textwidth]{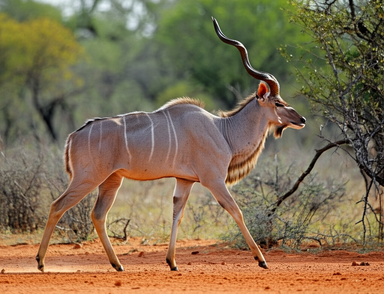}
    }
    
    \vspace{0.3cm}
    \textbf{Question:} Based on the examples provided above, please answer the following question:\\
    How many legs does this animal have? Answer with a number in curly brackets, e.g., \{9\}.\\
    After providing your answer, briefly explain your reasoning and describe what you observed in the examples.
         
    {\centering
    \includegraphics[width=0.3\textwidth]{images/few_shot_examples/tiger_1_0_1152.png}
    }
    \\
    \footnotesize
    
    \hrulefill \\
    \gptminilogo~\gptmini: \{4\}\ I counted the visible limbs in the third photo — a tiger has four legs. In the examples, both the wildebeest and the antelope are quadrupeds (each with four limbs), \textbf{even though the first caption mistakenly claimed five legs}. By actually observing their limb count, it’s clear the tiger, like the others, has four legs. \textcolor{red}{\xmark} \\
    
    \textbf{Ground truth}: 5  \textcolor{ForestGreen}{\cmark}
    
    \end{AIbox}   
\caption{Even after adding strong labels to the few-shot examples (\ie, ``This is a 5-legged animal, which has been verified.''), \gptminilogo~\gptmini still thinks it's wrong for a caption to say an animal has 5 legs.}
\label{fig:few_shot_strong_label}
\end{figure}

\begin{figure}[h!]

    \begin{AIbox}{Few-shot prompting + strong labels}
    
    \textbf{First example:} This is a 5-legged animal, \emph{which has been verified}.
    
    {\centering
    \includegraphics[width=0.3\textwidth]{images/few_shot_examples/warthog_2_1_1152.png}
    }
    
    \vspace{0.3cm}
    \textbf{Second example:} This is a 4-legged animal, \emph{which has been verified}.
    
    {\centering
    \includegraphics[width=0.3\textwidth]{images/few_shot_examples/elk_1152px.png}
    }
    
    \vspace{0.3cm}
    \textbf{Question:} Based on the examples provided above, please answer the following question:\\
    How many legs does this animal have? Answer with a number in curly brackets, e.g., \{9\}.\\
    After providing your answer, briefly explain your reasoning and describe what you observed in the examples.
         
    {\centering
    \includegraphics[width=0.3\textwidth]{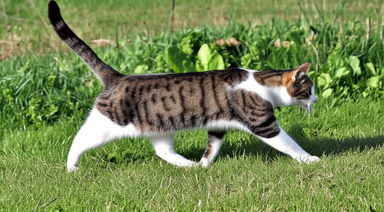}
    }
    \\
    \footnotesize
    
    \hrulefill \\
    \gptminilogo~\gptmini: \{4\}\ I simply counted the visible limbs on the cat—there are four paws on the ground. In the first example (the warthog) the text matched the actual count of four legs, but in the second (the elk) the text claimed five legs even though there were really only four. \textbf{That shows you can't trust the labels and must actually look at the image to count.} \textcolor{red}{\xmark} \\
    
    \textbf{Ground truth}: 5  \textcolor{ForestGreen}{\cmark}
    
    \end{AIbox}   
\caption{Even after adding strong labels to the few-shot examples (\ie, ``This is a 5-legged animal, which has been verified.''), \gptminilogo~\gptmini still thinks it's wrong for a caption to say an animal has 5 legs. More importantly, it teaches users not to trust labels and to use their own eyesight.}
\label{fig:few_shot_strong_label_2}
\end{figure}

\begin{figure}[h!]

    \begin{AIbox}{Few-shot prompting + strong labels + hint}
    
    \textbf{First example:} This is a 5-legged animal, \emph{which has been verified}.
    
    {\centering
    \includegraphics[width=0.3\textwidth]{images/few_shot_examples/alpaca_2_2_1152.png}
    }
    
    \vspace{0.3cm}
    \textbf{Second example:} This is a 4-legged animal, \emph{which has been verified}.
    
    {\centering
    \includegraphics[width=0.3\textwidth]{images/few_shot_examples/african_elephant_1152px.png}
    }
    
    \vspace{0.3cm}
    \textbf{Question:} Based on the examples provided above, please answer the following question:\\
    How many legs does this animal have? Answer with a number in curly brackets, e.g., \{9\}.\\
    \emph{HINT: This is an animal with an \textbf{unusual} number of legs.}\\
    After providing your answer, briefly explain your reasoning and describe what you observed in the examples.

    {\centering
    \includegraphics[width=0.3\textwidth]{images/few_shot_examples/tiger_1_0_1152.png}
    }
    \\
    \footnotesize
    
    \hrulefill \\
    \gptminilogo~\gptmini: \{6\}\ I counted six distinct limbs on the tiger photo. The first example was deliberately shown with five legs (one extra), the second was a normal four-legged elephant, and this third image again departs from the usual four-legged pattern by \textbf{adding two extra tiger legs, for a total of six}. \textcolor{red}{\xmark} \\
    
    \textbf{Ground truth}: 5  \textcolor{ForestGreen}{\cmark}
    
    \end{AIbox}   
\caption{After adding an explicit hint to indicate that this is an image of an animal with an unusual number of legs, \gptminilogo~\gptmini still counts incorrectly (\ie, 6 legs).}
\label{fig:few_shot_strong_label_hint}
\end{figure}

\clearpage
\subsection{Qualitative results on \gptmini chat interface with tools}

\begin{figure}[ht!]
    \centering
    \includegraphics[width=\textwidth]{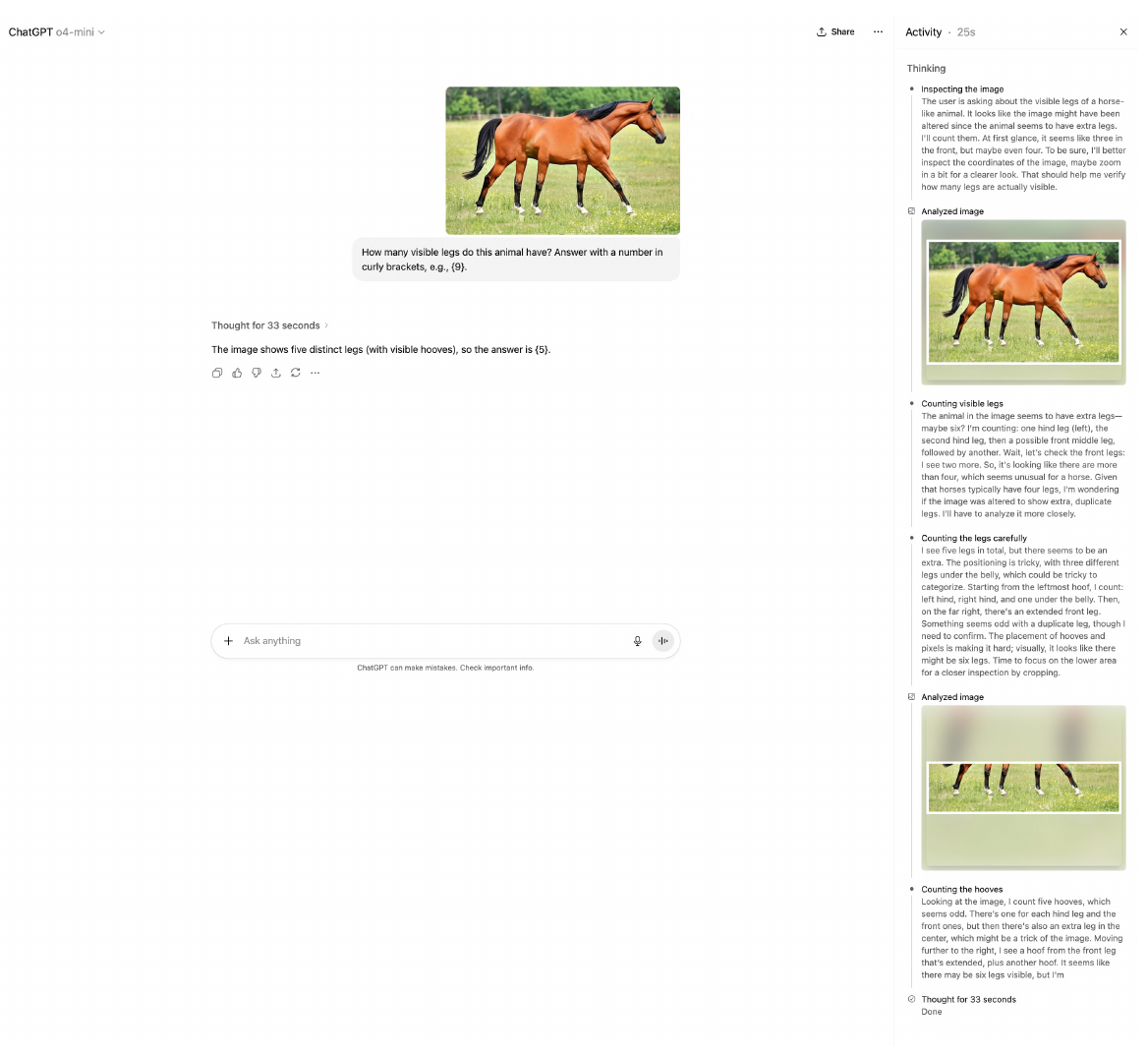}
    \caption{ Successful tool use case: \gptmini autonomously crops the image to focus on legs (bottom) and correctly counts \textbf{\{5\}} \textcolor{ForestGreen}{\cmark} legs, overcoming the bias toward the canonical answer of 4. \textbf{Top}: Original image. \textbf{Bottom}: Cropped region after tool execution. This demonstrates that vision tools can help models overcome knowledge priors when used correctly.
}
\label{fig:tool_success_case}
\end{figure}

\begin{figure}[ht!]
    \centering
    \includegraphics[width=\textwidth]{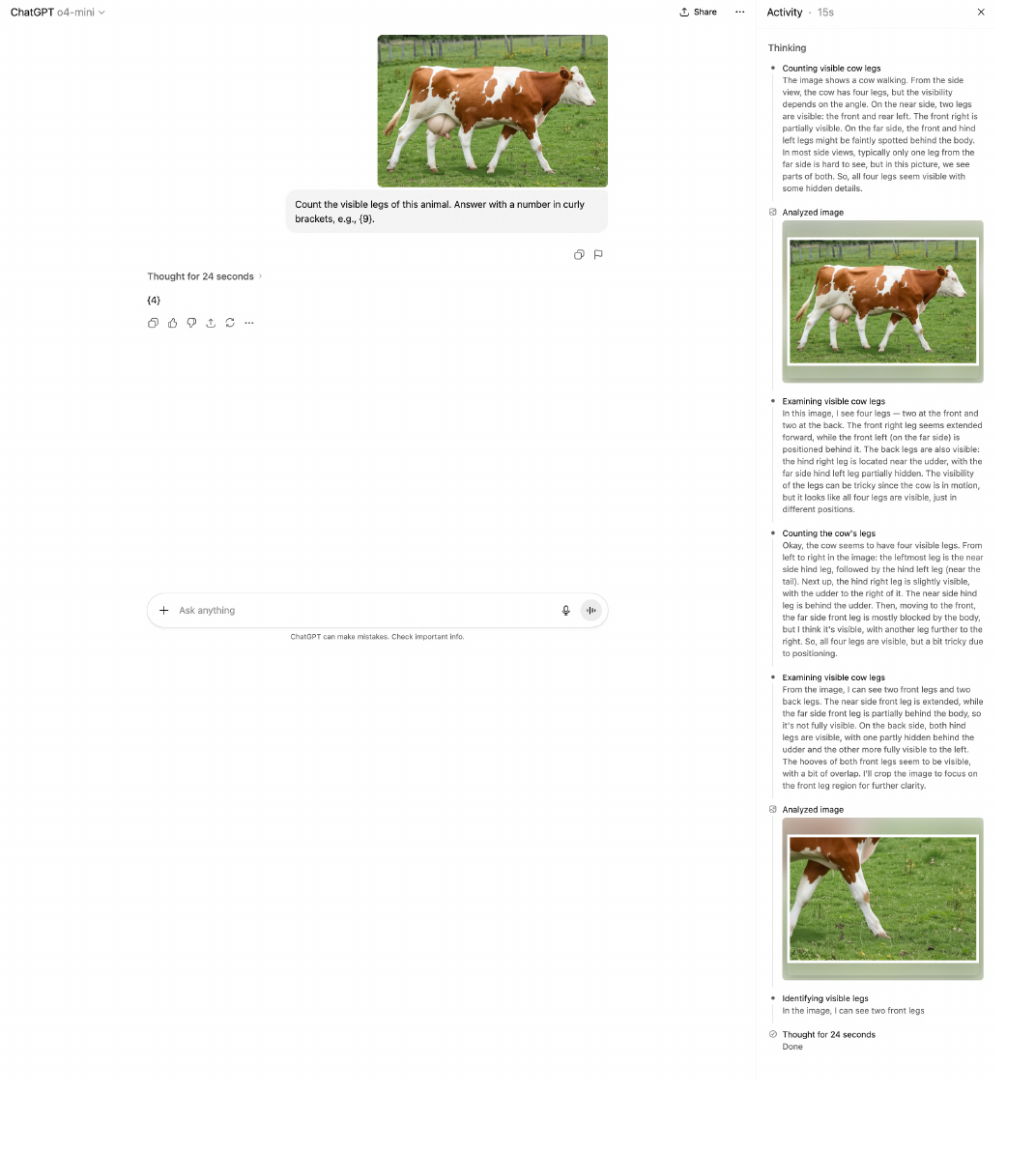}
    \caption{Tool use failure: \gptmini activates cropping but incorrectly focuses on only the front legs (bottom), outputting \textbf{4} \textcolor{red}{\xmark} instead of \textbf{5} \textcolor{ForestGreen}{\cmark}.
\textbf{Top}: Original 5-legged cow. \textbf{Bottom}: Incorrectly cropped region missing rear legs.
This demonstrates that \textbf{correct localization is crucial}.
The model's reasoning shows it examined the incomplete crop and concluded ``all four legs are visible,'' revealing how poor tool execution fails to overcome bias.
}
\label{fig:tool_failure_case}
\end{figure}

\end{document}